\documentclass[runningheads]{llncs}

    \usepackage{graphicx}
    \usepackage{tikz}
    \usepackage{comment}
    \usepackage{amsmath,amssymb} 
    \usepackage{color}
    \usepackage{graphicx}
    \usepackage{amsmath}
    \usepackage{amssymb}
    \usepackage{booktabs}
    \usepackage{times}
    \usepackage{epsfig}
    \usepackage{amsfonts}
    \usepackage{dsfont}
    \usepackage{multirow}
    \usepackage{enumitem}
    \usepackage{mathtools}
    \usepackage[style=plain]{floatrow}
    \usepackage{dirtytalk}
    \usepackage[normalem]{ulem}
    \usepackage{balance}
    
    \usepackage[bottom]{footmisc}
    \usepackage[super]{nth}
    \usepackage{subcaption}
    \usepackage{cite}
    \usepackage{comment}
    \usepackage{url}

    \DeclareMathOperator*{\argmin}{arg\,min}
    
    \makeatletter
    \newcommand{\printfnsymbol}[1]{%
      \textsuperscript{\@fnsymbol{#1}}%
    }
    \makeatother
    
\newif\ifdraft
\draftfalse
\drafttrue

\definecolor{orange}{rgb}{1,0.5,0}
\definecolor{violet}{RGB}{70,0,170}

\ifdraft
 \newcommand{\PF}[1]{{\color{red}{\bf PF: #1}}}
 
 \newcommand{\SH}[1]{{\color{blue}{\bf SH: #1}}}
 
 \newcommand{\SR}[1]{{\color{violet}{\bf SR: #1}}}
 
  \newcommand{\LC}[1]{{\color{cyan}{\bf LC: #1}}}
 
\else
 \newcommand{\PF}[1]{}
 
 \newcommand{\SH}[1]{}
 
 \newcommand{\SR}[1]{}
 
  \newcommand{\LC}[1]{}
 
\fi

\newcommand{\geomed}{\text{GeoMed}}

\def\Vec#1{{\boldsymbol{#1}}}
\def\Mat#1{{\boldsymbol{#1}}}

\newcommand{\mL}{\mathcal{L}}
\newcommand{\mU}{\mathcal{U}}

\newcommand{\bp}{\boldsymbol{p}}
\newcommand{\bx}{\boldsymbol{x}}

\newcommand{\bI}{\boldsymbol{I}}
\newcommand{\bX}{\boldsymbol{X}}
\newcommand{\bY}{\boldsymbol{Y}}

\def\ie{\emph{i.e.}}

\def\etal{\emph{et al}.}
    \usepackage[pagebackref=true,breaklinks=true,colorlinks,bookmarks=false]{hyperref}
    \usepackage[accsupp]{axessibility}

\begin{document}

\pagestyle{headings}
\mainmatter
\def\ECCVSubNumber{100}  

\title{On Triangulation as a Form of Self-Supervision for 3D Human Pose Estimation} 

\titlerunning{Robust Weighted Triangulation}
%

    \author{Soumava Kumar Roy\thanks{equal contribution}
    \and Leonardo Citraro\printfnsymbol{1} \and Sina Honari \and Pascal Fua }
    
    
    \authorrunning{Roy \etal}
    %
    
    \institute{Computer Vision Lab, EPFL, Switzerland \\ 
    \email{soumava.roy@epfl.ch ~~~~  leonardo.citraro@epfl.ch \\
    sina.honari@epfl.ch  ~~~~~~~~~  pascal.fua@epfl.ch}
    }

\maketitle
\begin{abstract}
   Supervised approaches to 3D pose estimation from single images are remarkably effective when labeled data is abundant. However, as the acquisition of ground-truth 3D labels is labor intensive and time consuming, recent attention has shifted towards semi- and weakly-supervised learning. Generating an effective form of supervision with little annotations still poses major challenge in crowded scenes. In this paper we propose to impose multi-view geometrical constraints by means of a weighted differentiable triangulation and use it as a form of self-supervision when no labels are available. We therefore train a 2D pose estimator in such a way that its predictions correspond to the re-projection of the triangulated 3D pose and train an auxiliary network on them to produce the final 3D poses. We complement the triangulation with a weighting mechanism that alleviates the impact of noisy predictions caused by self-occlusion or occlusion from other subjects. We demonstrate the effectiveness of our semi-supervised approach on Human3.6M and MPI-INF-3DHP datasets, as well as on a new multi-view multi-person dataset that features occlusion.
   \end{abstract}

\section{Introduction}
\label{sec:intro}
Supervised approaches in capturing human 3D pose are now remarkably effective, provided that enough annotated training data is available~\cite{Li14e,Li15a,Pavlakos16,Popa17,Sun18d,Tekin16b,Zhou17f,Zhou16b}. However, there are many scenarios that involve unusual activities for which not enough annotated data can be obtained. Semi-supervised or unsupervised methods are then required~\cite{Li19c,Pavllo19,Chen17h,Tung17a,Kanazawa19b,Chen19g}. A promising subset of those rely on constraints imposed by multi-view geometry: If a scene is observed by several cameras, there is a single {\it a priori} unknown 3D pose whose projection is correct in all views~\cite{Rhodin18a,Kocabas19,Mitra20}. This is used to provide a supervisory signal with limited need for manual annotations. There are many venues, such as a sports arena, that are equipped with multiple cameras, which can be easily used for this purpose.

However, some of these semi-supervised approaches are sensitive to occlusions and noisy predictions~\cite{Rhodin18a,Mitra20}. Others rely on a non-differentiable implementation of triangulation, which precludes making it a full-fledged component of the learning pipeline~\cite{Kocabas19}, while some others require full labels to apply a learnable triangulation~\cite{Iskakov19}. 
In this paper, we propose an approach that overcomes these limitations. To this end, we start from the observation that there are now many off-the-shelf pre-trained 2D pose estimation models that do not necessarily perform well in new environments but can be fine-tuned to do so. Our approach therefore starts with one of these that we run on all multi-view images whose results can be triangulated. The model is then trained so that the prediction in all views correspond to the re-projection of the pseudo 3D pose target. In particular, we give to the 2D prediction of each view a weight, that is derived from the level of agreement of that view with respect to all other views. In doing so, we apply a principled triangulation weighting mechanism to obtain pseudo 3D targets that can discard erroneous views, hence helping the model bootstrap in semi-supervised setups. Our approach can be used in any multi-view context, without restriction on camera placement. At inference time, our retrained network can be used on single-view images and have their output lifted to 3D by an auxiliary network. 

We demonstrate the effectiveness of our approach on the traditional Human3.6M~\cite{Ionescu14a} and MPI-INF-3DHP~\cite{Mehta17a} datasets. To highlight its ability to handle more crowded scenes, we test the robustness of our approach on a new multi-person dataset of an amateur basketball match. 

In short, our contributions are:
\begin{itemize}
\item a self-supervised multi-view consistency loss based on differentiable triangulation that takes the 2D network output as its input, without requiring annotations;
\item a weighting strategy that mitigates the effect of occlusion and noisy predictions;
\item a new multi-view multi-person datasets of an amateur basketball match featuring occlusion and difficult light conditions.
\end{itemize}
We will make the dataset publicly available along with our code. 


\section{Related Work}

With the advent of deep learning, direct regression of 2D and 3D poses from images has become the dominant approach for human pose estimation~\cite{Li14e,Newell16,Mehta17b,Rogez17,Rogez18}. Recent variations on this theme use volumetric representations \cite{Iskakov19}, lift 2D poses to 3D~\cite{Martinez17a,Iqbal20}, use graph convolutional network (GCN) \cite{Ci19,Cai19,Zeng20,Liu20f,Zou21} or generate and refine pseudo labels from video sequences \cite{Li19c,Pavllo19}. Given enough annotated training data, these methods are now robust and accurate. Thus, in recent years, the attention has shifted from learning 3D human poses in fully supervised setups to semi- or weakly-supervised ones \cite{Pavlakos17,Kanazawa19b,Wang19m,Pavlakos19b,Wandt19,Chen19g}, with a view to drastically reduce the required amount of annotated images.

In many venues such as sport arenas, there are multiple cameras designed to provide different viewpoints, often for broadcasting purposes. Hence, a valid approach is to exploit the constraints that multi-view geometry provides for training purposes, while at inference time, the trained network operates on single views. For example, in~\cite{Kocabas19},  a single-view 3D pose estimator is trained using pseudo labels generated by triangulating the predictions of a pre-trained 2D pose estimator in multiple images. However, as the 2D pose estimator is fixed, pseudo label errors caused by noisy 2D detections are never corrected.

In~\cite{Rhodin18a,Iqbal20}, a regressor is trained to produce 3D poses from single-views, while guaranteeing consistency of the predictions across multiple views. A problem with these approaches is that the predictions can be consistent but wrong, which makes it necessary to use some annotated 3D or 2D data. The method of~\cite{Rhodin18b} relies on a similar setup but reduces the required amount of annotated data by using a multi-view unsupervised pre-training strategy that learns a low-dimensional latent representation from which the 3D poses can be inferred using a simple regressor. However, due to the low-dimensionality of the learned latent space, the structural information from the image is lost, which makes it less accurate in practice.
In~\cite{Kundu20}, an encoder-decoder based self-supervised training algorithm is used to disentangle the appearance and the pose features of the same subject extracted from two different images. It also uses a part-based 2D puppet model as an additional form of prior knowledge regarding the 2D human skeleton structure and appearance to guide the learning process. This results in a very complex pipeline that requires a great deal of additional data during training without delivering superior performance as we demonstrate in the result section.

Triangulation is at the heart of our approach and is also used extensively in the fully-supervised method of~\cite{Iskakov19}, which uses 3D labels to refine the triangulation process, and of \cite{Remelli20a} where 2D poses are fused in a latent space. While \cite{Remelli20a} uses a standard triangulation that does not weigh different views, ~\cite{Iskakov19} learns a weight for each view, which is designed for a fully-supervised setup in order to have an effective weighting scheme. In contrast, ours is derived from the agreement of 2D views; therefore, it is better suited for semi- and weakly-supervised setups because the weights are computed from multi-view consistency rather than learned. In the result section, we show that the triangulation of \cite{Iskakov19} applied to our semi-supervised framework results in less accurate poses when the amount of labeled data is limited.

\begin{figure*}[!t]
    \begin{center}
    \includegraphics[width=1.0\linewidth]{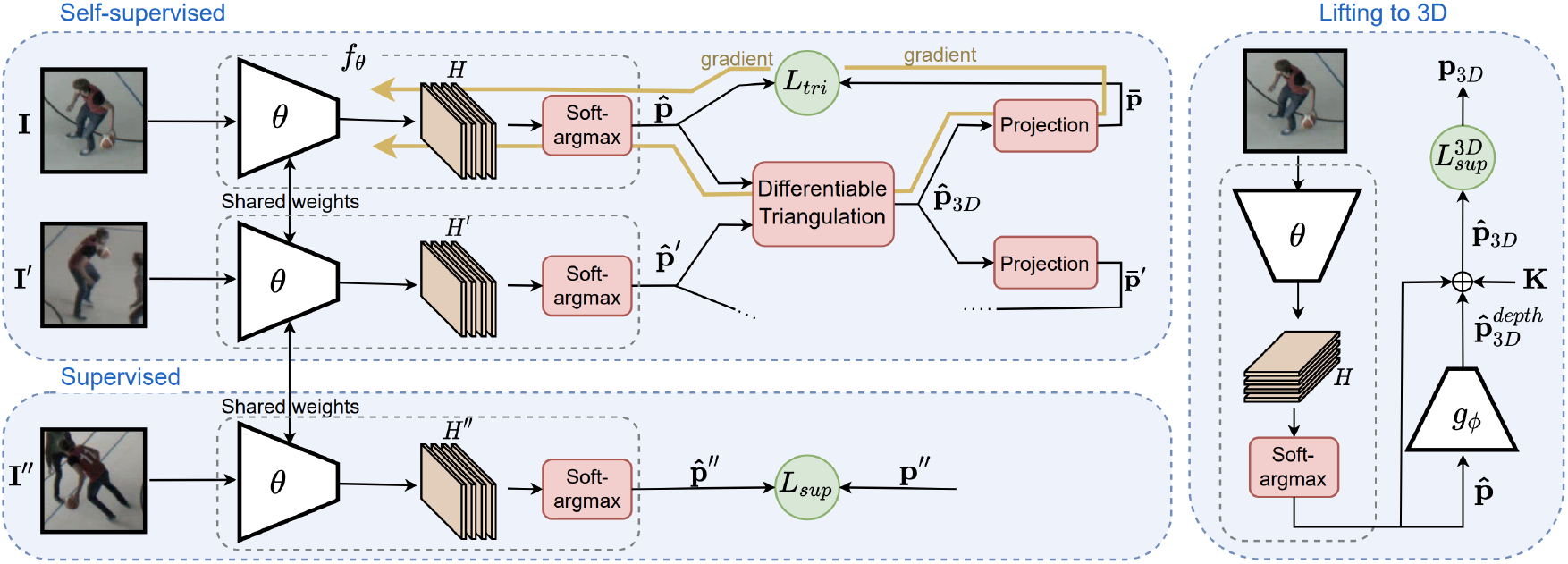} 
    \end{center}
    \vspace{-5mm}
    \caption{\small {\bf Network Architecture.} It comprises a detection network that outputs 2D joint locations in individual views and a lifting network that predicts root relative distances 
    to the camera which are then turned to 3D poses using the 2D ones and the intrinsic camera parameters. They are trained jointly using a small number of images with corresponding 3D annotations and a much larger set of images acquired using multiple cameras. During training, the prediction networks feed their 2D detections into a differentiable triangulation layer that estimates 3D joint locations. These are re-projected into the images to minimize the discrepancy between the projections and the original detections, along with a supervised loss that operates on the frames for which annotations are available. The estimated 3D joint locations are also used to train the lifting network.}
    \label{fig:arch}
\end{figure*}

\section{Method}
\label{sec:method}

Our goal is to train a deep neural network to estimate 3D human poses from single images using as little labeled data as possible. 
To this end, we acquire multiple views of our subjects and train a 2D pose estimator to detect joint locations consistently across views.
This {\it detection network} is complemented by a {\it lifting network} that infers 3D poses from 2D ones. 
During training, we check for consistency across views of the unlabeled images by triangulating the 2D detections, then we re-project the results into the images and use the triangulated 3D points to train the lifting network. 

Both networks are trained in a semi-supervised way on our target dataset, that is, we assume that the ground-truth 2D and 3D poses are known only for a small subset of the entire dataset. To train the detection network, in addition to a small set of 2D labels, we use a robust differentiable triangulation layer that takes as input the predicted 2D poses for the unlabeled images in individual views and produces the 3D estimates. These are used as pseudo labels to check for consistency of the 2D pose estimates and to train both the detection and the lifting networks. The complete pipeline is depicted by Figure~\ref{fig:arch} and is end-to-end trainable. Making the triangulation process robust to mis-detections in some of the views is a key component of our approach because some joints are not visible in some views, especially in crowded scenes. 

\subsection{Notations}
Let $\mL = \{[\bI^{l,c}]_{c=1}^{N_C}, \bp^l_{3D}\}_{l=1}^{N_L}$ be a sequence of labeled RGB images $\Mat{I}^{l,c} \in \mathbb{R}^{h \times w \times 3}$ taken using $N_C$ different cameras, and $\bp^l_{3D}$ be the ground-truth 3D pose, where $c$ denotes the index of cameras and $l$ is the index of the labeled images. Similarly, let $\mU = \{[\bI^{u,c}]_{v=1}^{N_C}\}_{u=1}^{N_U}$, where $u$ denotes the unlabeled, be a larger sequence of multi-view images without associated 3D pose labels. We take $\bp^l_{3D}$ to be the entire set of body joints $\{\bX^l_j\}_{j=1}^{N_J}$, where $\bX^l_j \in \mathbb{R}^3$ denotes the 3D coordinates of joint $j$ for the $l$ sample. Similarly, let $\bp^l_{2D}$ be $\{\bx^{l,c}_j\}_{j=1}^{N_J}$, where $\bx^{l,c}_j \in \mathbb{R}^2$ is the projection of $\bX^l_j$ in view $c$. In the remainder of this section, we drop the $l$, $u$, and $c$ notations when there is no ambiguity. 
 
Let  $f_{\Mat{\theta}}$  and $g_{\Mat{\phi}}$ denote the detection and lifting network with weights $\Mat{\theta}$ and $\Mat{\phi}$ respectively. The detection network $f_{\Mat{\theta}}$ takes $\bI^c$ as input and returns a 2D pose estimate in that view ${\hat \bp_{2D}^c}=\{\hat{\bx}^{c}_j\}_{j=1}^{N_J}$. The lifting network $g_{\Mat{\phi}}$ takes as input such a 2D pose predictions and returns root relative distances to the camera which are then turned to a 3D pose estimate ${\hat \bp}_{3D}$ using the known intrinsic camera parameters $\textbf{K}$. 

Finally, let $\Pi$ be a function that takes as input the known camera parameters and a set of image points, one per view, and triangulates to a location in the world ${\hat \bX}_j = \Pi(\{\hat \bx^{c}_j \}_{c=1}^{N_C} )$.

\subsection{Training the Detection and Lifting Networks}
\label{sec:losses}

We train the detection network $f_{\Mat{\theta}}$  and the lifting network $g_{\Mat{\phi}}$ by minimizing two loss functions that we define below, which we describe in more details in the following subsections.

\subsubsection{Detection Loss $L^f(\theta)$.}

We take it to be the sum of a supervised term computed on $\mL$ and an unsupervised one computed on $\mU$. We write it as
\begin{align}
L^f(\theta) & = L_{sup}(\theta; \mathcal{L}) + L_{tri}(\theta; \mathcal{U}) \nonumber \; ,\\
L_{sup}(\theta; \mathcal{L}) &= \sum_{l=1}^{N_L} \sum_{j=1}^{N_J} \sum_{c=1}^{N_C} \left\| \Vec{\hat x}^{l,c}_j - \bx^{l,c}_j \right\|^2_2 \; , \label{eq:projLosses} \\
L_{tri}(\theta; \mathcal{U}) &= \sum_{u=1}^{N_U} \sum_{j=1}^{N_J} \sum_{c=1}^{N_C} \left\| \hat{\bx}^{u,c}_j-{{\bar \bx}^{u,c}_j}\right\|^2_2 \; , \nonumber
\end{align}
where ${\bar \bx}^{u,c}_j$ denotes the projection of $\hat \bX^u_j$ obtained by triangulating the ${\hat \bx}^{u,c}_j$ predicted by the network. $L_{sup}$ is the supervised MSE loss between the predicted 2D poses and the ground-truth ones. $L_{tri}$ is the self-supervised loss that is at its minimum when the detected 2D poses are consistent with the projections of the triangulated poses. $L_{tri}$ pools information from the multiple views to provide a supervisory signal to our model without using any labels. All the image locations within the losses are normalized in the range $[-1,1]$ with respect to their subject's bounding box.

As shown in Figure~\ref{fig:arch}, $L_{tri}$ provides two sets of gradients during back-propagation. One from the re-projected triangulated pose directly to the detection model and one that flows into the triangulation. Blocking either one of these results in a different behavior that makes this self-supervised loss less stable in practice. We provide more details and experiments to highlight this in the supplementary.

\subsubsection{Lifting Loss $L^g(\phi; f_{\theta})$.}

Once $f_{\Mat{\theta}}$ is trained, we use its predictions on $\mathcal{U}$ and the triangulation operation $\Pi$ to create a new set with pseudo 3D poses $\mathcal{\hat U}  = \{[\Mat{I}^{u,c}]_{c=1}^{N_C},\Vec{\hat p}^{u}_{3D}\}_{u=1}^{N_U}$. We consider the lifting loss to be 
\begin{equation}
L^g(\phi; f_{\theta}) = L_{semi-sup}^{3D}(\phi; f_{\theta}, \mathcal{L}^\frown \mathcal{\hat U})
\label{eq:liftLoss}
\end{equation}
with respect to $\phi$. $L_{semi-sup}^{3D}$ is the MSE loss between the predicted 3D poses and ground-truth or triangulated ones. The parameters of the 2D pose estimator $f_{\theta}$ are fixed and provide the 2D poses as input to $g_{\phi}$. $^\frown$ denotes the concatenation operation on $\mathcal{L}$ and $\mathcal{\hat U}$. 

\subsection{Self-Supervision by Triangulation}
\label{sec:self-supervised}

Computing $L_{tri}(\theta; \mathcal{U})$ of Eq.~\ref{eq:projLosses} requires triangulating each joint $j$ in the set of 2D locations $\{\hat{\bx}^{c}_j\}_{c=1}^{N_C}$. In theory, the most reliable way to triangulate is to perform non-linear minimization~\cite{Hartley04}. However, because this computation is part of our deep learning, it must be differentiable, which is not easy to achieve for this kind of minimization. We therefore rely on the  Direct Linear Transform (DLT)~\cite{AbdelAziz71}, which involves transforming the projection equations to a system of linear ones whose solution can be found using Singular Value Decomposition~\cite{Golub96}. This is less accurate but easily differentiable. 

In practice, occlusions and mis-detections pose a much more significant challenge than any loss of accuracy due to the use of DLT. It is inevitable that some joints are hidden in some views and their estimated 2D locations $\hat{\bx}^{u,c}_j$ are erroneous, which results in inaccurate triangulations. To prevent this, we introduce a weighing strategy that reduces the influence of mis-detections in such problematic views. We apply these weights in DLT and re-write the unsupervised loss of Eq.~\ref{eq:projLosses} as
\begin{align}
L_{tri}(\theta; \mathcal{U}) &= \sum_{u=1}^{N_U} \sum_{j=1}^{N_J} \sum_{c=1}^{N_C} w_j^{u,c} \|\hat{\bx}^{u,c}_j-{{\bar \bx}^{u,c}_j}\|^2_2 \; , 
\end{align}
where $w_j^{u,c}$ denotes the reliability of the estimate of the location of joint $j$ in view $c$. To estimate these weights we use the purely geometrical approach based on robust statistics described below.

\subsubsection{Robust Localization.}

To robustly estimate the 3D location of a joint  $j$, we first compute candidate 3D locations using all pairs of 2D detections  $(\hat{\bx}^{c}_j,\hat{\bx}^{c'}_j)$ in pairs of views. Let 
\begin{align}
	\label{eqn_pairwise_triangulation}
	T_j & = \{ \hat \bY_j^{c,c'} = \Pi(\bx^{c}_j,\bx^{c'}_j) | (c,c') \in \binom{N_C}{2}\} \; , \\
	\tilde \bX_j &= \geomed(T_j) \nonumber
\end{align}
where $T_j$ is a {\it detection cluster} and $\tilde \bX_j$ is the center of the cluster.
$\Pi$ denotes the triangulation, $\binom{N_C}{2}$ denotes the set of all pairs of views, and $\geomed$ refers to calculating the geometric median of $T_j$. 
Given $T_j$ and $\tilde \bX_j$, we take the weights to be
\begin{align}
	\label{eqn:pairwise_distance}
	w^{c}_j &= \textrm{Median} (W^c_j) \\
	W^c_j  &= \{ w_j^{c,c'} = \exp(-\frac{\Vert \tilde \bX_j - \hat \bY^{c,c'}_j \Vert^2_2}{\sigma^2}) | c' \in C_j \backslash \{c\} \} \nonumber
\end{align}
where $W^c_j$ is a set of weights for joint j$^{th}$ from view $c$ to all other views. These encode the distance of each candidate location in $T_j$ to its center and are bounded in the range $[0,1]$ using a Gaussian function. $\sigma$ is a parameter of the model that is  set according to the amount of noise present in the candidate joints. Intuitively, when a joint estimate in $T_j$ is an outlier, its weights should be closer to zero, otherwise they should be close to one. In all our experiments we set it to $10$ millimeters.

\subsubsection{Joint Selection.}
\label{sec:jointSelect}

The above formulation relies on $\tilde \bX_j$, obtained in Eq.~\ref{eqn_pairwise_triangulation}, being reliable, which is usually true when more than $50 \%$ of the 2D detections are correct. However, this is not always the case. To handle this, at each training iteration, we assign a score to each $T_j$ and momentarily discards the joints whose score is below a certain threshold. 
To compute this score, we use the normalized Within-Cluster-Sum of Squared Errors (WSS) as follows:
\begin{equation}
WSS_j = \frac{1}{|T_j|}\sum_{\hat \bY_{j}^{c,c'} \in T_j} \Vert \hat \bY_{j}^{c,c'} - \tilde \bX_j \Vert^2_2 \; .
\label{eq:WSS}
\end{equation}

WSS is a measure of homogeneity defined within an agglomeration often used in clustering methods. Intuitively, when the image points are erroneous, the joint estimates composing the cluster $T_j$ are distant from one another resulting in a high $WSS_j$. Therefore, if $WSS_j$ is greater than a manually defined value during training, we momentarily discard the joint. As the training progresses, the model improves and allows the previously discarded joints to contribute again towards the training of the model. In all our experiments we set this threshold to $20$ millimeters that we found empirically.

In summary, the weights computed using our purely geometric approach of Eq.~\ref{eqn:pairwise_distance} allow the triangulation to output a more robust estimate of the true location of the joint. That is the center of the cluster $\tilde \bX_j$. Instead, Eq.~\ref{eq:WSS} allows to discard the joints that are considered unreliable, thus reducing their negative impact on the model. In Section \ref{sec:triangle_robustness}, we demonstrate how our approach produces more accurate triangulations when less than $50 \%$ of the views are affected by the noise, while obtaining equal results to a standard triangulation approach otherwise.


\subsection{Implementation Details}

\subsubsection*{Lifting 2D Poses to 3D.}

The second stage of our pipeline ``lifts" the predictions of our finetuned 2D pose estimator to 3D poses in the camera coordinate system. We adopt a commonly used pose representation \cite{Sun18d, Pavlakos18a, Iqbal18,Iqbal20} namely 2.5D $\Mat{p}^{2.5D} = \{ (u_j, v_j, d^{root} + d^{rel}_j) \}_{j=1}^{N_J} $ where $u_j$ and $v_j$ are the components of joint j$^{th}$ in the undistorted image space, $d^{root}$ is a scalar representing the depth of the root joint with respect to the camera and $d^{rel}_j$ is the relative depth of each joint to the root. The advantage of using this representation lies in the fact that the 2D pose $\{ (u_j, v_j) \}_{j=1}^{N_J} $ is spatially coincident with the image, which allows to fully exploit the characteristics of convolutional neural networks. In addition, the 2D pose estimator can be further improved using additional in-the-wild 2D pose datasets.

To obtain the 3D poses, we first train a multi-layer neural network as in \cite{Martinez17a} to output root relative depths $d^{rel}_j = g_{\phi}(f_{\theta}(\Mat{I}))$. Then, we recover the 3D pose in the camera or in the world coordinate system using the inverse of the projection equations
\begin{equation}
    (d^{root}+d^{rel}_j)\begin{pmatrix}
           u_j \\
           v_j \\
			1
         \end{pmatrix} 
         = \Mat{K} 
         \begin{pmatrix}
           X^C_j \\
           Y^C_j \\
			(d^{root}+d^{rel}_j)
         \end{pmatrix}
         = \Mat{K} \left[ \Mat{R}
         \begin{pmatrix}
           X_j \\
           Y_j \\
		   Z_j
         \end{pmatrix}    
         + \Vec{t}      
          \right]
\end{equation}
where $X^C$ and $Y^C$ are the first two components of a joint in camera space, $(X_j,Y_j,Z_j)$ is the joint in the world coordinate and $\Mat{K} \in \mathbb{R}^{3 \times 3}$, $\Mat{R} \in \mathbb{R}^{3 \times 3}$, and $\Vec{t} \in \mathbb{R}^{3 \times 1}$ are the intrinsic matrix, rotation matrix, and translation vector of the camera respectively.
The above formulation assumes that the depth of the root joint $d^{root}$ is known. In earlier works~\cite{Rhodin18a, Martinez17a, Sun18d}, the position of the root is taken to be the ground-truth one or not used at all simply because the dataset itself is provided in camera coordinate. We follow the same protocol and use the ground-truth depth of the root, however, an approximation can be obtain analytically as shown in \cite{Iqbal18}.

A simple multi-layer neural network can effectively learn the distribution of relative depths from the joint coordinate of the 2D poses, however, the image can still provide useful information to disambiguate certain cases. For this reason, we use an additional convolutional neural network $h_{\gamma}$, parameterized by $\gamma$, that takes an image $\bI$ as input and outputs a small feature vector. We then concatenate this feature vector with the joint coordinates of a 2D pose and feed it into the multi-layer neural network to output the relative depth $d^{rel}_j = g_{\phi}(f_{\theta}(\Mat{I})^\frown h_{\gamma}(\Mat{I}))$. This additional network contributes to distinguish classes of poses such as when subjects are standing, sitting and lying that can be more easily captured from the images themselves. In our experiments we consider $h_{\gamma}$ to be a ResNet50~\cite{He16a} pretrained on ImageNet~\cite{Russakovsky15} and replace its last linear layer with an MLP that shrinks the size of the output feature from $2048$ to $16$.

\subsubsection*{Detection and Lifting models.}
We use Alphapose \cite{Fang17a} trained on Crowdpose dataset \cite{Li18l} as our 2D pose estimator $f_{\Mat{\theta}}$ where we replace the original non-differentiable Non-Maximum Suppression (NMS) by a two-dimensional {\it Soft-argmax} as in \cite{Sun18d}. For the lifting network $g_{\Mat{\phi}}$ we used a multi-layer perceptron network with 2 hidden layers with a hidden size of $2048$. The first two layers are composed of a linear layer followed by ReLU activation and $10 \%$ dropout while the last layer is a linear layer. The lifting network takes as input the feature vector obtained by concatenating the joint coordinates of a 2D pose and the $16$ dimensional feature obtained from $h_{\gamma}$. $h_{\gamma}$ takes the cropped image resized to $256\times256$ as the input. The 2D pose is defined in the undistorted image space and is normalized in the range $[-1,1]$ with respect to the bounding box.

\subsubsection*{Training Procedure.}
We resize the input crops to $320 \times 256$ pixels by keeping the original aspect ratio and apply random color jittering for regularization.
In all our experiments we pretrained the 2D detection network first on the small set of labeled set and then we finetuned it on both labeled and unlabeled sets with our losses. During training, we used mini-batches of size $24$ samples where $8$ samples come from the labeled set and the others from the unlabeled one. We use Adam optimizer \cite{Kingma15} with constant learning rate of $1e-4$ for the pre-training and $1e-5$ for the fine-tuning. We do not use weight decay regularization. All the 3D losses were computed on poses expressed in meters and the image coordinates of the 2D losses are all normalized in the range $[-1,1]$ with respect to the bounding boxes.
\section{Experiments}
\label{sec:exp}

We primarily report our results in the \textbf{semi-supervised} learning setup, where we have 3D annotations only for a subset of the training images. We use the projection of these 3D annotations to train $f_{\Mat{\theta}}$ and they are directly used to supervise the training of $g_{\Mat{\phi}}$. For the unannotated samples, we use the 3D poses triangulated from 2D estimations and their projections to train both $g_{\Mat{\phi}}$ and $f_{\Mat{\theta}}$.

\begin{table}[t]
	\captionof{table}{{\small Quantitative results Human3.6M (Semi-supervised)}}
	\scalebox{1.2}{
	\begin{tabular}{l|c|c|c}
		\toprule
		\multicolumn{4}{c}{10\% of All Data} \\
		\midrule
		\textbf{Method} & \textbf{MPJPE}~$\downarrow$ & \textbf{NMPJPE}~$\downarrow$ & \textbf{PMPJPE}~$\downarrow$ \\
		\midrule
		Kundu \textit{et.al.}~\cite{Kundu20} &  - & - & 50.8 \\
		\textbf{Ours} & \textbf{56.9} & \textbf{56.6} & \textbf{45.4} \\
		\bottomrule
		\multicolumn{4}{c}{Only S1} \\
		\toprule
		Rhodin \etal~\cite{Rhodin18b}  & 131.7 & 122.6 & 98.2 \\
		Pavlako \etal~\cite{Pavlakos19b}  & 110.7 & 97.6 & 74.5 \\
		Li  \etal~\cite{Li19c}  & 88.8 & 80.1 & 66.5 \\
		Rhodin  \etal~\cite{Rhodin18a}  & - & 80.1 & 65.1 \\
		Kocabas  \etal~\cite{Kocabas19}  & - & 67.0 & 60.2 \\
		Pavllo \etal~\cite{Pavllo19} & 64.7 & 61.8 & - \\
		Iqbal \etal~\cite{Iqbal20} & 62.8 & \textbf{59.6} & 51.4 \\
		Kundu  \etal~\cite{Kundu20} & - & - & 52 \\
		\midrule
		Ours & \textbf{60.8} & 60.4 & \textbf{48.4} \\
		\bottomrule
	\end{tabular}
}	
\label{tab:h36m_semi} 
\end{table}

\subsection{Datasets and Metrics}
\label{sec:datasets}
We validate our proposed framework on two large 3D pose estimation datasets and a new multi-view dataset featuring an amateur basketball match to study the influence of occlusions in a crowded scene. We will release it publicly.

\paragraph{Human3.6M~\cite{Ionescu14a}.}
It is the most widely used indoor dataset for single and multi-view 3D pose estimation. It consists of $3.6$ million images captured form $4$ calibrated cameras. As in most published papers, we use subjects S$1$, S$5$, S$6$, S$7$, S$8$ for training and S$9$, S$11$ for testing. In the semi-supervised setup we follow two separate protocols:
	\begin{enumerate}
		\item Subject S$1$ is the only source of supervision, while other training subjects are treated as unlabeled, as done in  \cite{Rhodin18a, Pavlakos19b, Li18c}.
		\item Only $10 \%$ of uniformly sampled training samples are considered annotated and the rest as unlabeled \cite{Kundu20}. 
	\end{enumerate}

\paragraph{MPI-INF-3DHP~\cite{Mehta17a}. }

This dataset contains both constrained indoor and some complex outdoor images for single person 3D pose estimation. It features $8$ subjects performing $8$ different actions, which are recorded using $14$ different cameras, thereby covering a wide range of diverse 3D poses. We follow the standard protocol and use the 5 chest-height cameras only. In the semi-supervised setup, as for the Human3.6M dataset, we exploit annotations  for subject  S$1$ of the training set.

The sampling rate for the unlabeled set is set to $5$ for both the above mentioned datasets.

\paragraph{SportCenter.} We filmed an amateur basketball match using $8$ fixed and calibrated cameras. $2$ of the $8$ cameras are mounted on the roof of the court, which makes them useful for location tracking but less useful for pose estimation.  The images feature a variable number of subjects ranging from $10$ to $13$. They are either running, walking, or standing still. The players are often occluded either by others or by various objects, such as the metal frames of the nets. There are also substantial light variations that makes it even more challenging. We computed the players' trajectories for the whole sequence and manually annotated a subset with 2D poses. Thereafter, we obtained the 3D poses by triangulating the manually annotated 2D detections. The dataset comprises $315,000$ images out of which $560$ are provided with $3,740$ 2D poses and $700$ 3D poses. We use two subjects for testing and the remaining subjects are used for training. In total, we have $140$ annotated 3D poses for the test phase, while the remaining $560$ annotated 3D poses are used for training in the semi-supervised setup. 

\paragraph{Metrics.} 

We report the Mean Per-Joint Position Error (MPJPE), the normalized NMPJPE, and the procrustes aligned PMPJPE in millimeters. The best score is always shown in \textbf{bold}.

\paragraph{Calibration and Data Association.}

Our self-supervised loss function assumes the \emph{intrinsic} and \emph{extrinsic} camera parameters to be known. In practice, they can be  obtained using Bundle Adjustment~\cite{Triggs00}, which is a well established technique that is implemented by numerous software packages. In other words, this requirement is much less onerous than having to manually annotate 3D ground-truth poses. For our experiments, we used the camera parameters provided in the Human3.6M~\cite{Ionescu14a} and MPI-INF-3DHP~\cite{Mehta17a} datasets and performed Bundle Adjustment to obtain the camera parameters for the SportCenter dataset. We also assume that the ground-location of each person to be given. This lets us compute the bounding boxes for each person by assuming an average person height and projecting a cylinder of that height in each view. For Human3.6M and MPI-INF-3DHP we use the ground-truth locations provided in each of the datasets respectively, while, for the SportCenter dataset, we computed the trajectories using \cite{Lenz15b,Wang19f} and manually fixed the error in the highly crowded frames, which represent about 20\% of the total.

%
\begin{table}[!t]
    \captionof{table}{{\small Quantitative results MPI-INF-3DHP (Semi-supervised)}}
\scalebox{1.2}{
	\begin{tabular}{l|c|c|c}
		\toprule		
		\textbf{Method} & \textbf{MPJPE}~$\downarrow$ & \textbf{NMPJPE}~$\downarrow$ & \textbf{PMPJPE}~$\downarrow$\\ 
		\midrule
		\multicolumn{4}{c}{Only S1} \\
		\toprule
		Rhodin  \etal~\cite{Rhodin18a}  & - & 121.8 & -   \\
		Kocabas  \etal~\cite{Kocabas19} & - & 119.9 & -  \\
		Iqbal \etal~\cite{Iqbal20}  & 113.8 & 102.2 & - \\
		\midrule 
		Ours & \textbf{102.2}  & \textbf{99.6} & 93.6 \\
		\bottomrule
	\end{tabular}
}	

\label{tab:mpii_semi}
\end{table}
\begin{table*}[!t]
\caption{\small Quantitative results on SportCenter (semi-supervised) using different triangulation approaches.}
\label{tab:sports_center_semi_supervised}
\scalebox{0.9}{
	\begin{tabular}{l|ccc|ccc}
		\toprule
		\textbf{Method} & \multicolumn{3}{c|}{\textbf{Multiview}} & \multicolumn{3}{c}{\textbf{Single View}} \\
		\cmidrule{1-7}
		& \textbf{MPJPE} & \textbf{NMPJPE} & \textbf{PMPJPE} & \textbf{MPJPE} & \textbf{NMPJPE} & \textbf{PMPJPE} \\
		\midrule
         Ours non-differentiable w/o weights  &  109.7 & 107.7 & 97.6 & 142.9 & 140.4 &108.6  \\ 
         Ours non-differentiable  & 83.0  & 79.3  & 70.2 & 111.4 &  107.7 & 82.5 \\ 
         Ours w/o weights  & 80.5 & 78.4 & 66.7 & 118.5 & 116.5 & 95.4 \\
         Ours+Iskakov~\cite{Iskakov19} & 88.3 & 83.6 & 70.9 & 121.1 & 119.5 & 99 \\
		Ours & \textbf{66.9} & \textbf{65.5} & \textbf{55.4}  & \textbf{104.4} & \textbf{102.1} & \textbf{81.1}\\
		\bottomrule
	\end{tabular}
}
\end{table*}


\subsection{Quantitative Evaluation}
\label{sec:quant_eval}

We report results for our semi-supervised learning setup in Tables \ref{tab:h36m_semi}, \ref{tab:mpii_semi} and \ref{tab:sports_center_semi_supervised} for Human3.6M~\cite{Ionescu14a}, MPI-INF-3DHP~\cite{Mehta17a} and SportCenter datasets respectively. 

Considering only $10 \%$ of data as labeled for Human3.6M dataset in Table \ref{tab:h36m_semi},  our approach outperforms by a significant margin Kundu~\etal~\cite{Kundu20}. Similarly, we perform better than baselines when considering all samples of S$1$ as labeled (Only S1). The second best method is Kundu~\etal~\cite{Kundu20} but note that they rely on additional datasets such as in-the-wild YouTube videos and the MADS~\cite{Zhang17f} dataset, in addition to a part-based puppet to instill prior human skeletal knowledge in their overall learning framework, which we do {\it not}. In other words, we were able to eliminate  the need of such supplementary components by designing an effective geometrical reasoning based on a weighting strategy that mitigates the impact of noisy detections. 

Table~\ref{tab:mpii_semi} shows a comparative performance on the MPI-INF-3DHP dataset. Again, we outperform the competing methods without having to use any additional training datasets. 

\begin{figure*}[t]
    \centering
    \subfloat{{\includegraphics[trim={0.0cm 0 0 0},clip,height=0.32\linewidth, keepaspectratio]{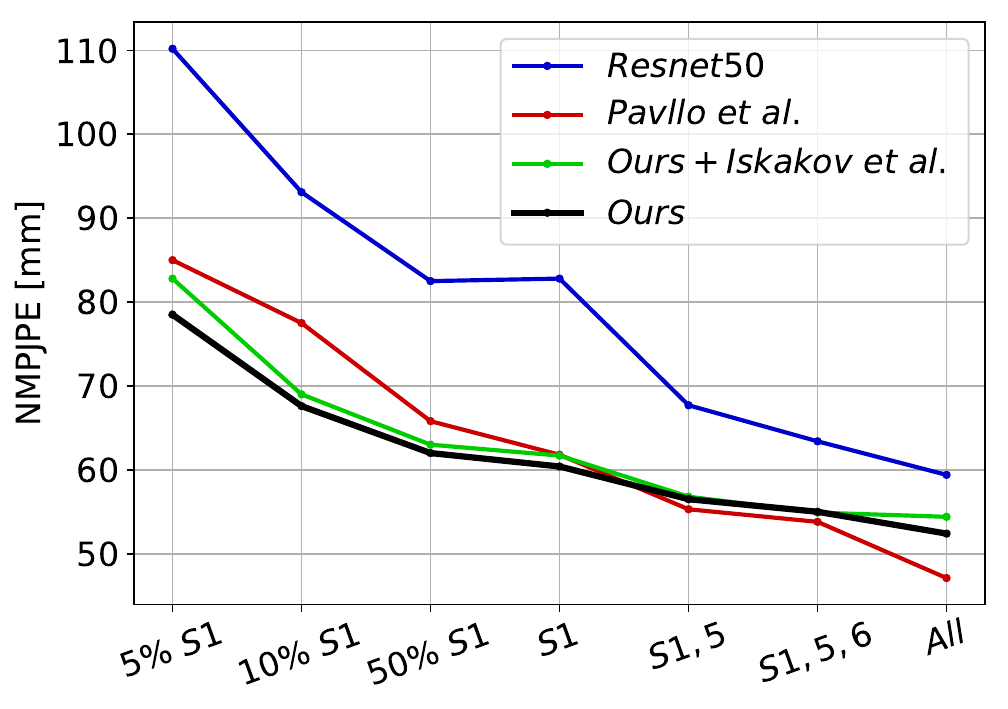} }}%
     \quad
    \subfloat{{\includegraphics[trim={0.0cm 0 0 0},clip,height=0.32\linewidth, keepaspectratio]{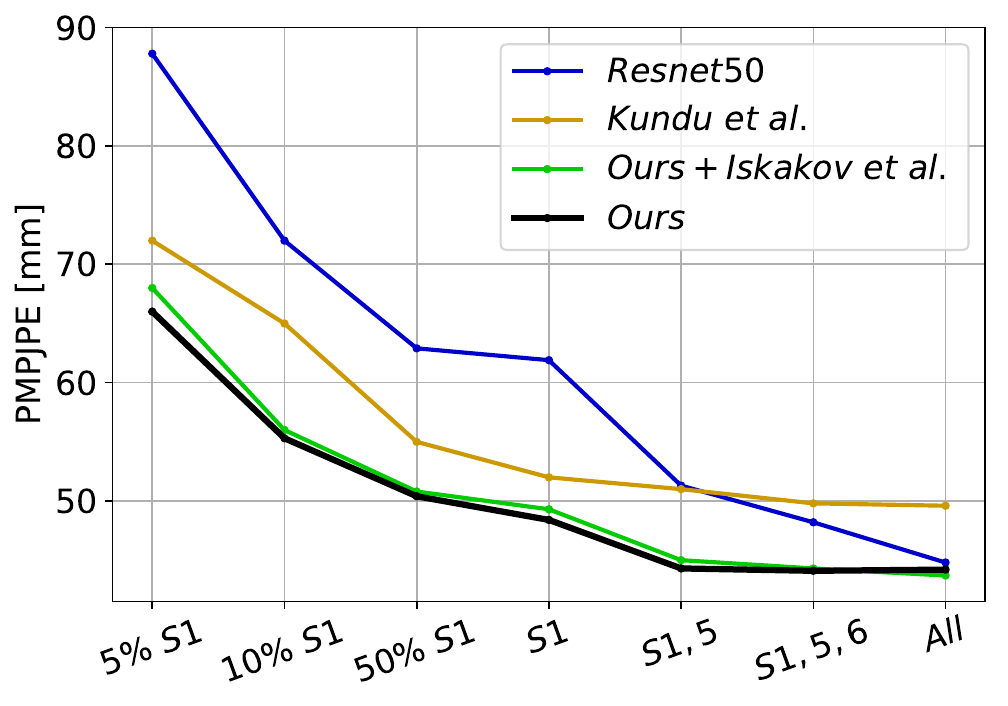} }}%
    \caption{\small Comparison with Pavllo~\etal~\cite{Pavllo19}, Kundu~\etal~\cite{Kundu20} and our semi-supervised variant with the triangulation of Iskakov~\cite{Iskakov19} on (left) NMPJPE and (right) PMPJPE metrics on Human3.6m with different amount of supervision.}
    \label{fig:plot_results}
\end{figure*}

In Figure~\ref{fig:plot_results}, we amplify on these results by showing NMJPE and PMPJPE performances as a function of the percentage of annotated training data we use. As observed we outperform the other methods in the semi-supervised setups with small number of annotations, in particular less than 100$\%$ of S1. In the fully-supervised setup, there are other methods such as ~\cite{Pavllo19} that perform better. This can be attributed to using temporal consistency in videos whereas we operate on single frames. Hence, we can safely assume that using temporal consistency would also boost our numbers and this is something we will investigate in future work.

\subsection*{Triangulation Components Impact}
\label{sec:triangle_component}
We evaluate the effectiveness of our approach and variants of it on the SportCenter dataset, which contains occlusions. To this end, we compare our triangulation approach ({\it Ours}) against three variants, namely (a) {\it Ours non-differentiable}, where we remove the gradient flowing through the triangulation process, (b) {\it Ours w/o weights} where we disable the weighting mechanism such that each view is given equal importance in the triangulation process and (c) {\it Ours non-differentiable w/o weights} which is a combination of the above two variants. In addition, we compare our approach to {\it Ours+Iskakov} where we use the weighting strategy proposed in~\cite{Iskakov19}. This entails training a neural network module to assign to each views weights when triangulating. In contrast, our approach to weighting relies purely on geometry, which is beneficial when only little labeled data is available. 

\begin{figure*}
    \centering
    
    \begin{subfigure}[b]{0.24\linewidth}        
        \centering
        \includegraphics[width=\linewidth]{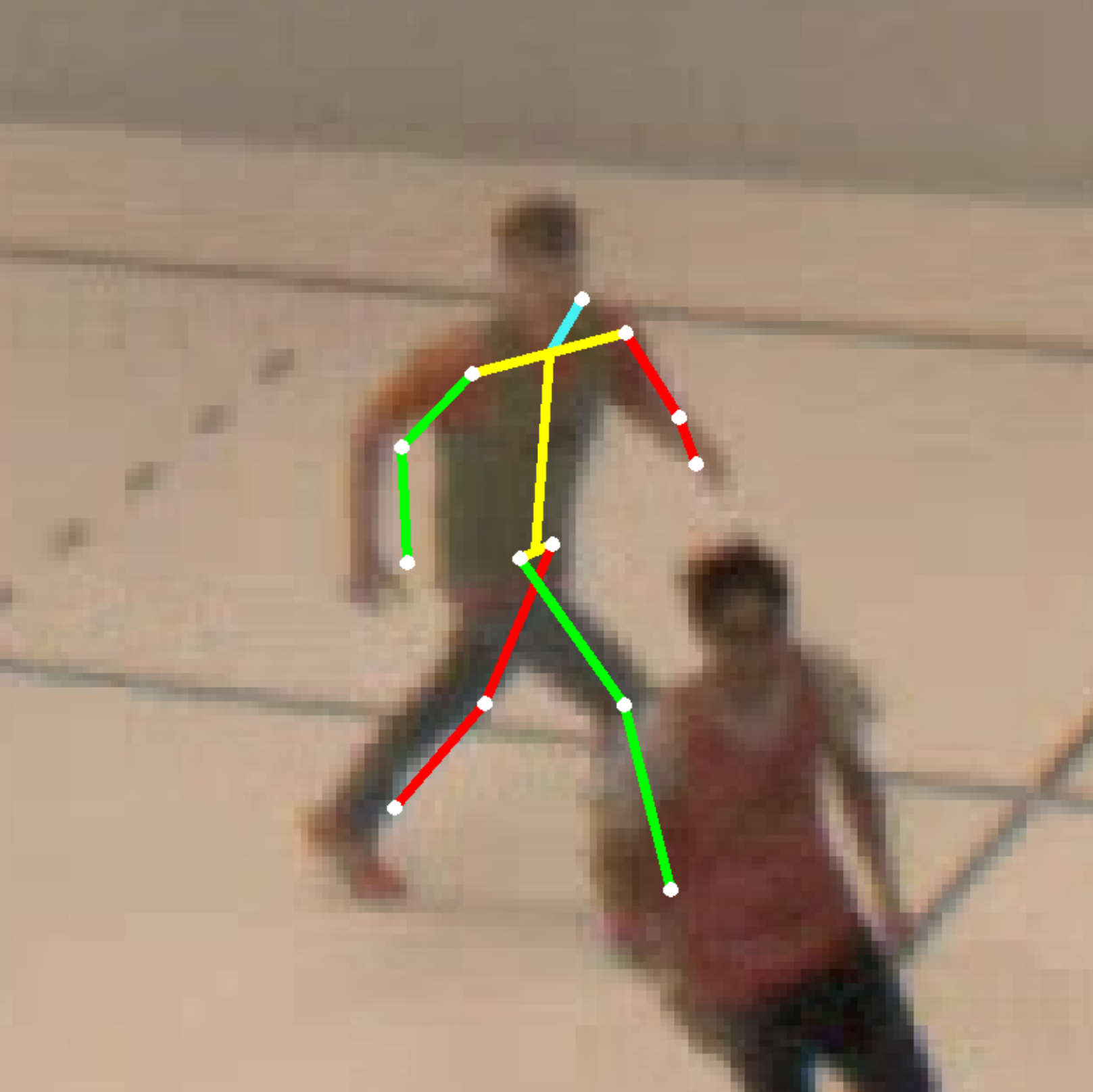}
    \end{subfigure}
    \begin{subfigure}[b]{0.24\linewidth}        
        \centering
        \includegraphics[width=\linewidth]{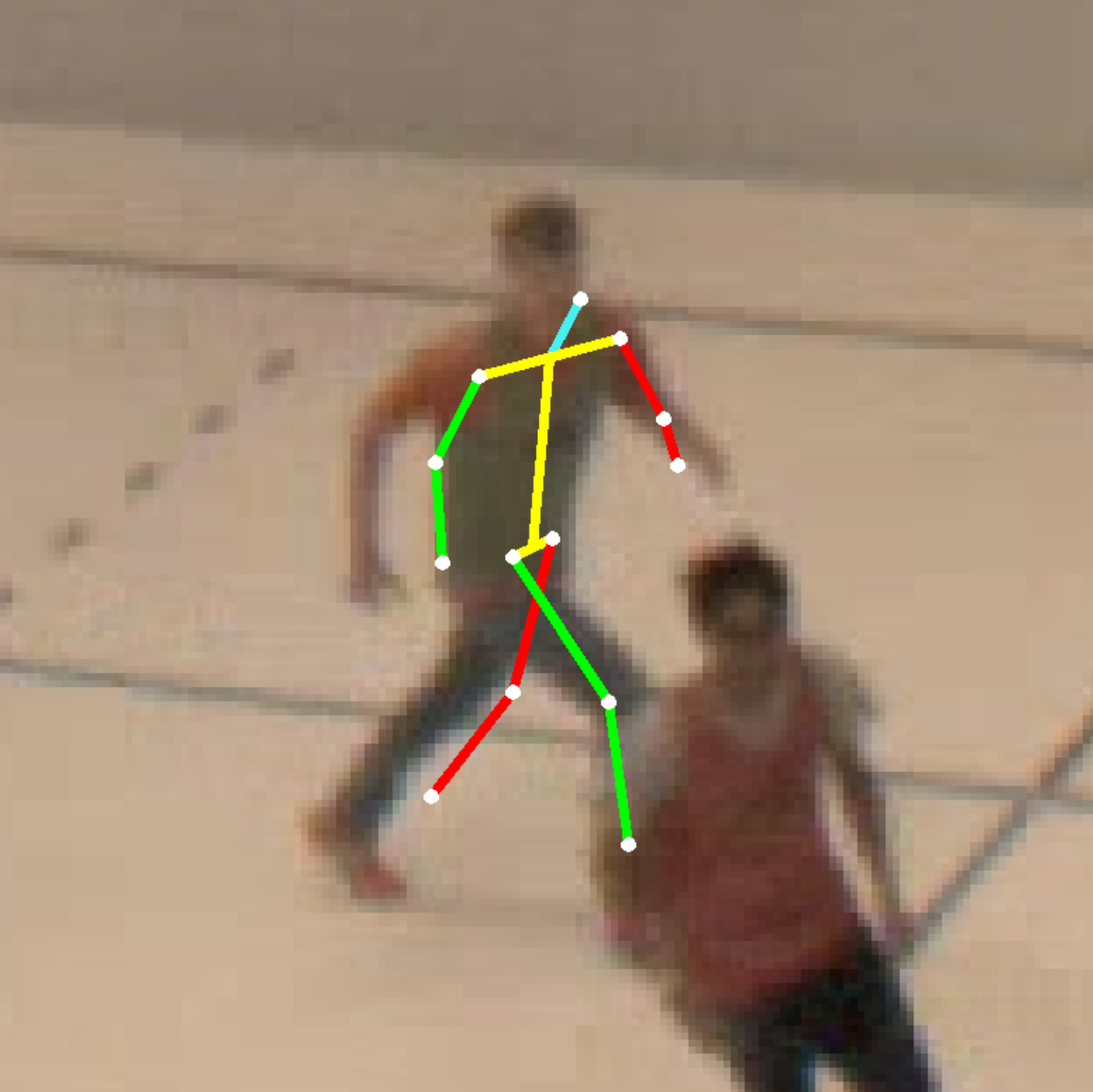}
    \end{subfigure}
    \begin{subfigure}[b]{0.24\linewidth}        
        \centering
        \includegraphics[width=\linewidth]{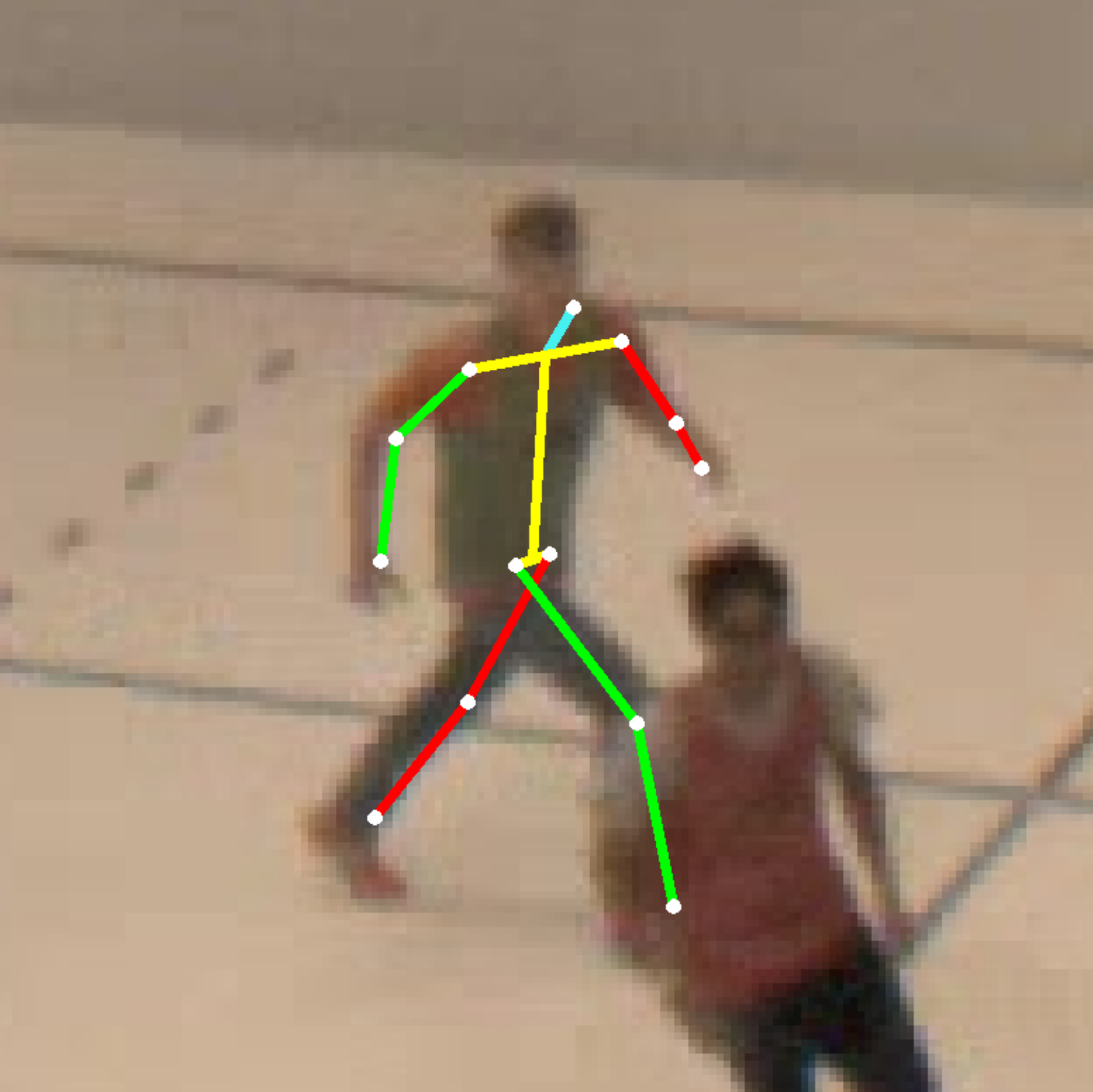}
    \end{subfigure}
    \begin{subfigure}[b]{0.24\linewidth}        
        \centering
        \includegraphics[width=\linewidth]{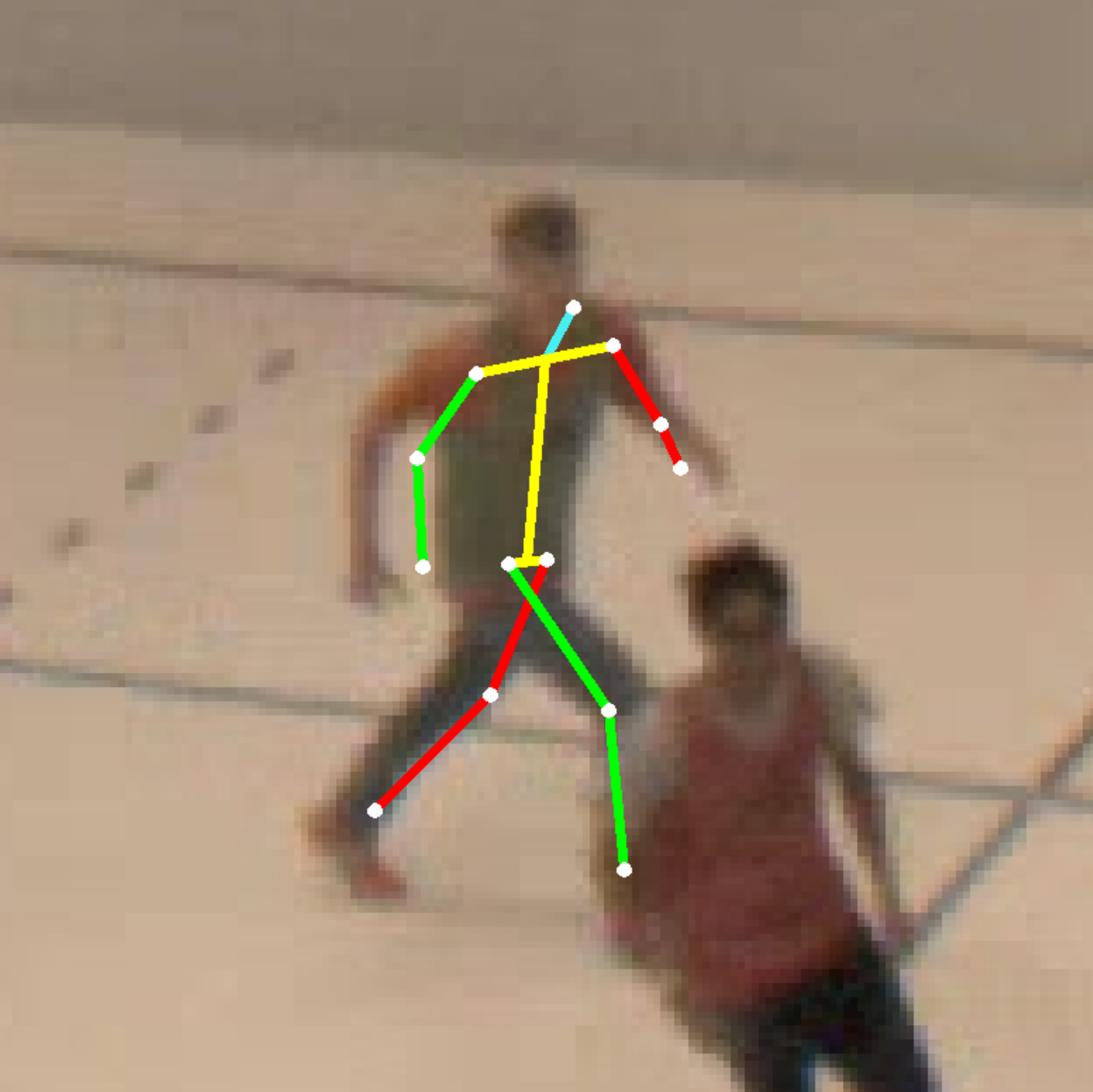}

    \end{subfigure} \\   \vspace{1mm}

    \begin{subfigure}[b]{0.24\linewidth}        
        \centering
        \includegraphics[width=\linewidth]{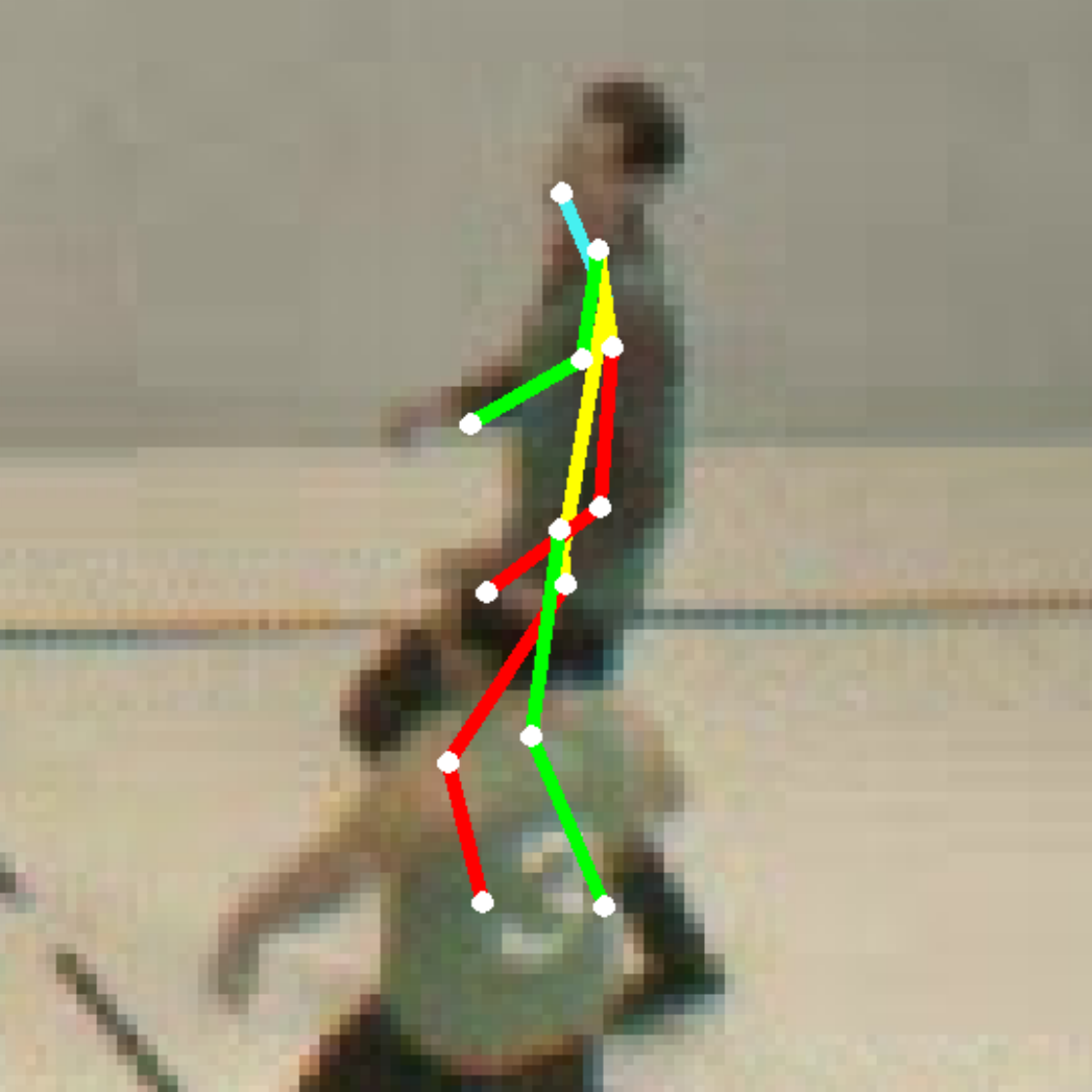}
    \end{subfigure}
    \begin{subfigure}[b]{0.24\linewidth}        
        \centering
        \includegraphics[width=\linewidth]{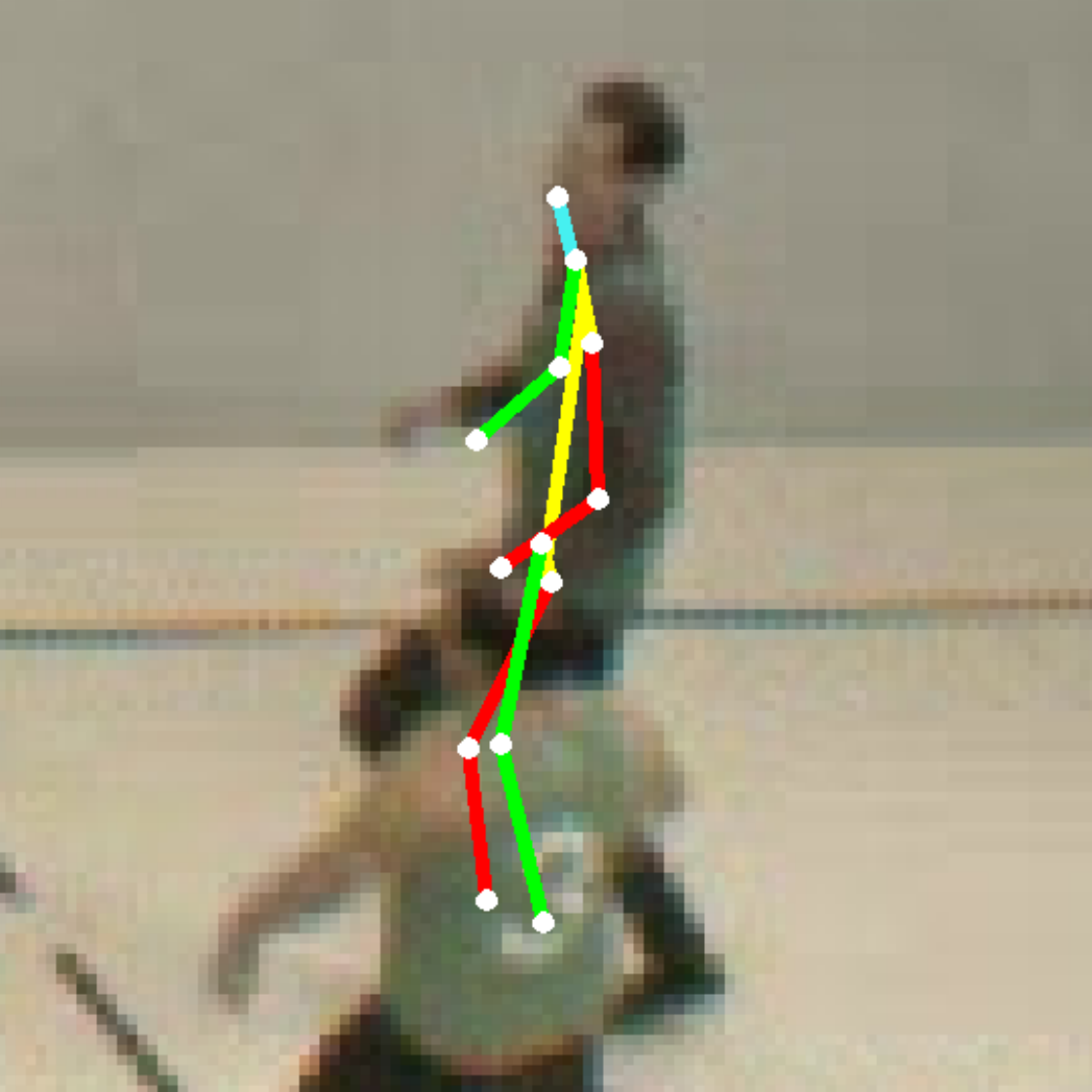}
    \end{subfigure}
    \begin{subfigure}[b]{0.24\linewidth}        
        \centering
        \includegraphics[width=\linewidth]{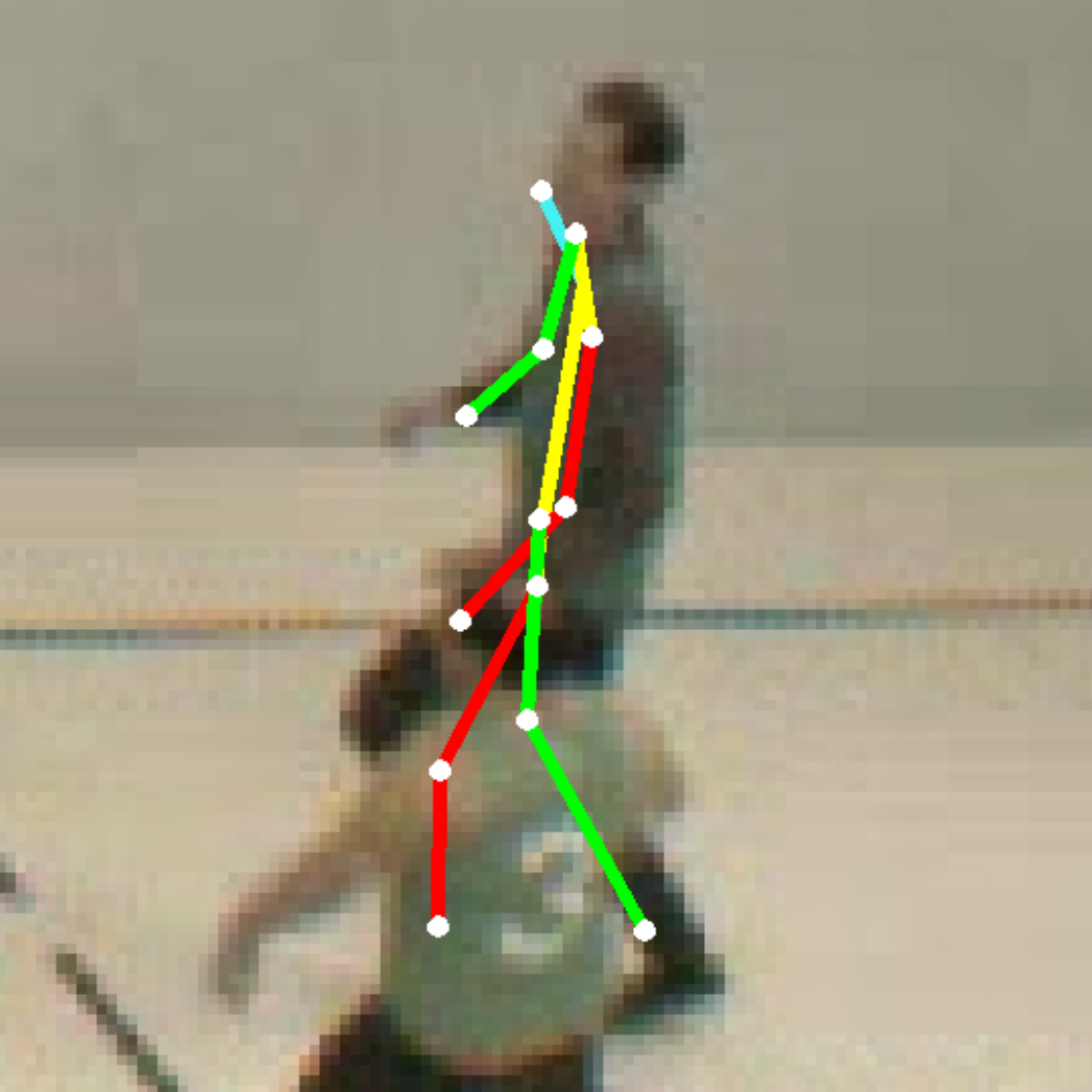}
    \end{subfigure}
    \begin{subfigure}[b]{0.24\linewidth}        
        \centering
        \includegraphics[width=\linewidth]{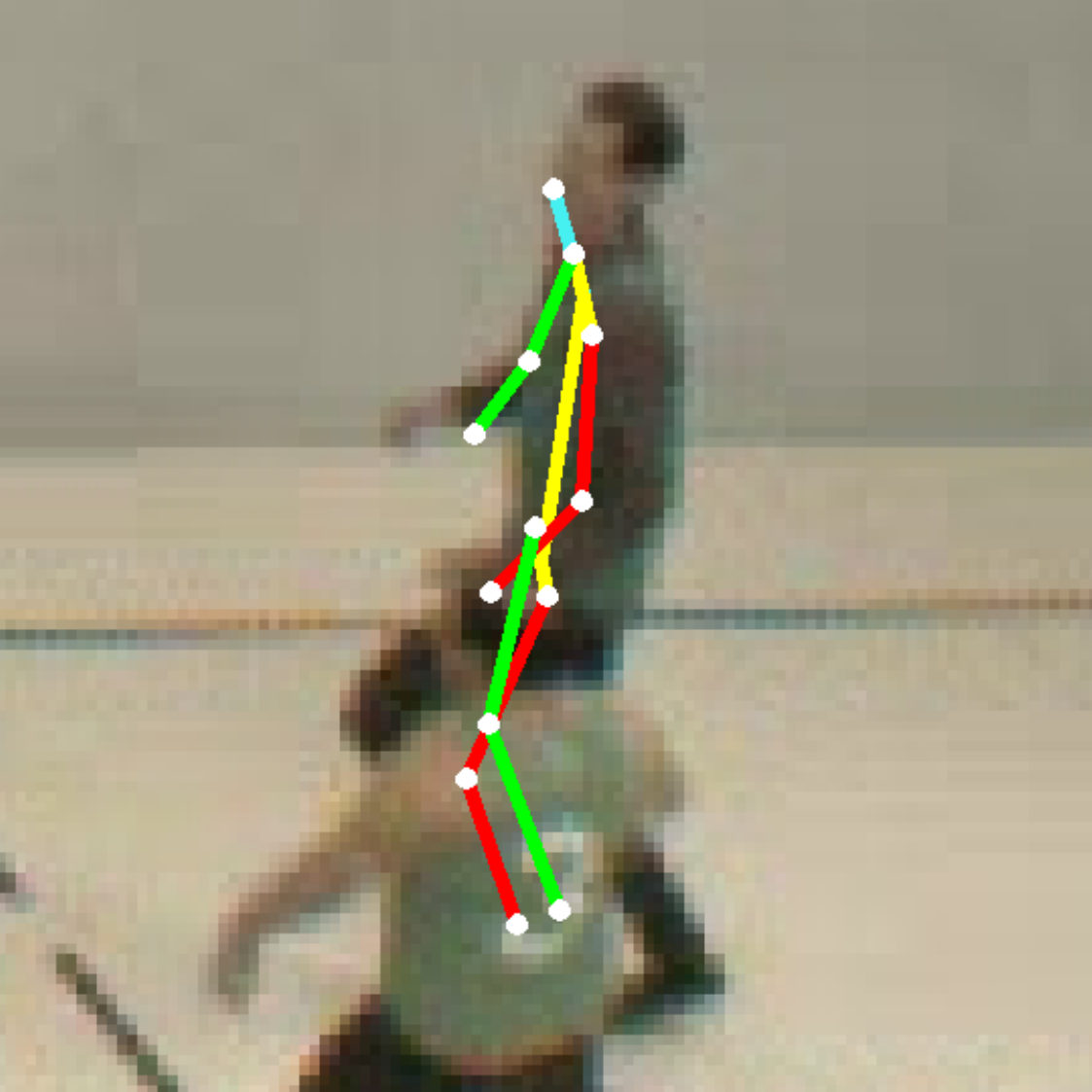}
    \end{subfigure} \\ \vspace{1mm}

    \begin{subfigure}[b]{0.24\linewidth}        
        \centering
        \includegraphics[width=\linewidth]{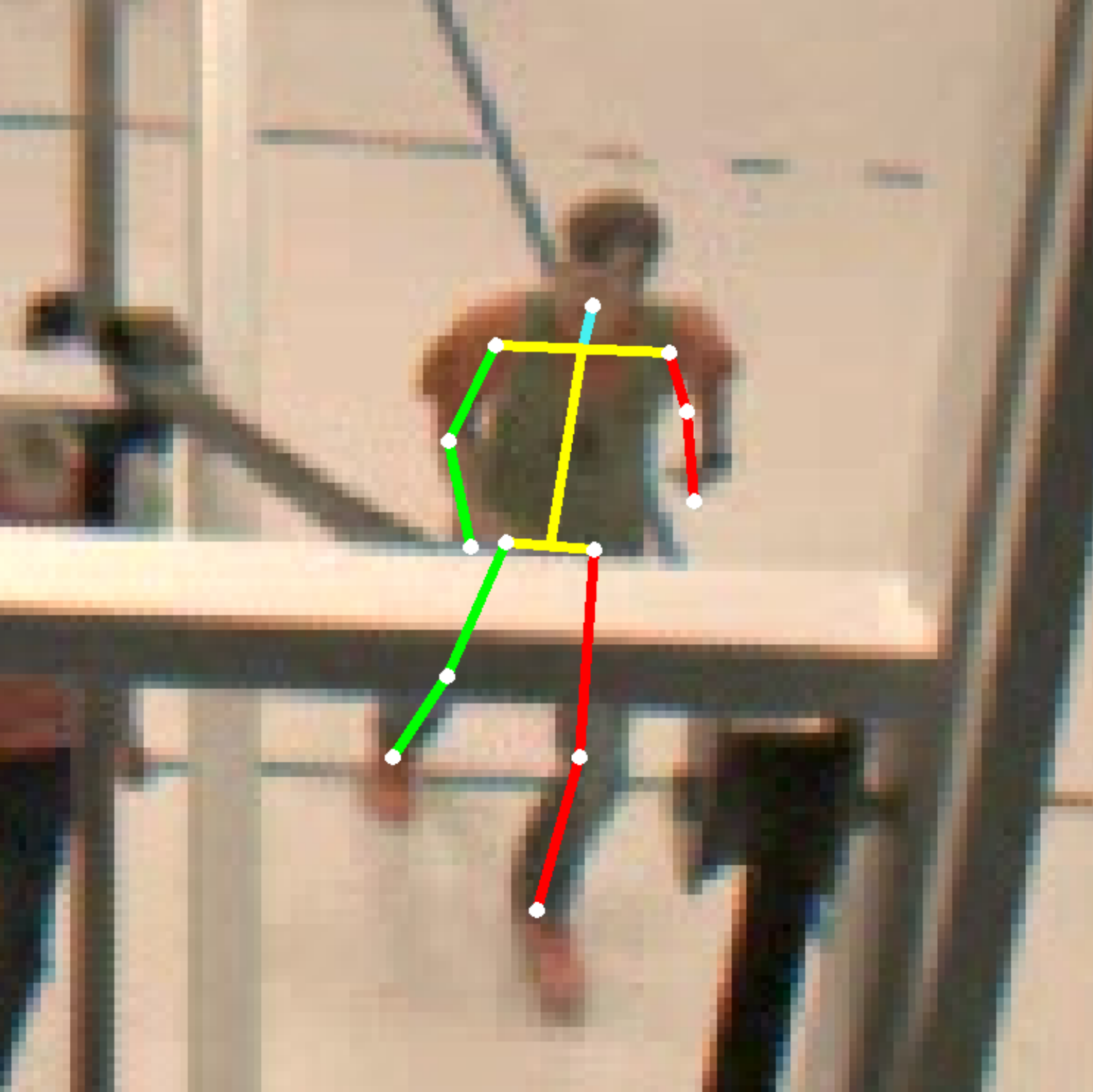}
    \end{subfigure}
    \begin{subfigure}[b]{0.24\linewidth}        
        \centering
        \includegraphics[width=\linewidth]{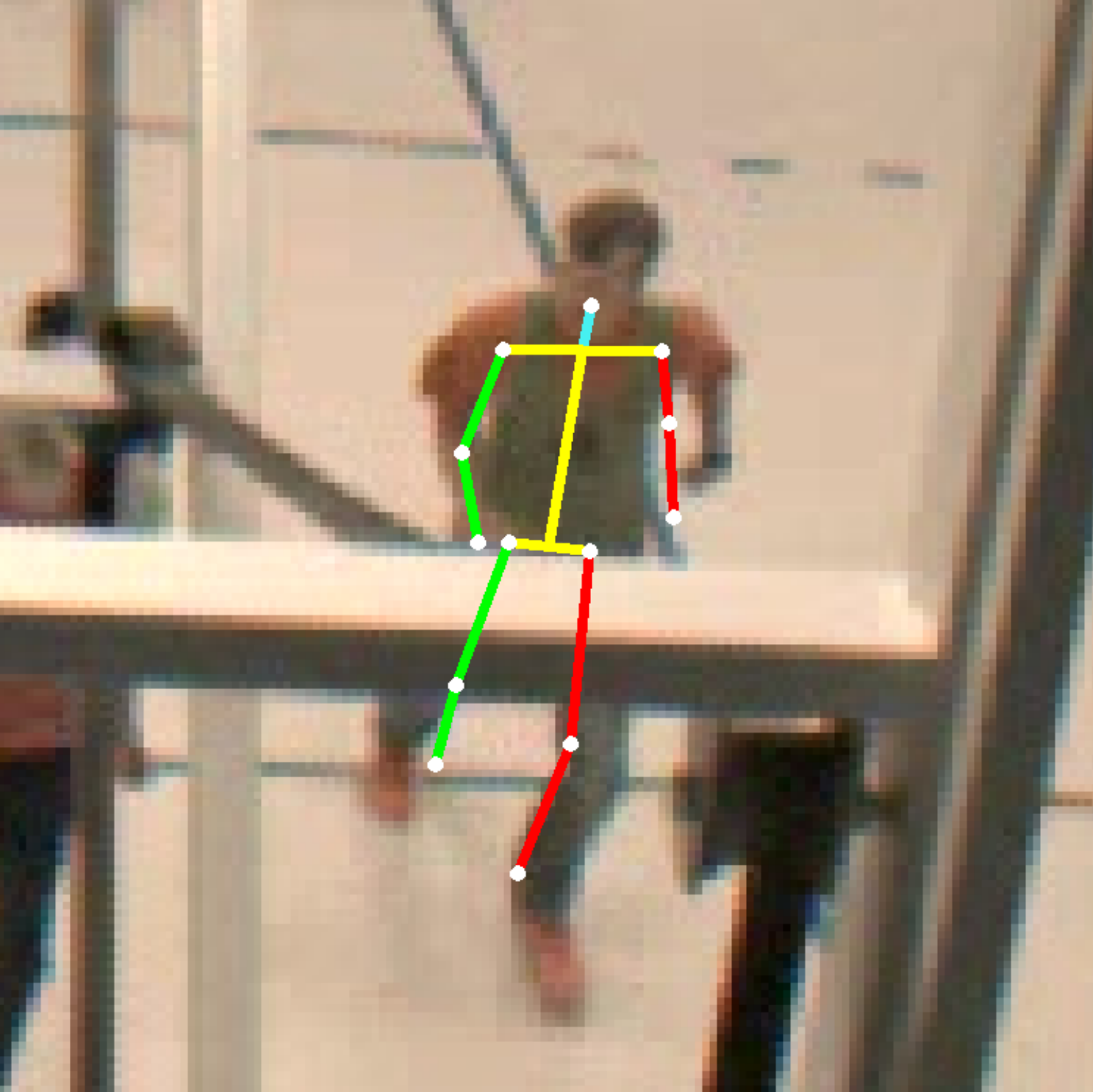}
    \end{subfigure}
    \begin{subfigure}[b]{0.24\linewidth}        
        \centering
        \includegraphics[width=\linewidth]{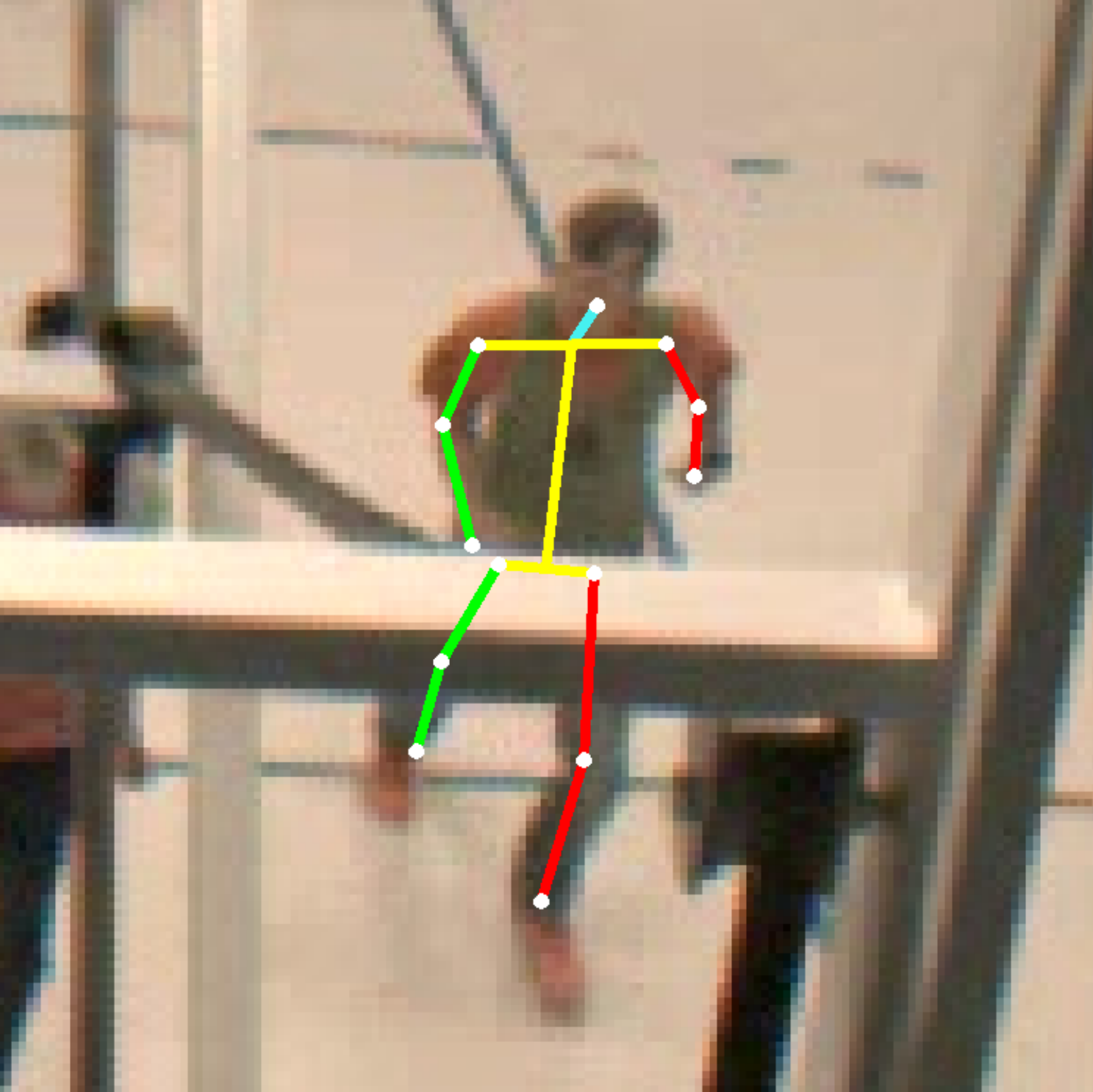}
    \end{subfigure}
    \begin{subfigure}[b]{0.24\linewidth}        
        \centering
        \includegraphics[width=\linewidth]{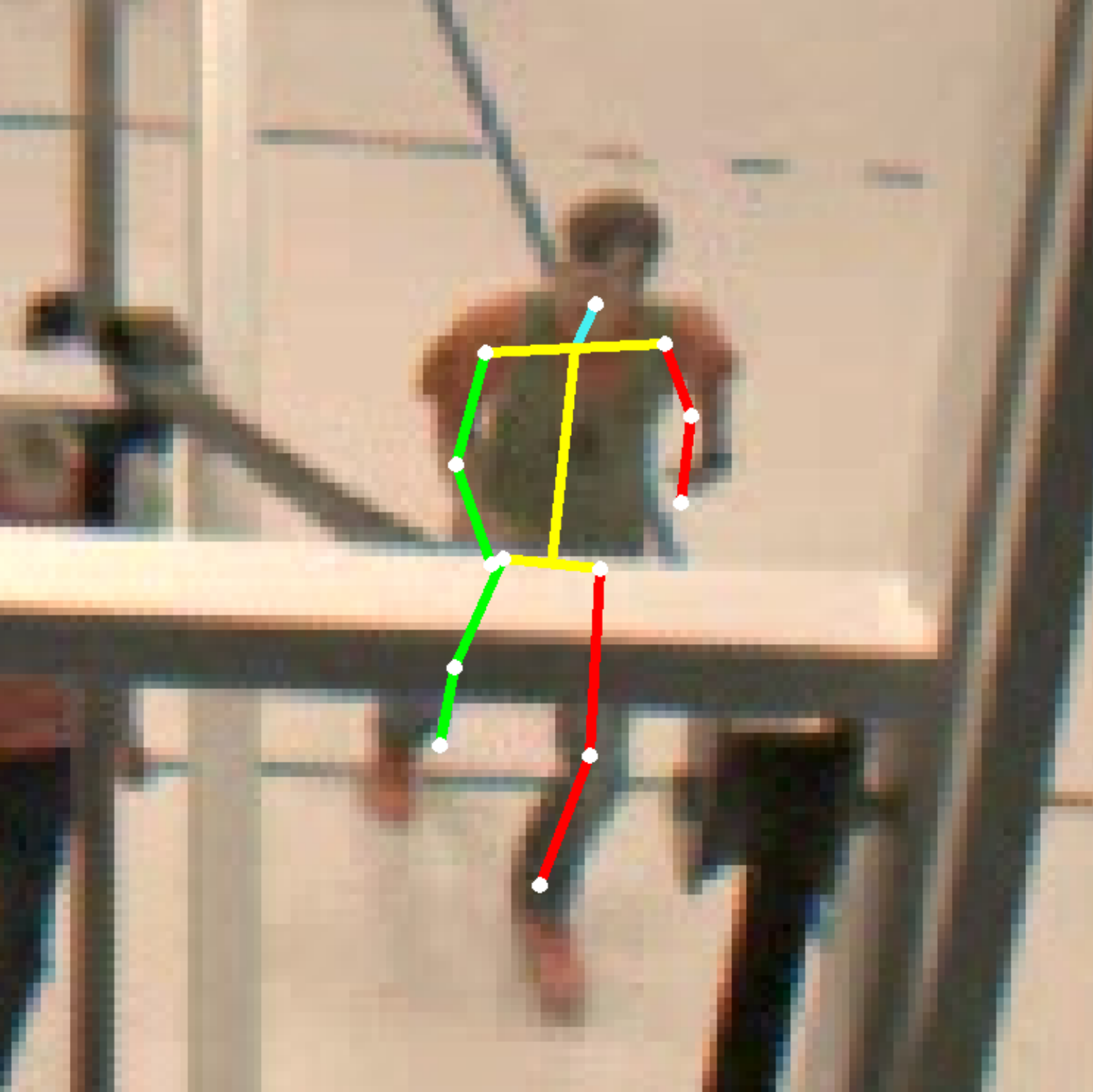}
    \end{subfigure} \\ \vspace{1mm}
    
    \begin{subfigure}[b]{0.24\linewidth}        
        \centering
        \includegraphics[width=\linewidth]{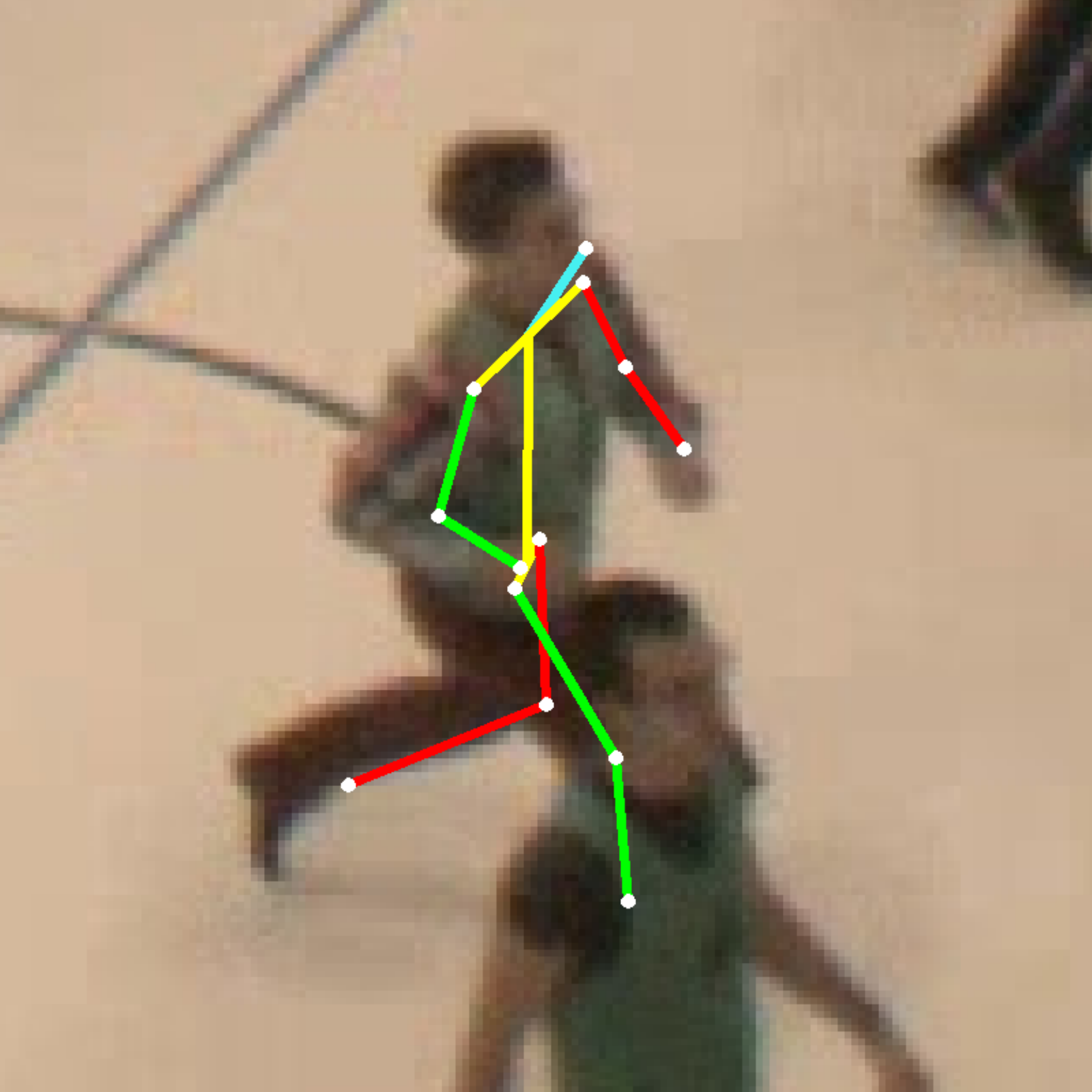}
        \caption{Ours}
    \end{subfigure}
    \begin{subfigure}[b]{0.24\linewidth}        
        \centering
        \includegraphics[width=\linewidth]{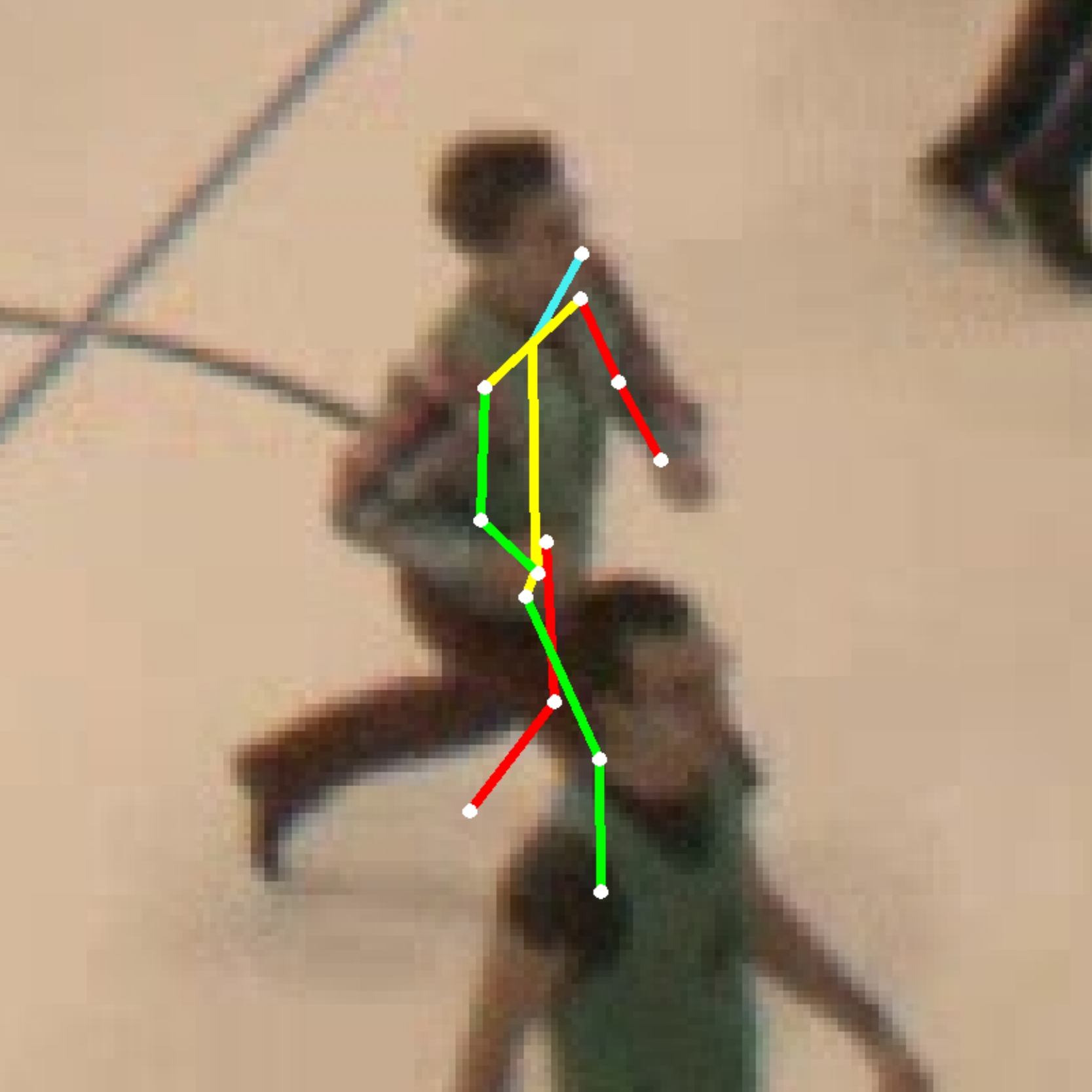}
        \caption{w/o w}
    \end{subfigure}
    \begin{subfigure}[b]{0.24\linewidth}        
        \centering
        \includegraphics[width=\linewidth]{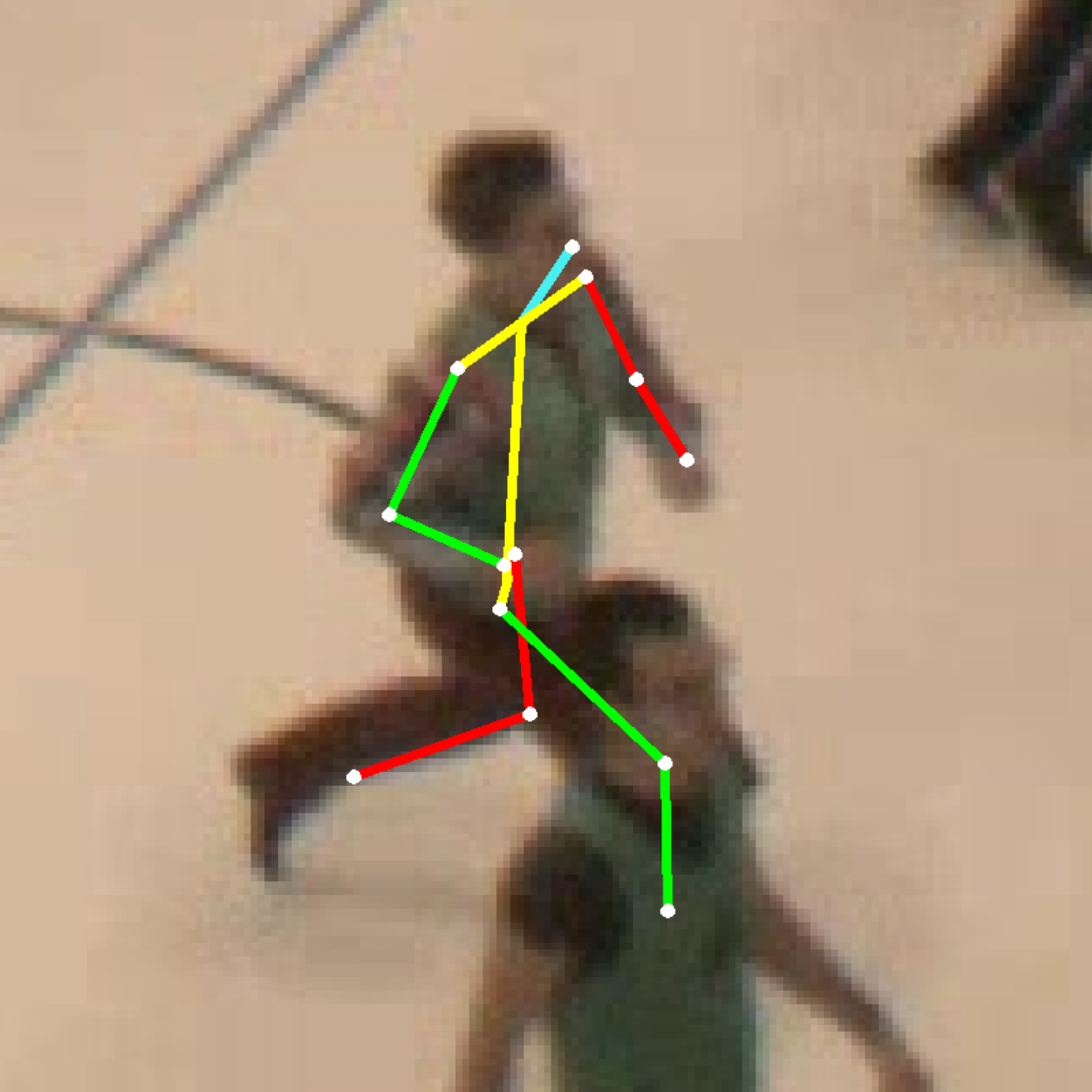}
        \caption{Ours SV}
    \end{subfigure}
    \begin{subfigure}[b]{0.24\linewidth}        
        \centering
        \includegraphics[width=\linewidth]{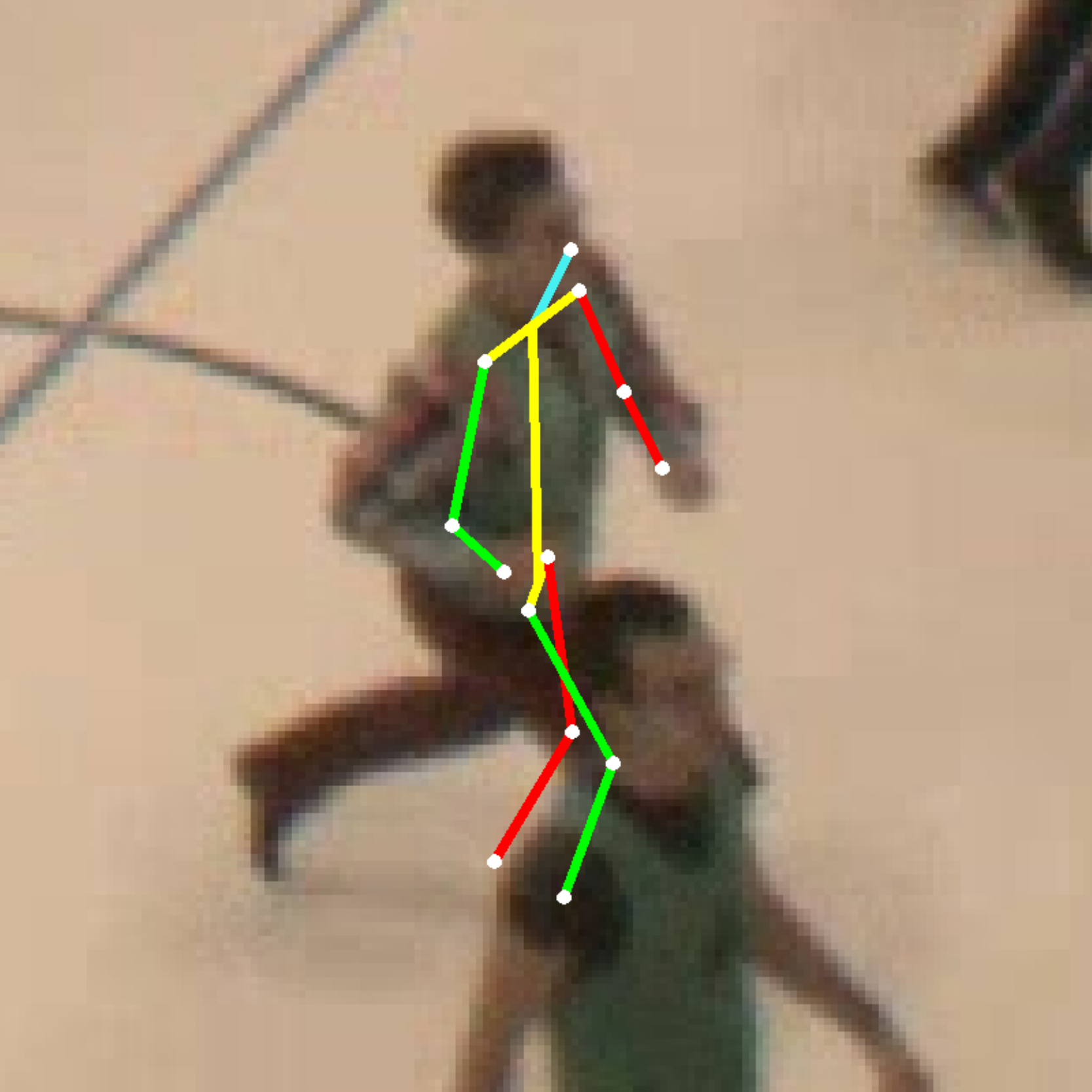}
        \caption{\small w/o w. SV}
    \end{subfigure} \\    
          
    \caption{\small Qualitative results on the SportCenter dataset. From left to right, multi-view triangulated pose with (a) our approach and (b) Standard DLT (without weighting mechanism). Single view predicted results of (c) our approach and (d) without weighting. It can be noted that our weighting strategy produces more robust 3D poses which provide better supervision on occluded samples.}
    \label{fig:occlusion_images}
\end{figure*}

\begin{figure*}
	\centering
	
	\begin{subfigure}[b]{0.23\linewidth}        
		\centering
		\includegraphics[width=\linewidth]{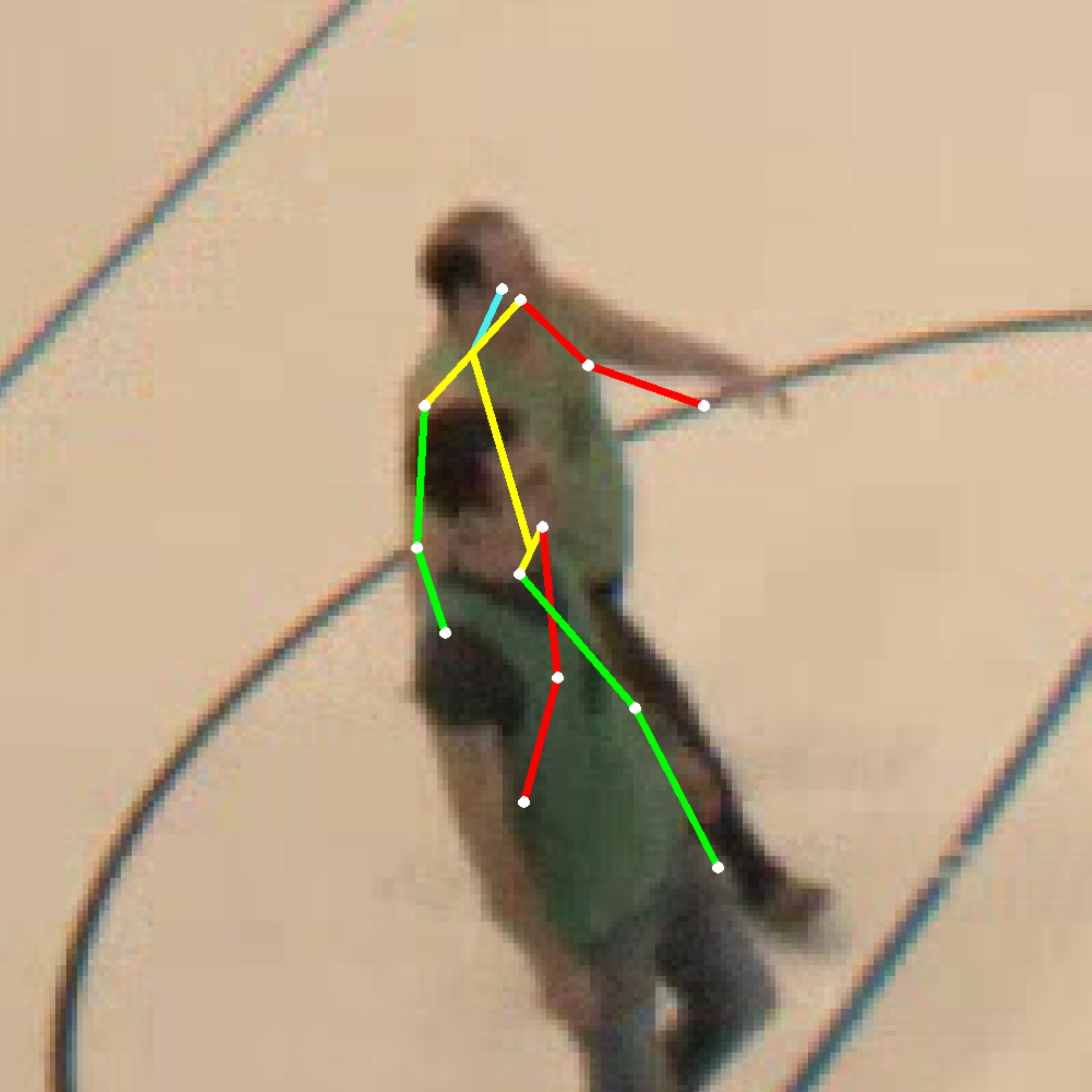}
	\end{subfigure}
	\begin{subfigure}[b]{0.23\linewidth}        
		\centering
		\includegraphics[width=\linewidth]{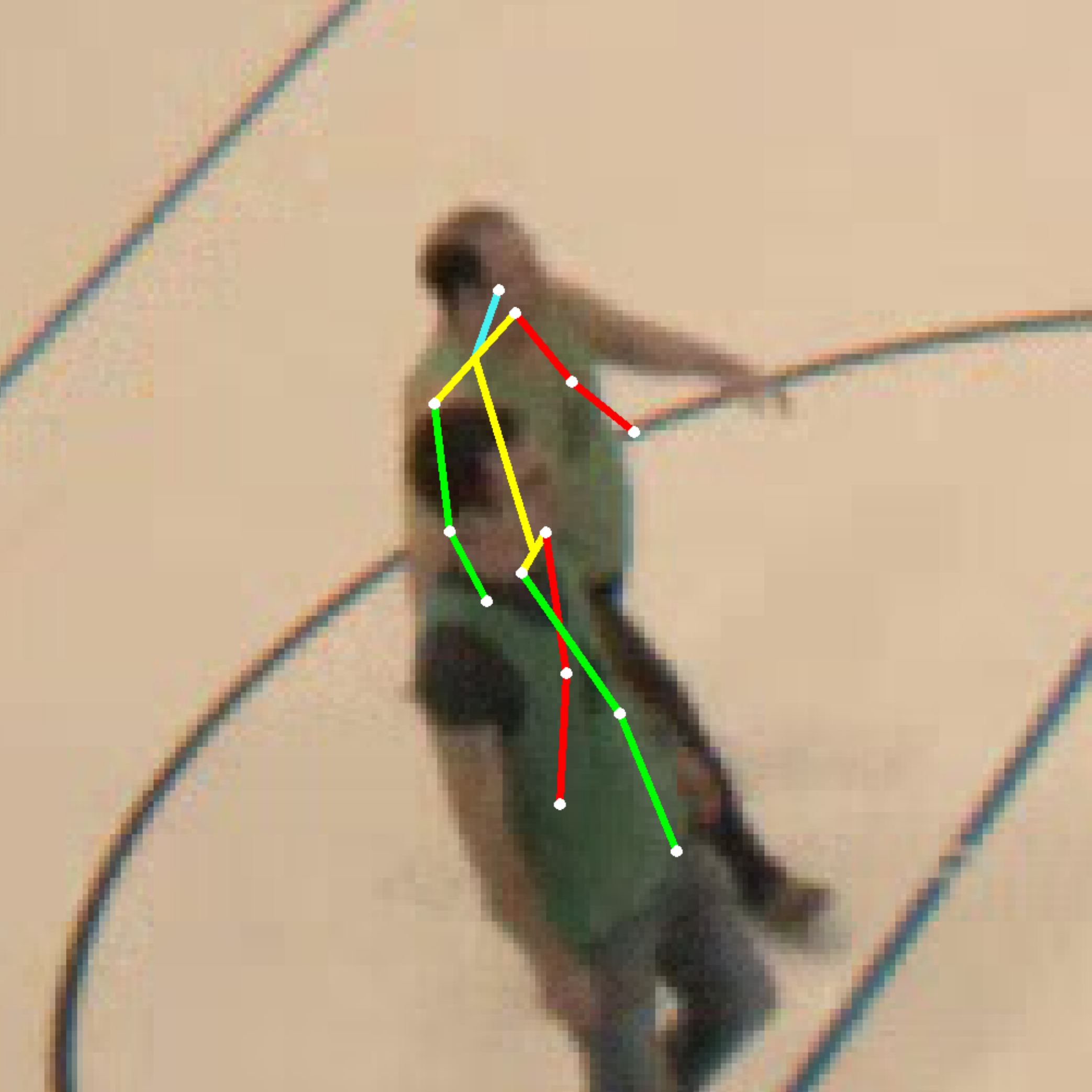}
	\end{subfigure}
	\begin{subfigure}[b]{0.23\linewidth}        
		\centering
		\includegraphics[width=\linewidth]{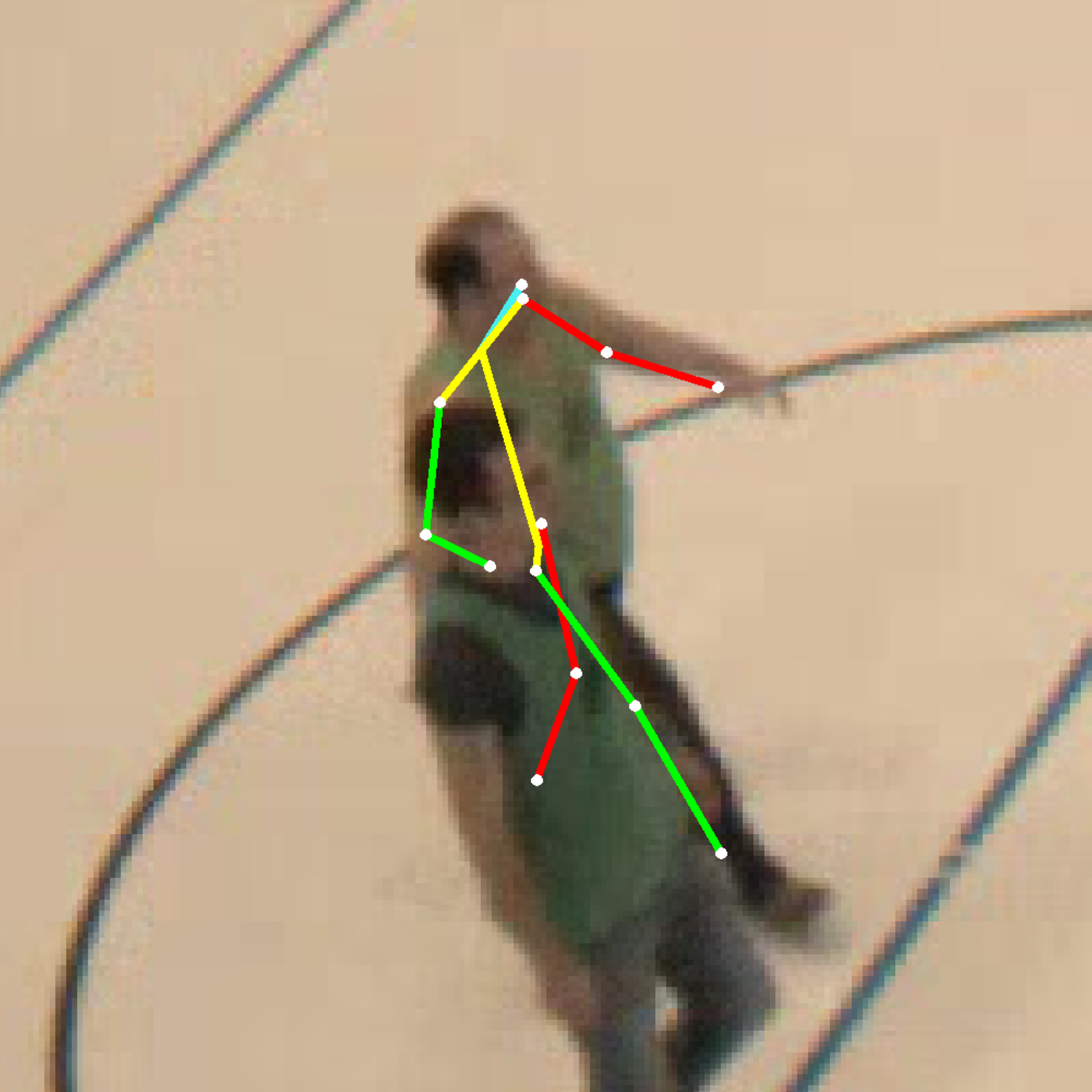}
	\end{subfigure}
	\begin{subfigure}[b]{0.23\linewidth}        
		\centering
		\includegraphics[width=\linewidth]{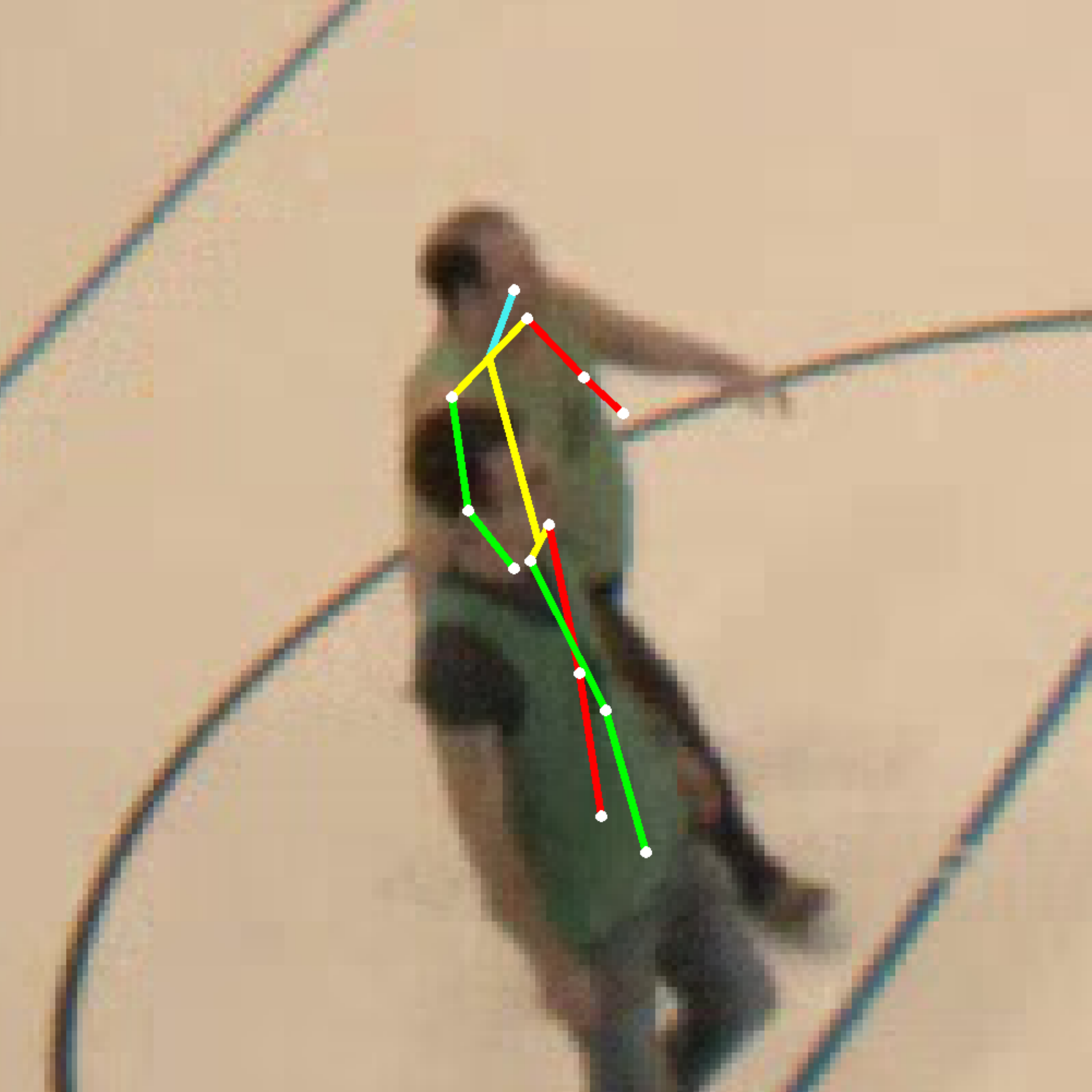}
	\end{subfigure} \\   \vspace{1mm}

	\begin{subfigure}[b]{0.23\linewidth}        
		\centering
		\includegraphics[width=\linewidth]{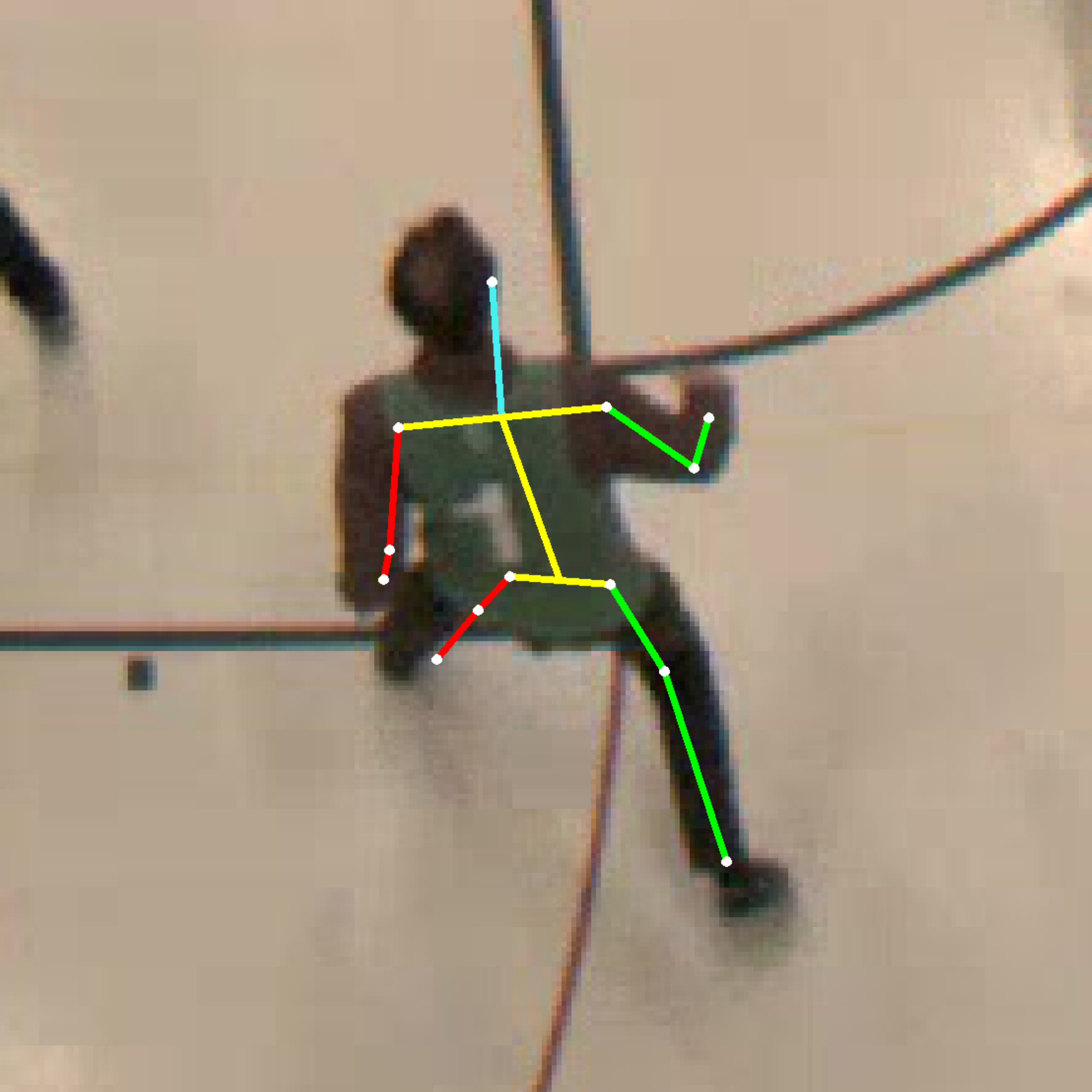}
	\end{subfigure}
	\begin{subfigure}[b]{0.23\linewidth}        
		\centering
		\includegraphics[width=\linewidth]{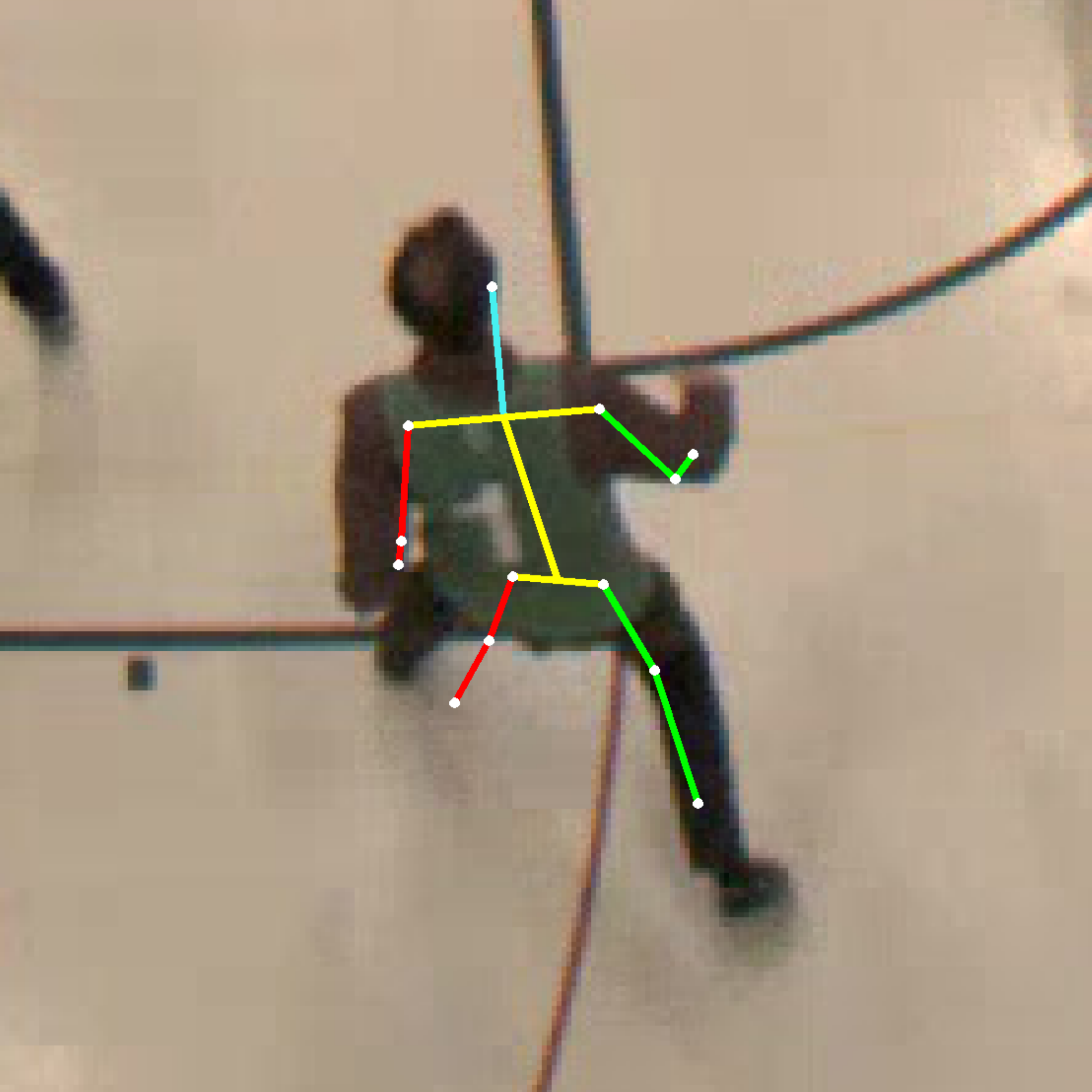}
	\end{subfigure}
	\begin{subfigure}[b]{0.23\linewidth}        
		\centering
		\includegraphics[width=\linewidth]{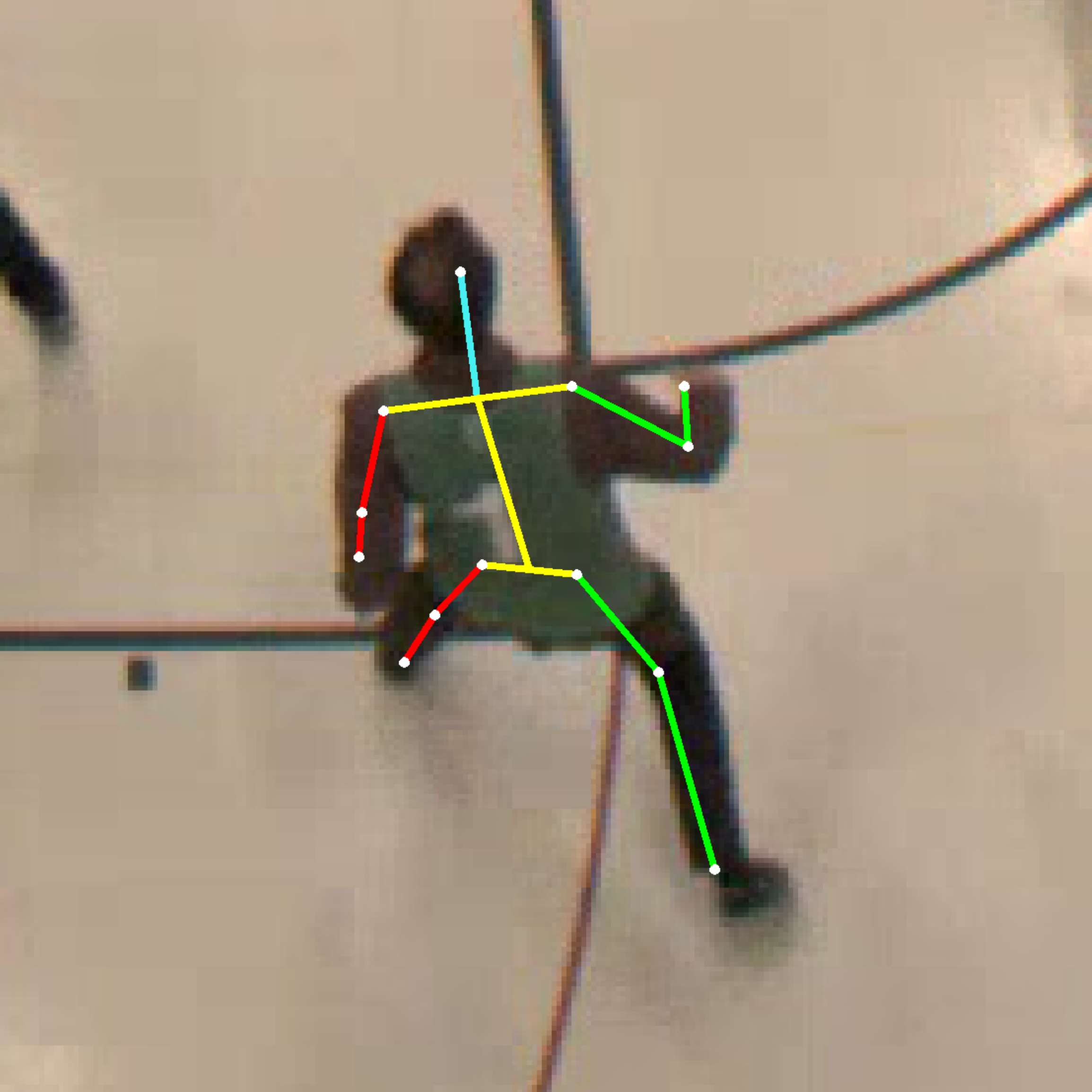}
	\end{subfigure}
	\begin{subfigure}[b]{0.23\linewidth}        
		\centering
		\includegraphics[width=\linewidth]{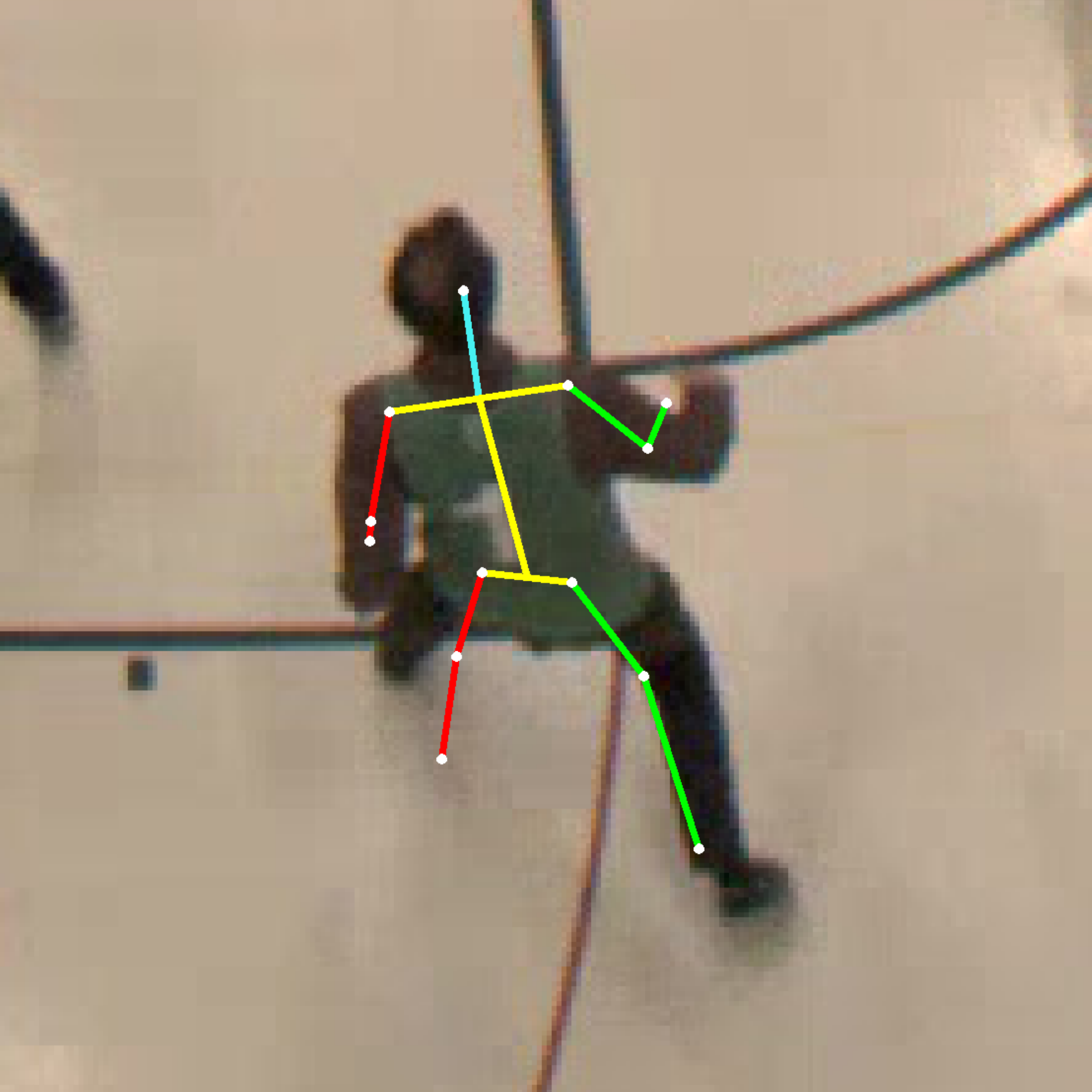}
	\end{subfigure} \\ \vspace{1mm}

	\begin{subfigure}[b]{0.23\linewidth}        
		\centering
		\includegraphics[width=\linewidth]{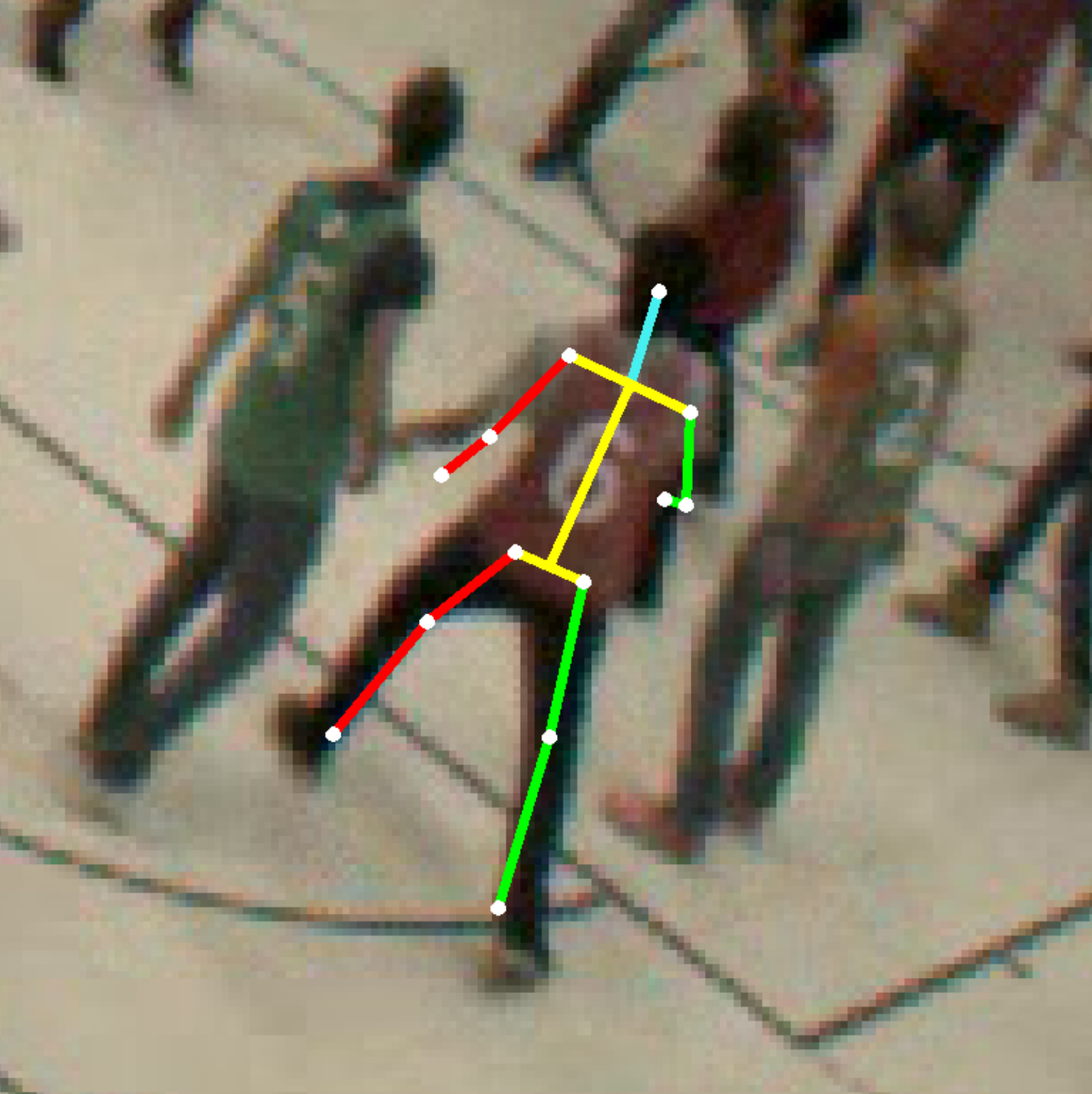}
	\end{subfigure}
	\begin{subfigure}[b]{0.23\linewidth}        
		\centering
		\includegraphics[width=\linewidth]{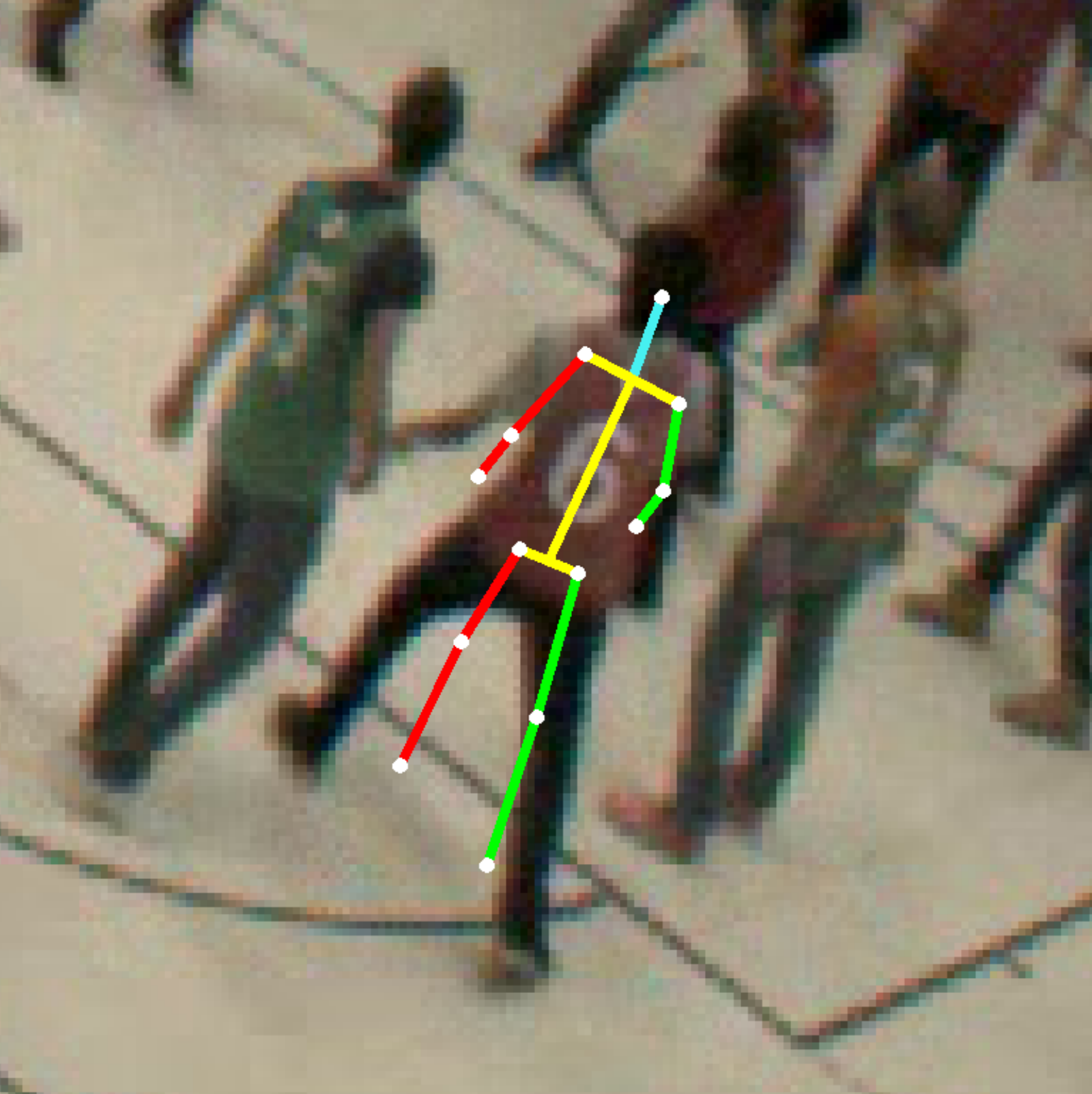}
	\end{subfigure}
	\begin{subfigure}[b]{0.23\linewidth}        
		\centering
		\includegraphics[width=\linewidth]{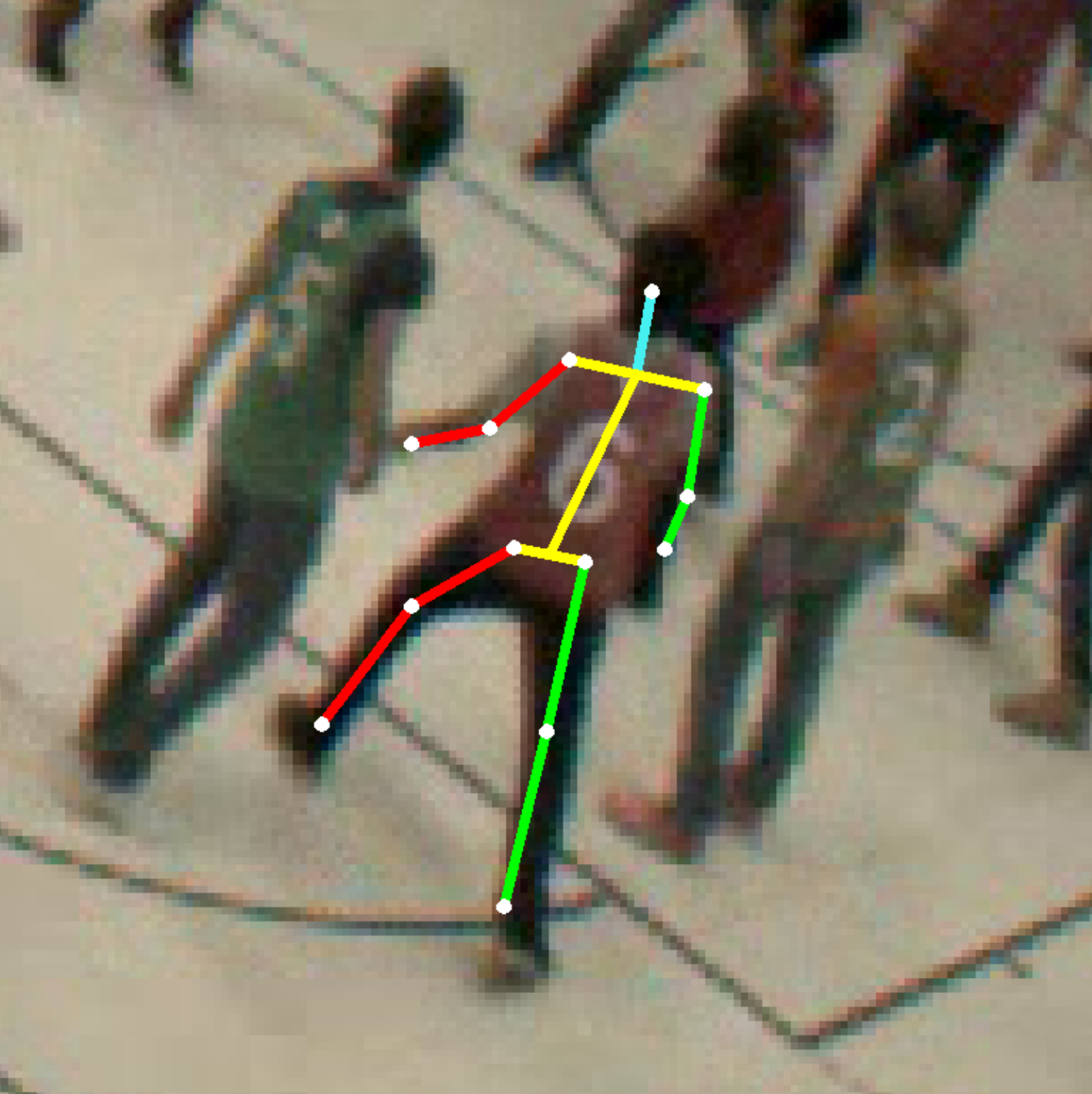}
	\end{subfigure}
	\begin{subfigure}[b]{0.23\linewidth}        
		\centering
		\includegraphics[width=\linewidth]{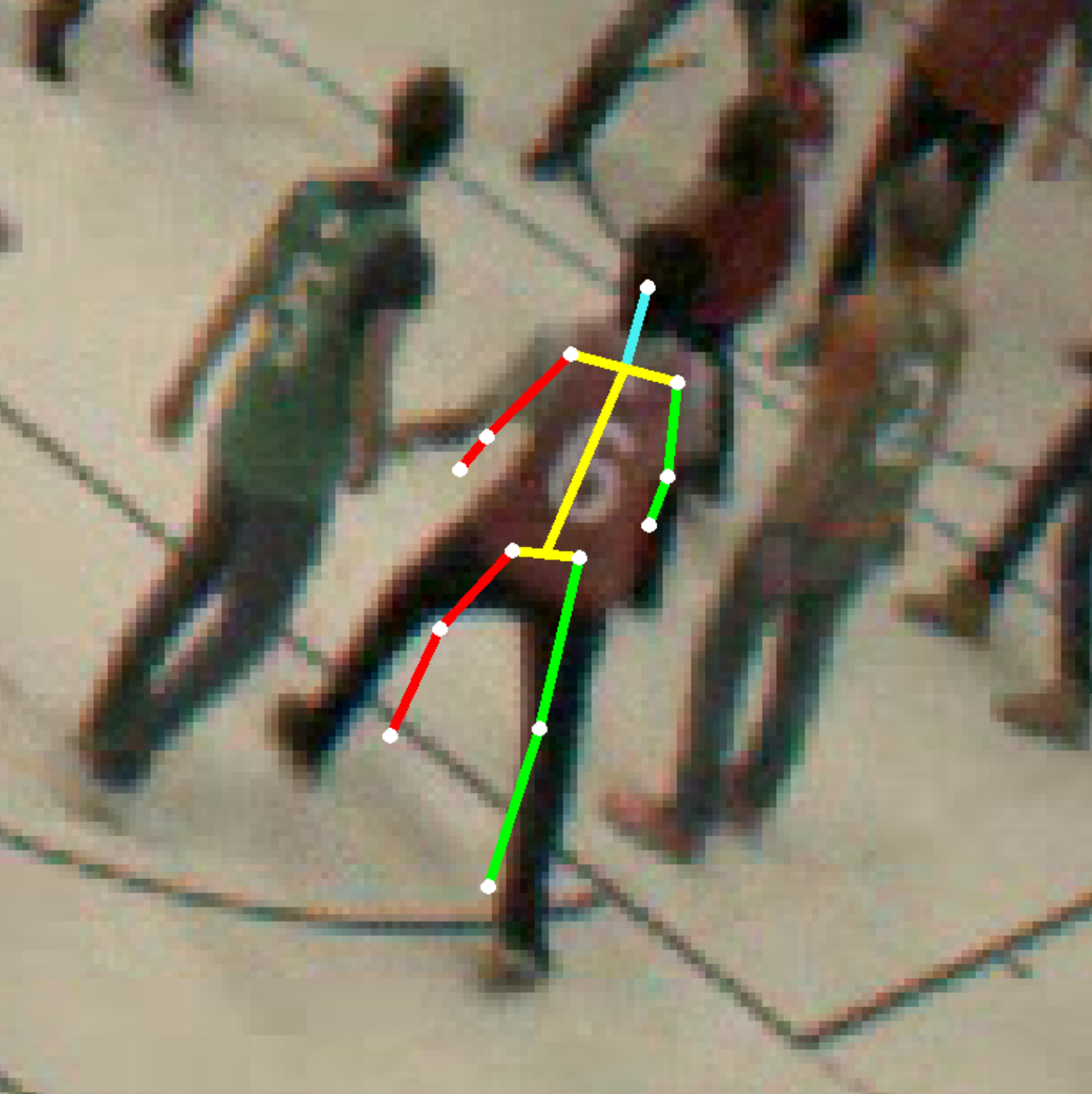}
	\end{subfigure} \\ \vspace{1mm}

	\begin{subfigure}[b]{0.23\linewidth}        
		\centering
		\includegraphics[width=\linewidth]{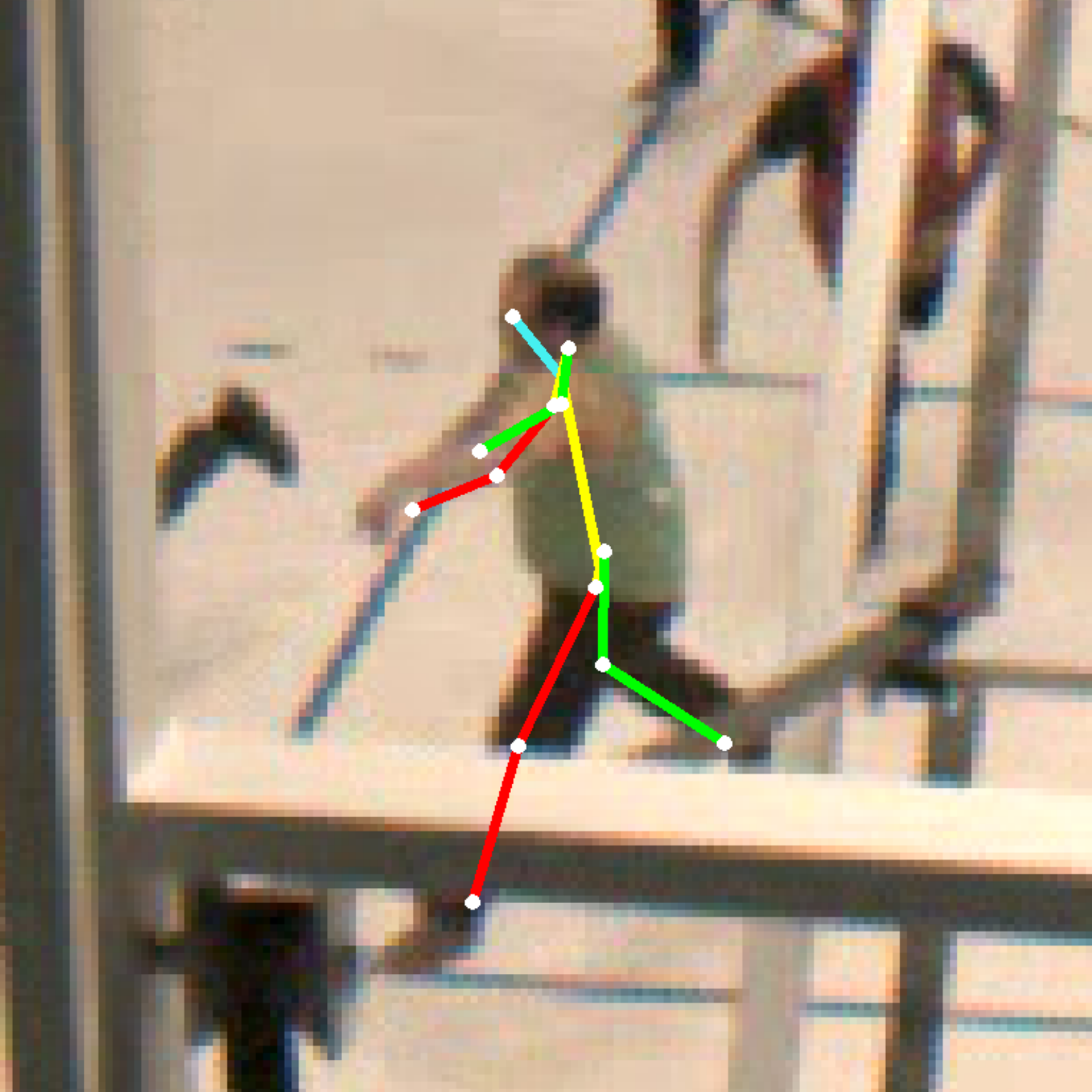}
	\end{subfigure}
	\begin{subfigure}[b]{0.23\linewidth}        
		\centering
		\includegraphics[width=\linewidth]{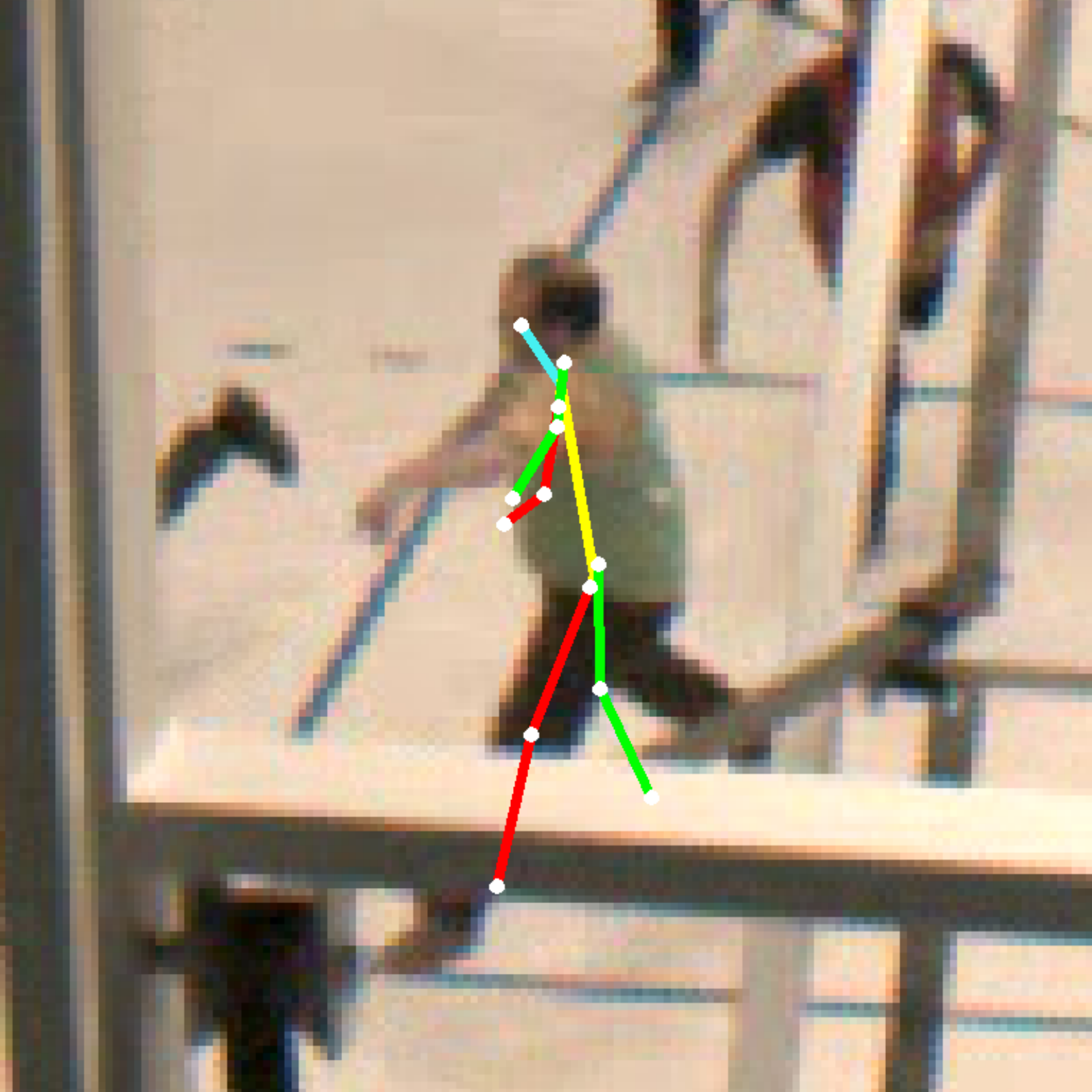}
	\end{subfigure}
	\begin{subfigure}[b]{0.23\linewidth}        
		\centering
		\includegraphics[width=\linewidth]{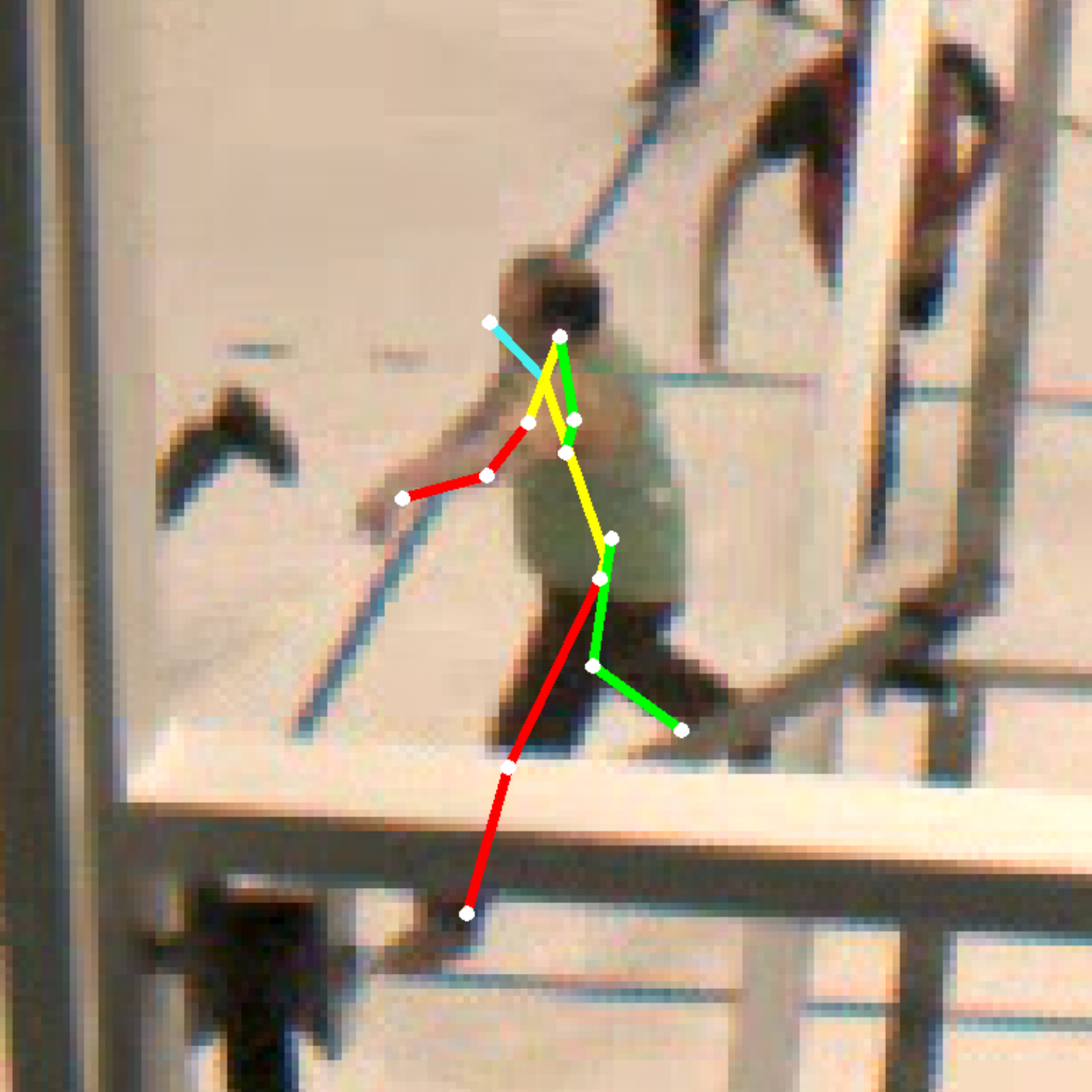}
	\end{subfigure}
	\begin{subfigure}[b]{0.23\linewidth}        
		\centering
		\includegraphics[width=\linewidth]{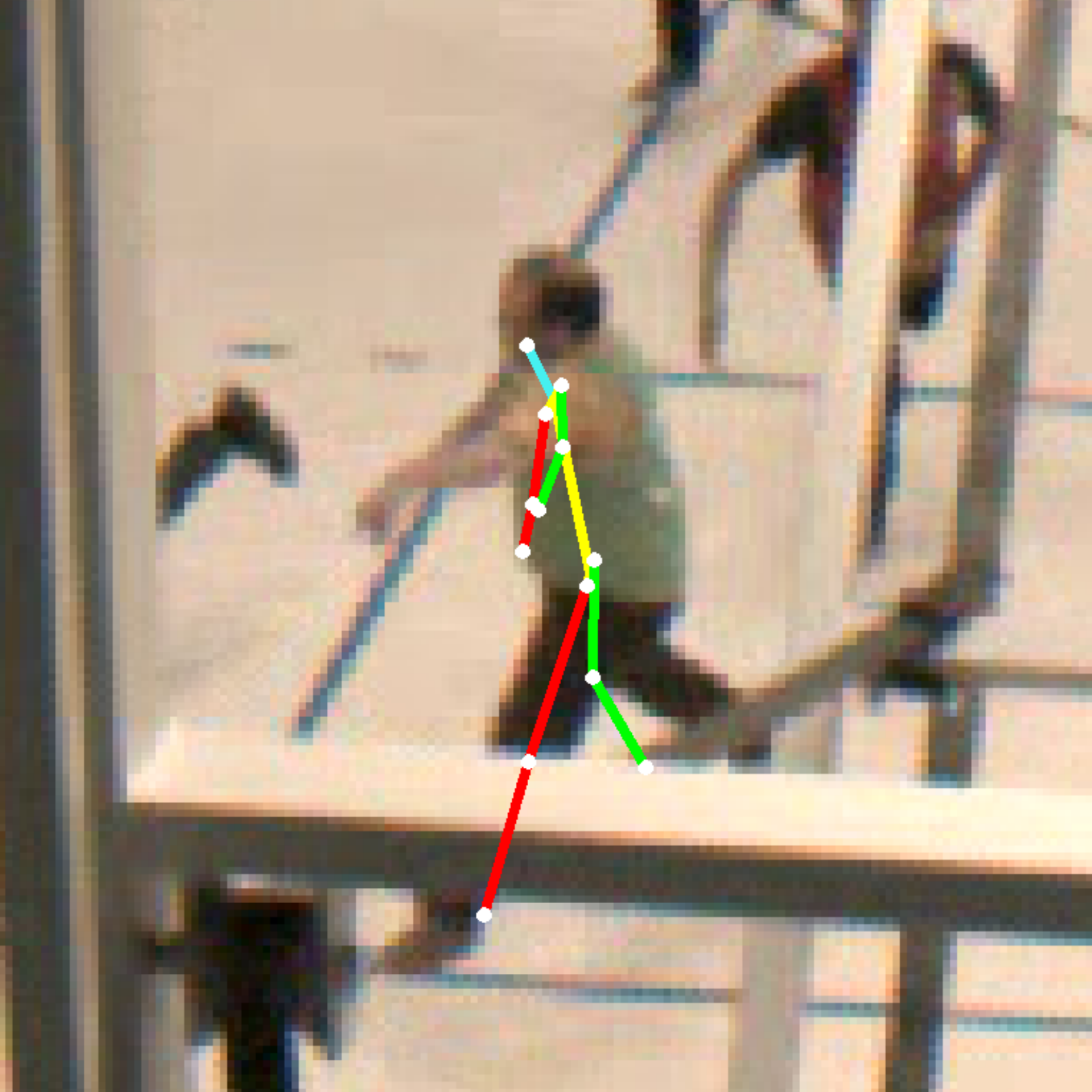}
	\end{subfigure} \\ \vspace{1mm}
	
	\begin{subfigure}[b]{0.23\linewidth}        
		\centering
		\includegraphics[width=\linewidth]{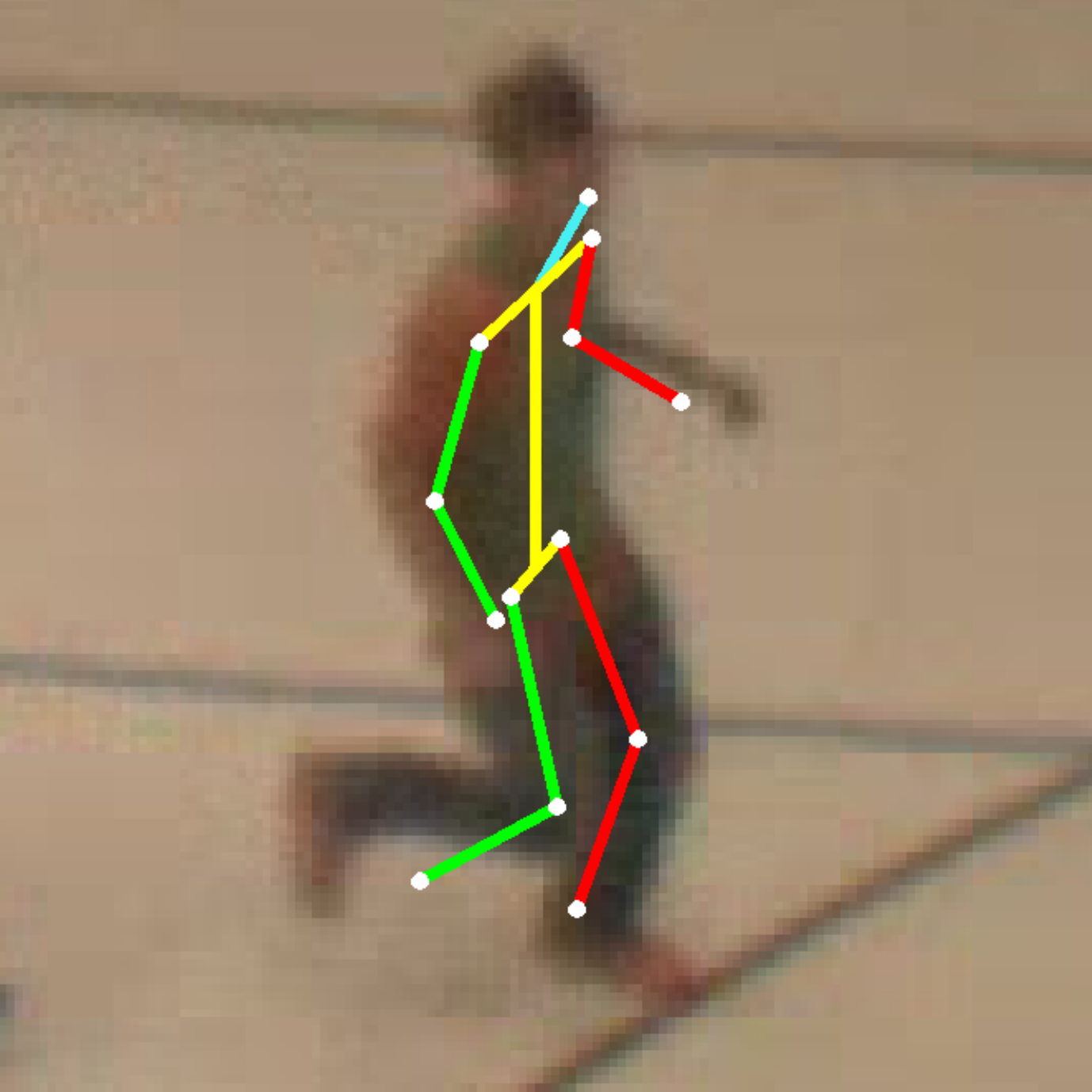}
		\caption{Ours}
	\end{subfigure}
	\begin{subfigure}[b]{0.23\linewidth}        
		\centering
		\includegraphics[width=\linewidth]{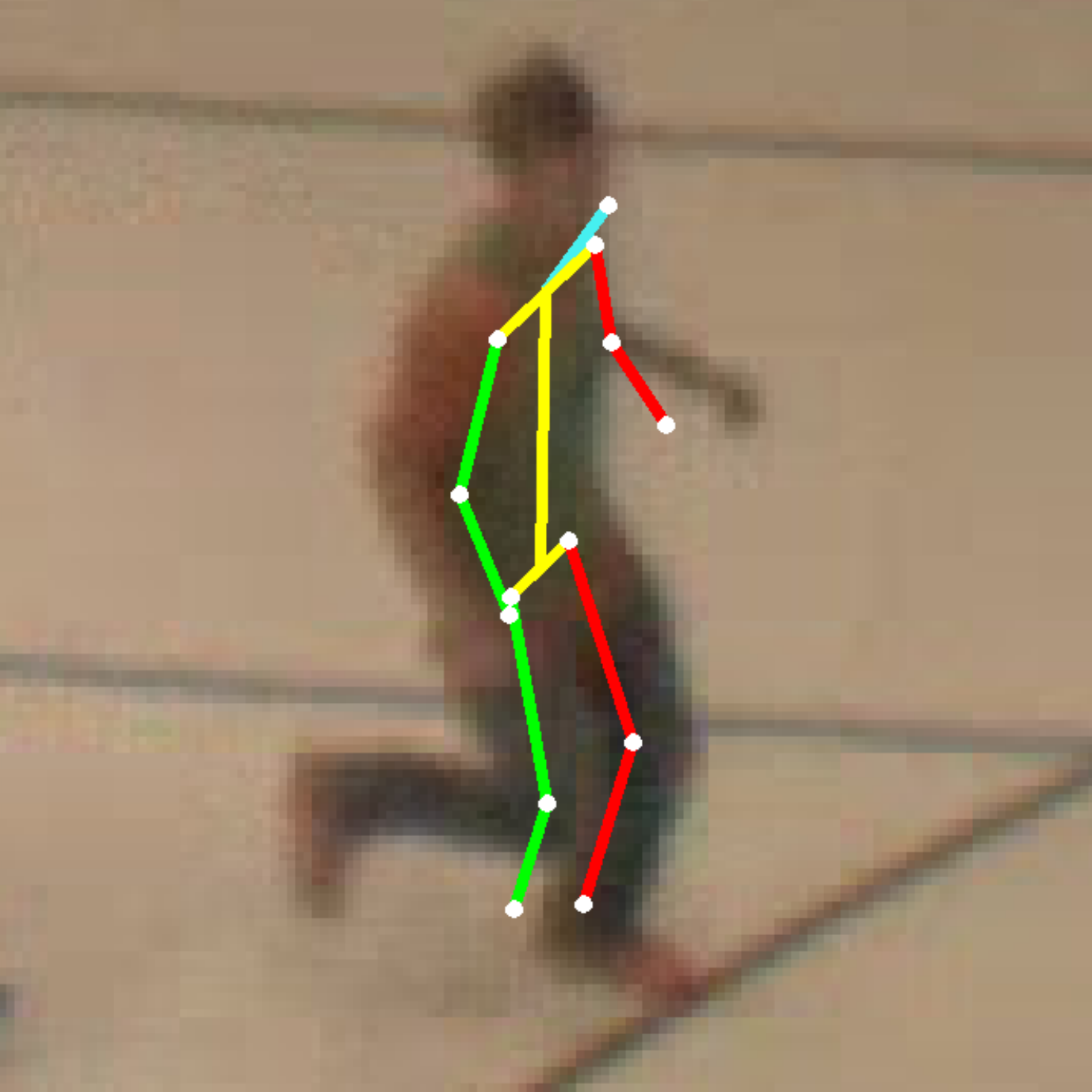}
		\caption{w/o w}
	\end{subfigure}
	\begin{subfigure}[b]{0.23\linewidth}        
		\centering
		\includegraphics[width=\linewidth]{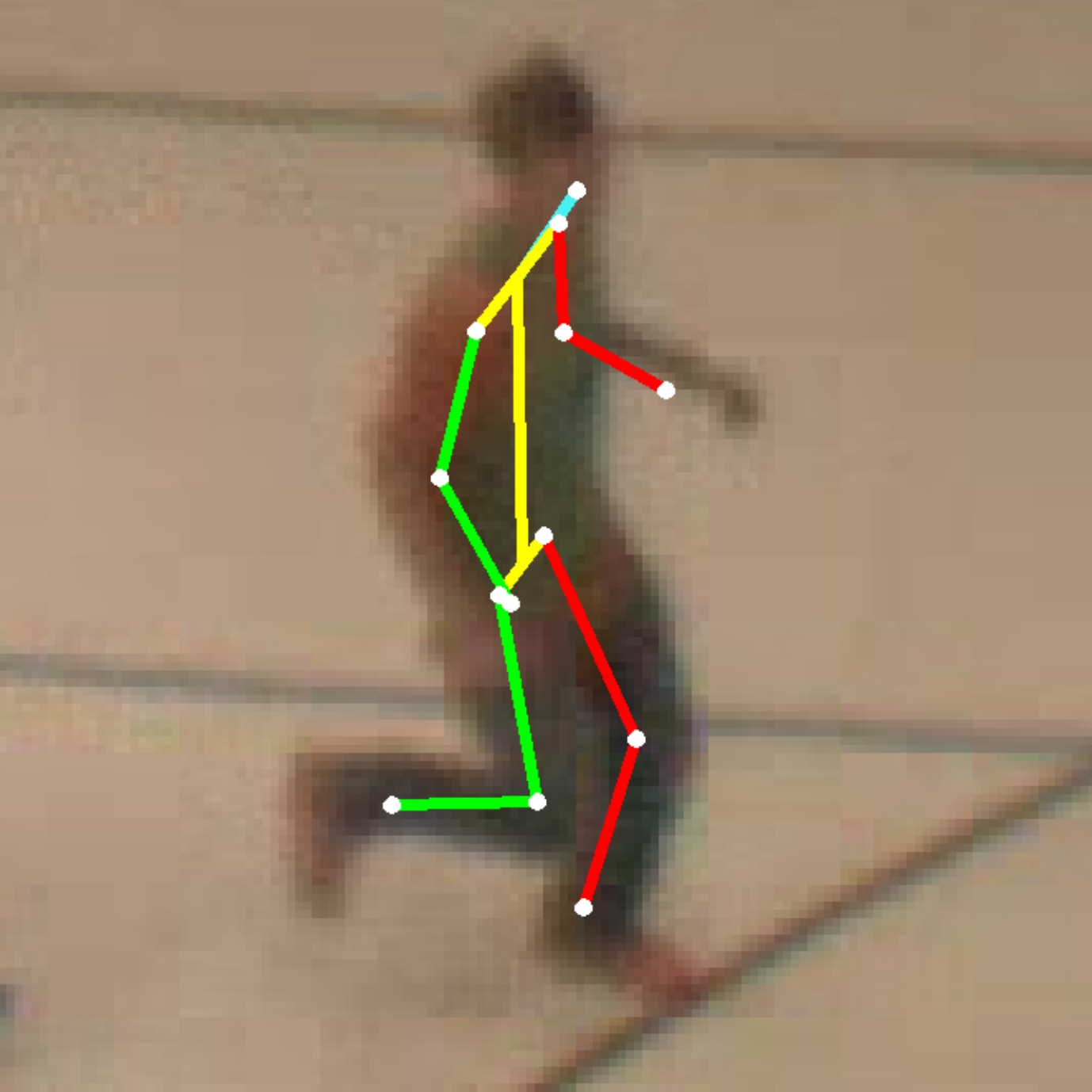}
		\caption{Ours SV}
	\end{subfigure}
	\begin{subfigure}[b]{0.23\linewidth}        
		\centering
		\includegraphics[width=\linewidth]{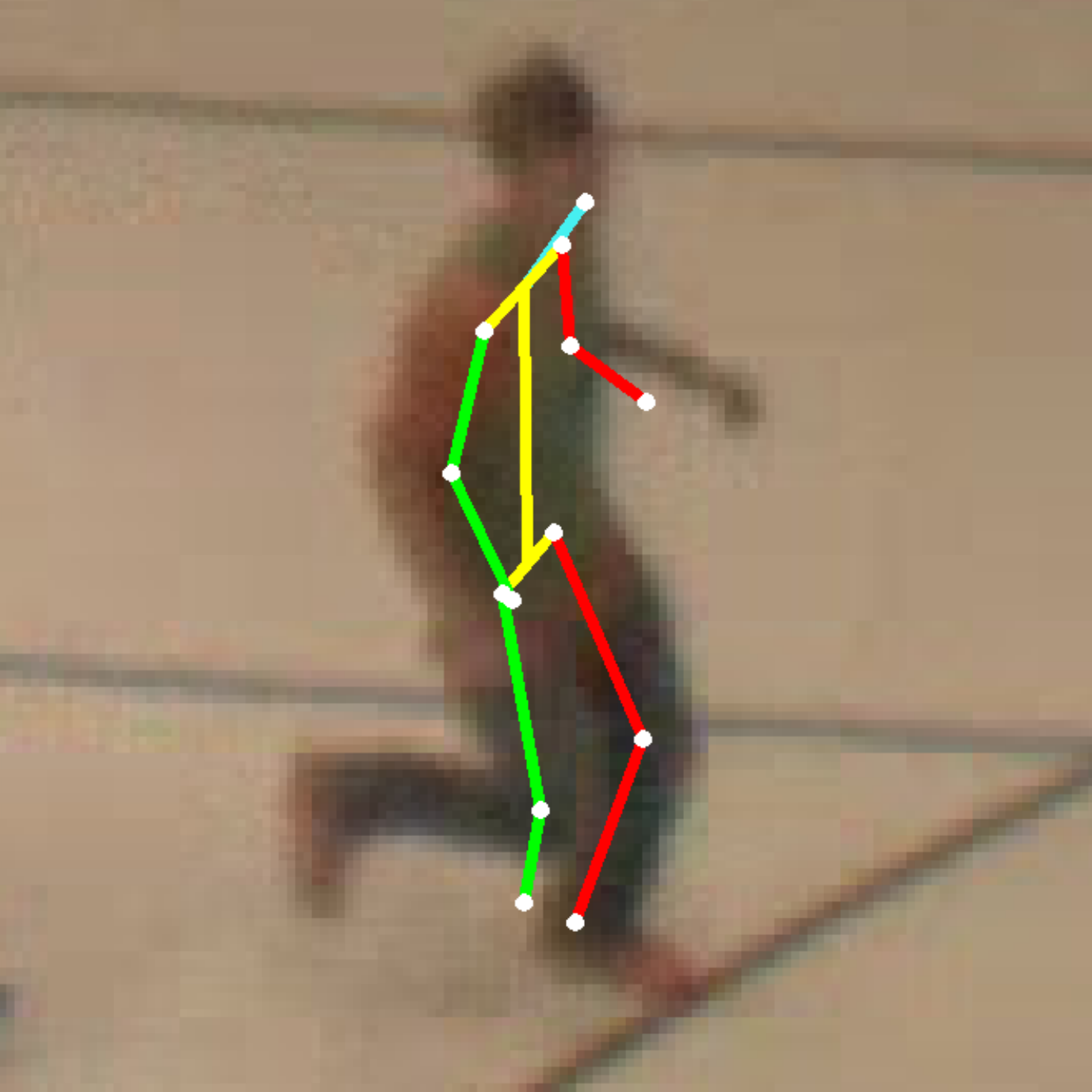}
		\caption{\small w/o w. SV}
	\end{subfigure}
	\vspace{-0.3cm}
	\caption{\small Qualitative results on the SportCenter dataset. From left to right, multi-view triangulated pose with (a) our approach and (b) Standard DLT (without weighting mechanism). Single view predicted results of (c) our approach and (d) without weighting. 
	}
	\label{fig:occlusion_images_2}
\end{figure*}

\begin{figure*}
	\centering
	
	\begin{subfigure}[b]{0.235\linewidth}        
		\centering
		\includegraphics[width=\linewidth]{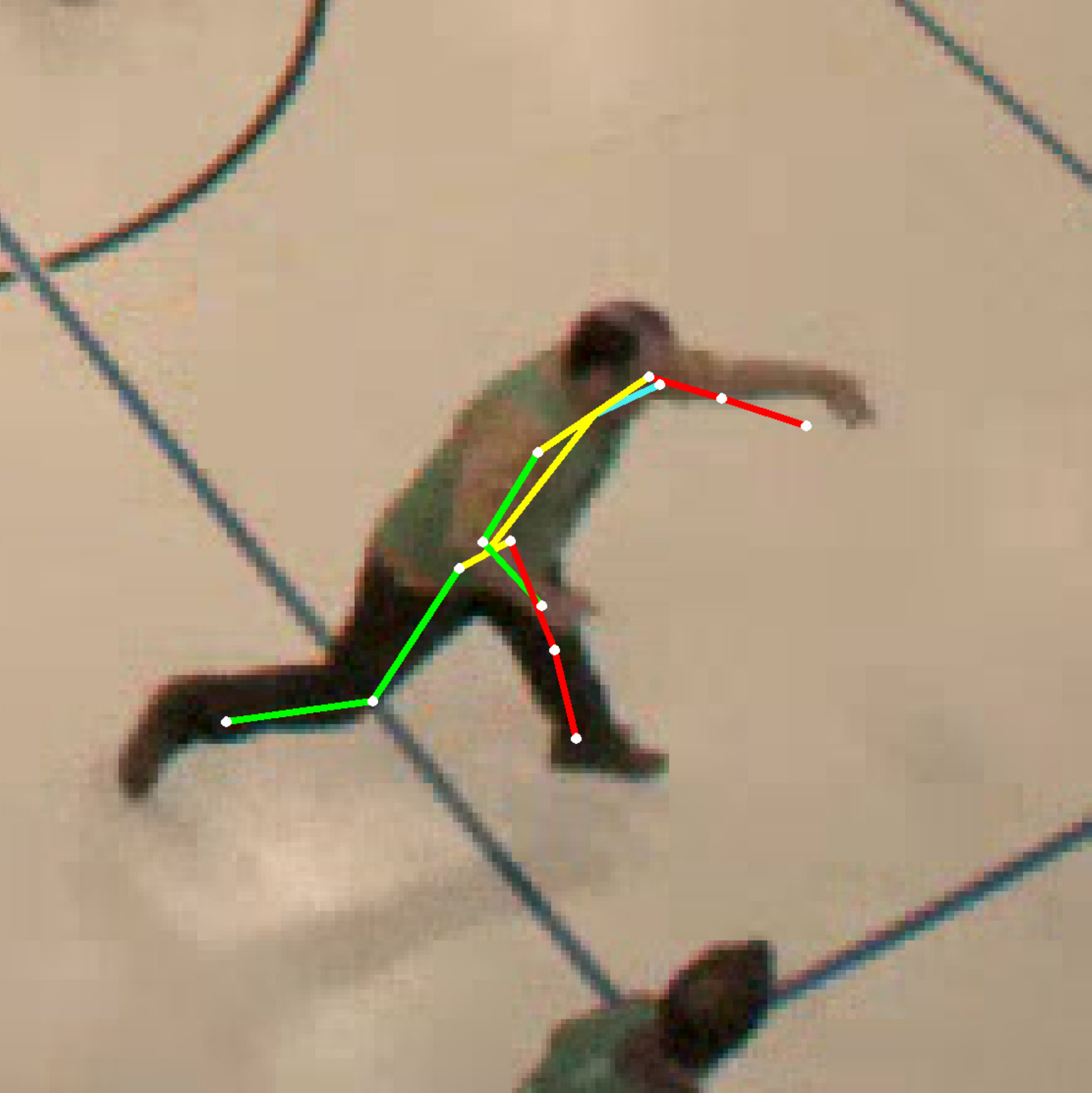}
	\end{subfigure}
	\begin{subfigure}[b]{0.235\linewidth}        
		\centering
		\includegraphics[width=\linewidth]{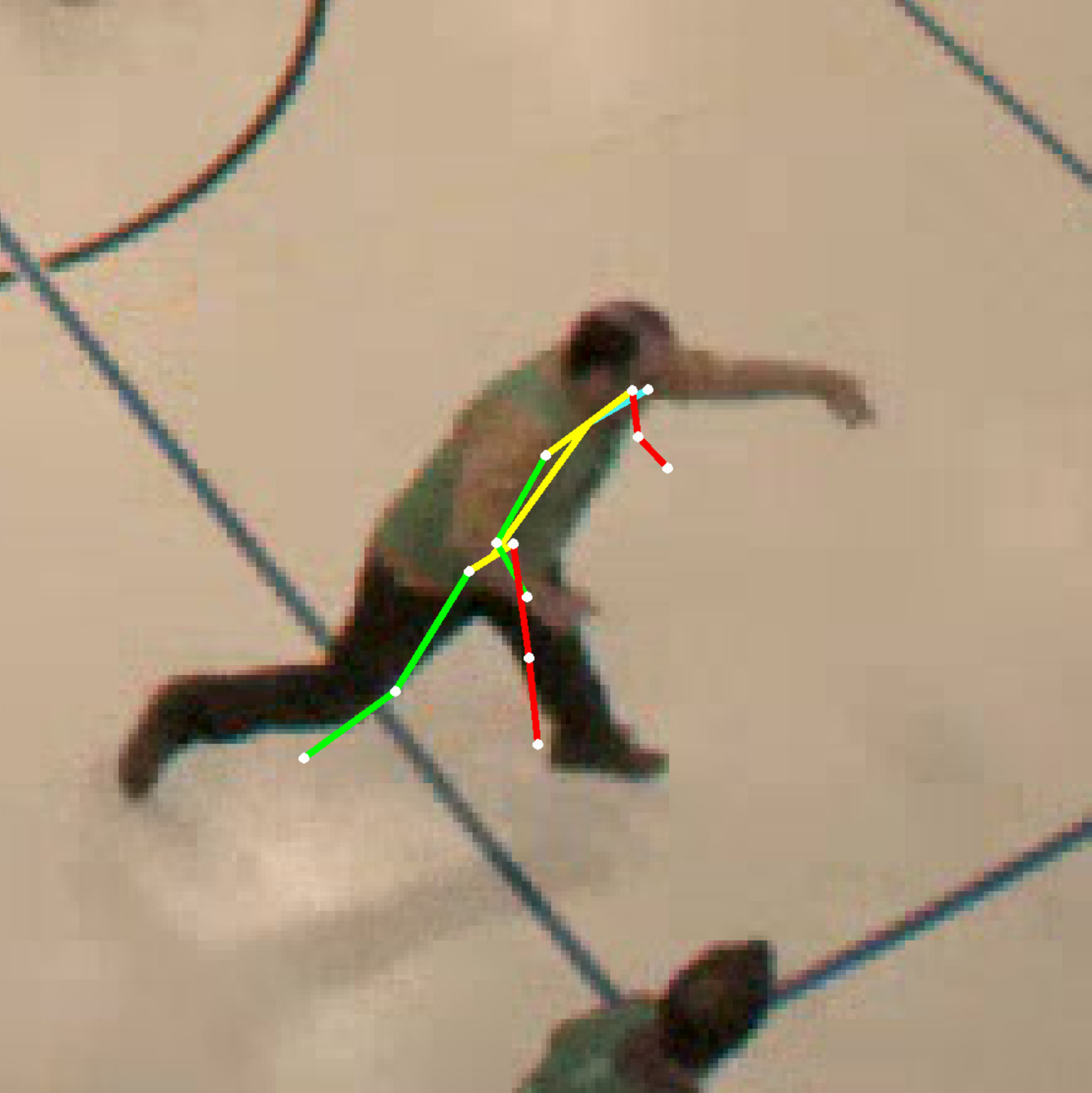}
	\end{subfigure}
	\begin{subfigure}[b]{0.235\linewidth}        
		\centering
		\includegraphics[width=\linewidth]{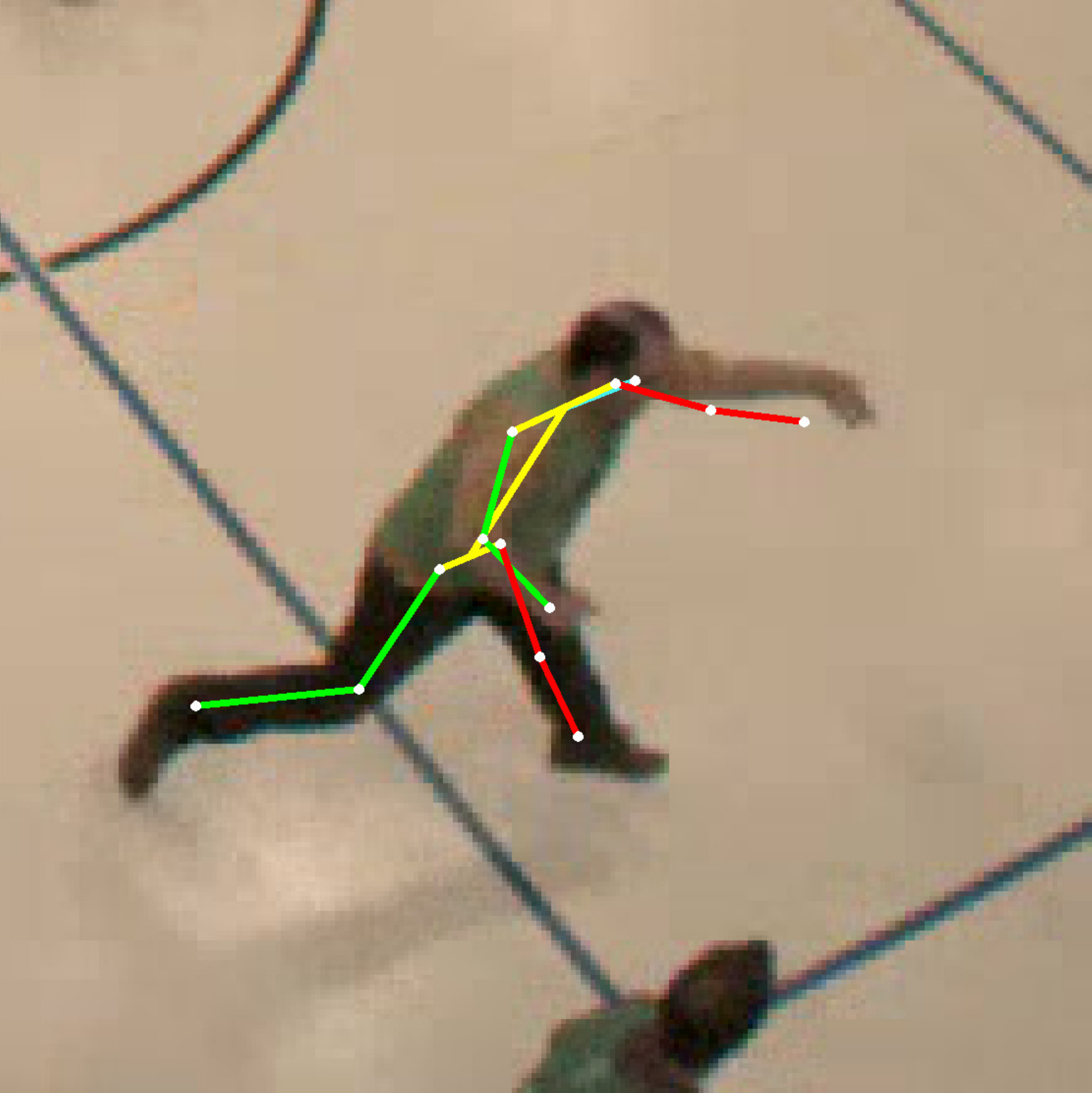}
	\end{subfigure}
	\begin{subfigure}[b]{0.235\linewidth}        
		\centering
		\includegraphics[width=\linewidth]{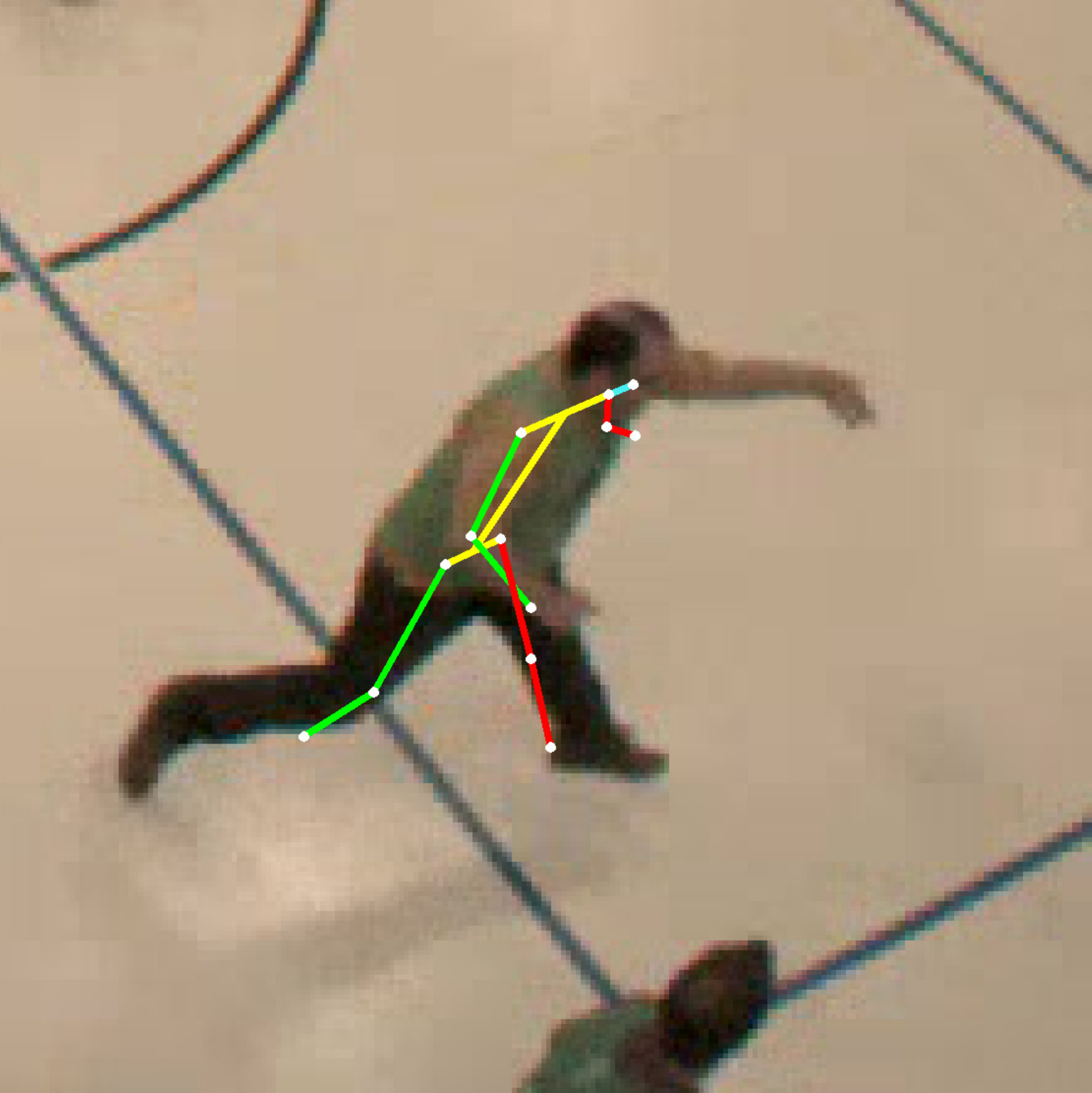}
	\end{subfigure} \\   
	
	\begin{subfigure}[b]{0.235\linewidth}        
		\centering
		\includegraphics[width=\linewidth]{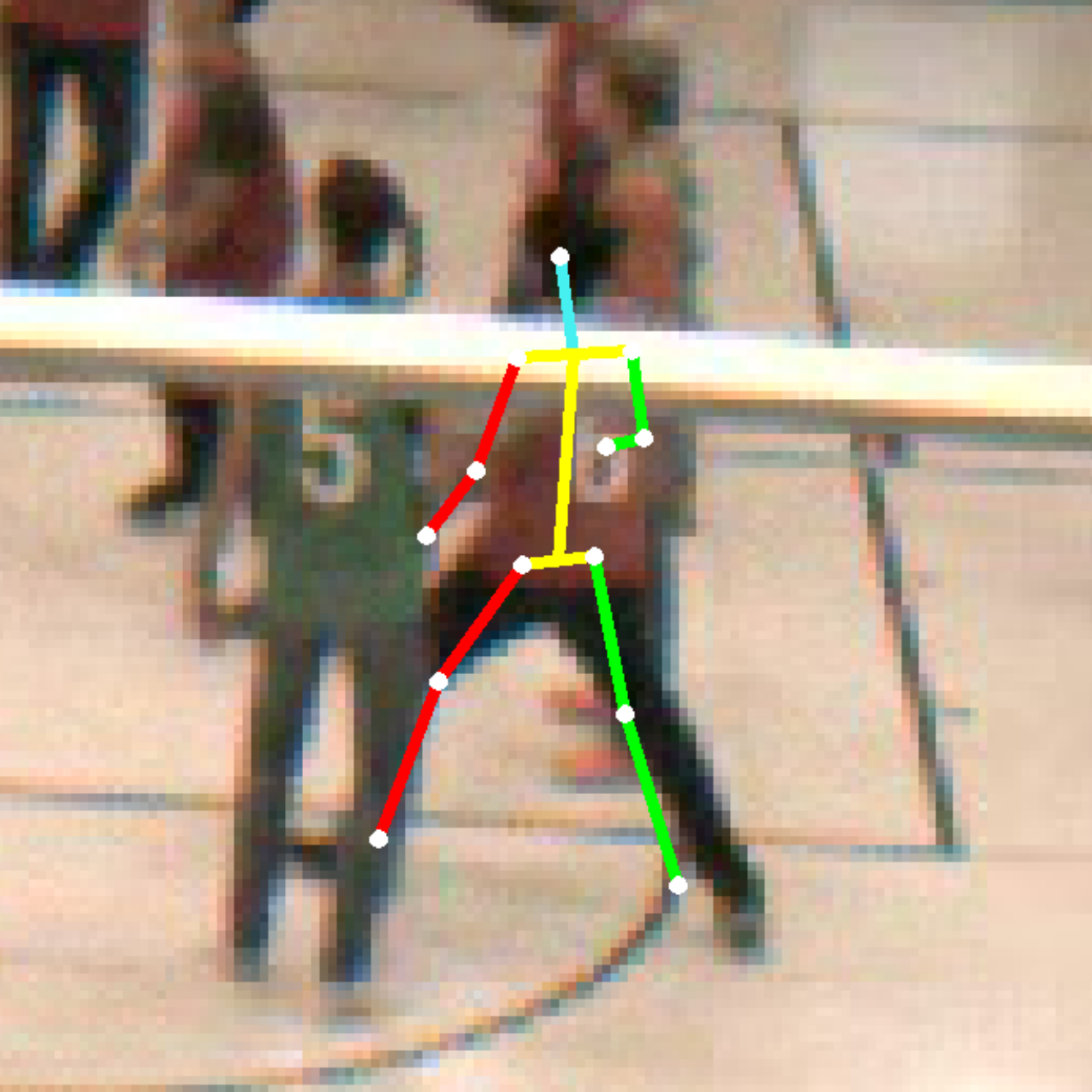}
	\end{subfigure}
	\begin{subfigure}[b]{0.235\linewidth}        
		\centering
		\includegraphics[width=\linewidth]{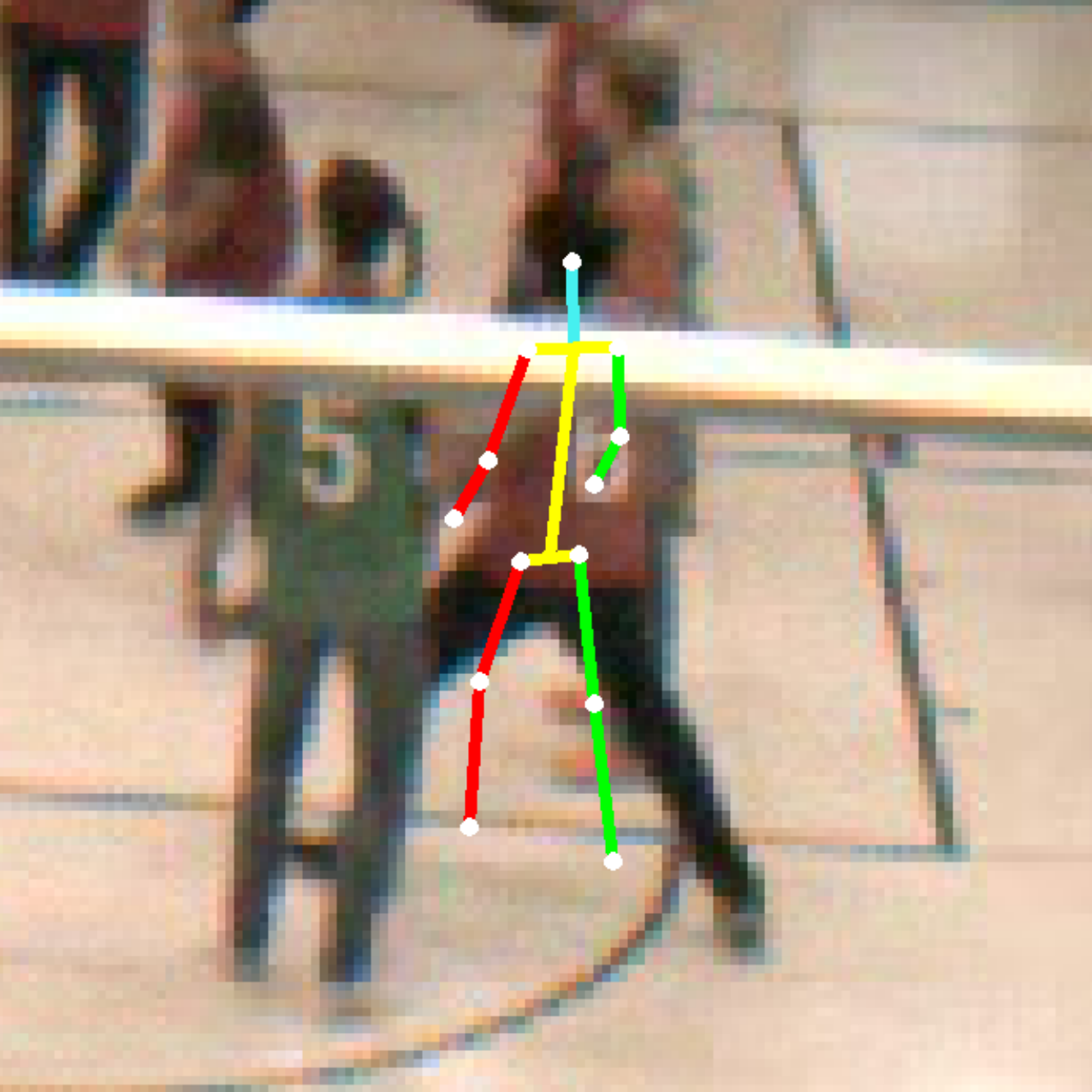}
	\end{subfigure}
	\begin{subfigure}[b]{0.235\linewidth}        
		\centering
		\includegraphics[width=\linewidth]{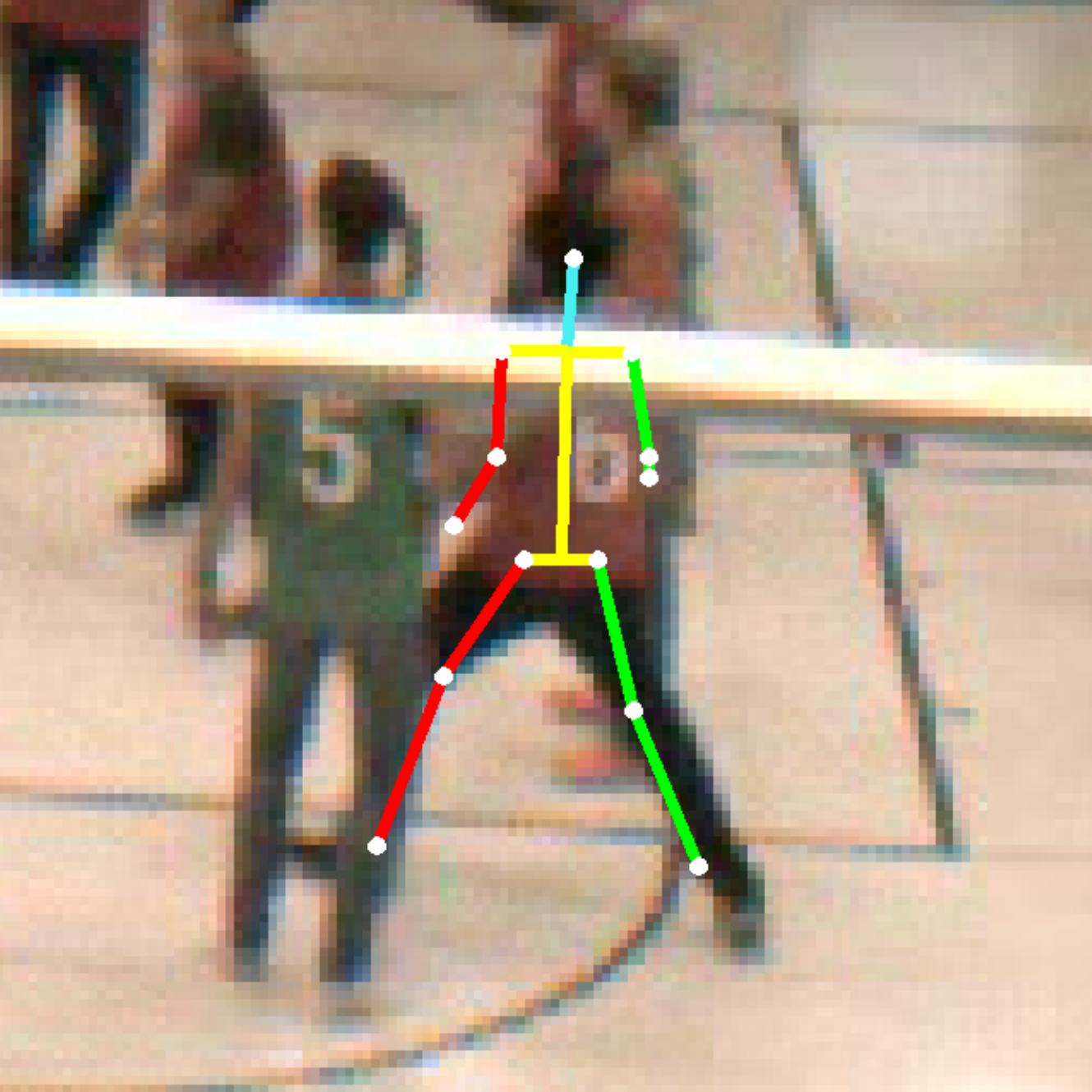}
	\end{subfigure}
	\begin{subfigure}[b]{0.235\linewidth}        
		\centering
		\includegraphics[width=\linewidth]{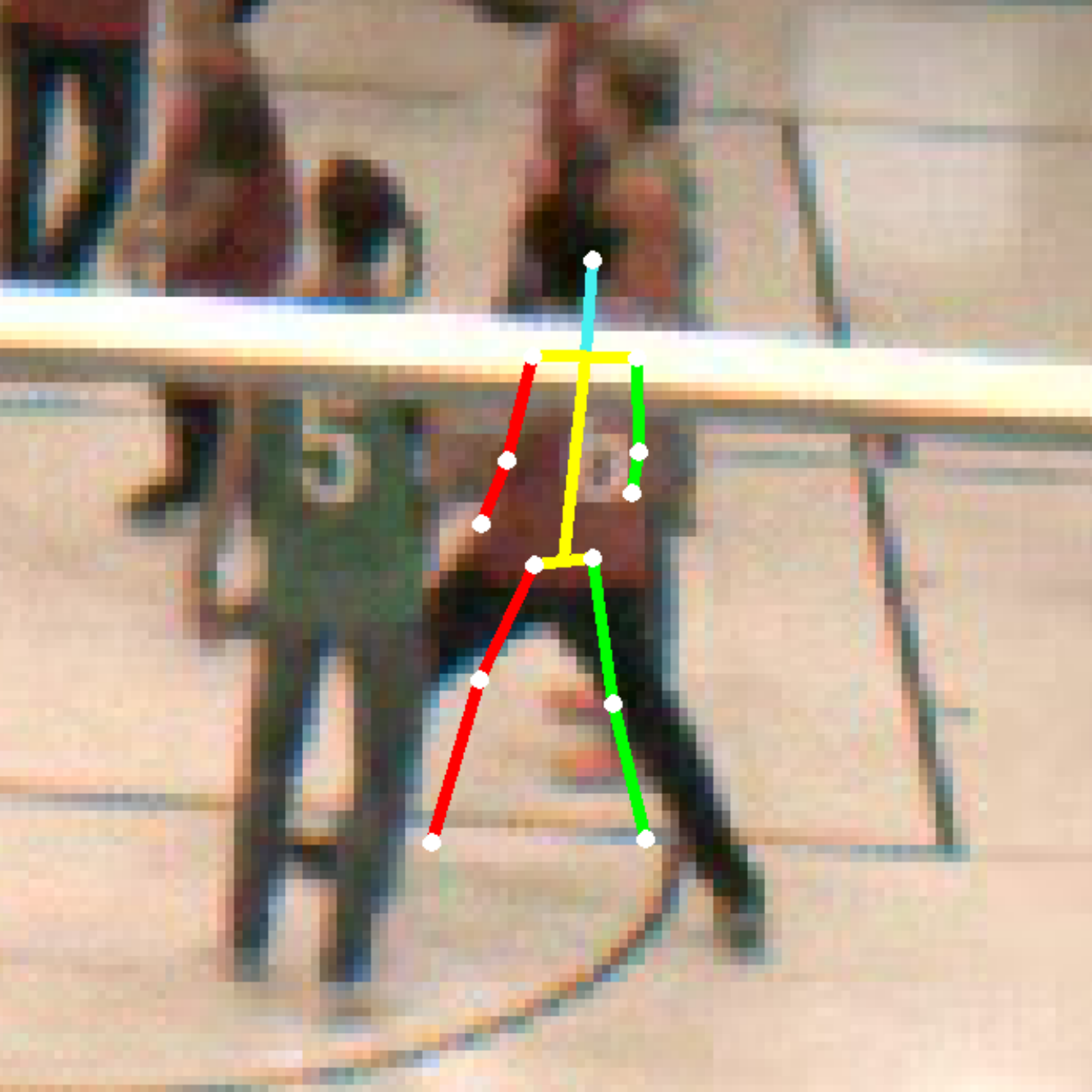}
	\end{subfigure}  \\   \vspace{-1mm}
	
	\begin{subfigure}[b]{0.235\linewidth}        
		\centering
		\includegraphics[width=\linewidth]{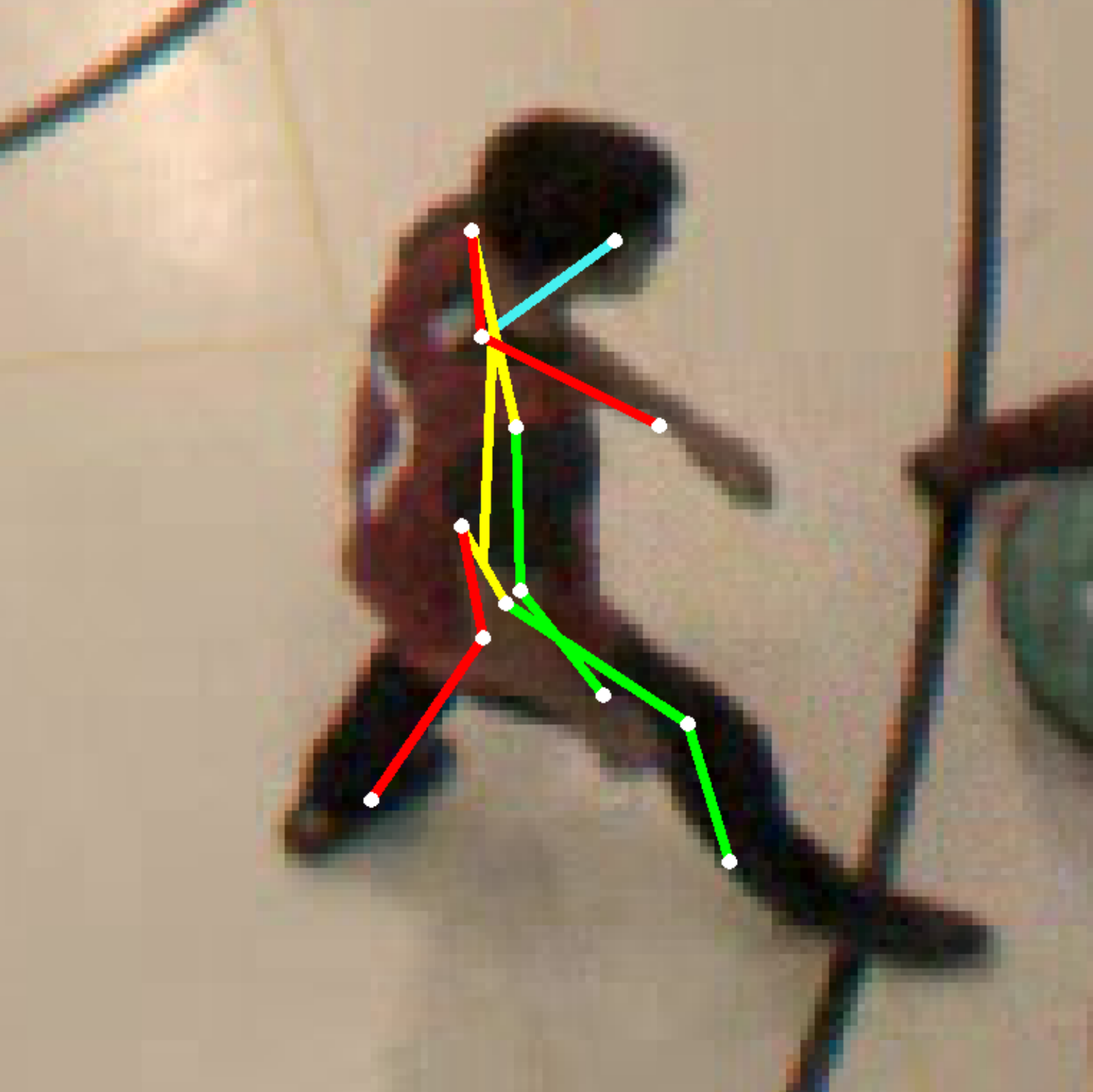}
	\end{subfigure}
	\begin{subfigure}[b]{0.235\linewidth}        
		\centering
		\includegraphics[width=\linewidth]{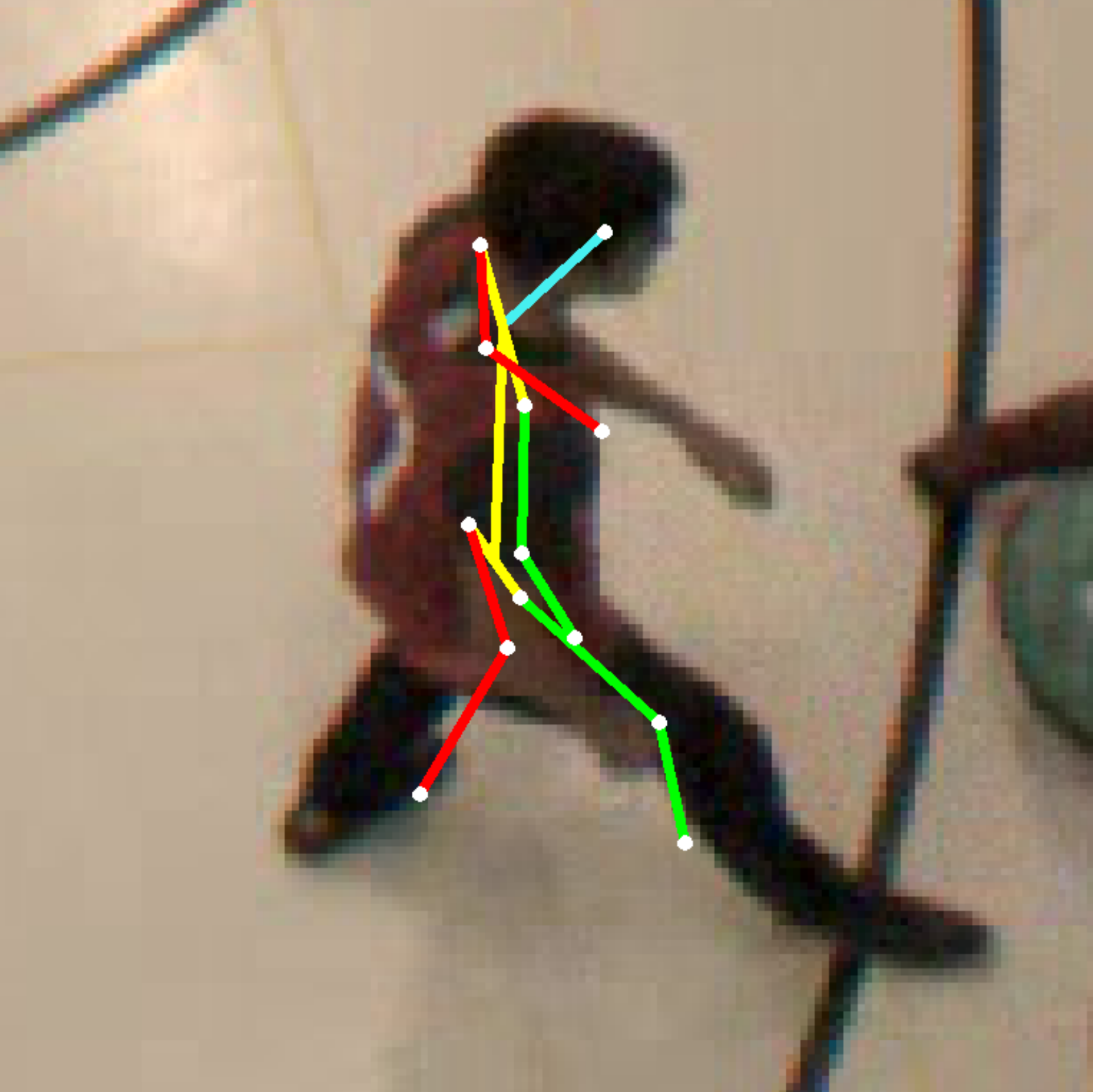}
	\end{subfigure}
	\begin{subfigure}[b]{0.235\linewidth}        
		\centering
		\includegraphics[width=\linewidth]{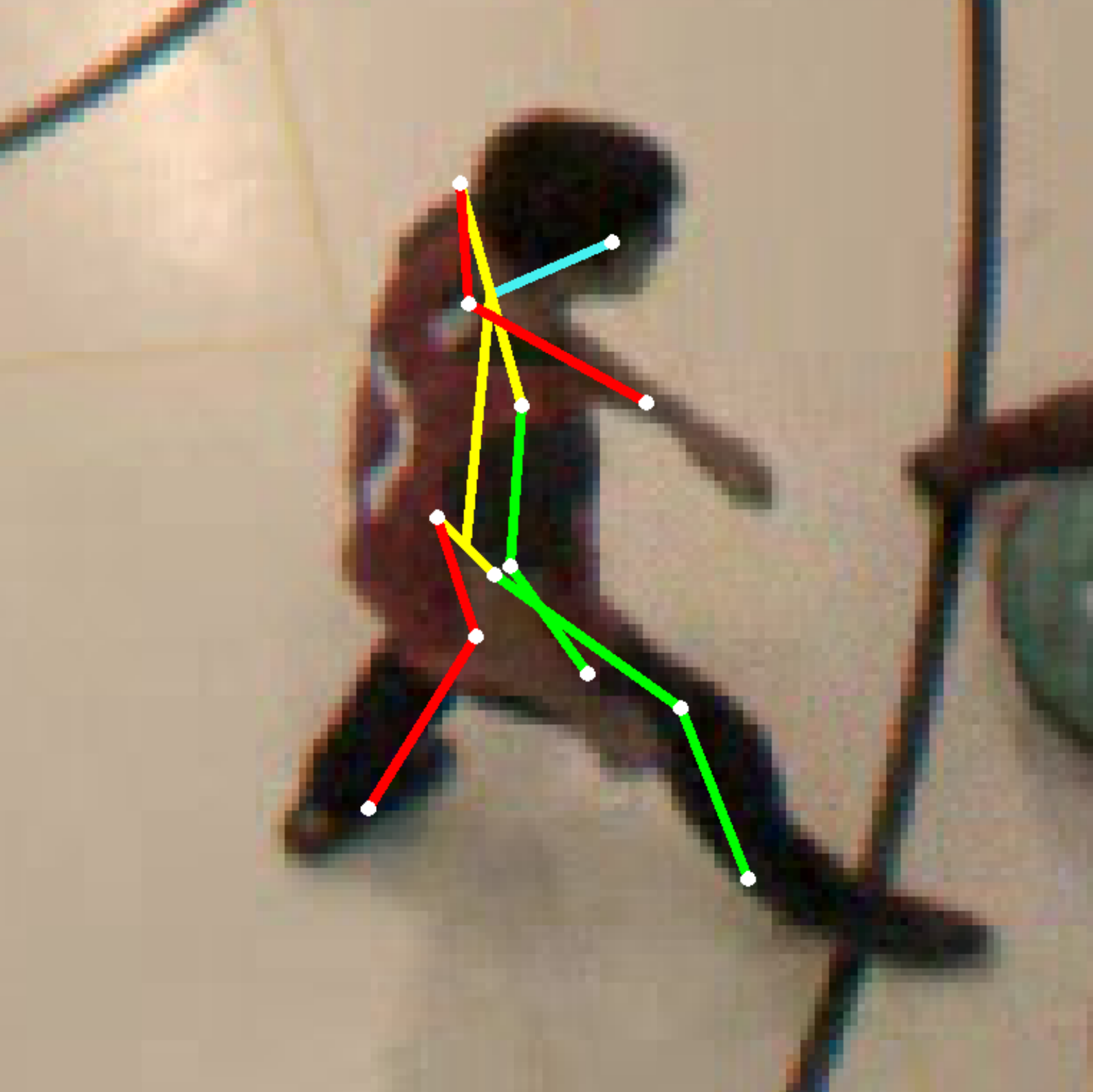}
	\end{subfigure}
	\begin{subfigure}[b]{0.235\linewidth}        
		\centering
		\includegraphics[width=\linewidth]{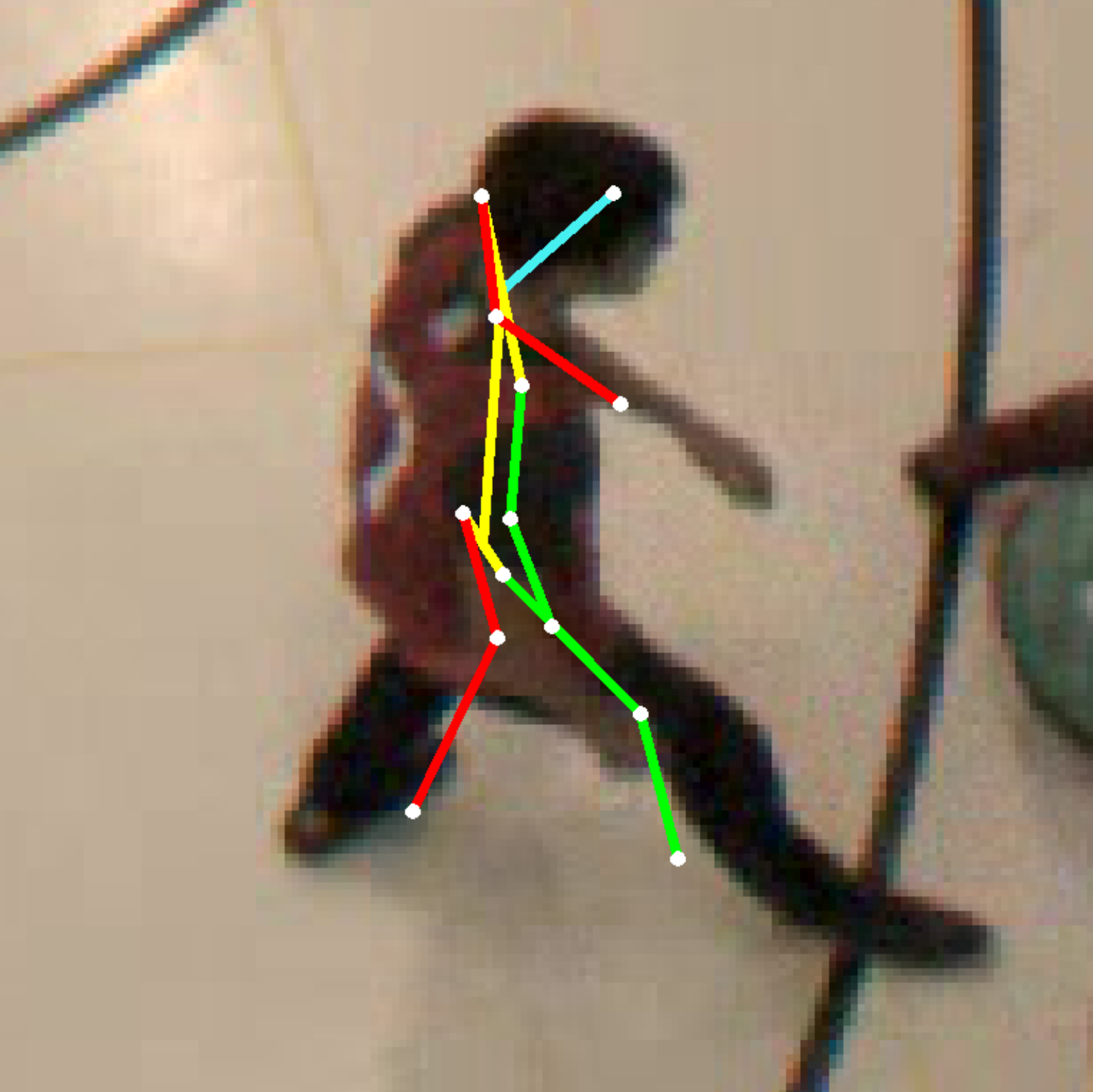}
	\end{subfigure} \\ \vspace{-1mm}

	\begin{subfigure}[b]{0.235\linewidth}        
		\centering
		\includegraphics[width=\linewidth]{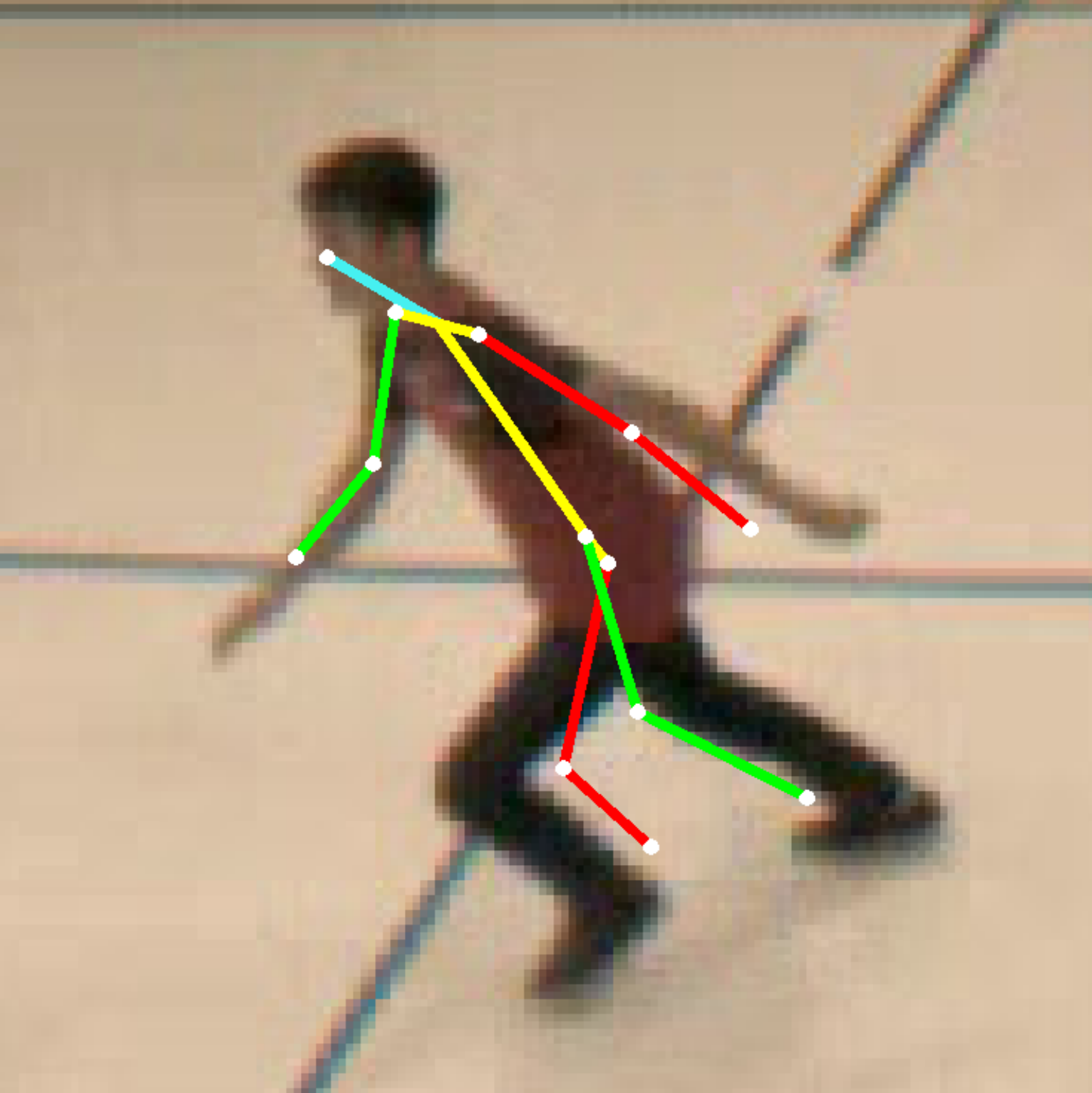}
	\end{subfigure}
	\begin{subfigure}[b]{0.235\linewidth}        
		\centering
		\includegraphics[width=\linewidth]{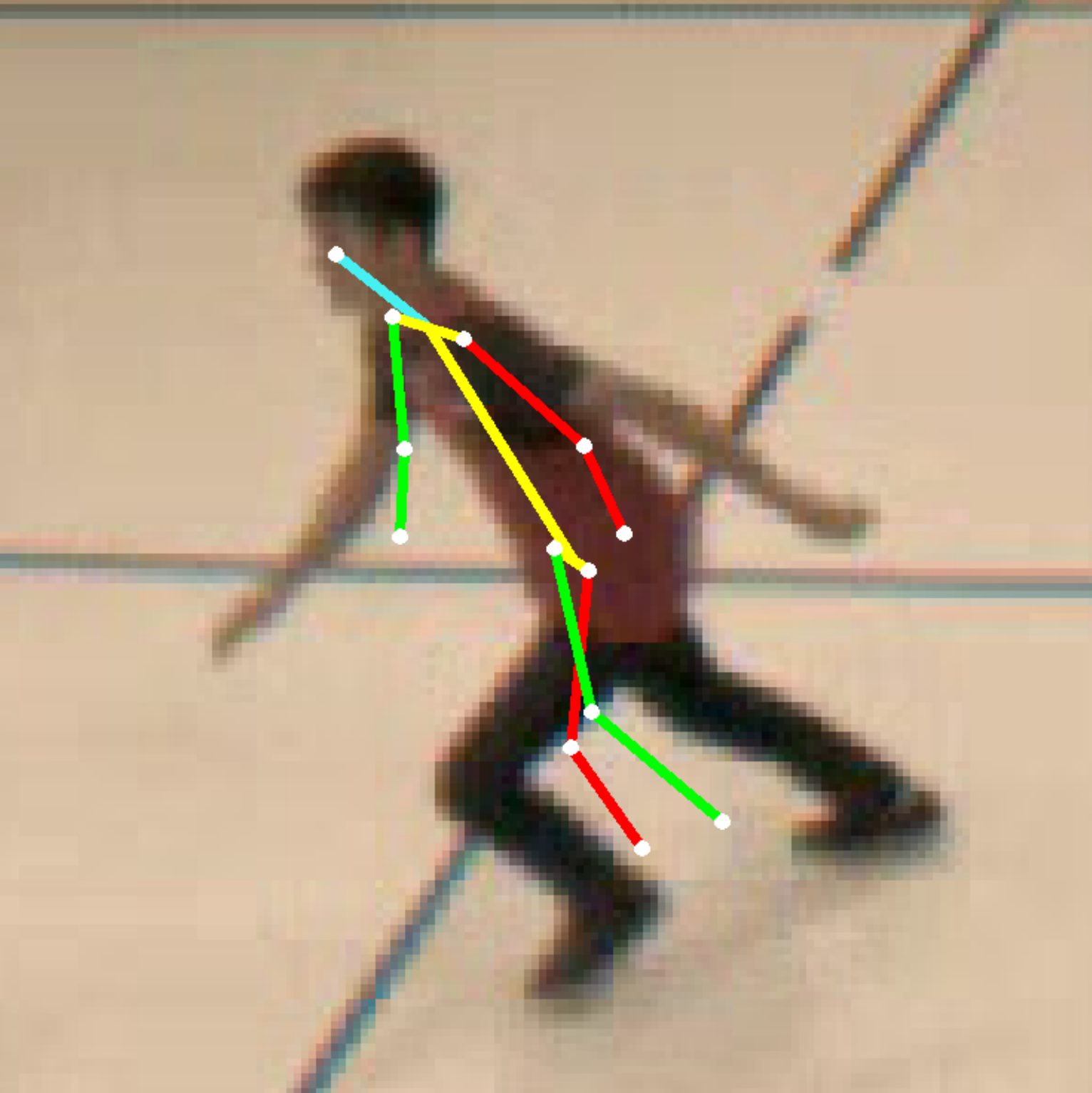}
	\end{subfigure}
	\begin{subfigure}[b]{0.235\linewidth}        
		\centering
		\includegraphics[width=\linewidth]{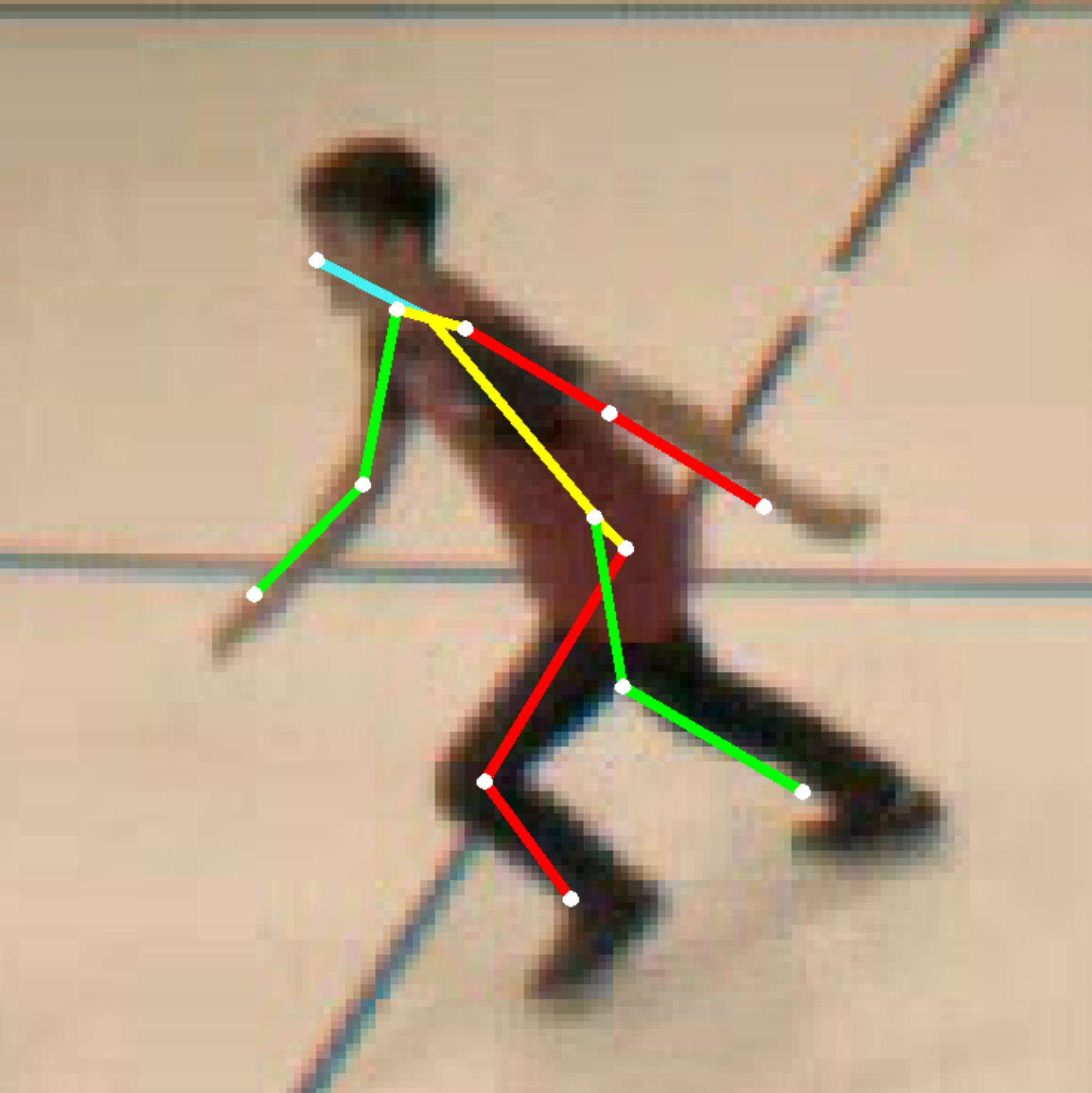}
	\end{subfigure}
	\begin{subfigure}[b]{0.235\linewidth}        
		\centering
		\includegraphics[width=\linewidth]{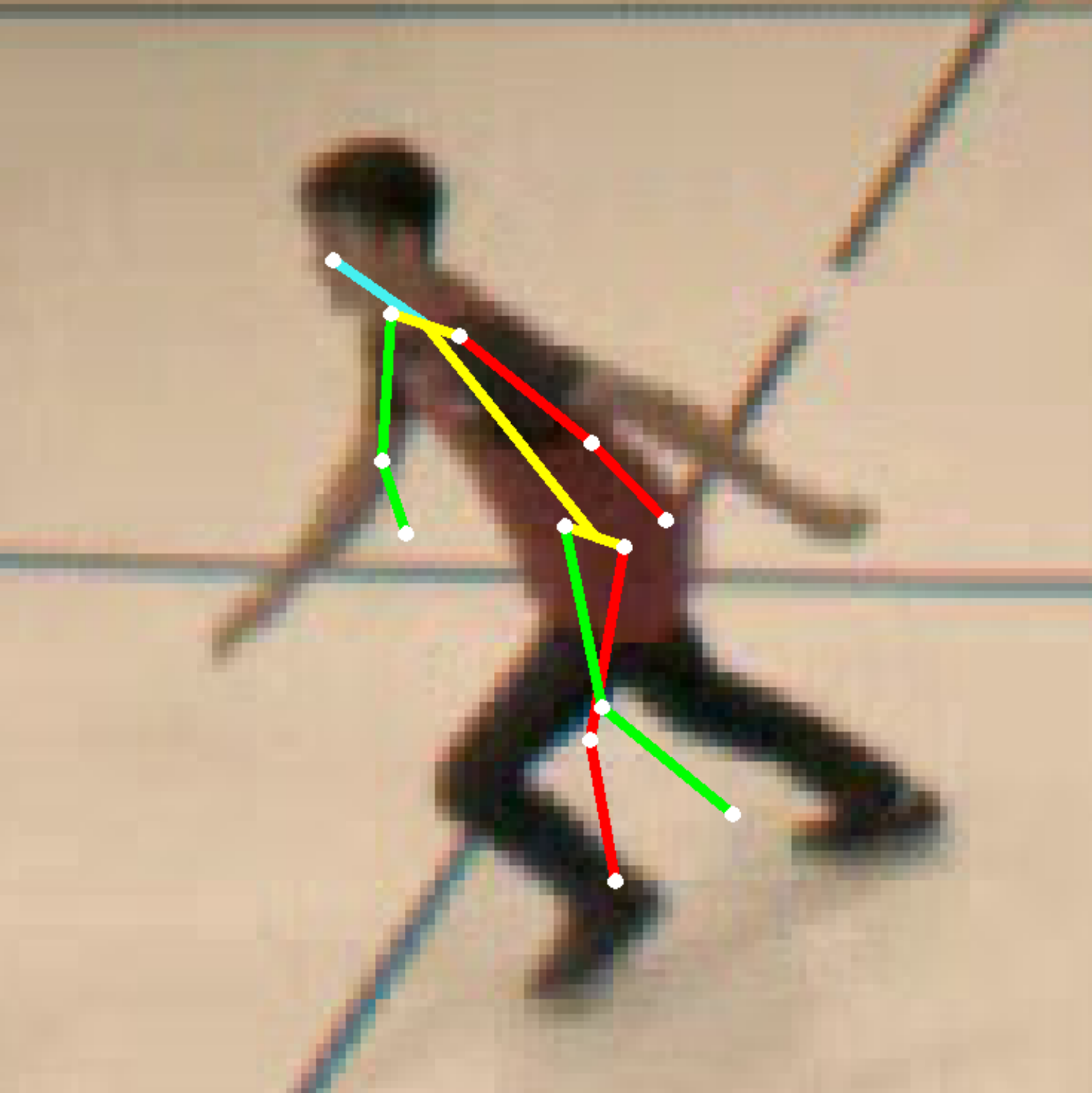}
	\end{subfigure} \\ \vspace{-3mm}
	
	\begin{subfigure}[b]{0.235\linewidth}        
		\centering
		\includegraphics[width=\linewidth]{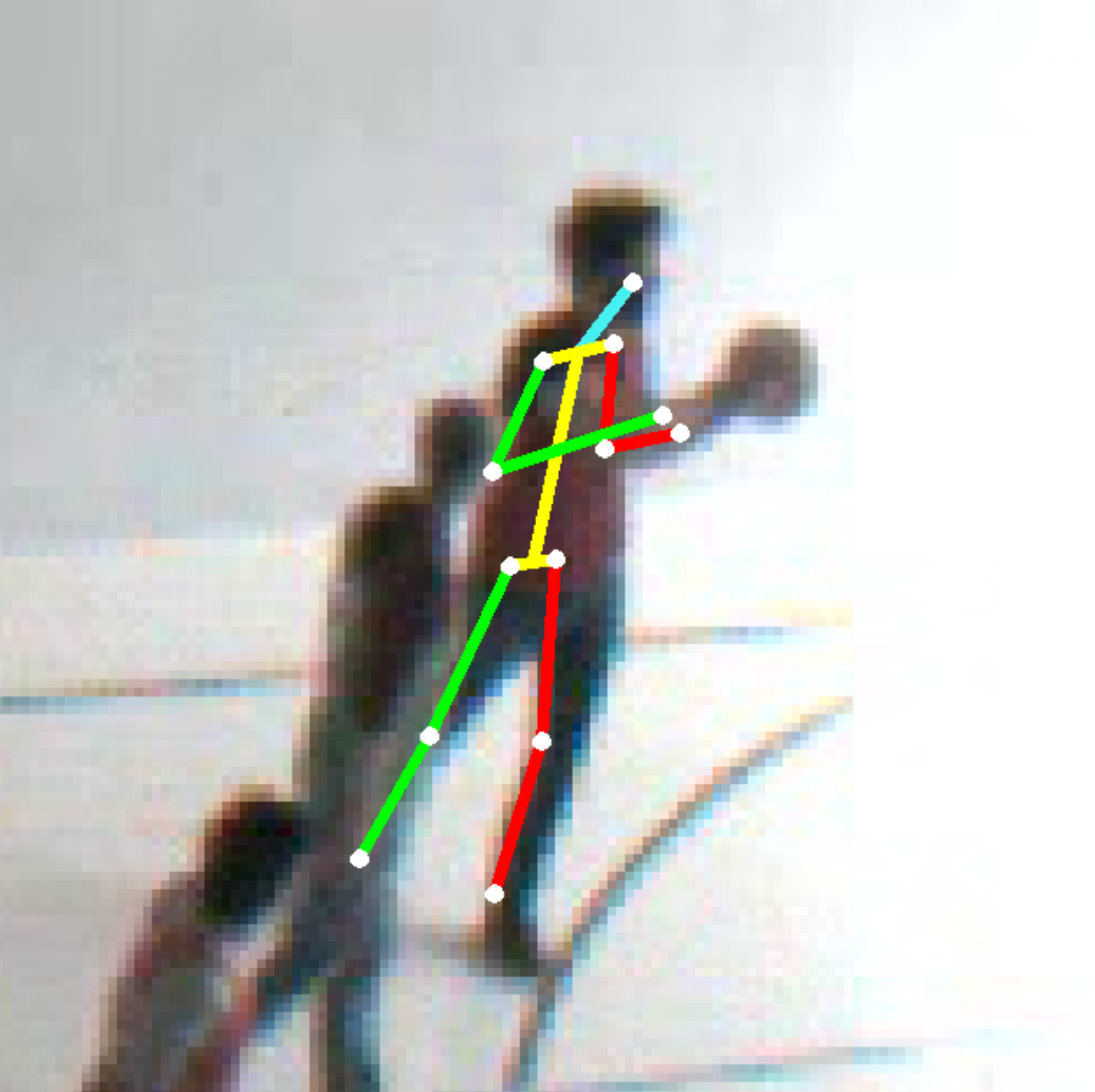}
		\caption{Ours}
	\end{subfigure}
	\begin{subfigure}[b]{0.235\linewidth}        
		\centering
		\includegraphics[width=\linewidth]{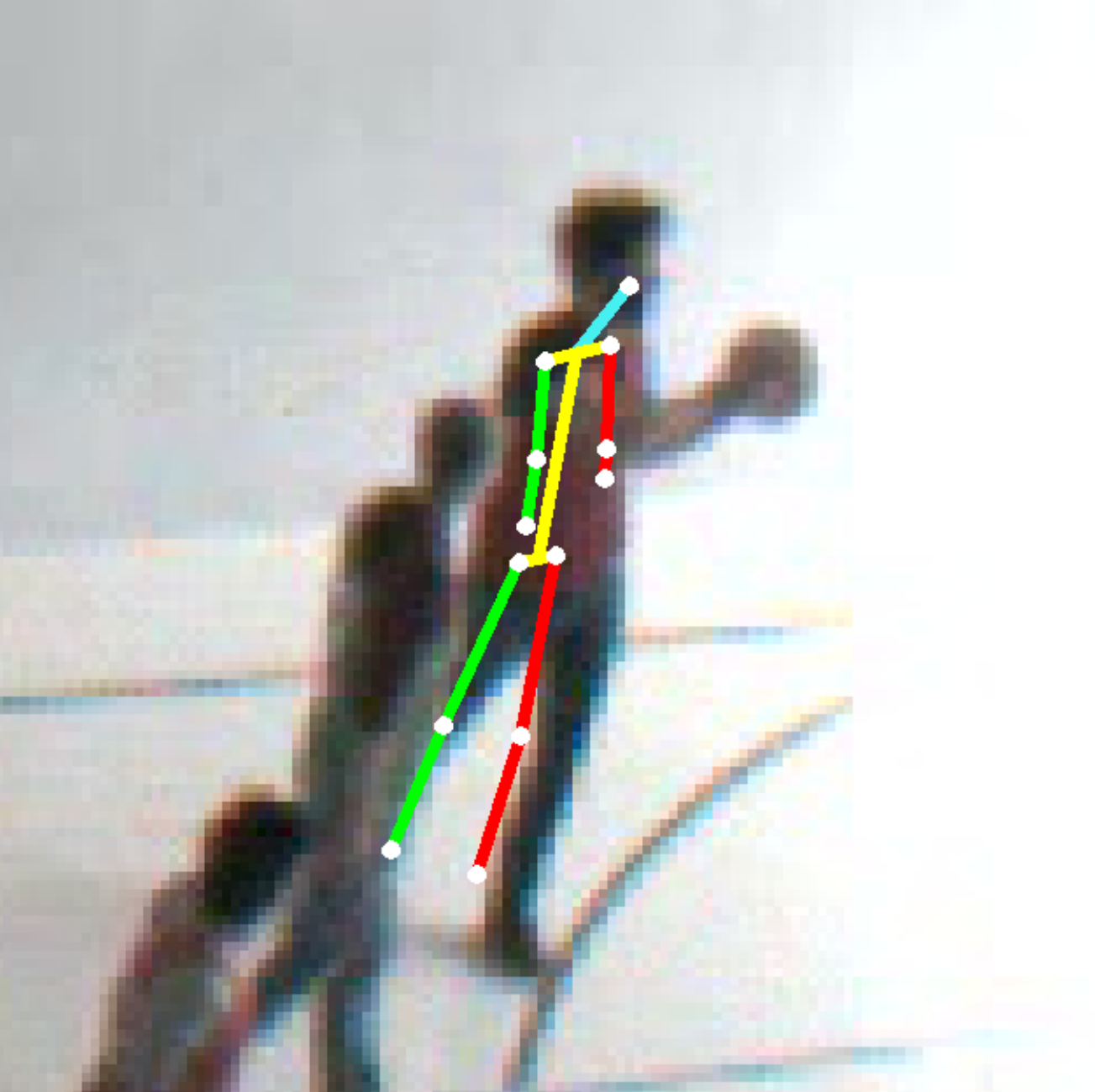}
		\caption{w/o w}
	\end{subfigure}
	\begin{subfigure}[b]{0.235\linewidth}        
		\centering
		\includegraphics[width=\linewidth]{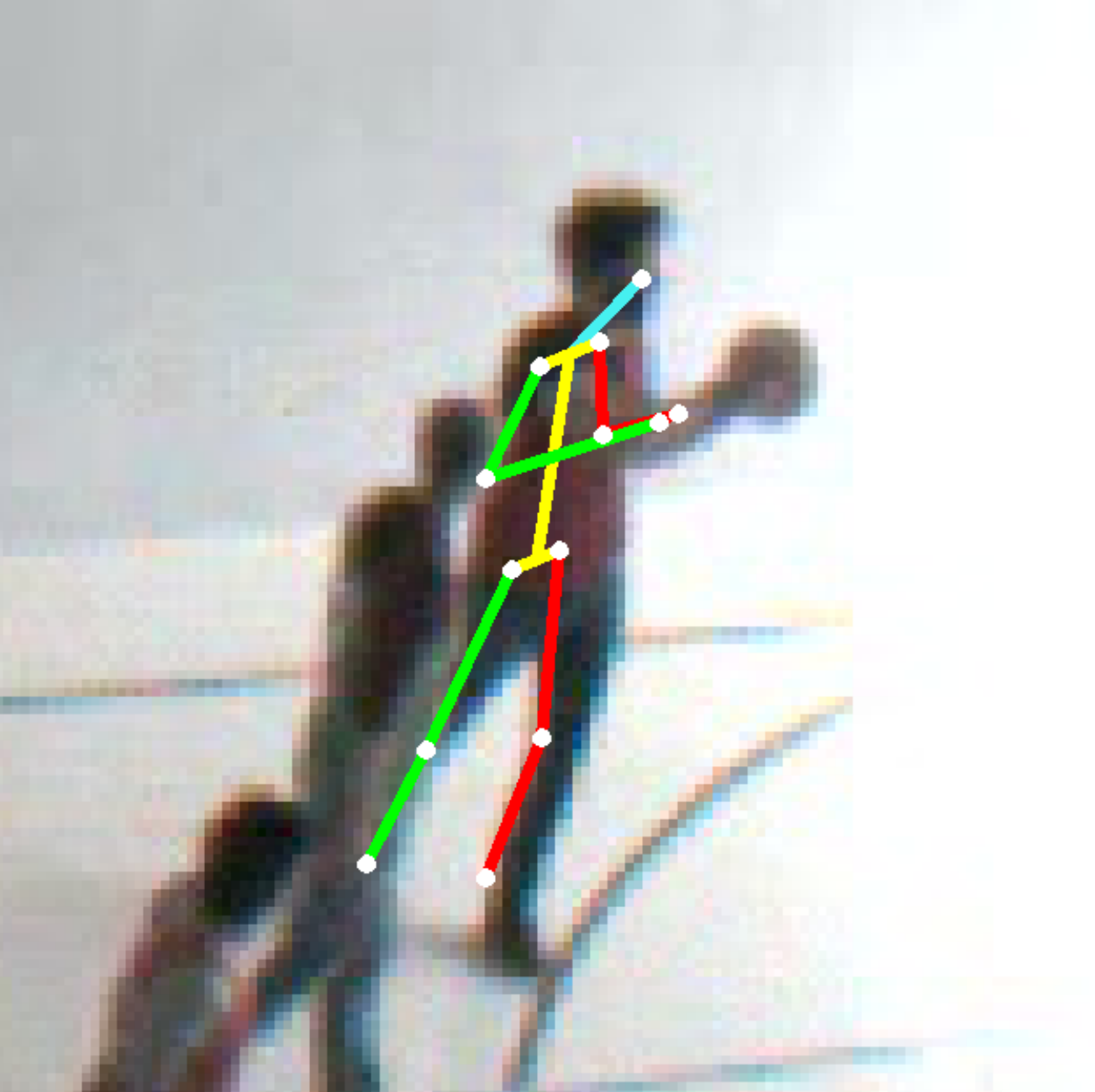}
		\caption{Ours SV}
	\end{subfigure}
	\begin{subfigure}[b]{0.235\linewidth}        
		\centering
		\includegraphics[width=\linewidth]{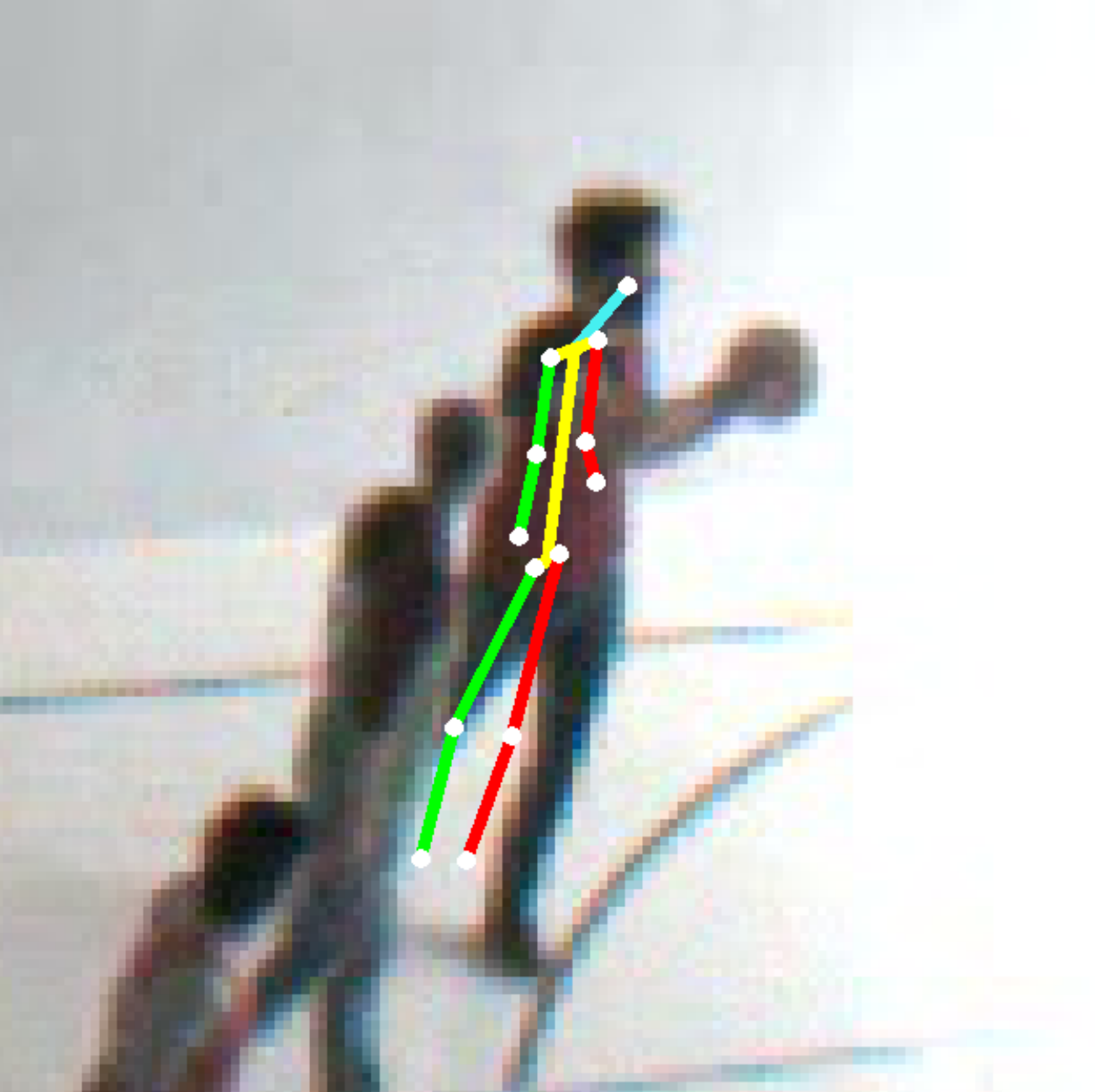}
		\caption{\small w/o w. SV}
	\end{subfigure}

	\vspace{-0.3cm}
	\caption{\small Qualitative results on the SportCenter dataset. From left to right, multi-view triangulated pose with (a) our approach and (b) Standard DLT (without weighting mechanism). Single view predicted results of (c) our approach and (d) without weighting. 
	}
	\label{fig:occlusion_images_3}
\end{figure*}

\begin{figure*}
	\centering
	
	\begin{subfigure}[b]{0.24\linewidth}        
		\centering
		\includegraphics[width=\linewidth]{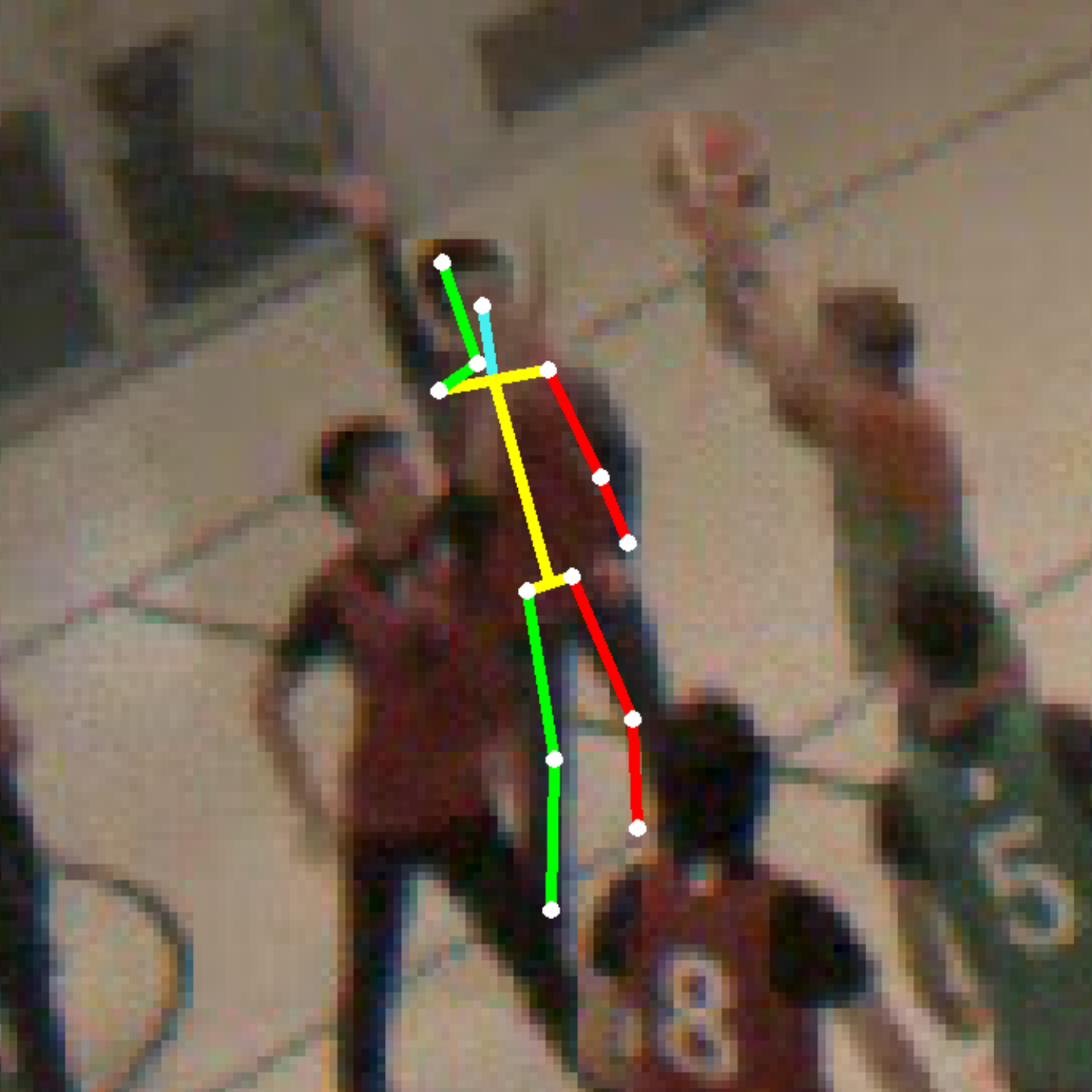}
	\end{subfigure}
	\begin{subfigure}[b]{0.24\linewidth}        
		\centering
		\includegraphics[width=\linewidth]{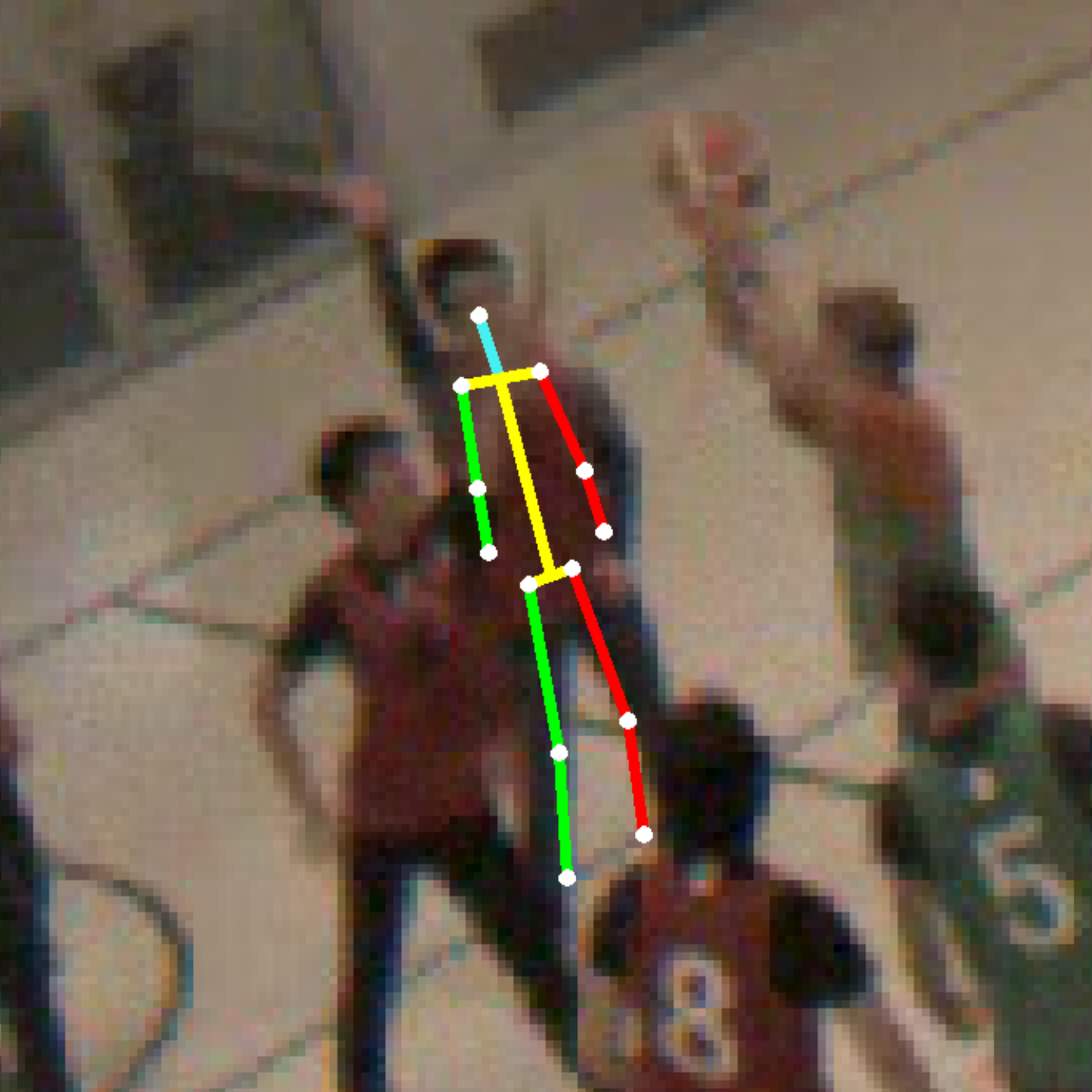}
	\end{subfigure}
	\begin{subfigure}[b]{0.24\linewidth}        
		\centering
		\includegraphics[width=\linewidth]{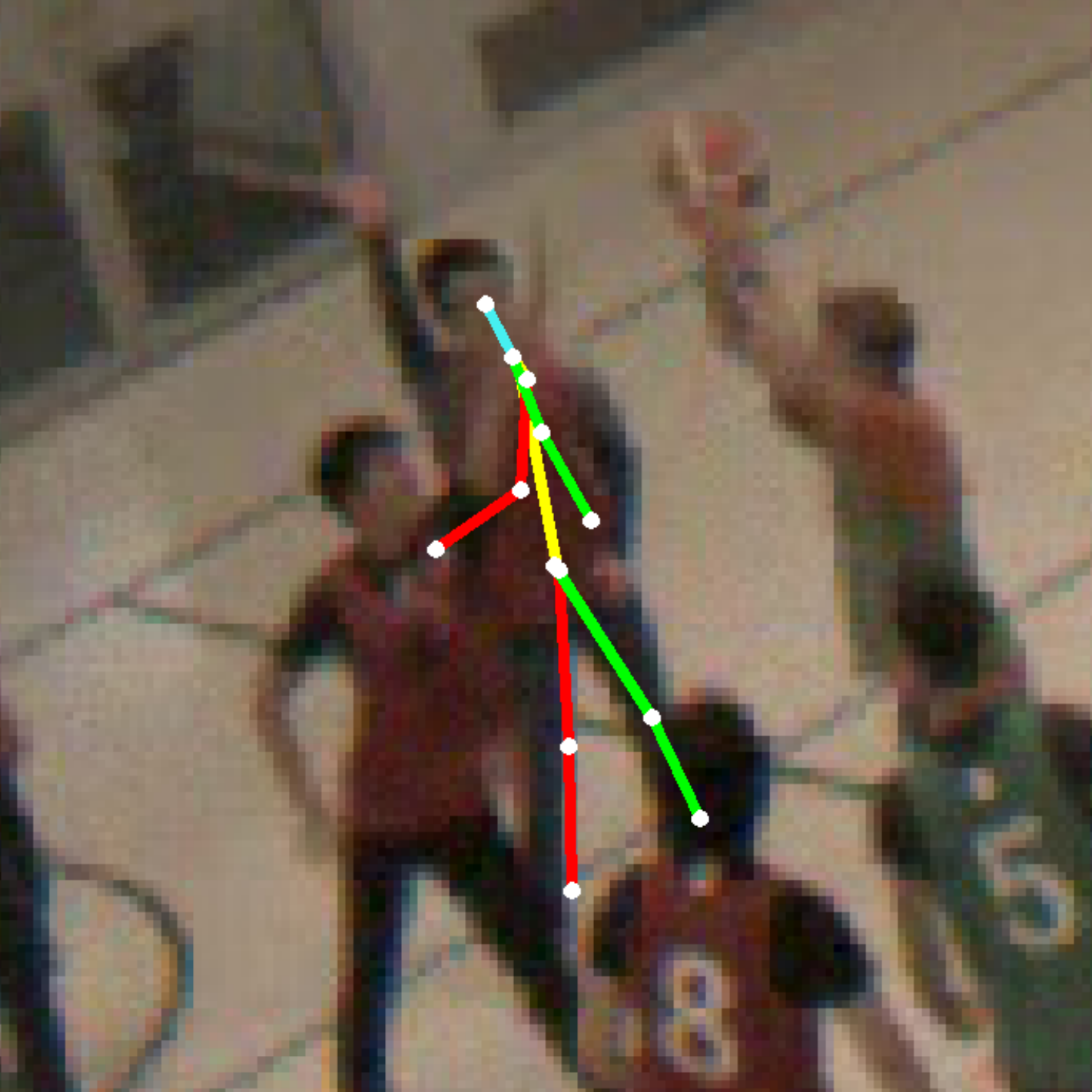}
	\end{subfigure}
	\begin{subfigure}[b]{0.24\linewidth}        
		\centering
		\includegraphics[width=\linewidth]{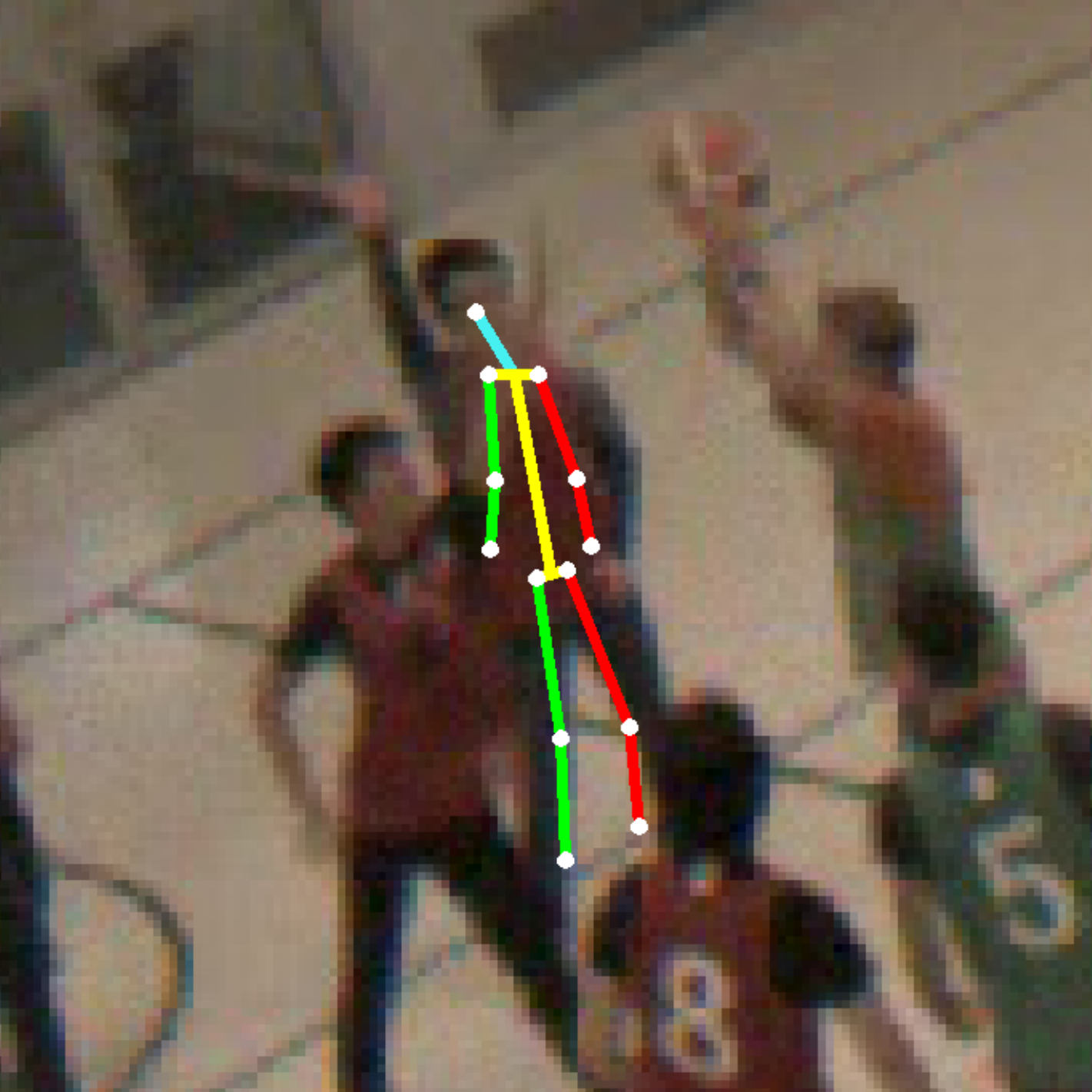}
	\end{subfigure} \\   \vspace{1mm}
	
	\begin{subfigure}[b]{0.24\linewidth}        
		\centering
		\includegraphics[width=\linewidth]{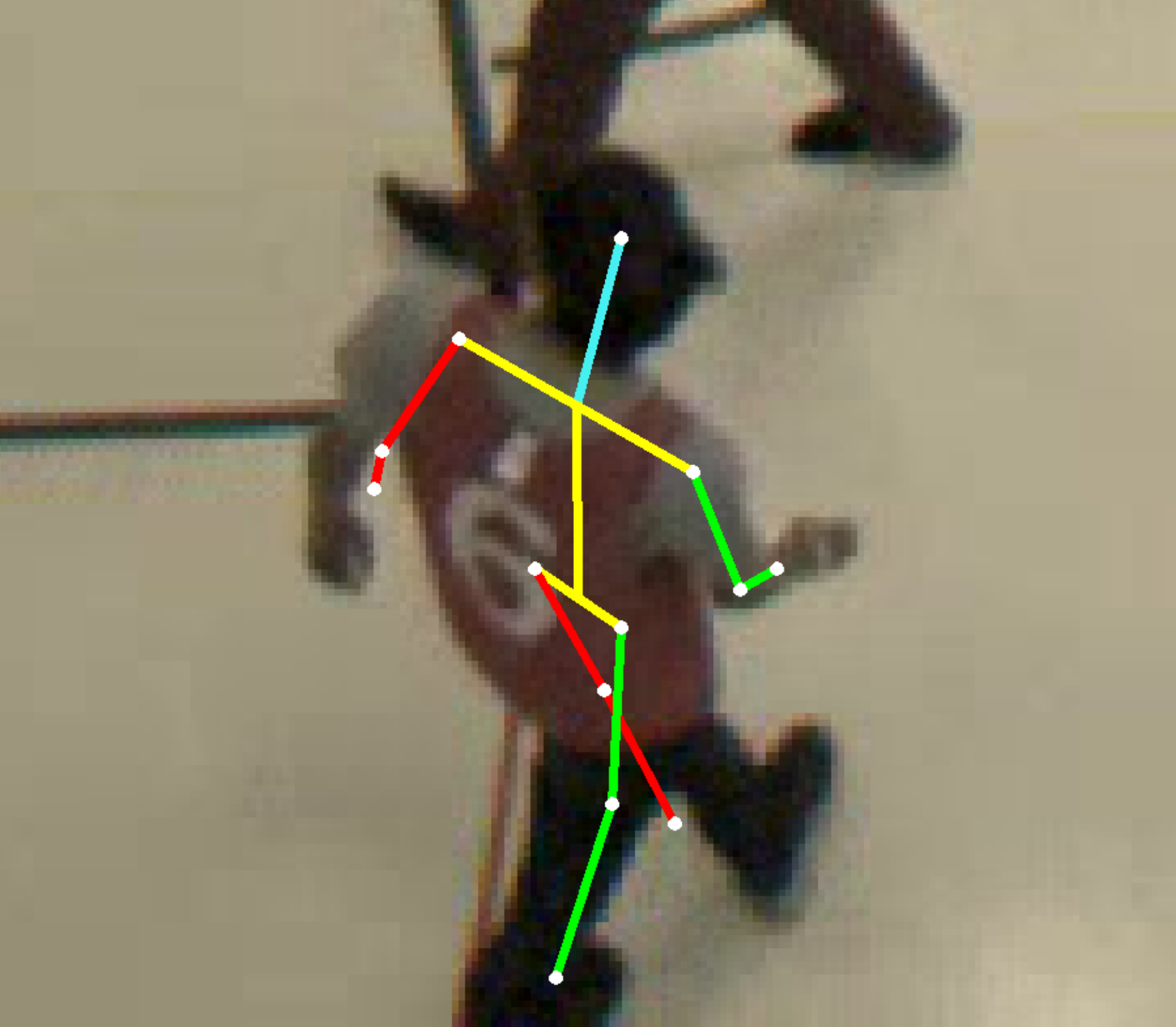}
	\end{subfigure}
	\begin{subfigure}[b]{0.24\linewidth}        
		\centering
		\includegraphics[width=\linewidth]{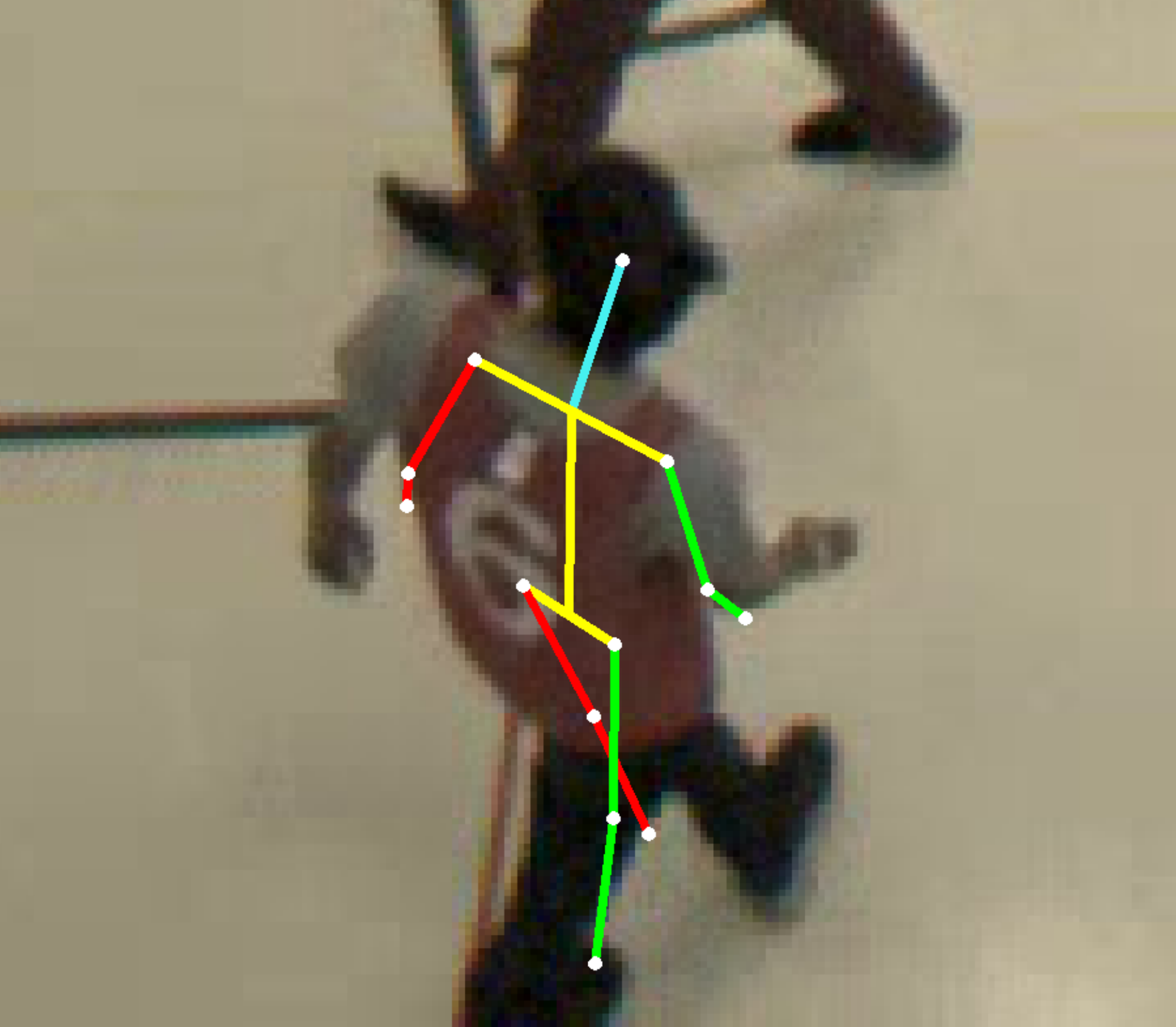}
	\end{subfigure}
	\begin{subfigure}[b]{0.24\linewidth}        
		\centering
		\includegraphics[width=\linewidth]{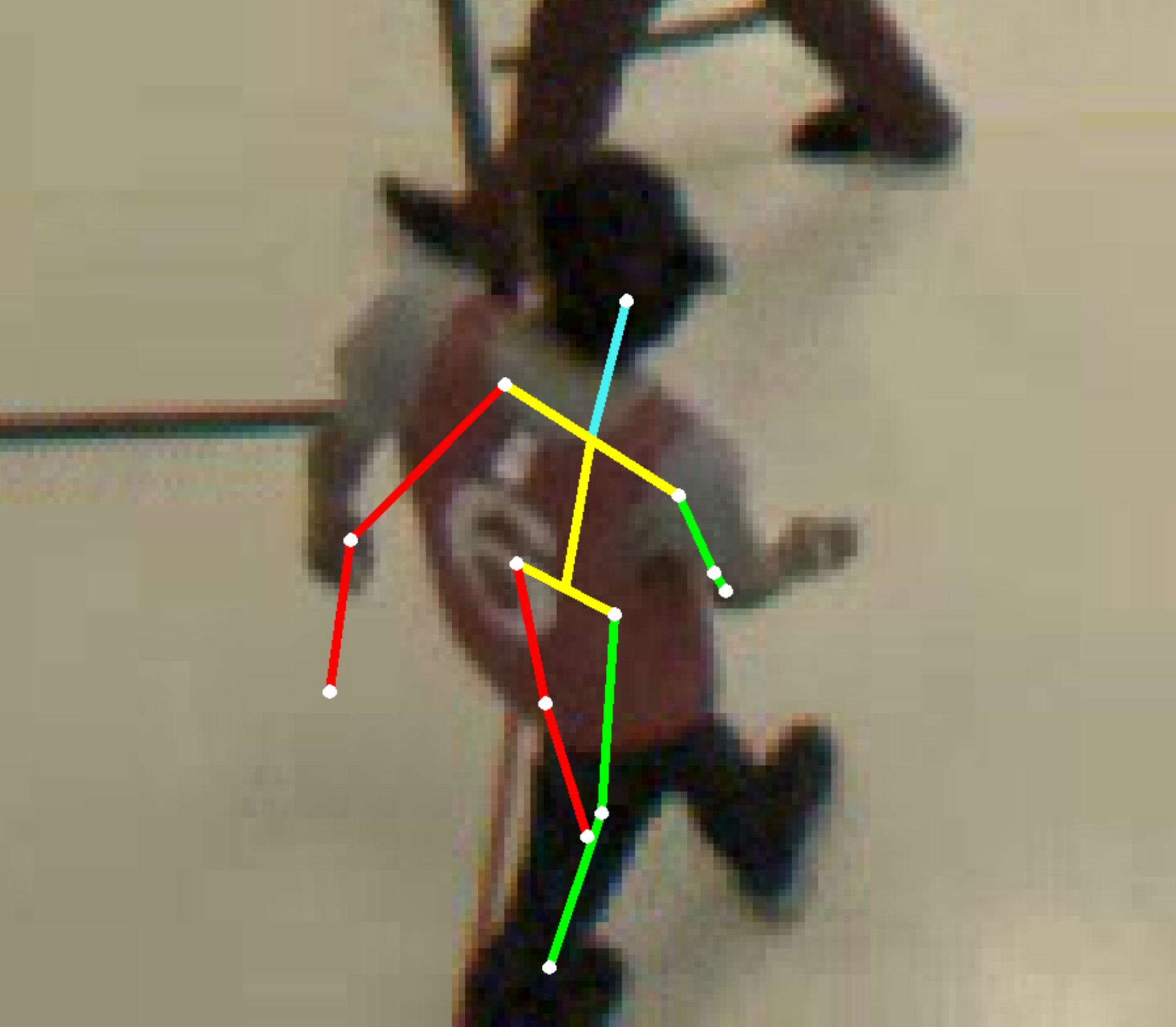}
	\end{subfigure}
	\begin{subfigure}[b]{0.24\linewidth}        
		\centering
		\includegraphics[width=\linewidth]{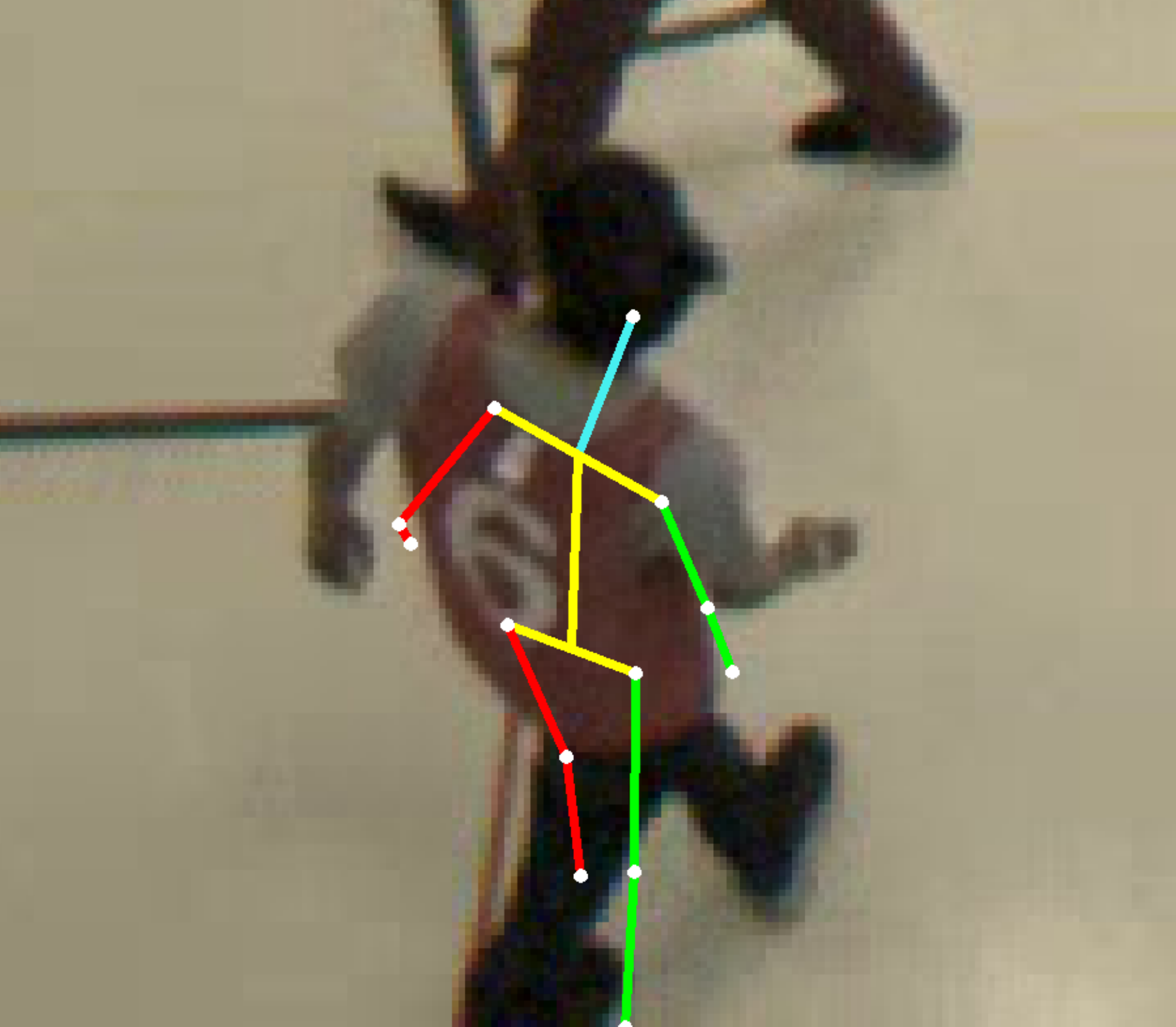}
	\end{subfigure}  \\ 
	
	\begin{subfigure}[b]{0.24\linewidth}        
		\centering
		\includegraphics[width=\linewidth]{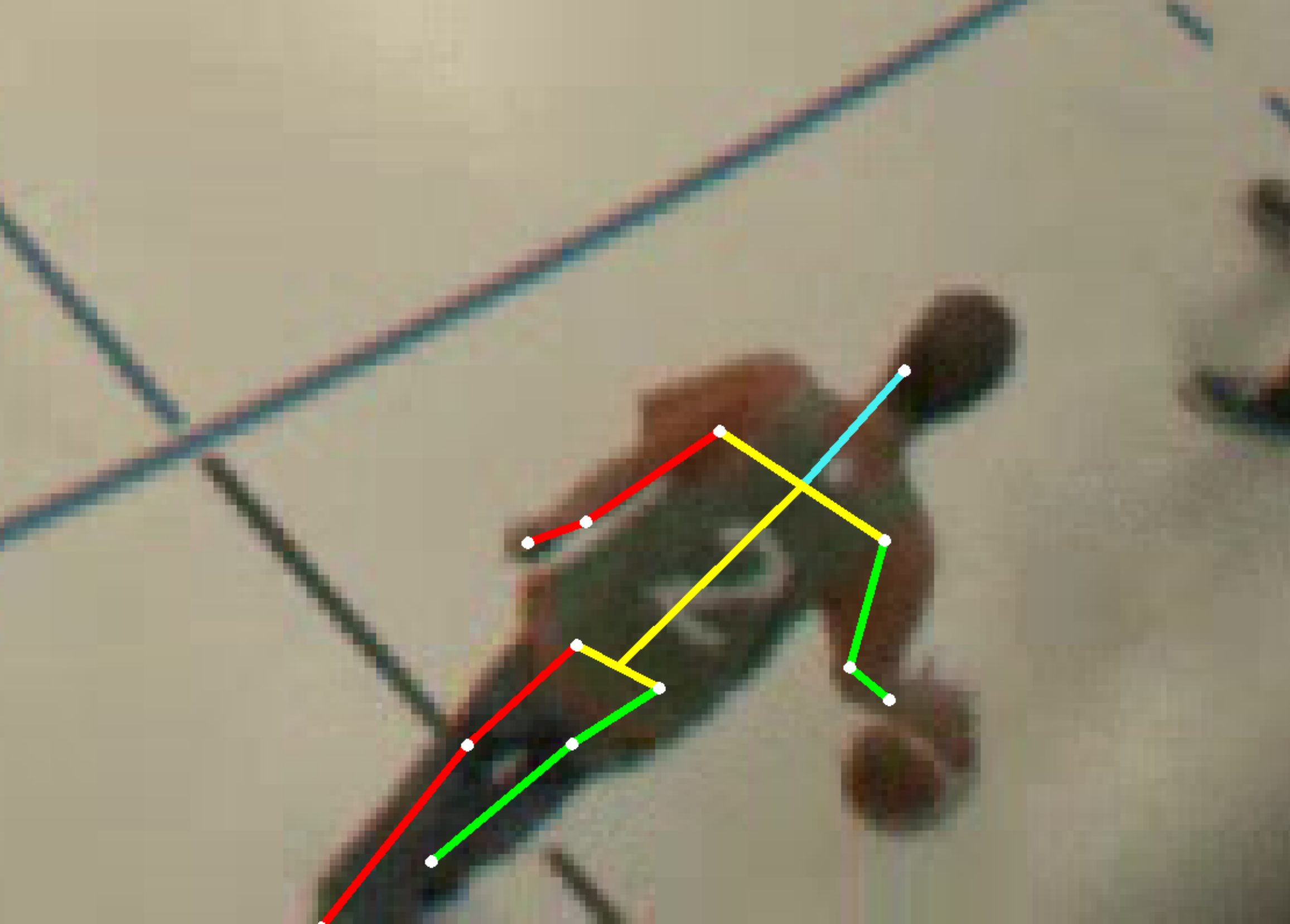}
	\end{subfigure}
	\begin{subfigure}[b]{0.24\linewidth}        
		\centering
		\includegraphics[width=\linewidth]{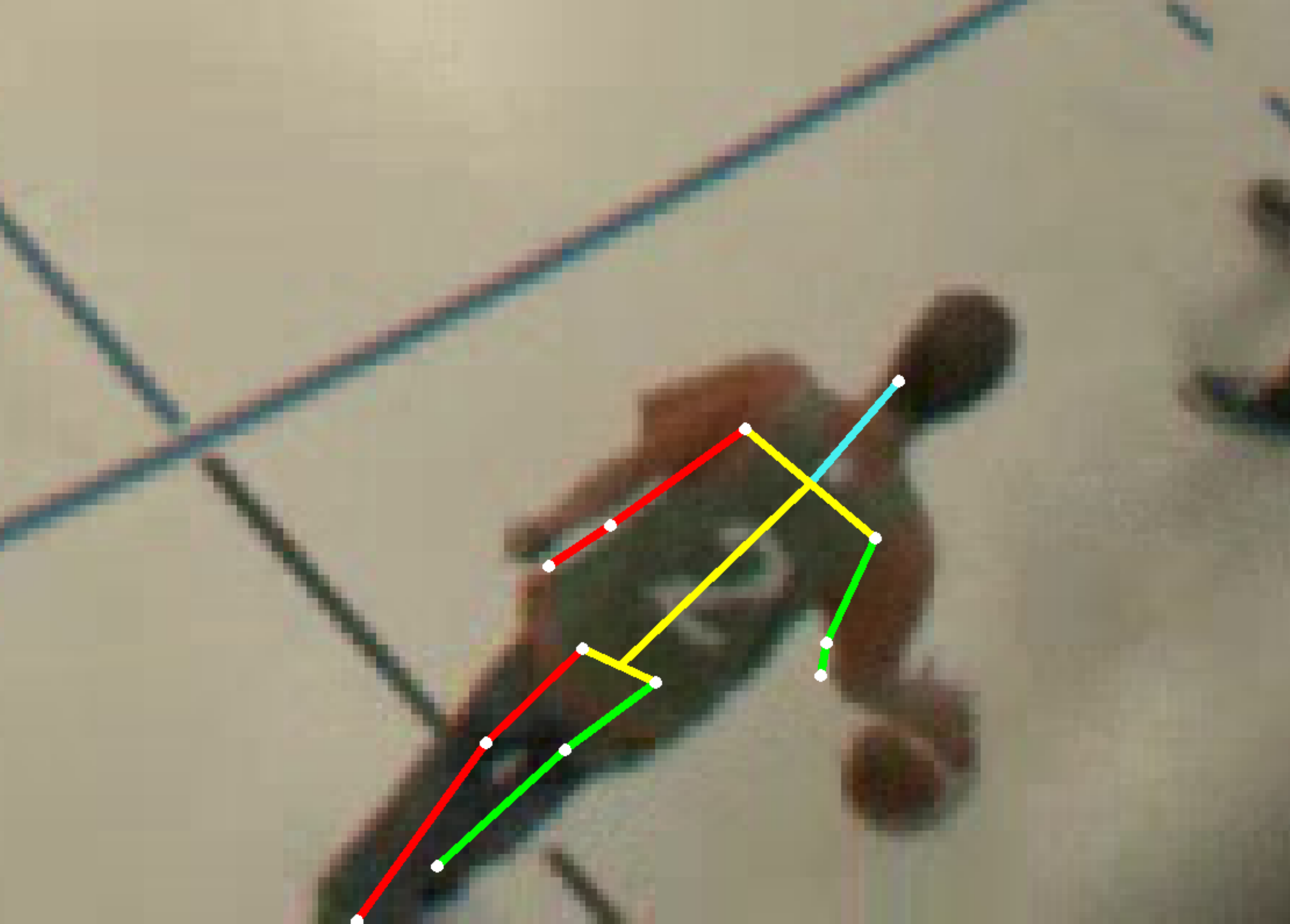}
	\end{subfigure}
	\begin{subfigure}[b]{0.24\linewidth}        
		\centering
		\includegraphics[width=\linewidth]{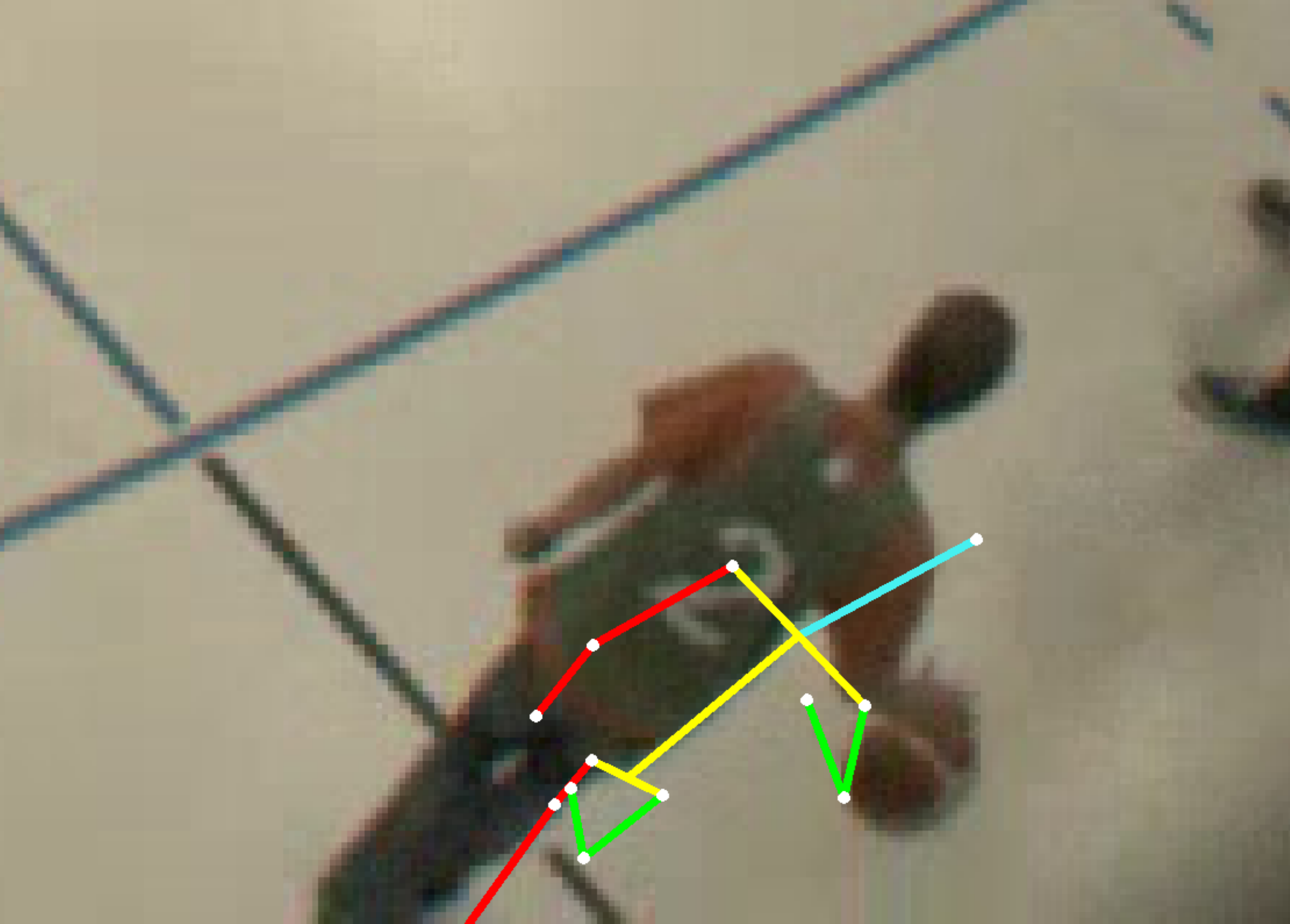}
	\end{subfigure}
	\begin{subfigure}[b]{0.24\linewidth}        
		\centering
		\includegraphics[width=\linewidth]{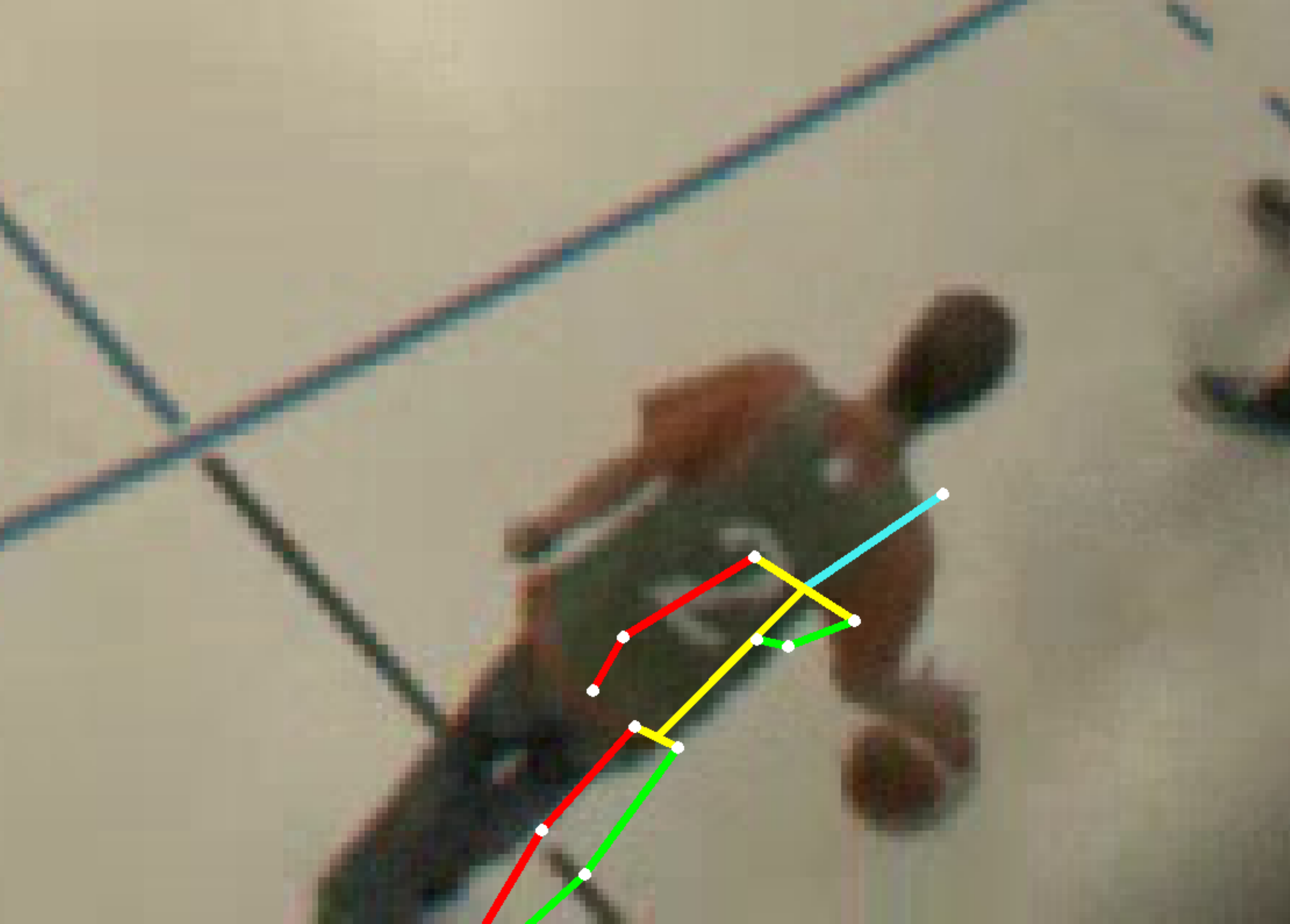}
	\end{subfigure} \\ \vspace{1mm}

	\begin{subfigure}[b]{0.24\linewidth}        
		\centering
		\includegraphics[width=\linewidth]{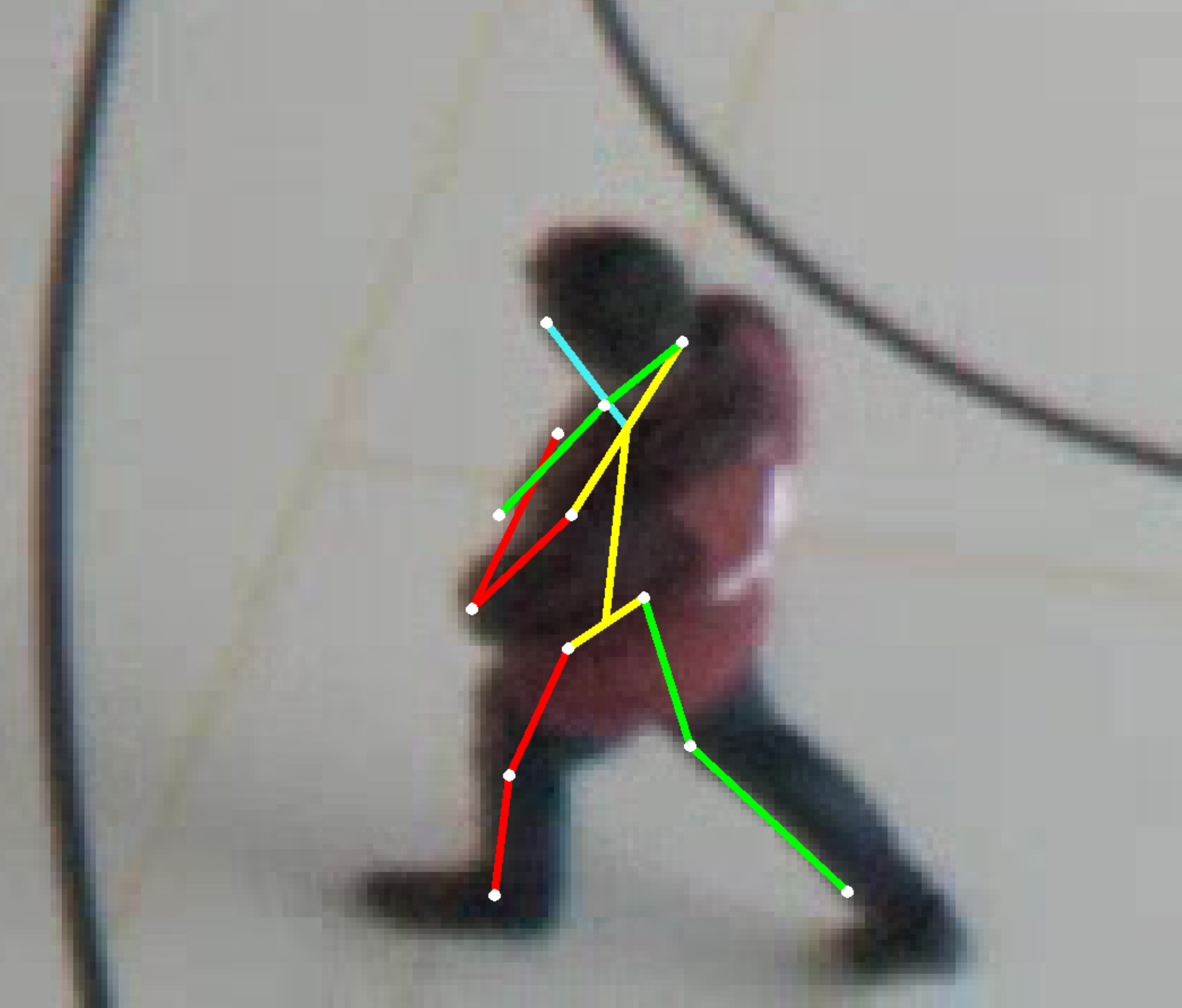}
	\end{subfigure}
	\begin{subfigure}[b]{0.24\linewidth}        
		\centering
		\includegraphics[width=\linewidth]{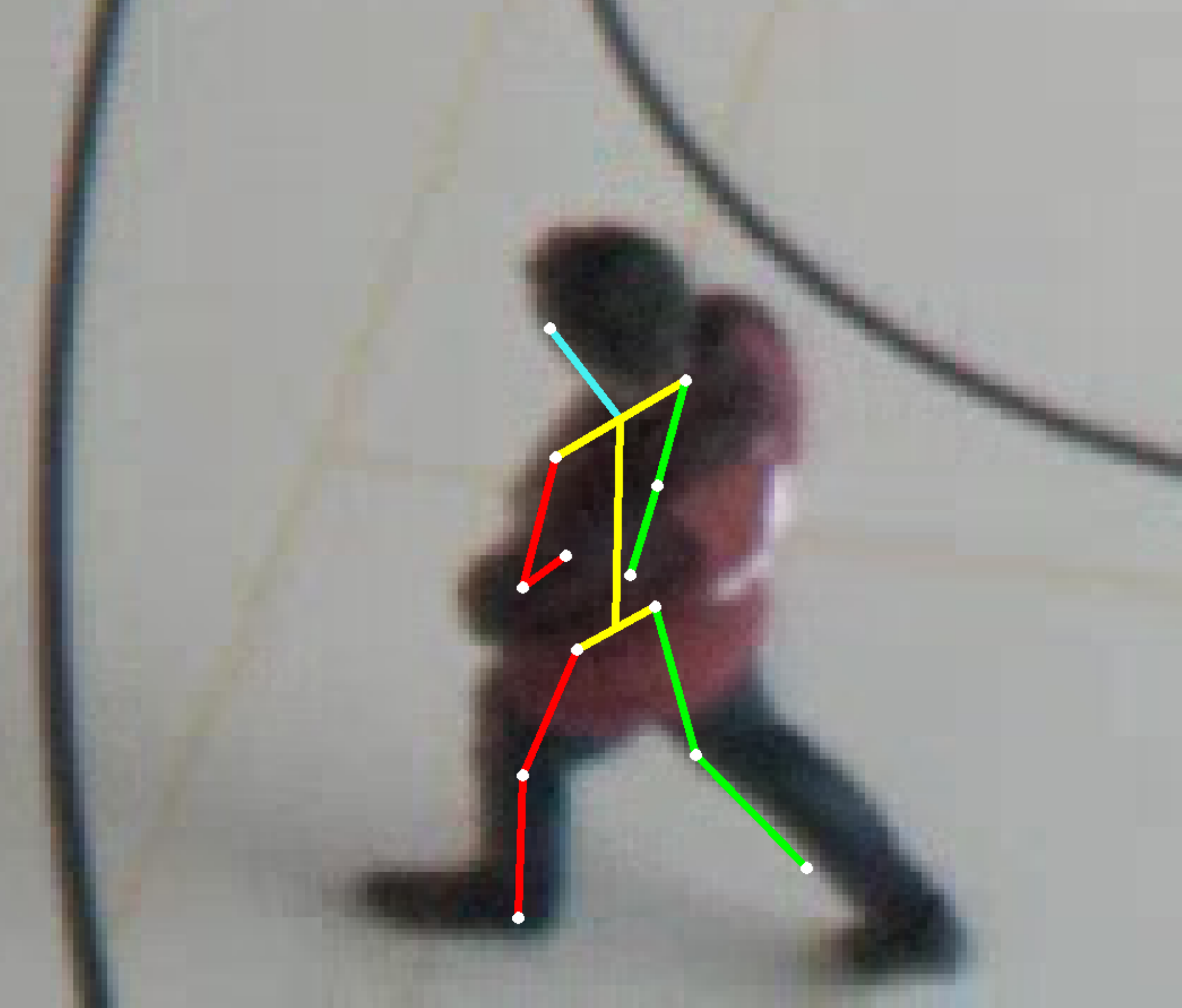}
	\end{subfigure}
	\begin{subfigure}[b]{0.24\linewidth}        
		\centering
		\includegraphics[width=\linewidth]{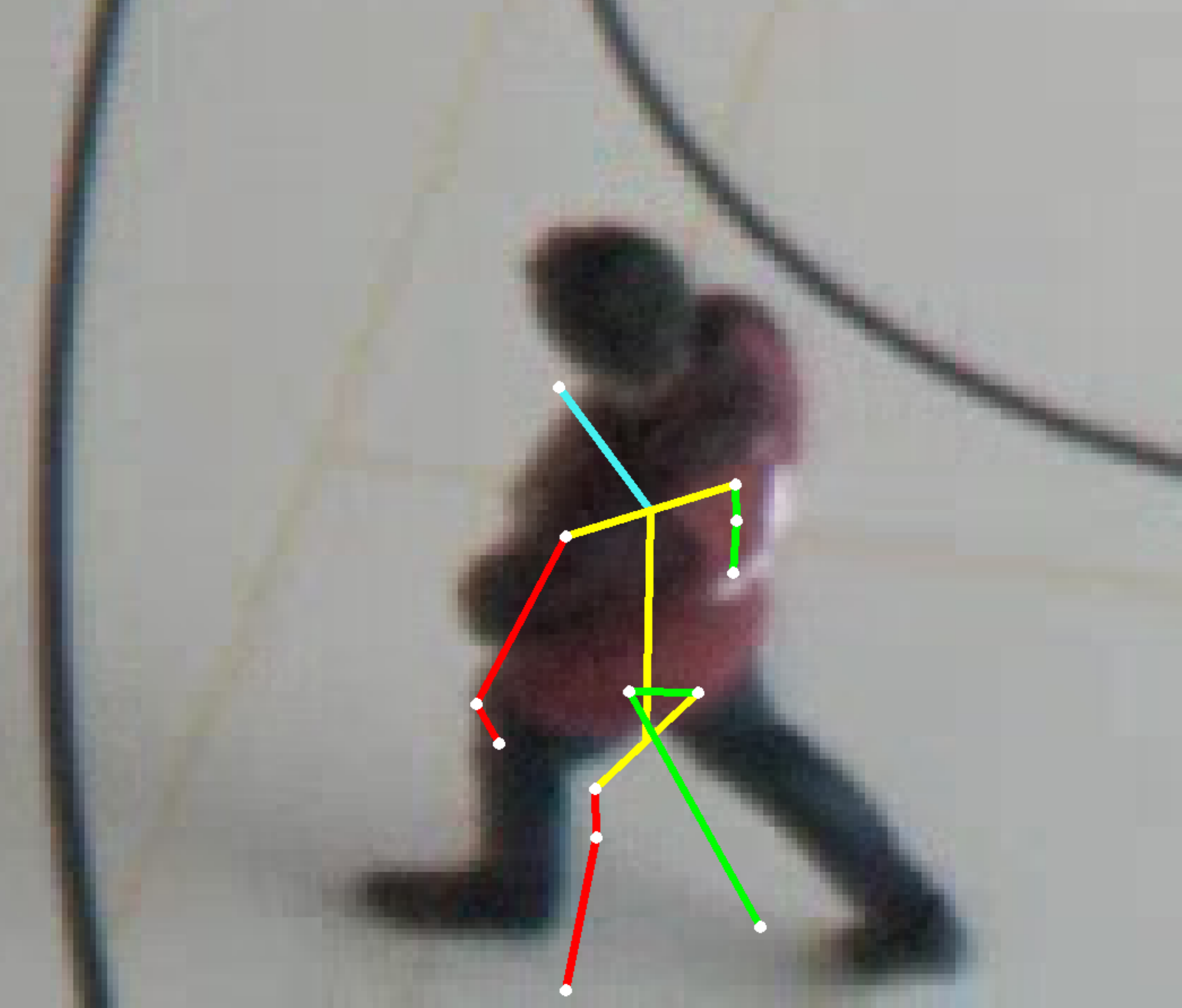}
	\end{subfigure}
	\begin{subfigure}[b]{0.24\linewidth}        
		\centering
		\includegraphics[width=\linewidth]{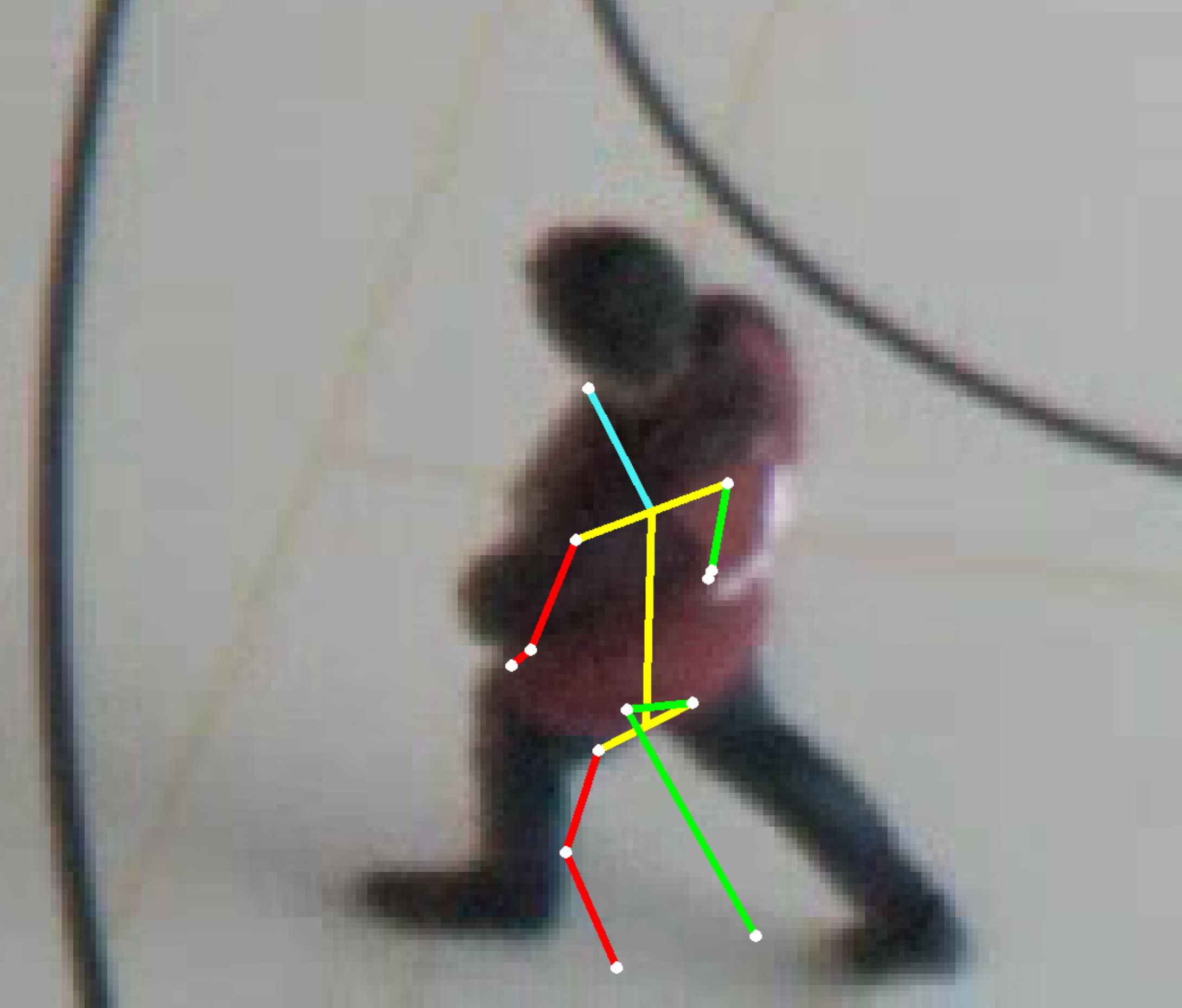}
	\end{subfigure} \\  \vspace{-1mm}
	
	\begin{subfigure}[b]{0.24\linewidth}        
		\centering
		\includegraphics[width=\linewidth]{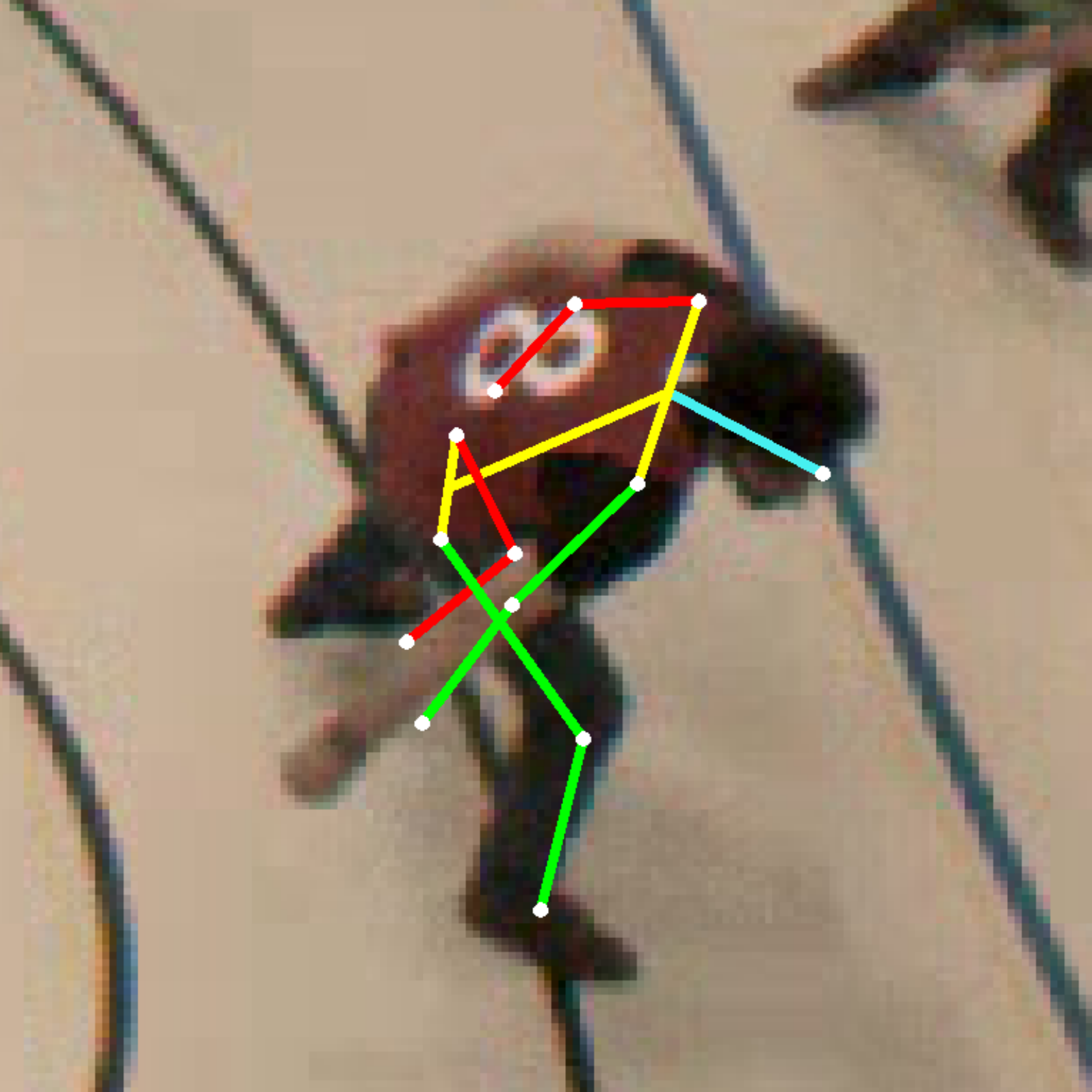}
		\caption{Ours}
	\end{subfigure}
	\begin{subfigure}[b]{0.24\linewidth}        
		\centering
		\includegraphics[width=\linewidth]{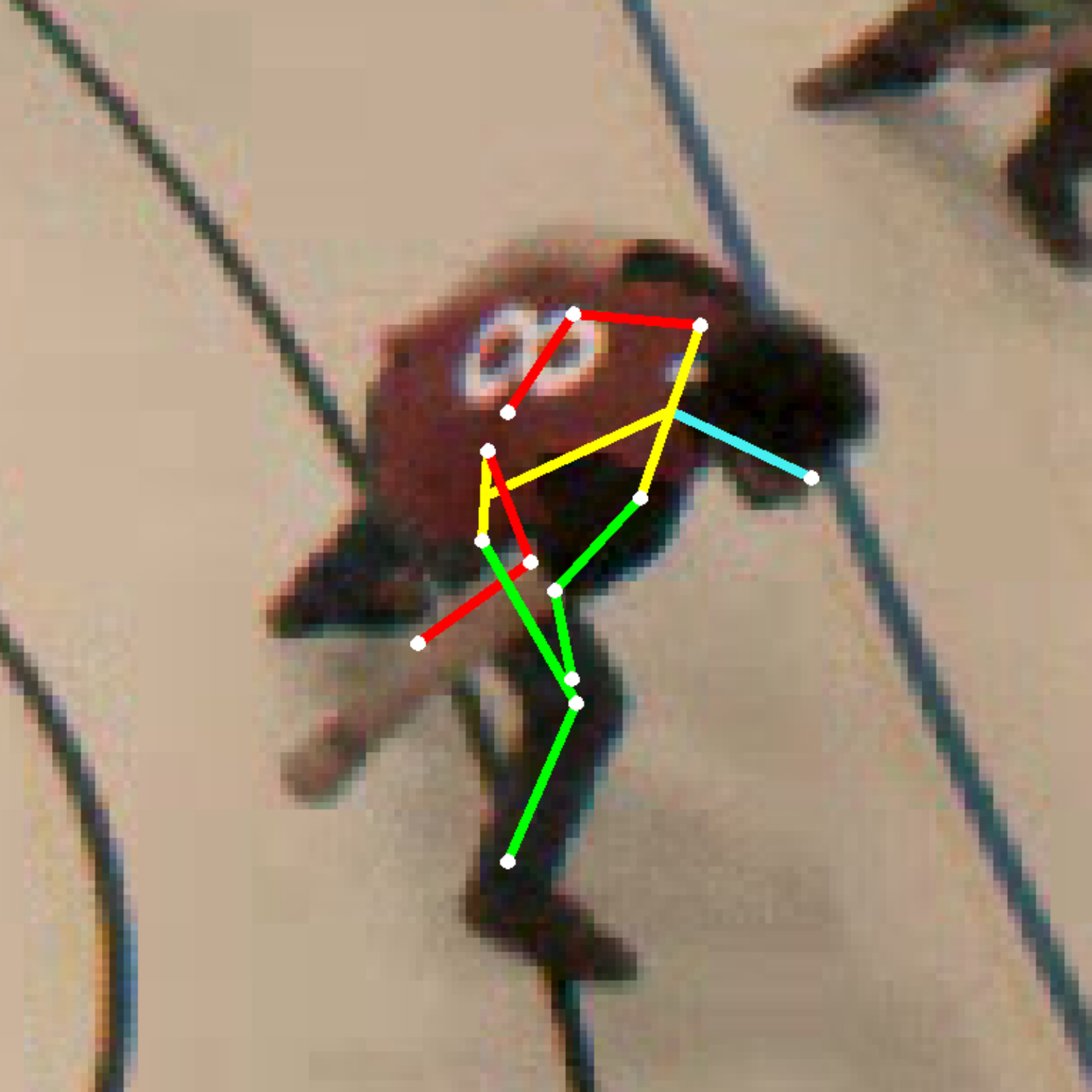}
		\caption{w/o w}
	\end{subfigure}
	\begin{subfigure}[b]{0.24\linewidth}        
		\centering
		\includegraphics[width=\linewidth]{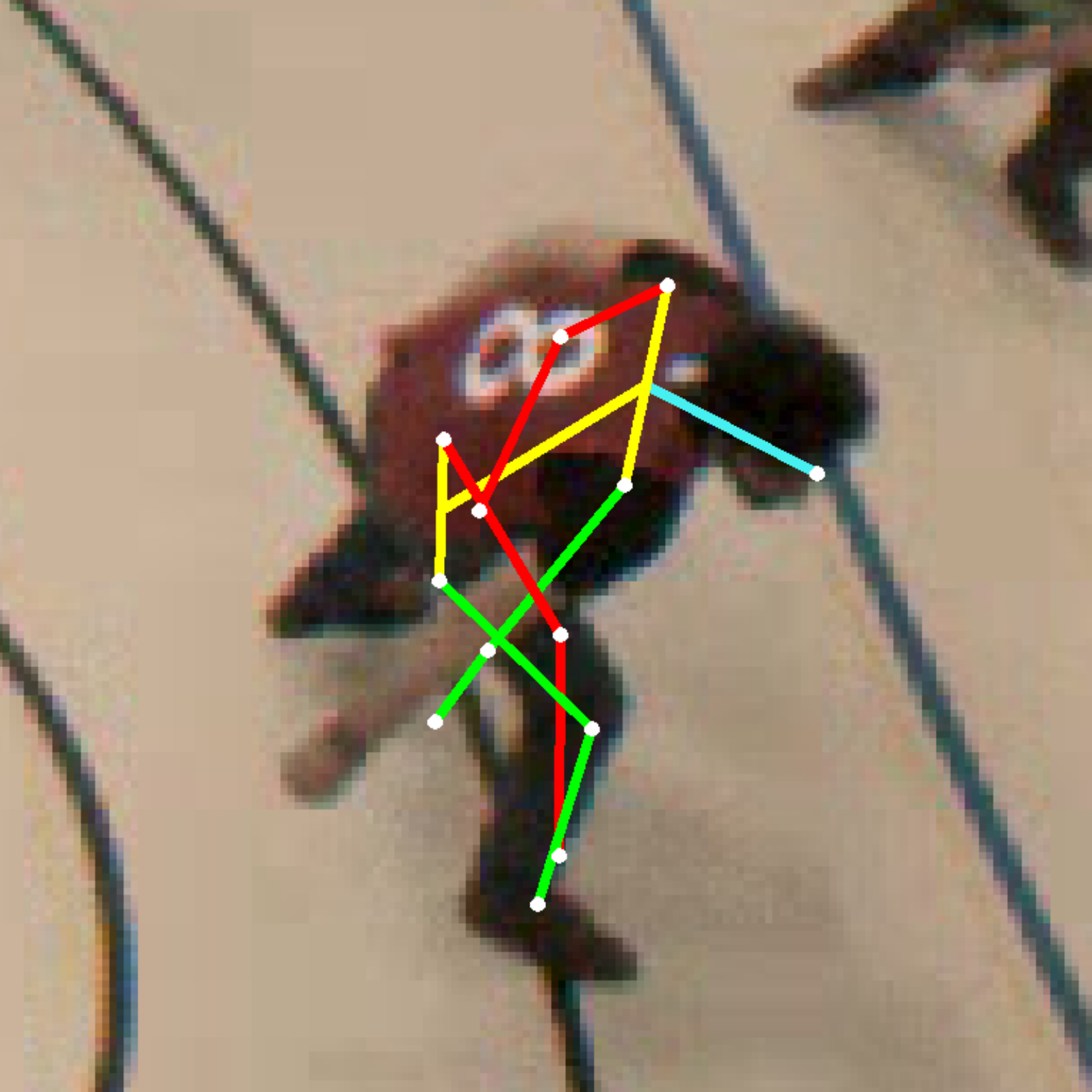}
		\caption{Ours SV}
	\end{subfigure}
	\begin{subfigure}[b]{0.24\linewidth}        
		\centering
		\includegraphics[width=\linewidth]{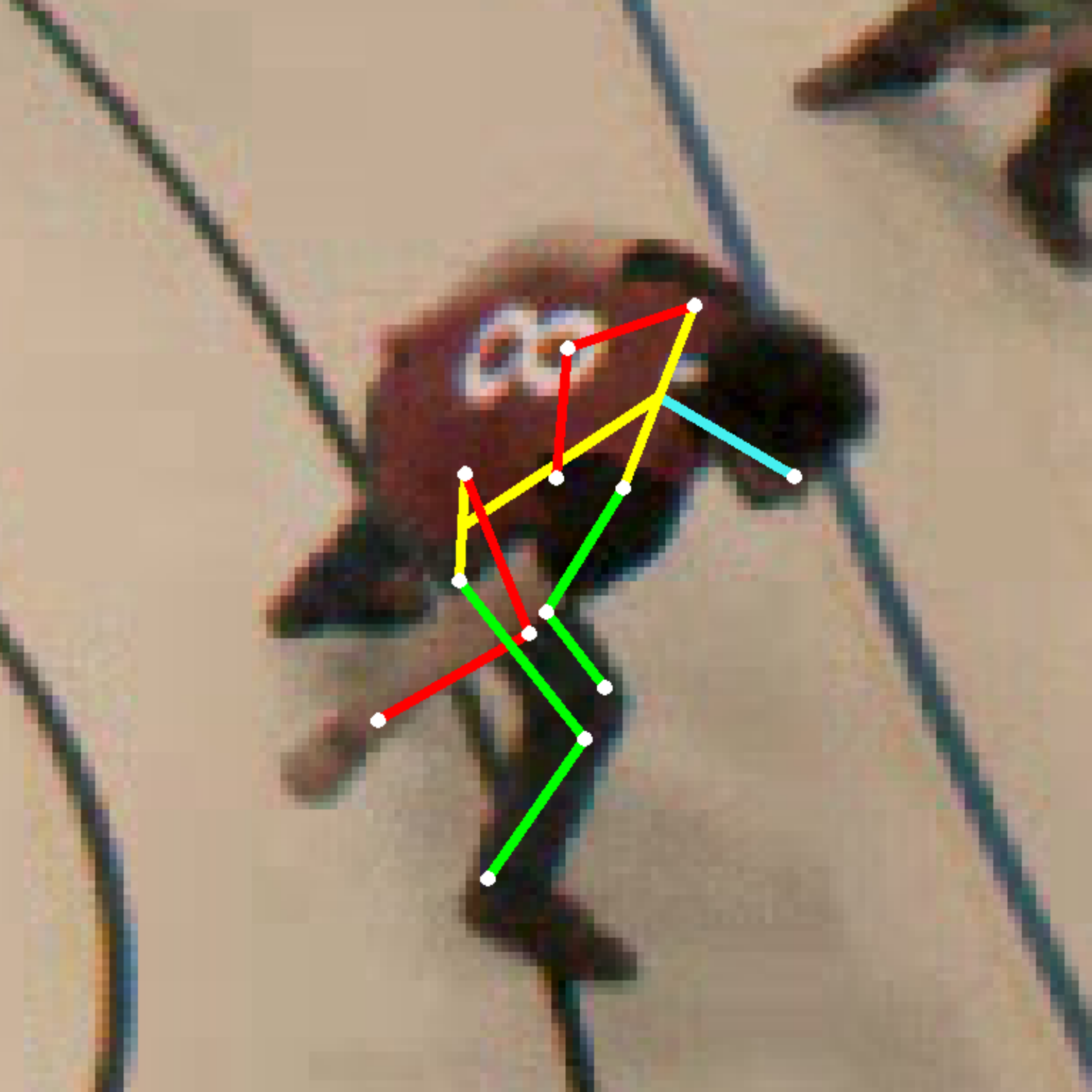}
		\caption{\small w/o w. SV}
	\end{subfigure}

	\vspace{-0.3cm}
	\caption{\small Failure cases where the lifting network $g_{\Mat{\phi}}$ fail to generate plausible 3D poses on the SportCenter dataset. From left to right, multi-view triangulated pose with (a) our approach and (b) Standard DLT (without weighting mechanism). Single view predicted results of (c) our approach and (d) without weighting.
	}
	\label{fig:occlusion_images_fail_1}
\end{figure*}

We use the same architecture for all these experiments to showcase the effect of the different weighting schemes. We report the multi-view (MV) and the single-view (SV) results in Table~\ref{tab:sports_center_semi_supervised}. Our approach performs better than {\it Ours w/o weights} because of its ability to produce reliable 3D poses from multiple 2D views even in the presence of occlusions. This is true both in the multi-view and single-view case. Using a differentiable triangulation also demonstrates to provide better supervision. Moreover, our approach also outperforms {\it Ours+Iskakov} on this task, presumably because the latter suffers from the lack of adequate amount of labeled data and the noisy 2D detections caused due the occlusions. Hence it fails to yield the optimal weights for triangulation purposes. Figure~\ref{fig:occlusion_images},~\ref{fig:occlusion_images_2} and~\ref{fig:occlusion_images_3} provide a qualitative comparison between our approach and a variant of ours with disabled weighting mechanism. It can be noted how our weighting strategy produces more precise 3D poses, which in turn provides better supervision for occluded samples. Some failure cases where the lifting network $g_{\Mat{\phi}}$ fails to generate plausible 3D poses is shown in Figure~\ref{fig:occlusion_images_fail_1}, which we aim to mitigate in the future by using a discriminator network on the predicted 3D poses by $g_{\Mat{\phi}}$ in addition to using a form of temporal consistency.

\subsection*{Triangulation Robustness Analysis}
\label{sec:triangle_robustness}
We analyze how the estimated 3D object locations, that are the results of applying triangulation, are affected by 2D localization error when comparing our triangulation approach with the standard one. To do so, we create a simulation using six views with real camera parameters. We first generate $N$ three-dimensional object locations and then obtain their corresponding ground-truth projections in each one of the views using the camera parameters. We then run experiments in which we randomly select a subset of cameras and gradually increase the amount of 2D noise applied to the projected points in these views. The noise that we apply to the 2D points is uniformly sampled on a circle of radius $r$ around the ground-truth 2D locations, where $r$ is the level of noise. For our triangulation approach with $WSS$ of Eq.~\ref{eq:WSS}, when $WSS$ is bigger then a threshold $\lambda=20$ millimeters we set all the weights to a small but equal weight. This is equivalent to using standard triangulation.
\begin{figure}[ht]
	\centering
	\includegraphics[width=0.6\textwidth]{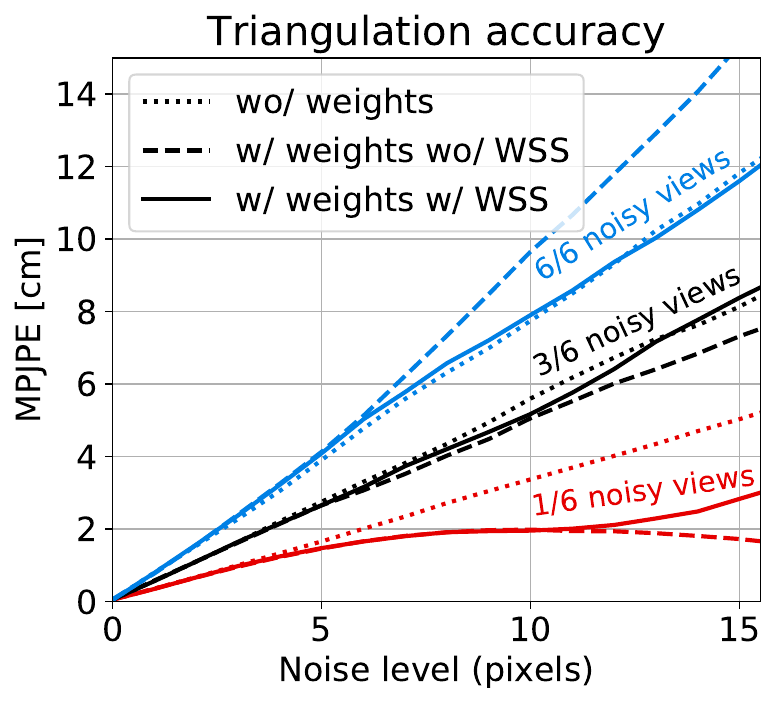}
	\caption{\small\textbf{Triangulation Error (lower is better)}. Triangulation error with respect to an increasing level of noise expressed in pixels and applied to a varying number of views. Our triangulation approach (w/ weights w/ WSS) performs better than standard triangulation (wo/ weights) when less then $50\%$ of the cameras are affected by the noise and on par otherwise. Without WSS (w/ weights wo/ WSS) instead, produces less accurate locations when most of the image points are erroneous.
	}
	\label{fig:triang_accuracy}
\end{figure}
In Figure \ref{fig:triang_accuracy} we report the MPJPE score between the triangulated results and the ground-truth object locations, where noise is applied to different number of cameras, for our approach ({\it w/ weights w/ WSS}) and variants with disabled weights ({\it wo/ weights}) and disabled $WSS$ ({\it w/ weights wo/ WSS}). Note that our triangulation produces more accurate results than the standard triangulation when less than $50 \%$ of the views are affected by the noise and on par otherwise. This is because the weights allow to reduce the impact of the noisy predictions while $WSS$ discards the joint when more than $50 \%$ of the views are unreliable.


\begin{table*}[t]
	\caption{\small Comparative study on the impact of unlabeled data for different levels of supervision~$\mL$. We report the MPJPE results in millimeters.}
	\label{tab:semi_sup_ablation}
	\scalebox{1.2}{
	\begin{tabular}{c|c|c|c|c|c|c|c}
		\toprule
	 $\mL$	& 1\% S1 & 5\% S1 & 10\% S1 & 50\% S1 & S1 & S1, S5 & S1, S5, S6\\
		\midrule
		Baseline & 135.3 & 110.5 & 94.0 & 83.4 & 83.9 & 69.2 & 65.1 \\
		\midrule
		Ours & 105.7 & 79.1 & 67.9 & 62.3 & 60.8 & 56.9 & 55.6 \\
		\bottomrule
	\end{tabular}
	}
\end{table*}

\begin{figure}
	\centering
	\includegraphics[width=0.6\textwidth]{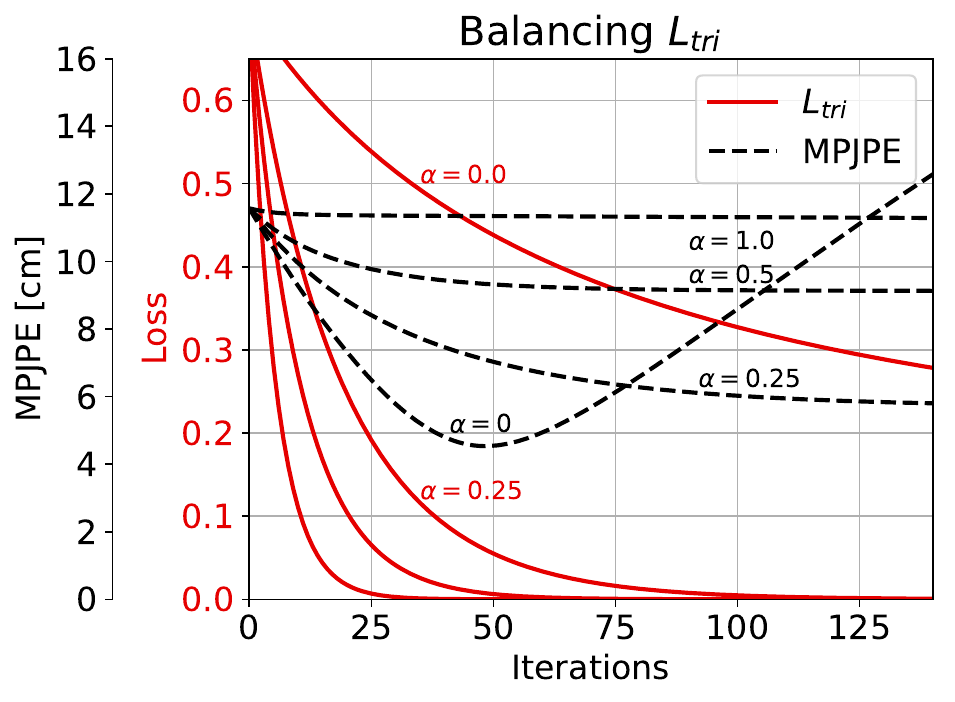}
	\captionof{figure}{\small \textbf{Balancing $L_{tri}$}. $L_{tri}$, described in Section~\textsection~\textcolor{red}{3.2.1}
	, measures the difference between the detected and re-projected 2D point estimates after triangulation. It provides two gradients flows that we control with parameter $\alpha$. MPJPE, on the other hand, shows the difference between the triangulated and ground-truth 3D points. We show the effect on the $L_{tri}$ loss and MPJPE for different values of $\alpha$ in a controlled test setup. When $\alpha=0$ the model can potentially reach the best MPJPE but then deviates quickly as the training continues. On the other hand, when $\alpha=1$ the model cannot eliminate the noise. A trade-off between the two makes the model stable while reduces the impact of noise.}
	\label{fig:balancing_Ltri}
\end{figure}

\subsection*{Stability of $L_{tri}$}
\label{sec:balancing_L_tri}

As discussed in Section~\textsection~\textcolor{red}{3.2.1}
, the self-supervised loss term $L_{tri}$ minimizes the difference between the predicted image points $\hat \bx$ and the projection of the triangulated one $\bar \bx$ using the Mean Squared Error (MSE). Since both are function of the network $f_{\theta}$, there are two gradients that affect the network weights~$\theta$. To better understand the contribution of each gradient we decompose $L_{tri}$ in two terms as shown in Eq. \ref{eq:Ltri_decomp}. In the first term, we consider $\bar \bx$ to be ground-truth data, that means that there is no gradient flowing in it. In the second term instead we consider $\hat \bx$ as to be the ground-truth. Note that, $\hat \bx$ and $\bar \bx$ are dependent on each other and can change from one iteration to another.

\begin{equation}
L_{tri}(\hat \bx, \bar \bx) = \alpha L(\hat \bx, \bar \bx_{GT})  + (1-\alpha)L(\hat \bx_{GT}, \bar \bx)
\label{eq:Ltri_decomp}
\end{equation}

When $\alpha=0$, the gradient flows only through the differentiable triangulation via $\bar \bx$. When $\alpha=1$ it flows only through the predictions $\Vec{\hat x}$, with the triangulated points considered as the ground-truth.
Since we want to exploit triangulation, setting $\alpha=0$ seems a reasonable choice, however, we demonstrate with a simulation (Figure \ref{fig:balancing_Ltri}) that it can be counterproductive.

To do so, we run an experiment in a controlled environment by isolating the self-supervised loss $L_{tri}$ from the rest of the pipeline; neural networks are not used here either. We first generate $N$ three-dimensional object locations and project them in each one of the views ($3$ in our case) to obtain ground-truth image locations $\bx$ using camera parameters. Then, to simulate the error of the 2D pose estimator, we inject a moderate amount (1 pixel) of Gaussian noise to all the image points and a higher amount (10 pixels) to one view only. Finally, we optimize the position of the noisy image locations $\hat \bx$ using $L_{tri}$ where at each iteration we compute $\bar \bx$ using the standard differentiable triangulation. The problem can be expressed as  $\hat \bx^* = \argmin_{\hat \bx} L_{tri}(\hat \bx, \bar \bx)$.

Figure \ref{fig:balancing_Ltri} depicts the loss values and the MPJPE between the triangulated points and the ground-truth ones for different values of $\alpha$. It can be noted that when $\alpha=0$ the MPJPE can reach its lowest value but, as the training continues, it deviates quickly. This is due to having the model aiming to put all the joint locations at the center of the image, which is a target that can minimize the triangulation loss without minimizing the MPJPE error. This means that minimizing the self-supervised loss does not correspond to minimizing the MPJPE. On the other hand, when $\alpha=1$ the error on the noisy image key-points is hardly reduced but is rather propagated to the other views leading to the degradation of the learnt model. When both terms in Eq.~\ref{eq:Ltri_decomp} are active, the image points that are precise deviate less and the MPJPE curve is more stable. Note that with additional supervision from labeled data all the MPJPE curves in Figure \ref{fig:balancing_Ltri} would improve over time. In our experiment we found that using $\alpha=0.5$ is a good compromise.

\subsection*{Exploiting the Unlabeled Data}
Here we study the impact of the unlabeled set of images while training the lifting network. Table~\ref{tab:semi_sup_ablation} compares the results of our semi-supervised approach ({\it Ours}) and that of a standard supervised approach ({\it Baseline}) as we progressively increase the amount of available 3D supervision on the Human3.6M~\cite{Ionescu14a} dataset for learning the lifting network $g_{\Mat{\phi}}$ in the single view setup. We first train the 2D pose estimator $f_{\Mat{\theta}}$ and then freeze its parameters and use it for both {\it Ours} and {\it Baseline}. In {\it Baseline}, we train $g_{\Mat{\phi}}$ directly on the labeled set of images $\mL$ for each setup using the following loss function:
\begin{equation}
	L^g(\phi; f_{\theta}) = L^{3D}(\phi; f_{\theta}, \mathcal{L})~~~,
\end{equation}
where $L^{3D}$ is the MSE loss between the predicted and the ground truth 3D poses on $\mL$. In {\it Ours}, in addition to the labeled set we also train the lifting network using the estimated triangulated 3D poses on the unlabeled data as in Eq.~\textcolor{red}{2} (of the main text). 
As expected, the performance of both networks improve as the amount of labeled data increases. In addition, {\it Ours} produces consistently better results than the {\it Baseline}. This proves the effectiveness of the pseudo labels generated using our proposed weighted triangulation method. 

%


\begin{table*}[t]
\caption{Quantitative results on the Second Spectrum dataset (semi-supervised) using different triangulation approaches.}
\label{tab:second_spectrum}
\begin{tabular}{c|ccc|ccc}
\toprule
		\textbf{Method} & \multicolumn{3}{c|}{\textbf{Multiview}} & \multicolumn{3}{c}{\textbf{Single View}} \\
		\cmidrule{1-7}
		& \textbf{MPJPE} & \textbf{NMPJPE} & \textbf{PMPJPE} & \textbf{MPJPE} & \textbf{NMPJPE} & \textbf{PMPJPE} \\
		\midrule
Ours w/o weights & 61.5 & 60.3 & 52.3 & 95.7 & 93.9 & 75.8 \\
Ours & \textbf{54.3} & \textbf{52.9} & \textbf{46.2} & \textbf{94.7} & \textbf{92.8} & \textbf{74.5} \\
\bottomrule
\end{tabular}
\end{table*}

\begin{table*}[t]
\caption{Quantitative reuslts on SportCenter dataset captured on a mobile camera (Single View).}
\label{tab:sports_center_semi_supervised_mobile}
\begin{tabular}{c|ccc|ccc}
\toprule
& \multicolumn{3}{c|}{Using $\Vec{\hat x}$ } & \multicolumn{3}{c}{Using $\bx$} \\
 \midrule
		\cmidrule{1-7}
	\textbf{Method} & \textbf{MPJPE} & \textbf{NMPJPE} & \textbf{PMPJPE} & \textbf{MPJPE} & \textbf{NMPJPE} & \textbf{PMPJPE} \\
		\midrule
Ours w/o weights & 199 & 197.3 & 140 & 138.3 & 138.4 & 88.1 \\
Ours + Iskakov~\cite{Iskakov19} & 197.7  & 193.7 & \textbf{130.1} & 131.8 & 132.5 & \textbf{82.8} \\
Ours & \textbf{188.9} & \textbf{186.2} & 130.5 & \textbf{129.1} & \textbf{126.8} & 85.6 \\

\bottomrule
\end{tabular}
\end{table*}

\subsection{Second Spectrum Dataset}
Here we evaluate and compare our proposed method against the \emph{Standard DLT} on a basketball dataset provided by Second Spectrum. The dataset consists of $15$ players and $3$ referees playing a game of basketball at the National Basketball Association (NBA). The videos are captured using $8$ calibrated cameras, with $4$ placed in each half of the court. We have considered $2$ (out of $15$) players and $1$ referee (out of $3$) to be the test set, while the remaining $13$ players and $2$ referees are considered as the train set. Each subject has its annotated 2D and 3D ground truth poses. In the semi-supervised learning setup, we have considered $2,160$ annotated images as the labeled set~$\mL$ and the unlabeled set~$\mU$ consists of $19,865$ images. The results for multiview and single view experiments are shown in Table~\ref{tab:second_spectrum}. As seen, we consistently outperform the \emph{Ours w/o weights} across all the three evaluation metrics for both the multiview and the single view experiments; thereby clearly demonstrating the importance and effectiveness of our proposed weighting strategy in handling noisy 2D detections and producing reliable triangulated 3D poses.

\subsection{Mobile Camera}
In this section, we evaluate and compare our proposed methods against (a) \emph{Ours w/o weights} and
(b) {\it Ours+Iskakov}~\cite{Iskakov19} on a video sequence of SportCenter dataset captured on a single mobile camera. The images are captured using a calibrated Iphone6. Like before, we evaluate the learnt models of Table~\ref{tab:sports_center_semi_supervised} on two test subjects, thereby resulting in a test set of $38$ images. The results using the predictions of the learnt 2D pose estimator model $f_{\theta}$ (\ie~$\Vec{\hat x}$) and the 2D ground truth annotations (\ie~$\bx$) are shown in Table~\ref{tab:sports_center_semi_supervised_mobile}. As expected, there exists a significant performance gap($\sim$ 60 mm) between using $\Vec{\hat x}$ and $\bx$ as the input to the lifting network $g_{\Mat{\phi}}$. We believe this gap in performance can be bridged provided sufficient amount of data obtained using a mobile camera is available to improve the performance of $f_{\theta}$. Having said that, our proposed method still remains the best performing method against the other two baseline methods which clearly shows the effectiveness of our weighting strategy to generate suitable triangulated 3D poses.


\section*{Qualitative Results}
\begin{figure*}[t]
	\begin{tabular}{l|l|l|l|l}
	
		\begin{tikzpicture}
		    \draw (0, 0) node[inner sep=0] {\includegraphics[width=0.16\linewidth, trim={5.3cm 0.5cm 5.3cm 0.3cm}, clip]{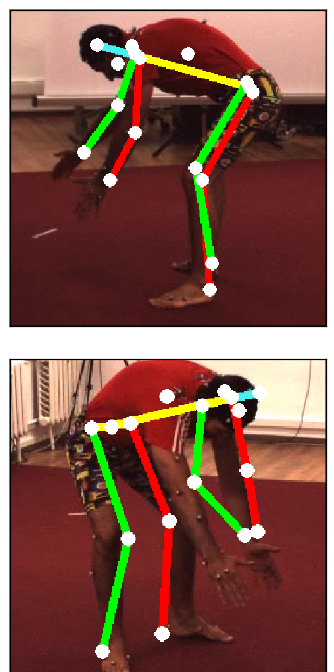}};
		    \draw (-0.98, 2.4) node [fill=white] {a)};
		    \draw (-0.98, -0.5) node [fill=white] {b)};
		\end{tikzpicture}	

		& 
		\includegraphics[width=0.16\linewidth, trim={5.3cm 0.5cm 5.3cm 0.3cm}, clip]{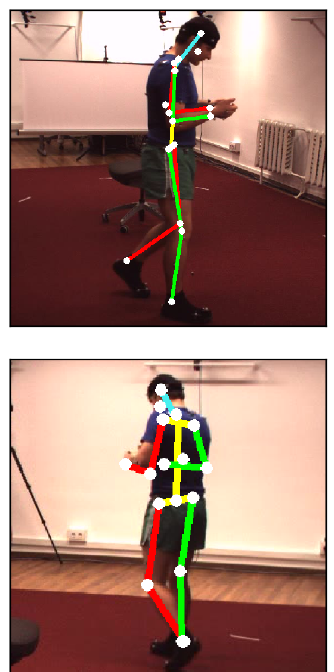} 
		&
		\includegraphics[width=0.16\linewidth, trim={5.3cm 0.5cm 5.3cm 0.3cm}, clip]{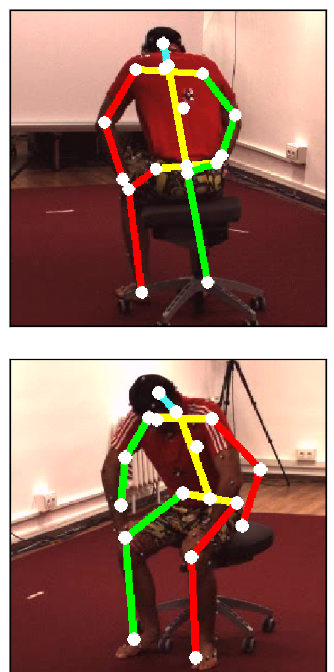} 
		&
		\includegraphics[width=0.16\linewidth, trim={5.3cm 0.5cm 5.3cm 0.3cm}, clip]{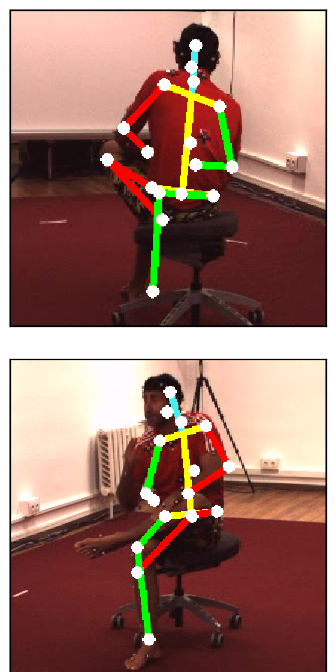} 
		&
		\includegraphics[width=0.16\linewidth, trim={5.3cm 0.5cm 5.3cm 0.3cm}, clip]{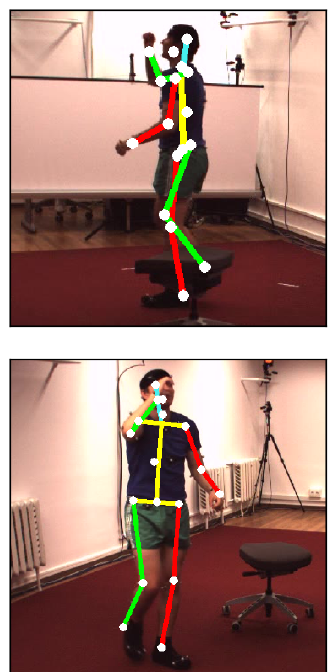} 
		\\
		\midrule \midrule

			\begin{tikzpicture}
				\draw (0, 0) node[inner sep=0] {\includegraphics[width=0.16\linewidth, trim={5.3cm 0.5cm 5.3cm 0.3cm}, clip]{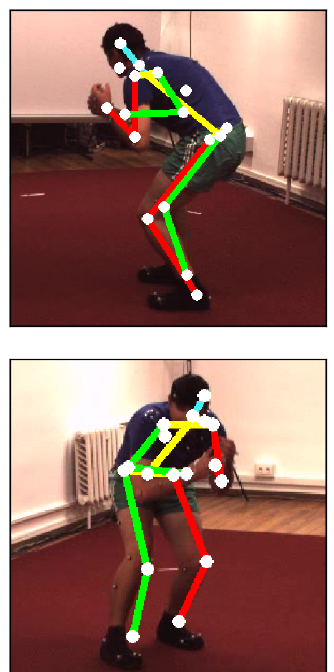} };
				 \draw (-0.98, 2.4) node [fill=white] {a)};
				  \draw (-0.98, -0.5) node [fill=white] {b)};
			\end{tikzpicture}        

		& 
		\includegraphics[width=0.16\linewidth, trim={5.3cm 0.5cm 5.3cm 0.3cm}, clip]{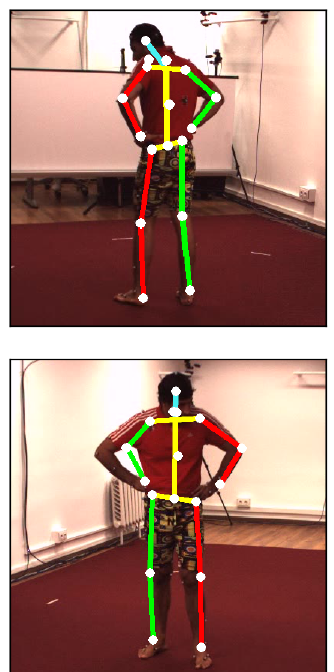} 
		& 
		\includegraphics[width=0.16\linewidth, trim={5.3cm 0.5cm 5.3cm 0.3cm}, clip]{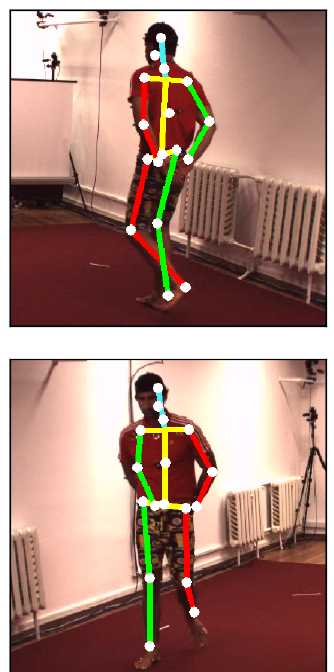} 
		&
		\includegraphics[width=0.16\linewidth, trim={5.3cm 0.5cm 5.3cm 0.3cm}, clip]{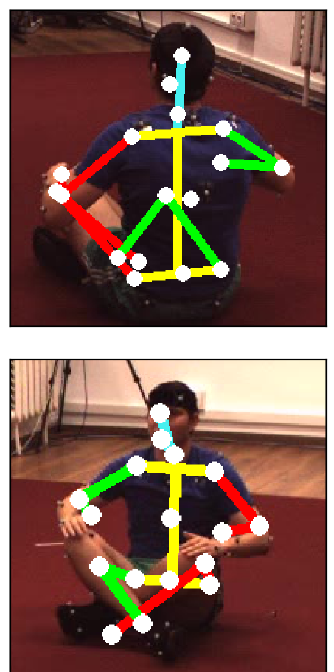} 
		&
		\includegraphics[width=0.16\linewidth, trim={5.3cm 0.5cm 5.3cm 0.3cm}, clip]{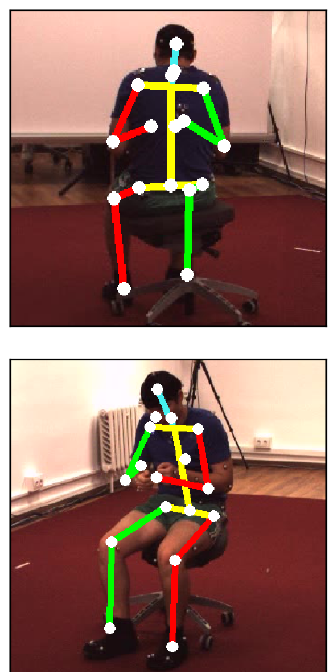}
		
	\end{tabular}
	\vspace{-2mm}
    \caption{\small {\bf Qualitative results for the test set of Human3.6M.} The superimposed pose in the top image (a) corresponds to the projection of the 3D prediction in this view from our single view model trained on 10\% of labeled data. This projected pose is equivalent to the prediction of the 2D pose estimator. The bottom image (b) depicts the same 3D pose but from another viewpoint. This allow us to visualise the depth information, which would be not visible otherwise.}
    \label{fig:visuals_h36m2}
\end{figure*}
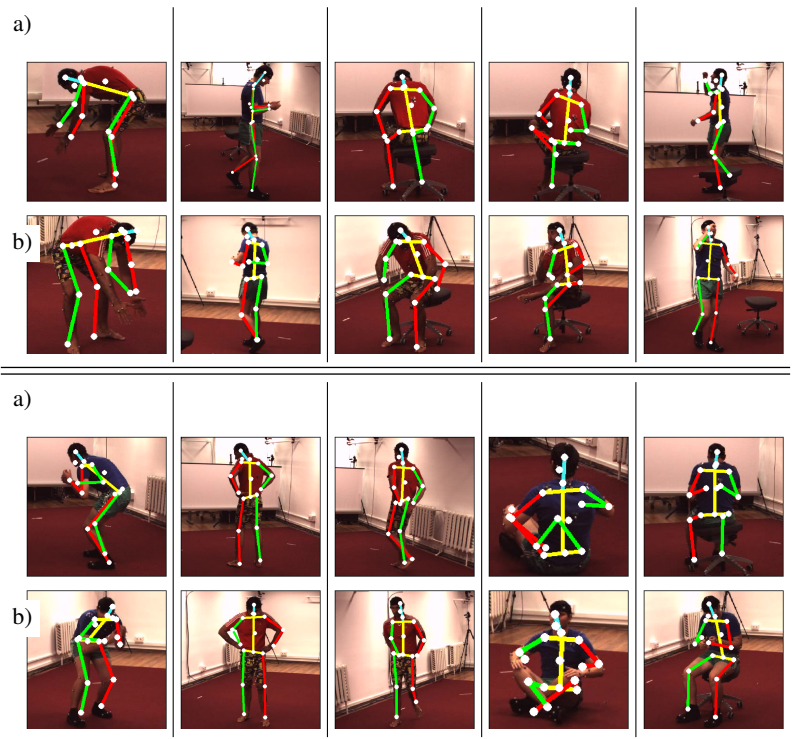

\begin{figure*}[t]
	\begin{tabular}{llll}
		\includegraphics[width=0.23\linewidth, trim={4.3cm 2.5cm 3.3cm 2.3cm}, clip]{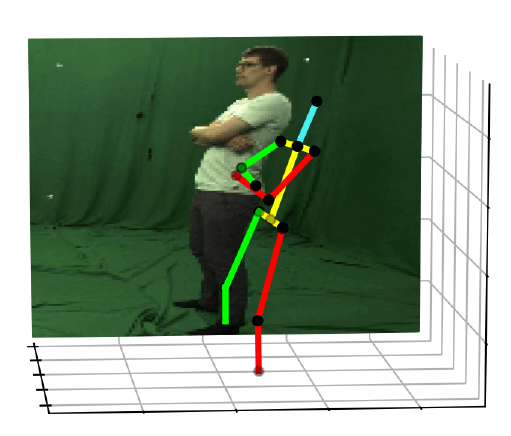} 
		& 
		\includegraphics[width=0.23\linewidth, trim={4.3cm 2.5cm 3.3cm 2.3cm}, clip]{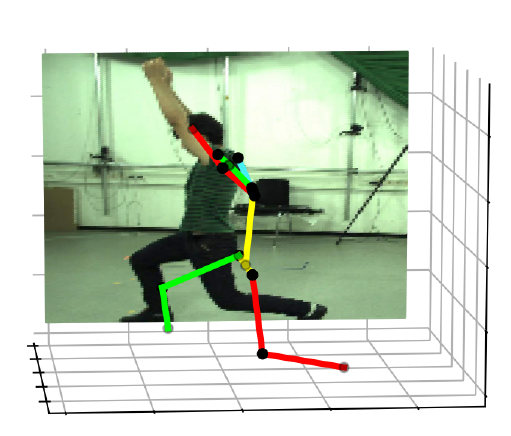} 
		&
		\includegraphics[width=0.23\linewidth, trim={4.3cm 2.5cm 3.3cm 2.3cm}, clip]{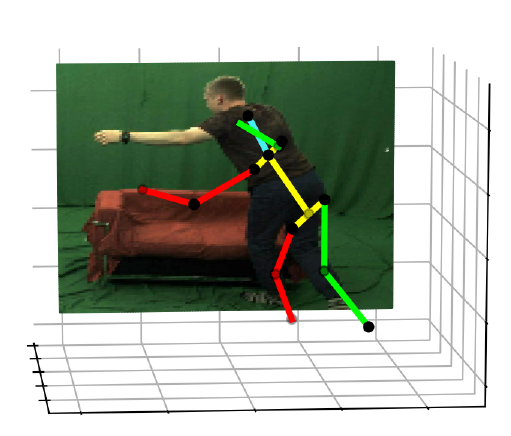} 
		&
		\includegraphics[width=0.23\linewidth, trim={4.3cm 2.5cm 3.3cm 2.3cm}, clip]{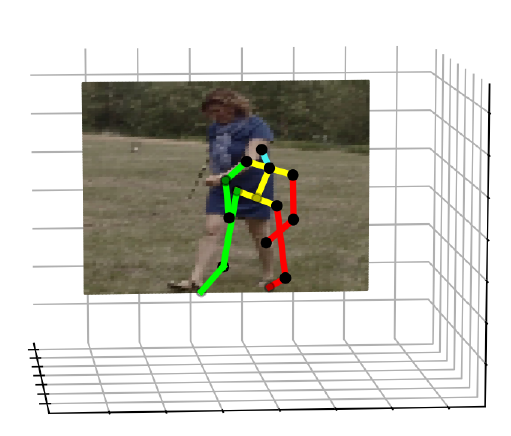} \\
		
		\includegraphics[width=0.23\linewidth, trim={4.3cm 2.5cm 3.3cm 2.3cm}, clip]{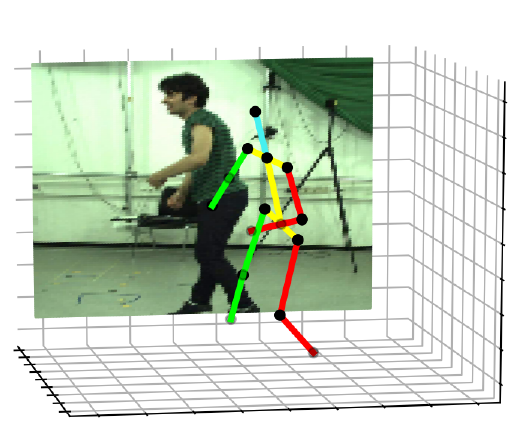} 
		& 
		\includegraphics[width=0.23\linewidth, trim={4.3cm 2.5cm 3.3cm 2.3cm}, clip]{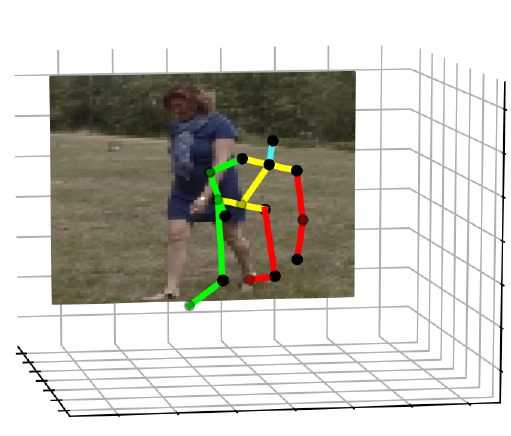} 
		&
		\includegraphics[width=0.23\linewidth, trim={4.3cm 2.5cm 3.3cm 2.3cm}, clip]{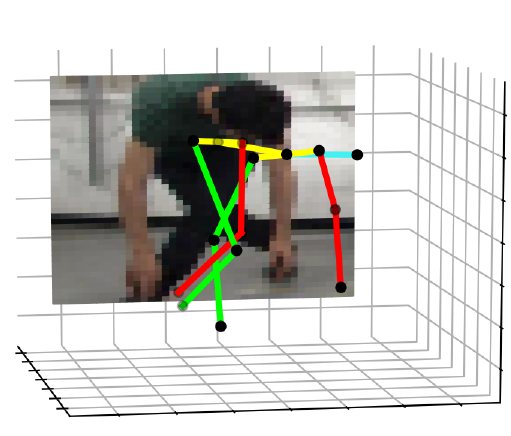} 
		&
		\includegraphics[width=0.23\linewidth, trim={4.3cm 2.5cm 3.3cm 2.3cm}, clip]{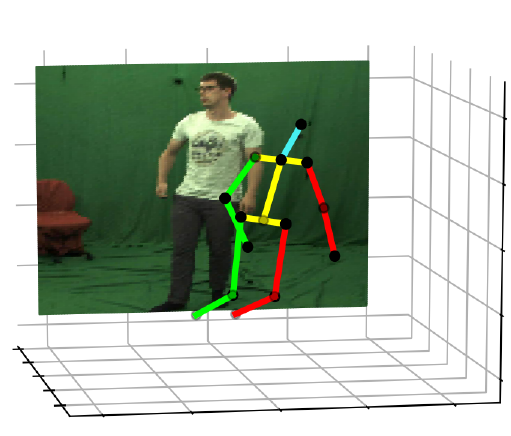} \\

		\includegraphics[width=0.23\linewidth, trim={4.3cm 2.5cm 3.3cm 2.3cm}, clip]{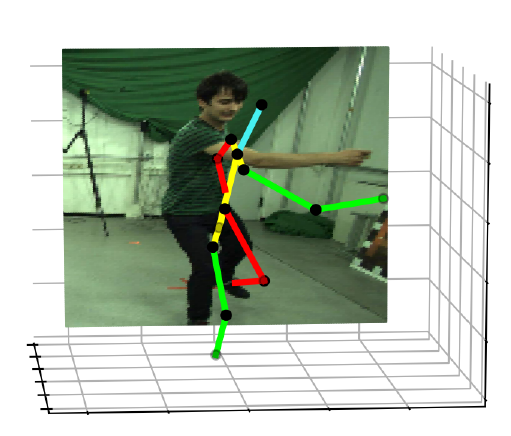} 
		& 
		\includegraphics[width=0.23\linewidth, trim={4.3cm 2.5cm 3.3cm 2.3cm}, clip]{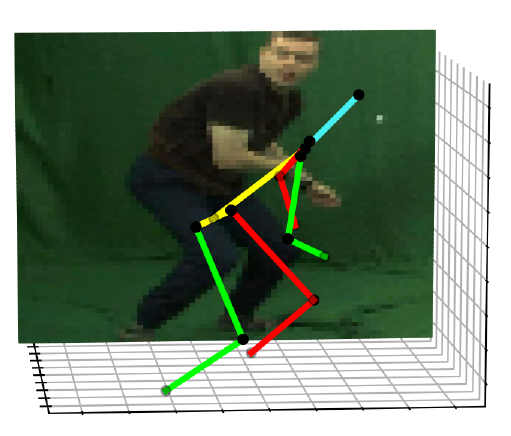} 
		&
		\includegraphics[width=0.23\linewidth, trim={4.3cm 2.5cm 3.3cm 2.3cm}, clip]{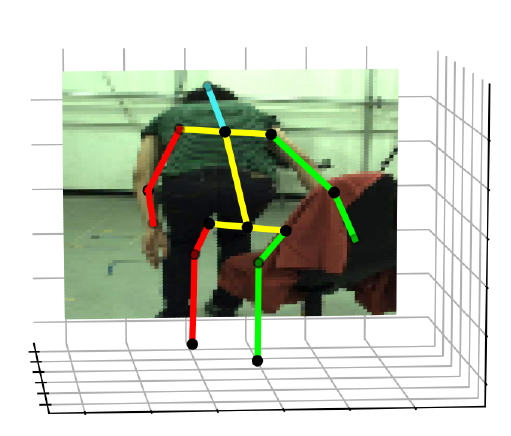} 
		&
		\includegraphics[width=0.23\linewidth, trim={4.3cm 2.5cm 3.3cm 2.3cm}, clip]{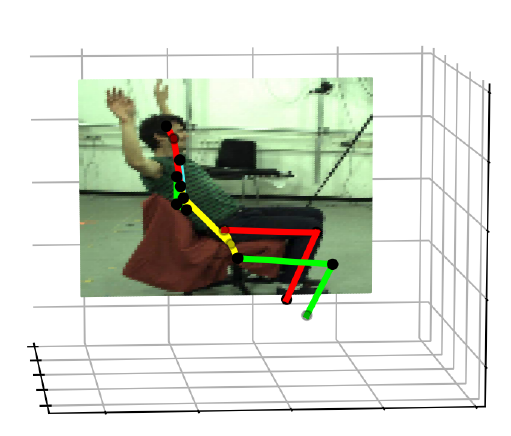}
		
	\end{tabular}
	\vspace{-2mm}
    \caption{\small {\bf Qualitative results for the test set of MPI-INF-3DHP.} We show the single view 3D predictions of our model trained under the semi-supervised setting.}
    \label{fig:visuals_3dhp}
\end{figure*}

\begin{figure*}[t!]
    \centering
    \begin{subfigure}[b]{0.16\linewidth}        
        \centering

\begin{tikzpicture}
    \draw (0, 0) node[inner sep=0] {\includegraphics[width=\linewidth]{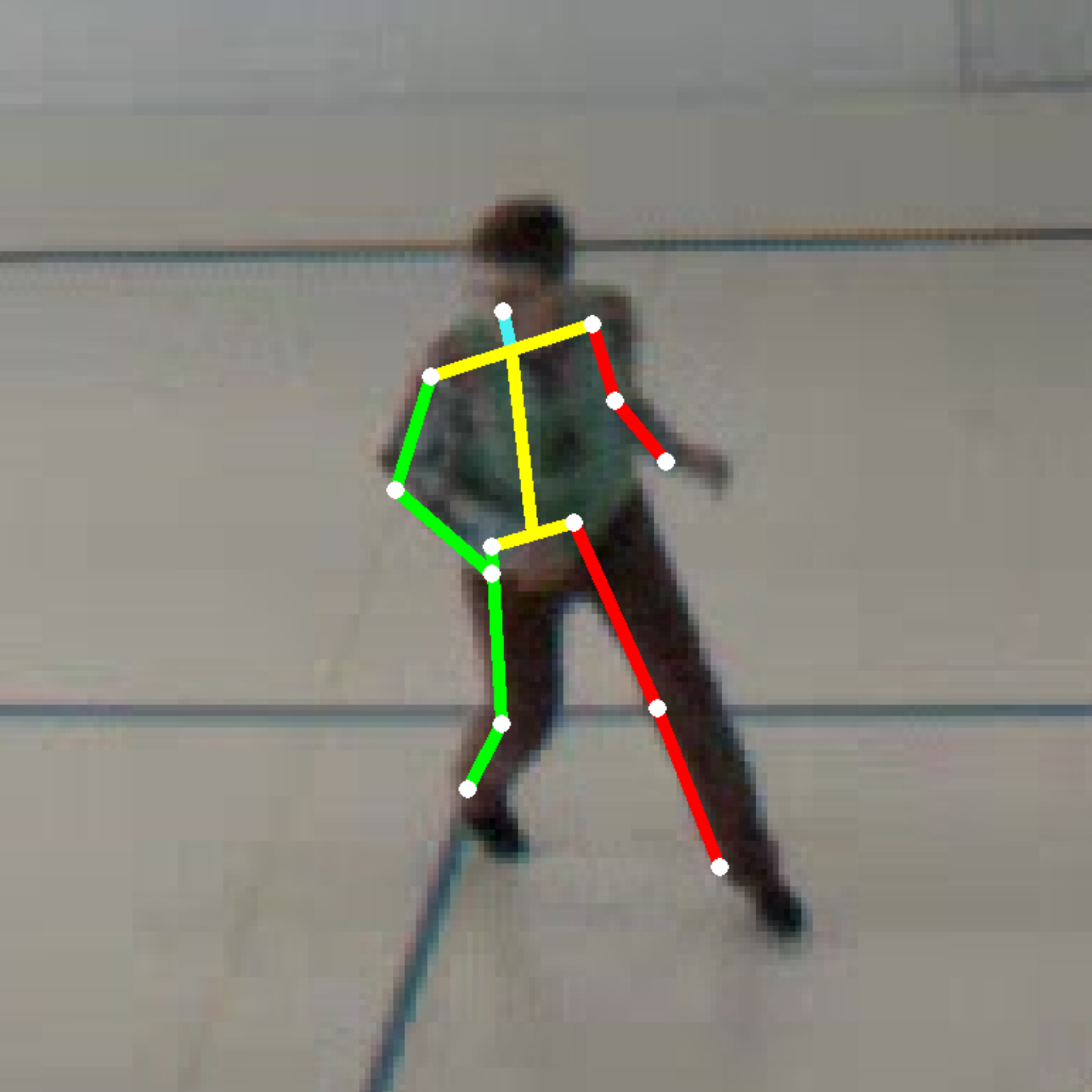}};
    \draw (-0.98, 1) node [fill=white] {a)};
\end{tikzpicture}        
        
    \end{subfigure}
    \begin{subfigure}[b]{0.16\linewidth}        
        \centering
        \includegraphics[width=\linewidth]{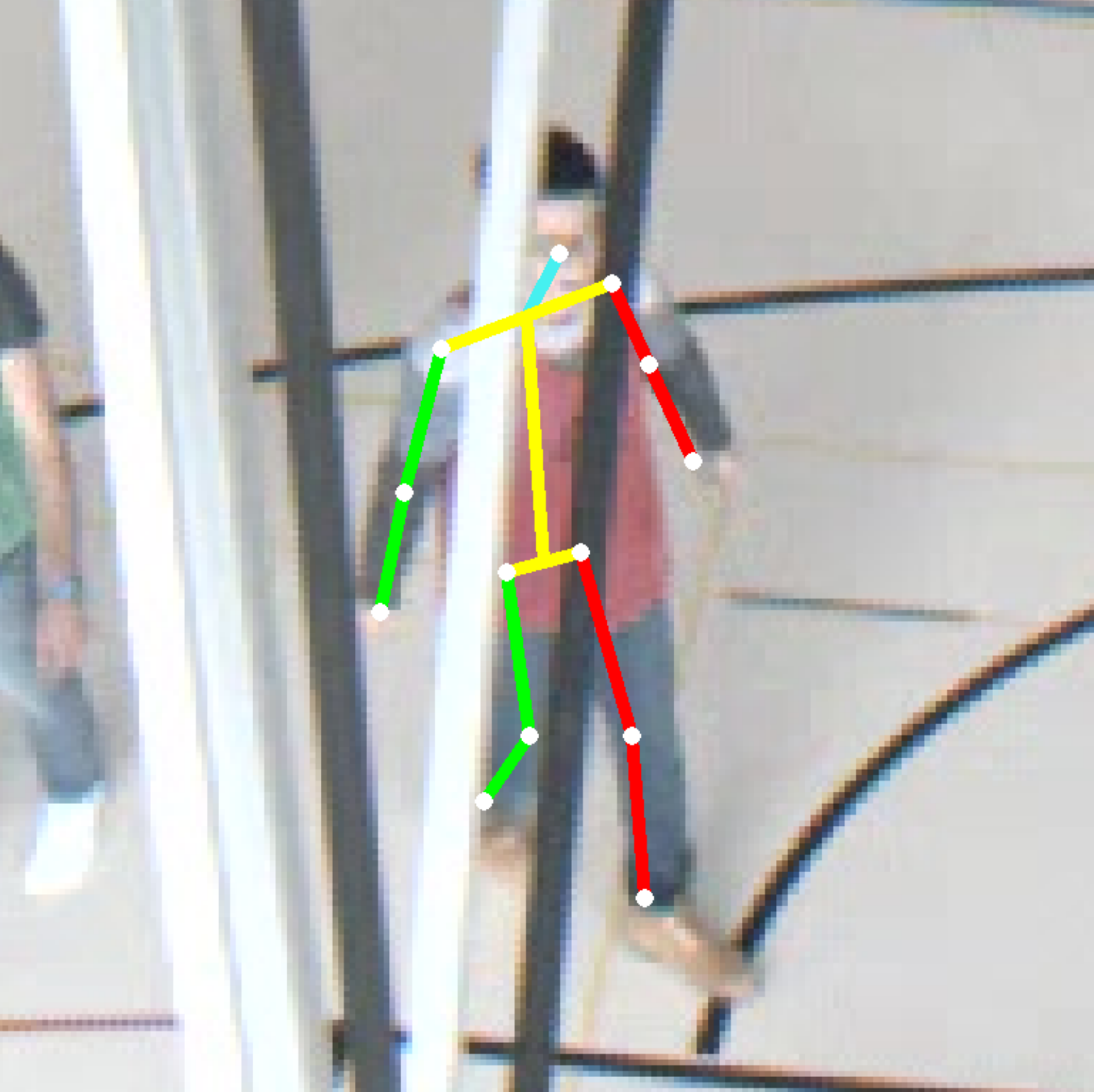}
    \end{subfigure}
    \begin{subfigure}[b]{0.16\linewidth}        
        \centering
        \includegraphics[width=\linewidth]{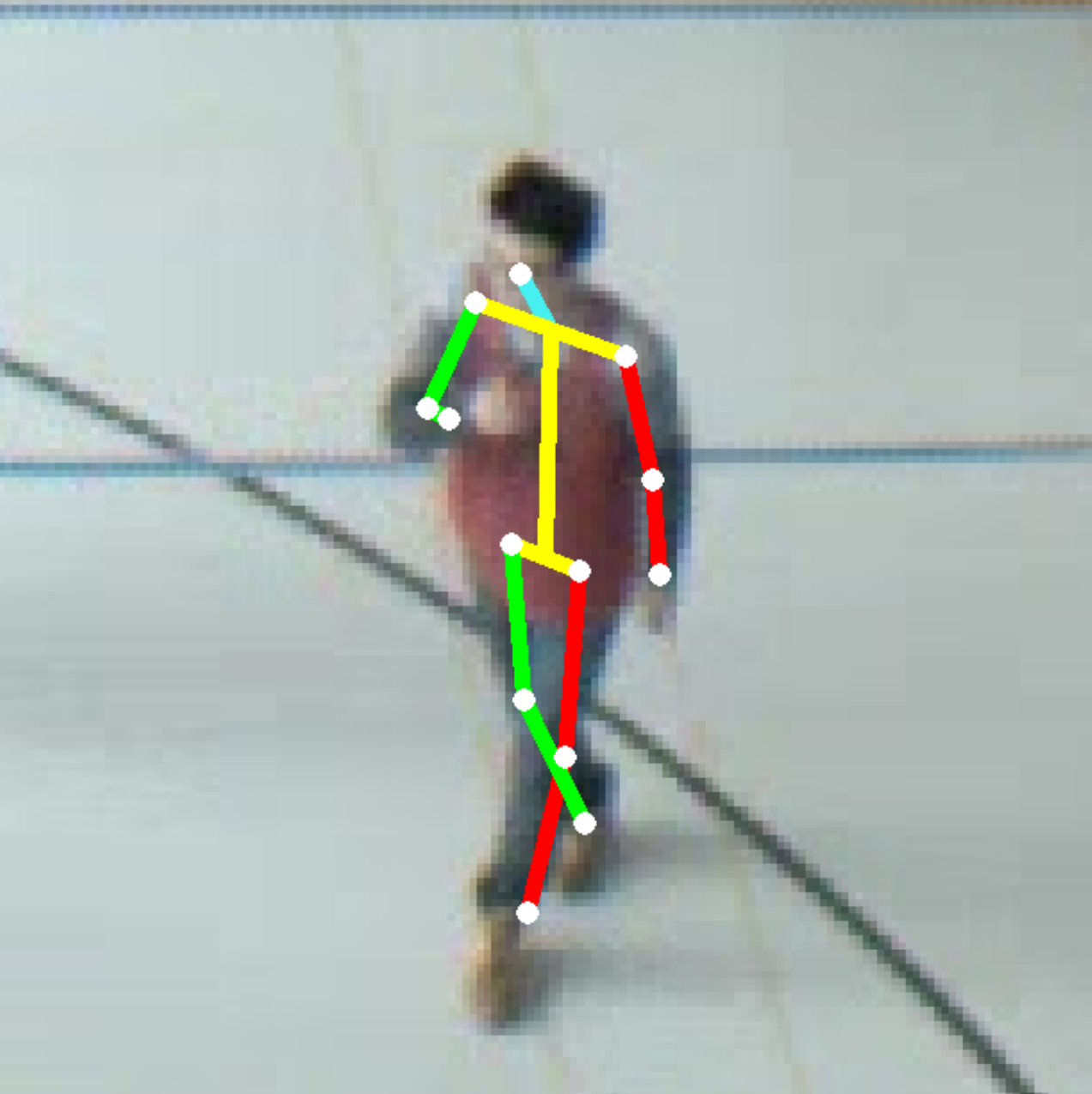}
    \end{subfigure}
    \begin{subfigure}[b]{0.16\linewidth}        
        \centering
        \includegraphics[width=\linewidth]{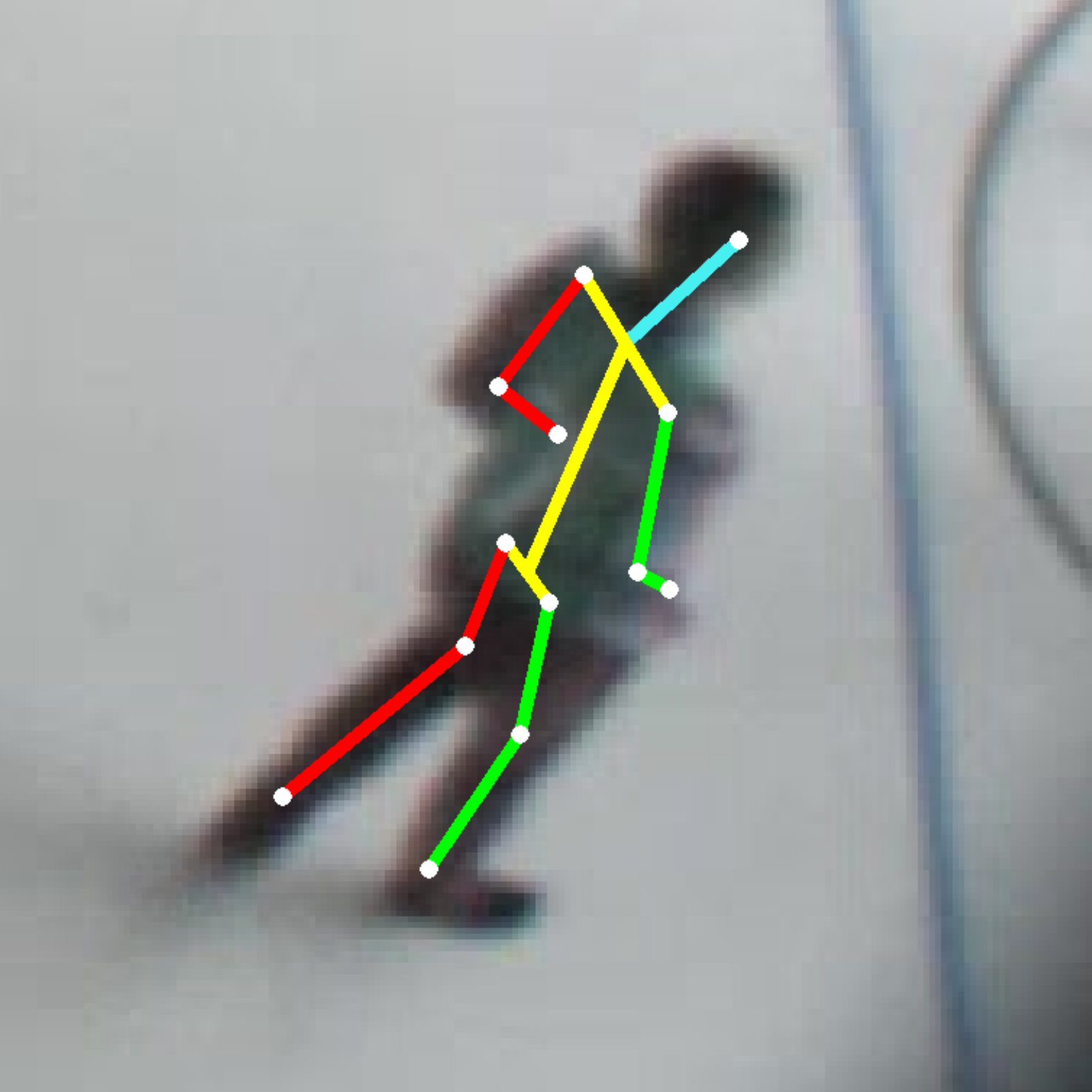}
    \end{subfigure}
    \begin{subfigure}[b]{0.16\linewidth}        
        \centering
        \includegraphics[width=\linewidth]{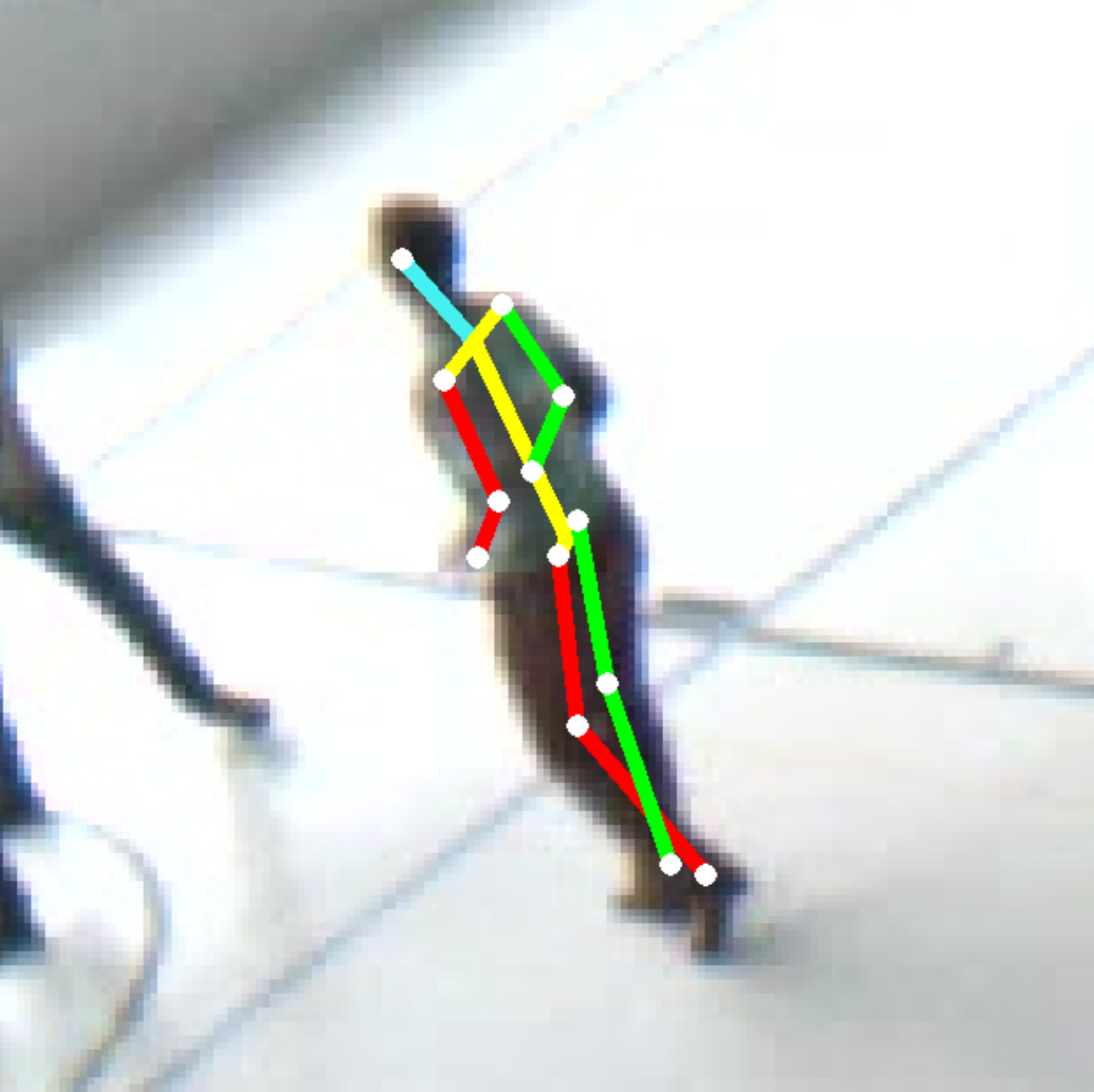}
    \end{subfigure}
    \begin{subfigure}[b]{0.16\linewidth}        
        \centering
        \includegraphics[width=\linewidth]{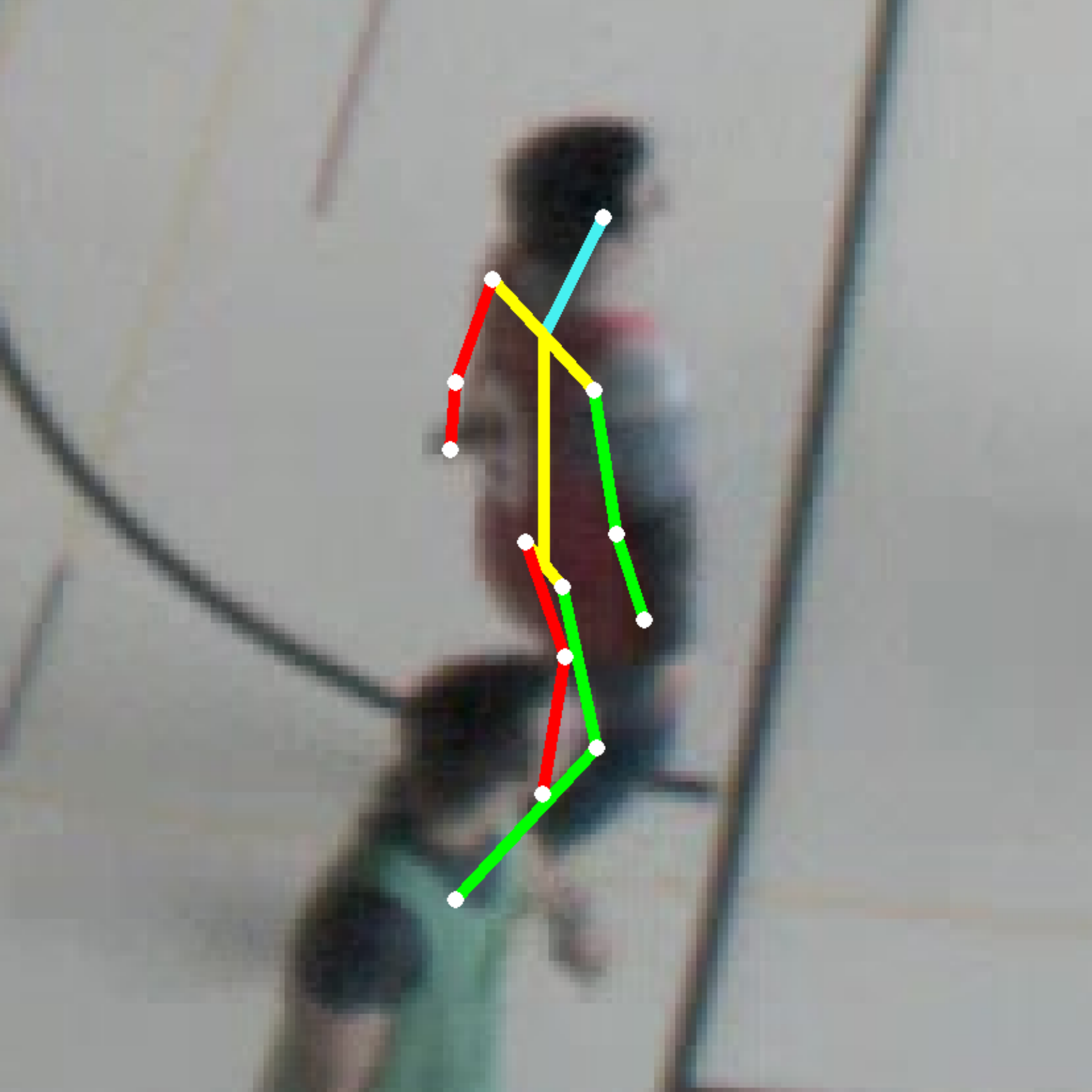}
    \end{subfigure} \\ \vspace{1mm}

    \begin{subfigure}[b]{0.16\linewidth}        
        \centering

\begin{tikzpicture}
    \draw (0, 0) node[inner sep=0] {\includegraphics[width=\linewidth]{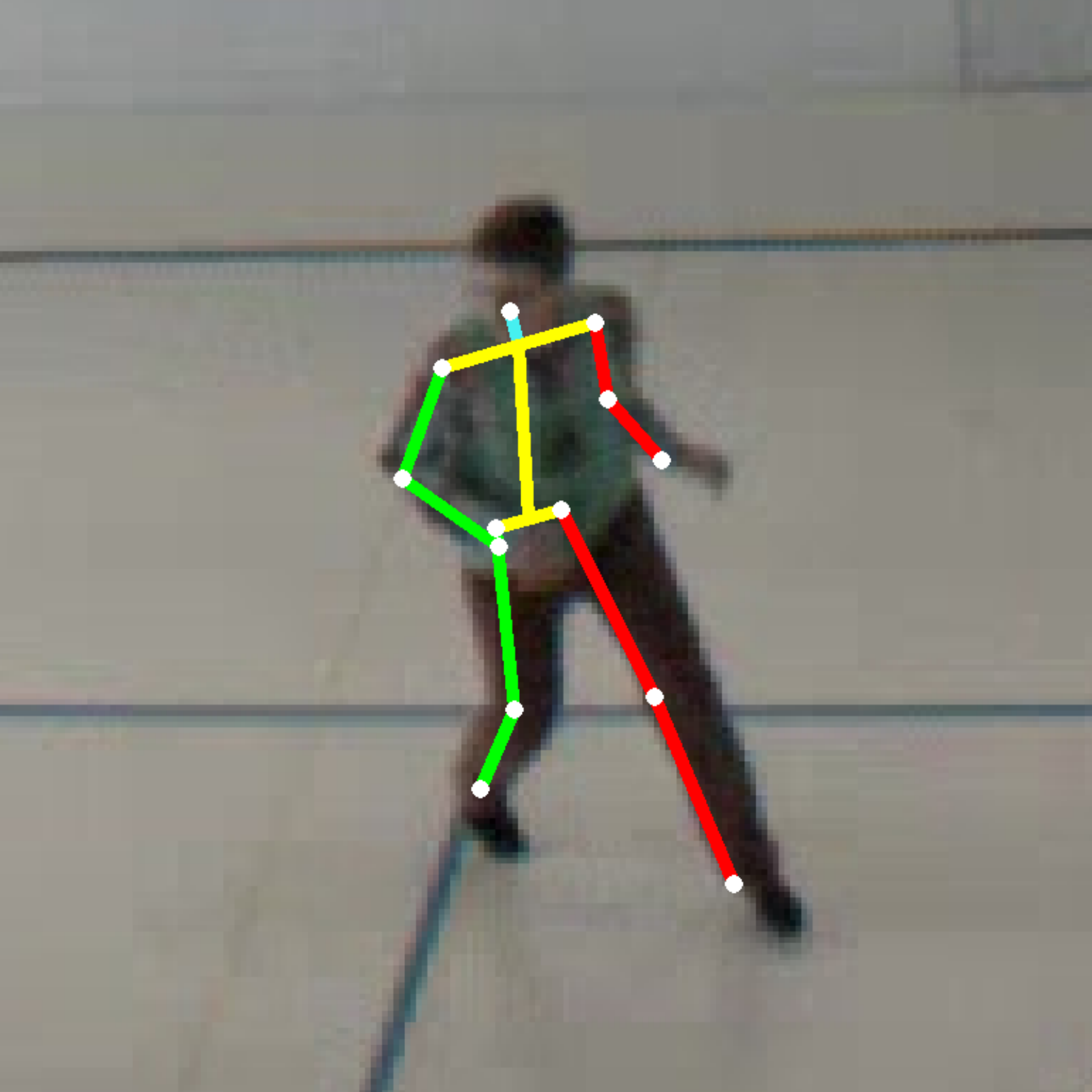}};
    \draw (-0.98, 1) node [fill=white] {b)};
\end{tikzpicture}          
        
    \end{subfigure}
    \begin{subfigure}[b]{0.16\linewidth}        
        \centering
        \includegraphics[width=\linewidth]{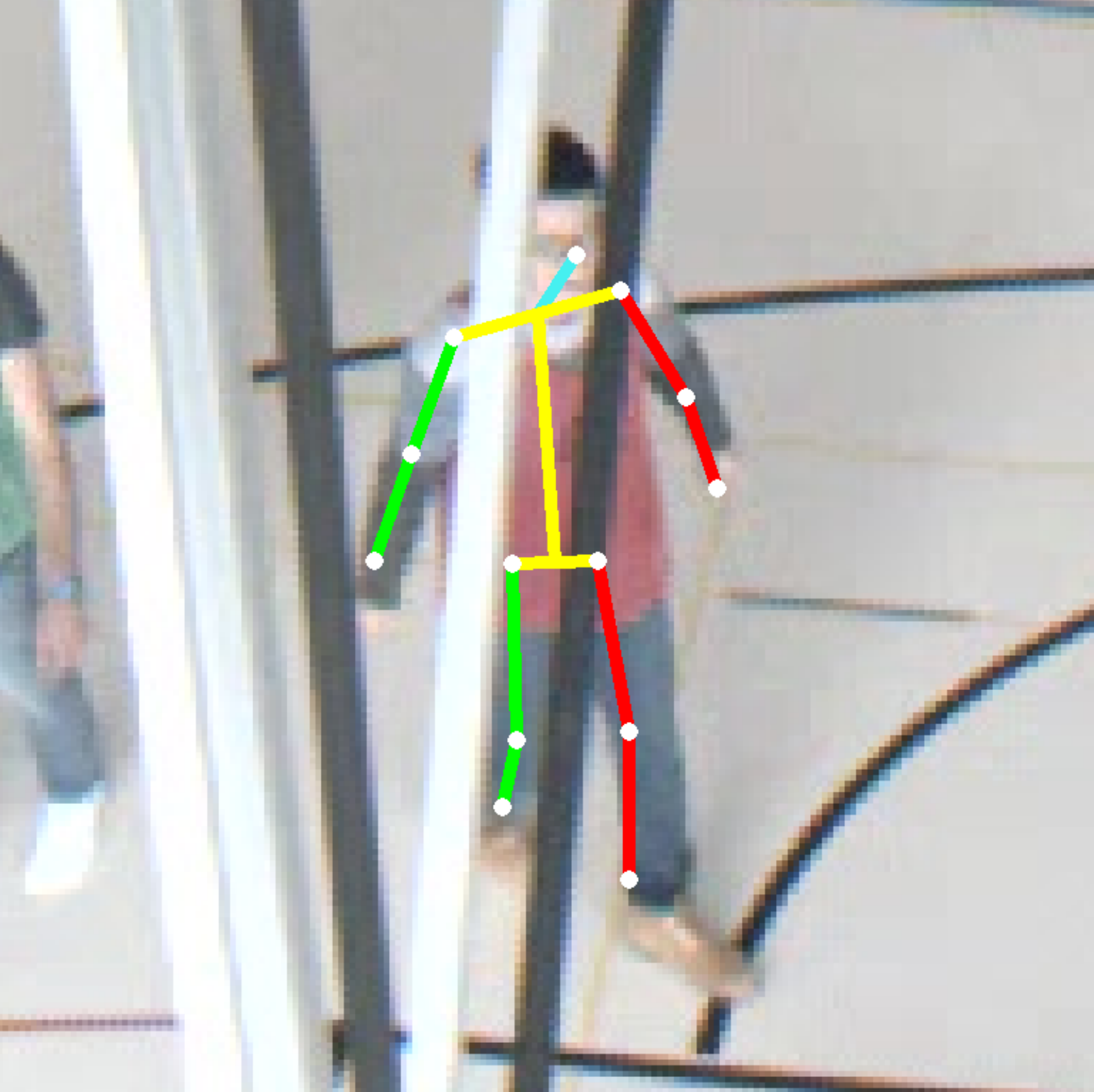}
    \end{subfigure}
    \begin{subfigure}[b]{0.16\linewidth}        
        \centering
        \includegraphics[width=\linewidth]{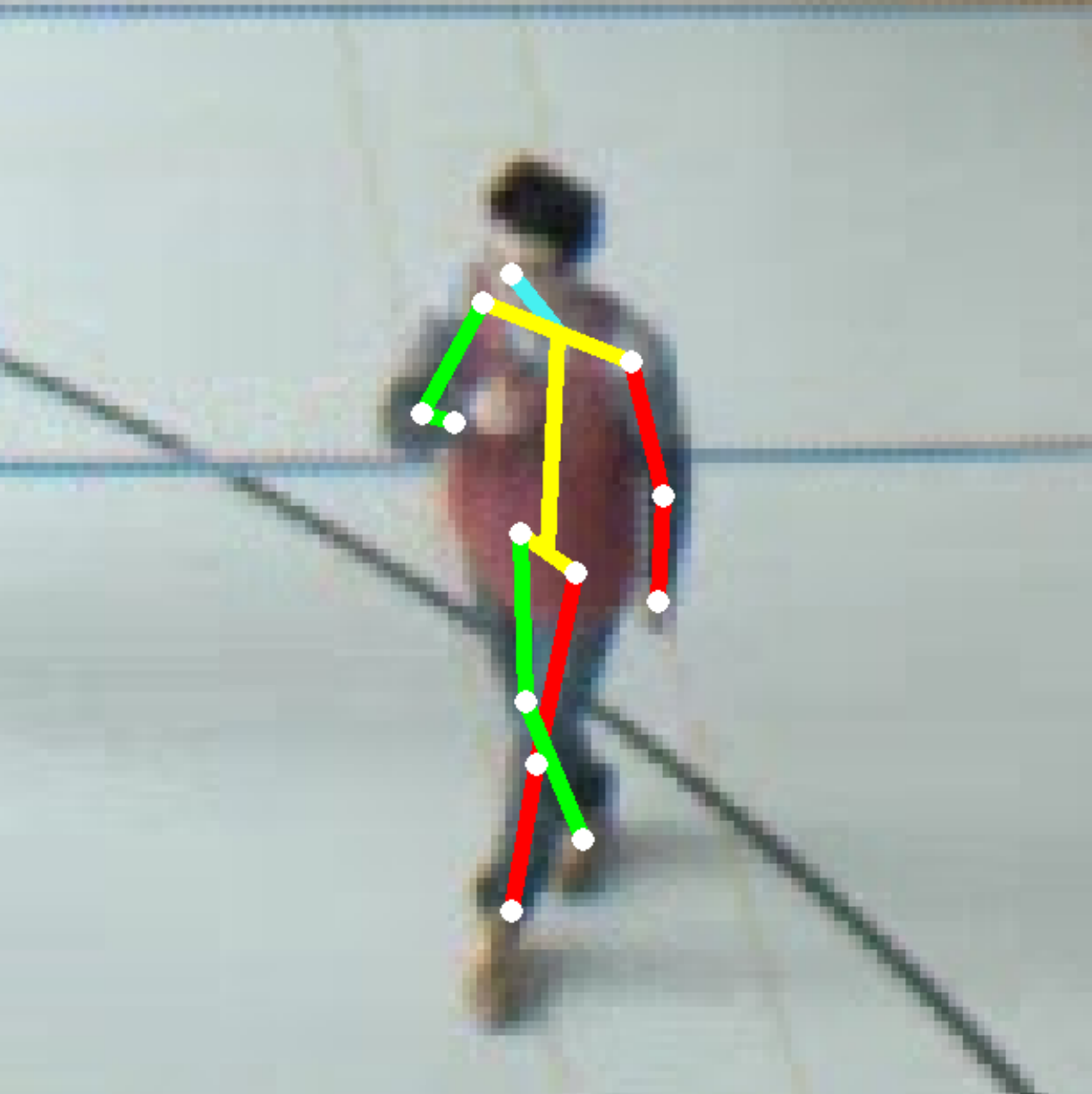}
    \end{subfigure}
    \begin{subfigure}[b]{0.16\linewidth}        
        \centering
        \includegraphics[width=\linewidth]{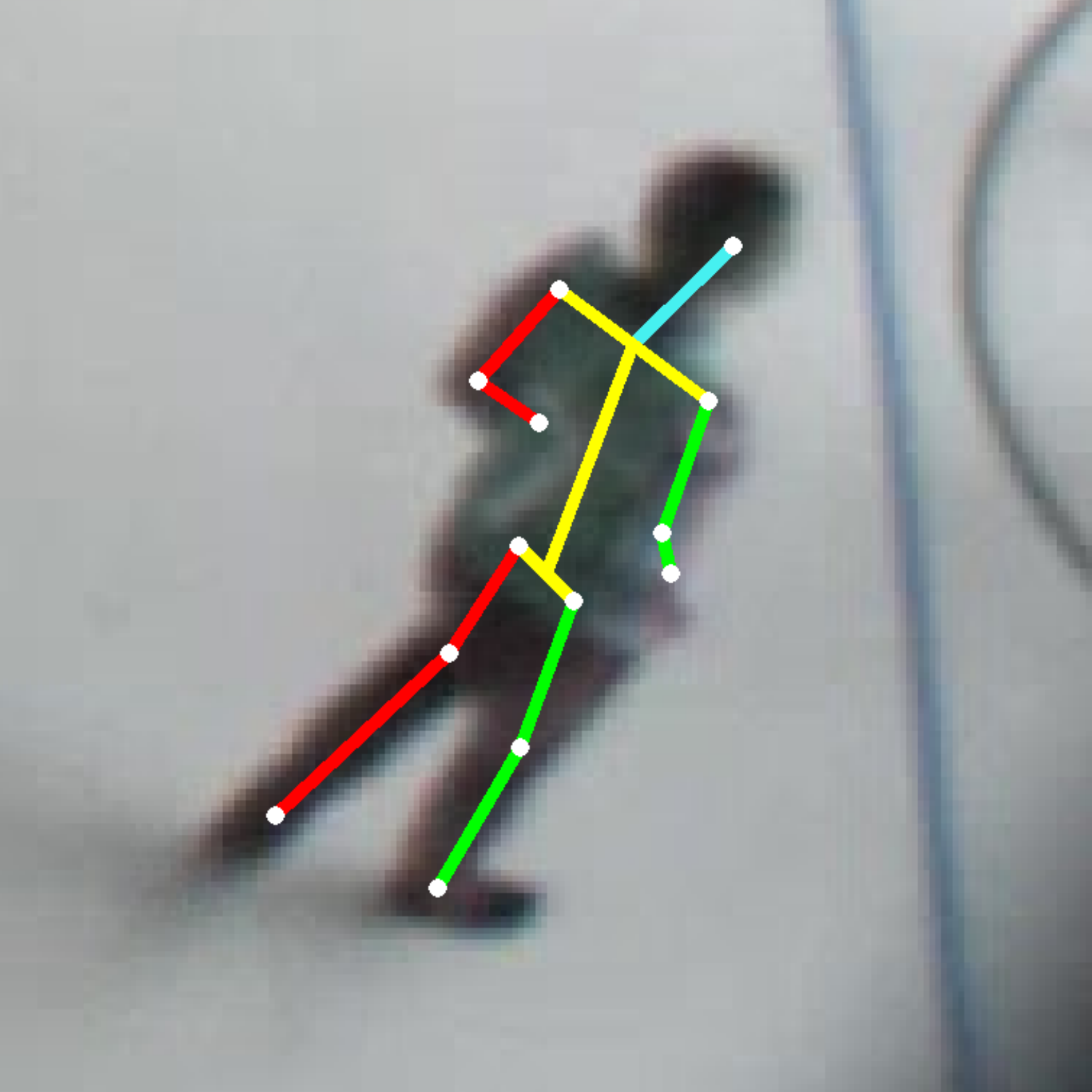}
    \end{subfigure}
    \begin{subfigure}[b]{0.16\linewidth}        
        \centering
        \includegraphics[width=\linewidth]{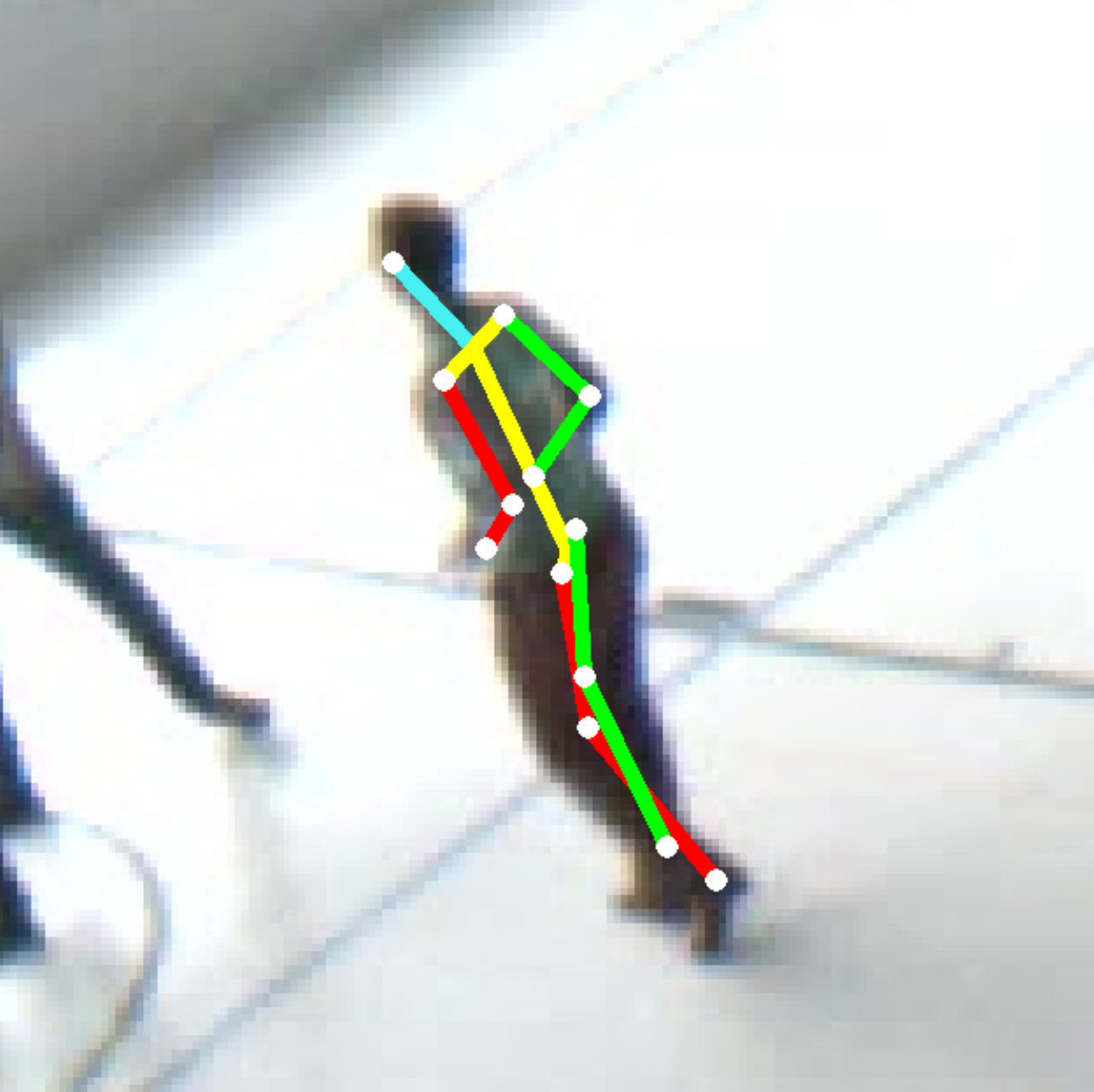}
    \end{subfigure}
    \begin{subfigure}[b]{0.16\linewidth}        
        \centering
        \includegraphics[width=\linewidth]{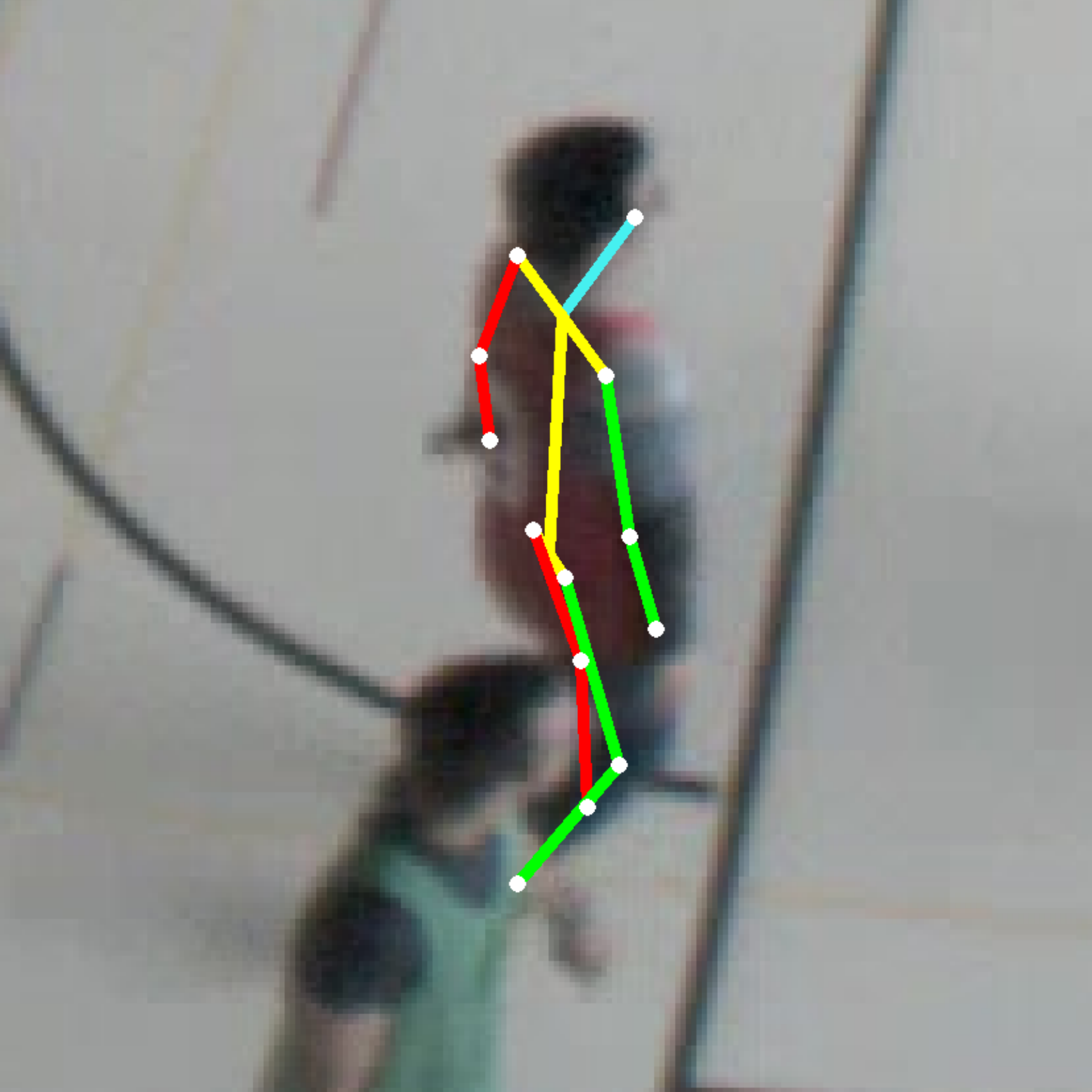}
    \end{subfigure} \\ \vspace{1mm}

    \begin{subfigure}[b]{0.16\linewidth}        
        \centering

\begin{tikzpicture}
    \draw (0, 0) node[inner sep=0] {\includegraphics[width=\linewidth]{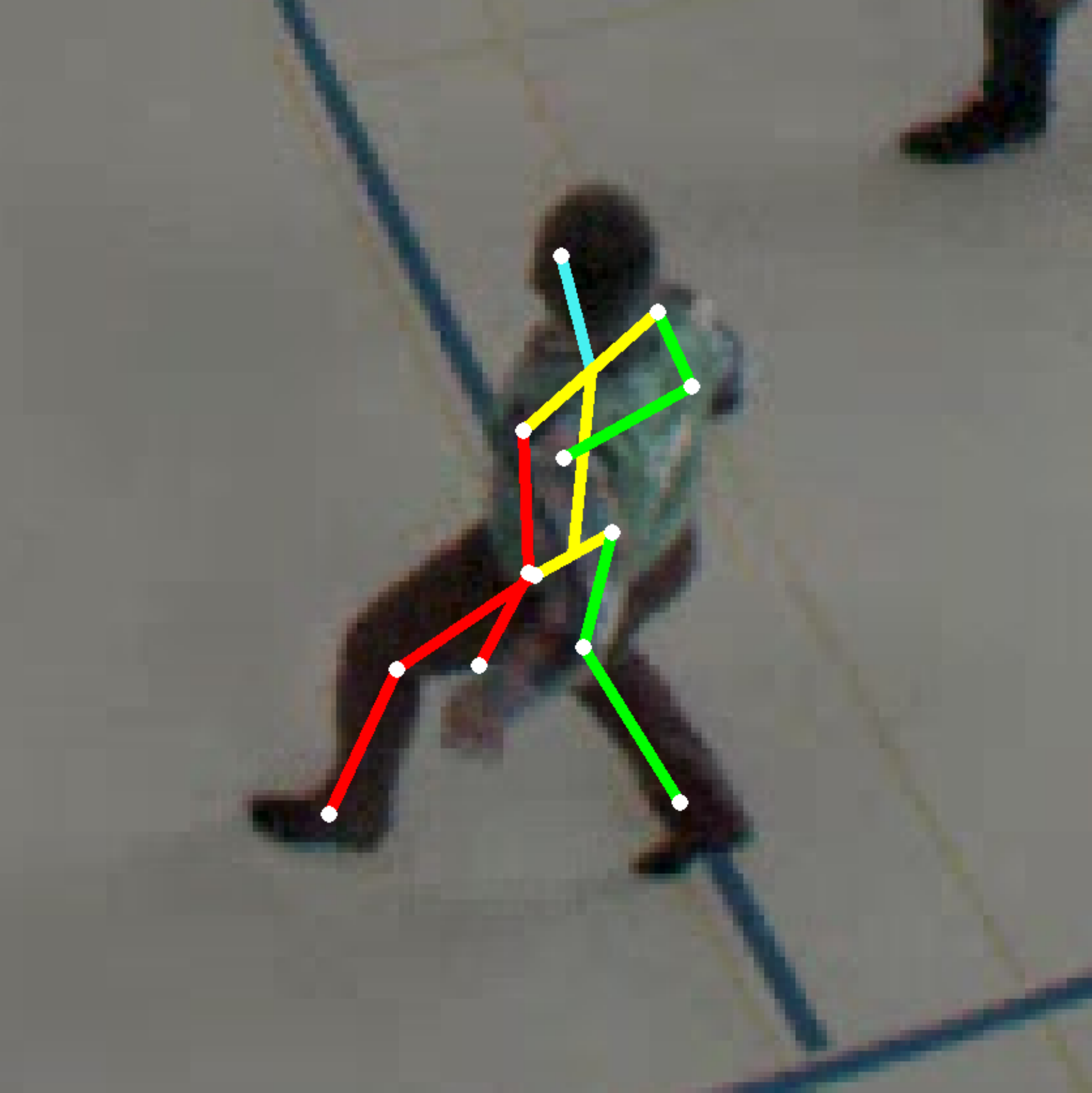}};
    \draw (-0.98, 1) node [fill=white] {c)};
\end{tikzpicture}        
        
    \end{subfigure}
    \begin{subfigure}[b]{0.16\linewidth}        
        \centering
        \includegraphics[width=\linewidth]{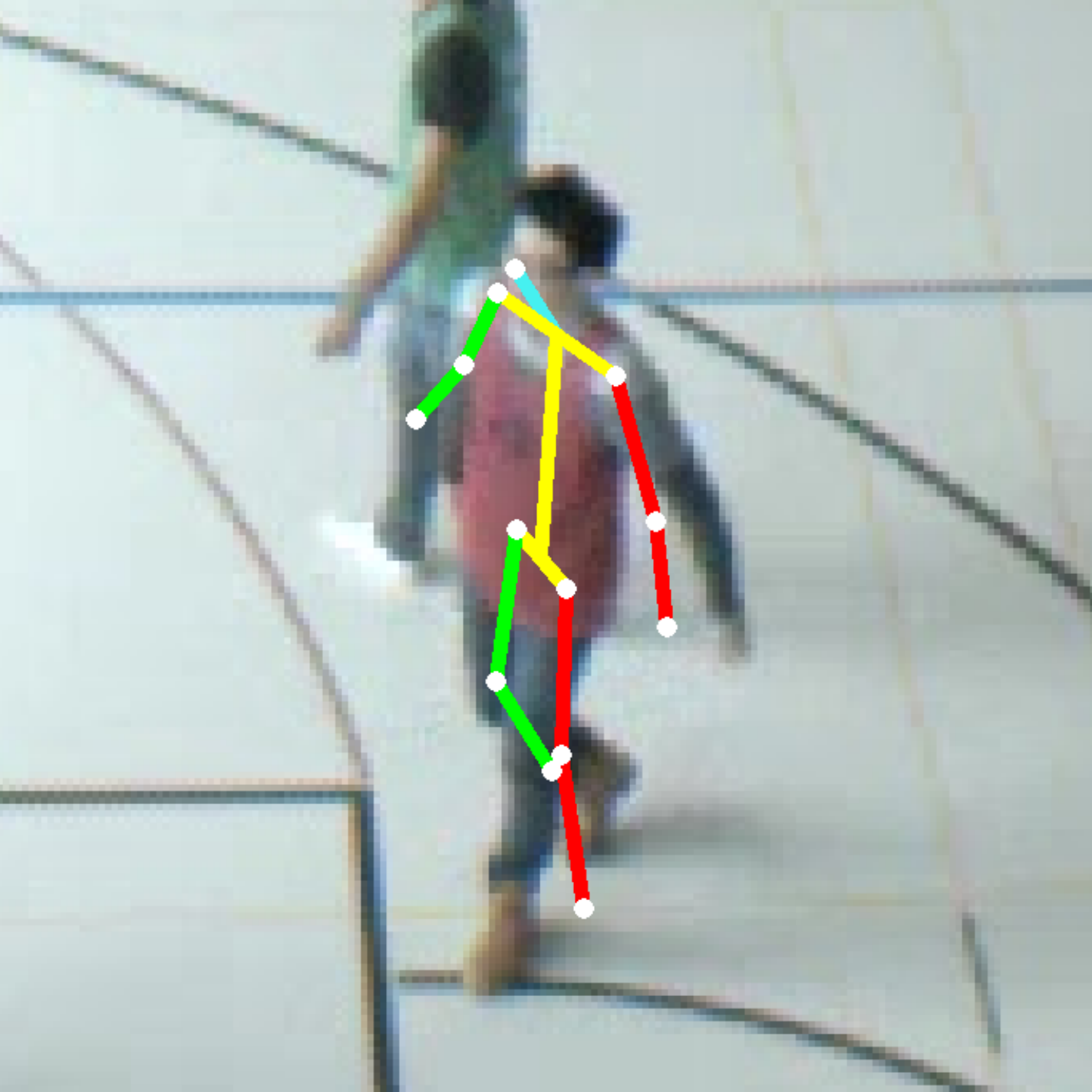}
    \end{subfigure}
    \begin{subfigure}[b]{0.16\linewidth}        
        \centering
        \includegraphics[width=\linewidth]{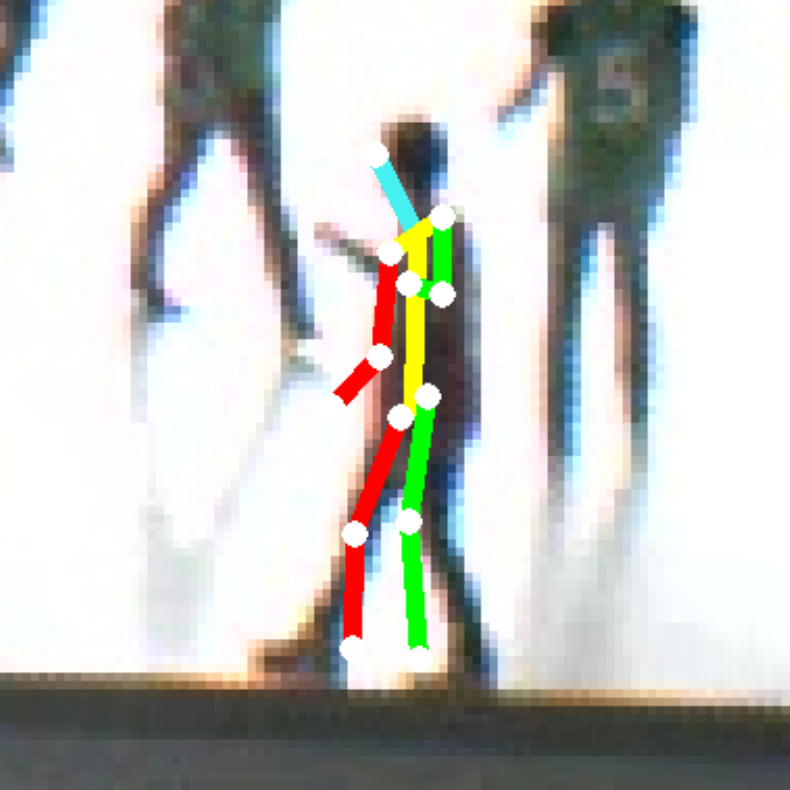}
    \end{subfigure}
    \begin{subfigure}[b]{0.16\linewidth}        
        \centering
        \includegraphics[width=\linewidth]{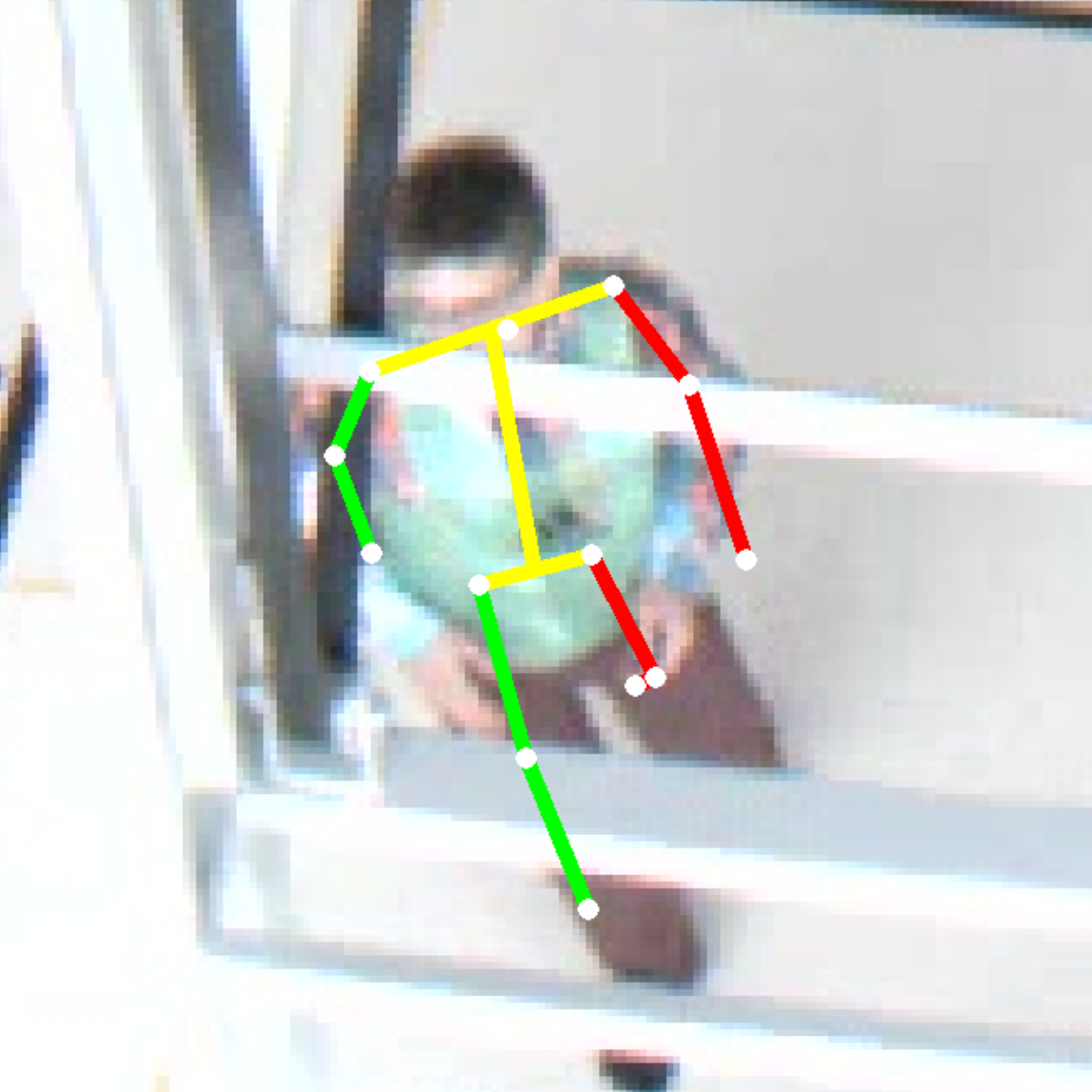}
    \end{subfigure}
    \begin{subfigure}[b]{0.16\linewidth}        
        \centering
        \includegraphics[width=\linewidth]{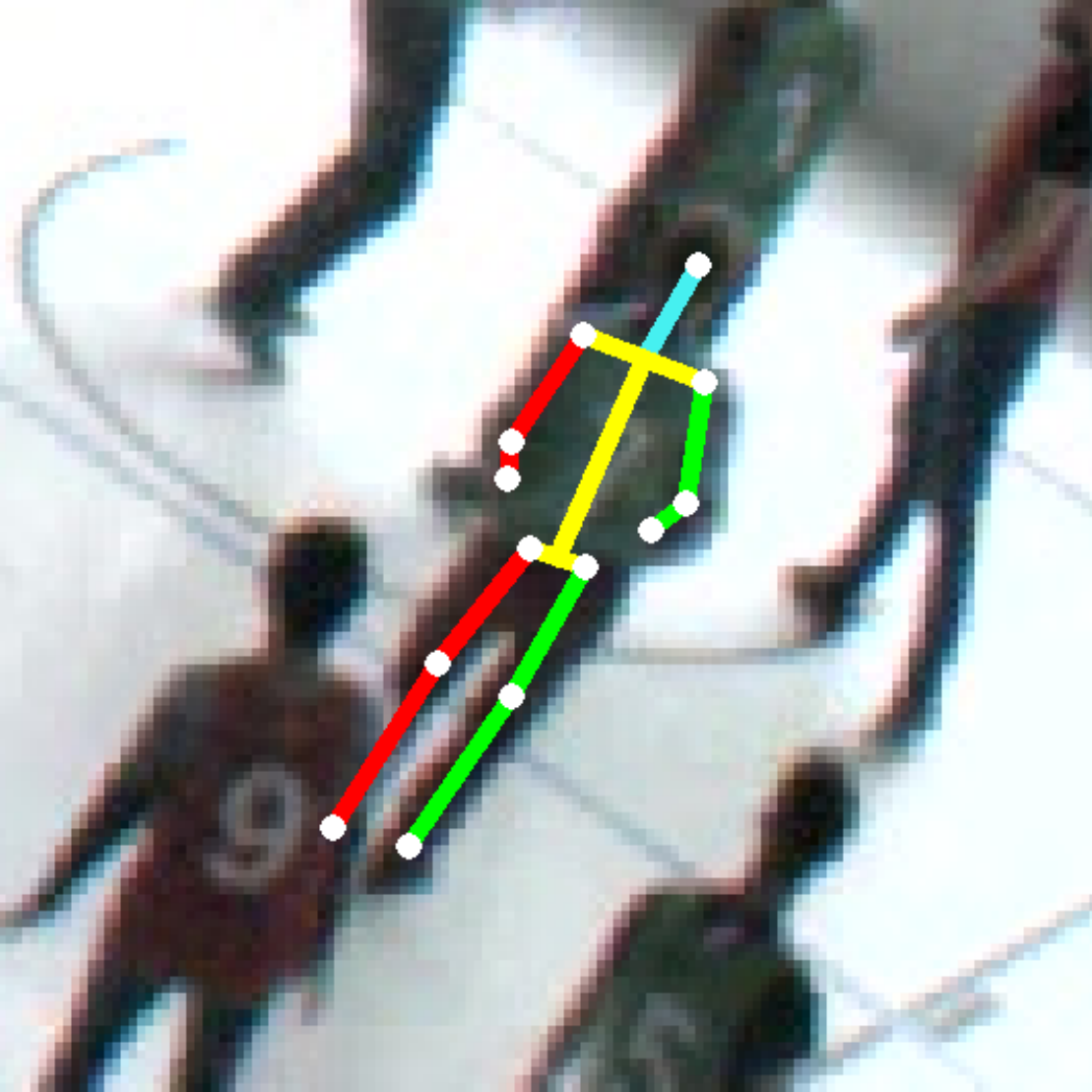}
    \end{subfigure}
    \begin{subfigure}[b]{0.16\linewidth}        
        \centering
        \includegraphics[width=\linewidth]{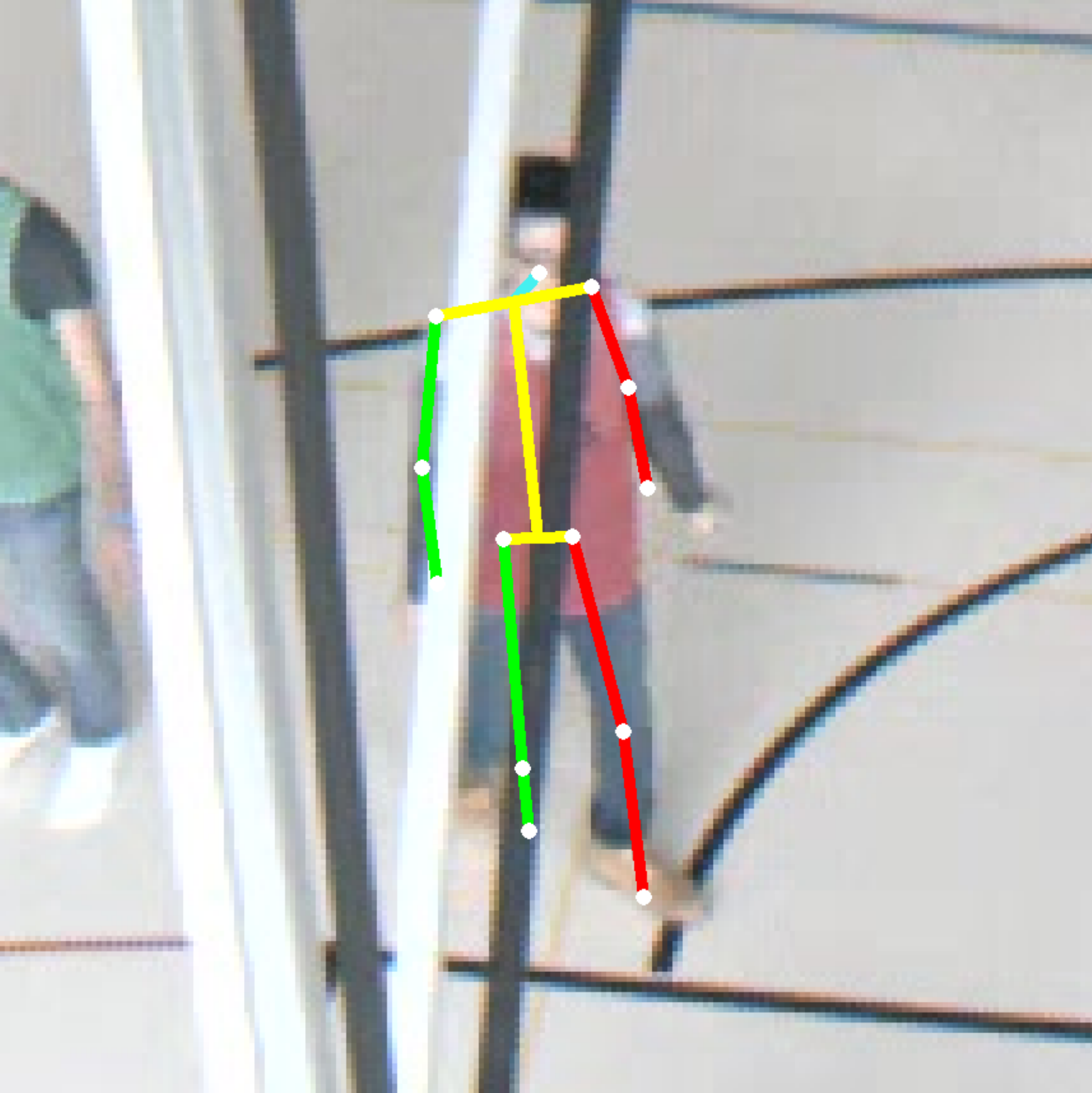}
    \end{subfigure}

    \caption{\small \textbf{Qualitative results on the test set of the SportCenter dataset.} From top to bottom, superimposed 3D poses results of (a) multi-view triangulation, (b) our single-view lifting approach, and (c) same 3D pose as b) but projected into another view to visualize the depth information.}
    \label{fig:visuals_sc_sup}
\end{figure*}

\begin{figure*}[t!]
    \centering
    \begin{subfigure}[b]{0.16\linewidth}        
        \centering

\begin{tikzpicture}
    \draw (0, 0) node[inner sep=0] {\includegraphics[width=\linewidth]{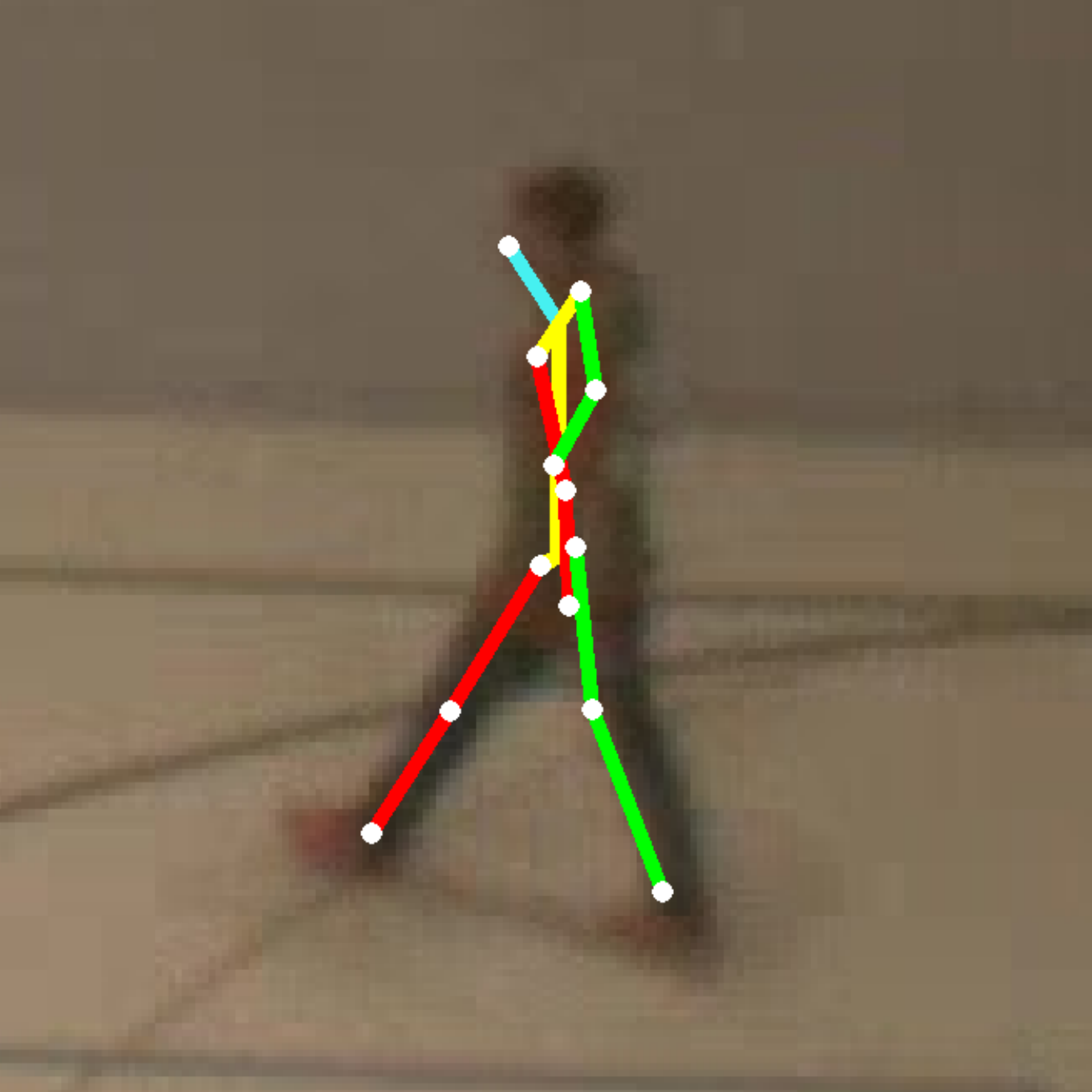}};
    \draw (-0.98, 1) node [fill=white] {a)};
\end{tikzpicture}        
        
    \end{subfigure}
    \begin{subfigure}[b]{0.16\linewidth}        
        \centering
        \includegraphics[width=\linewidth]{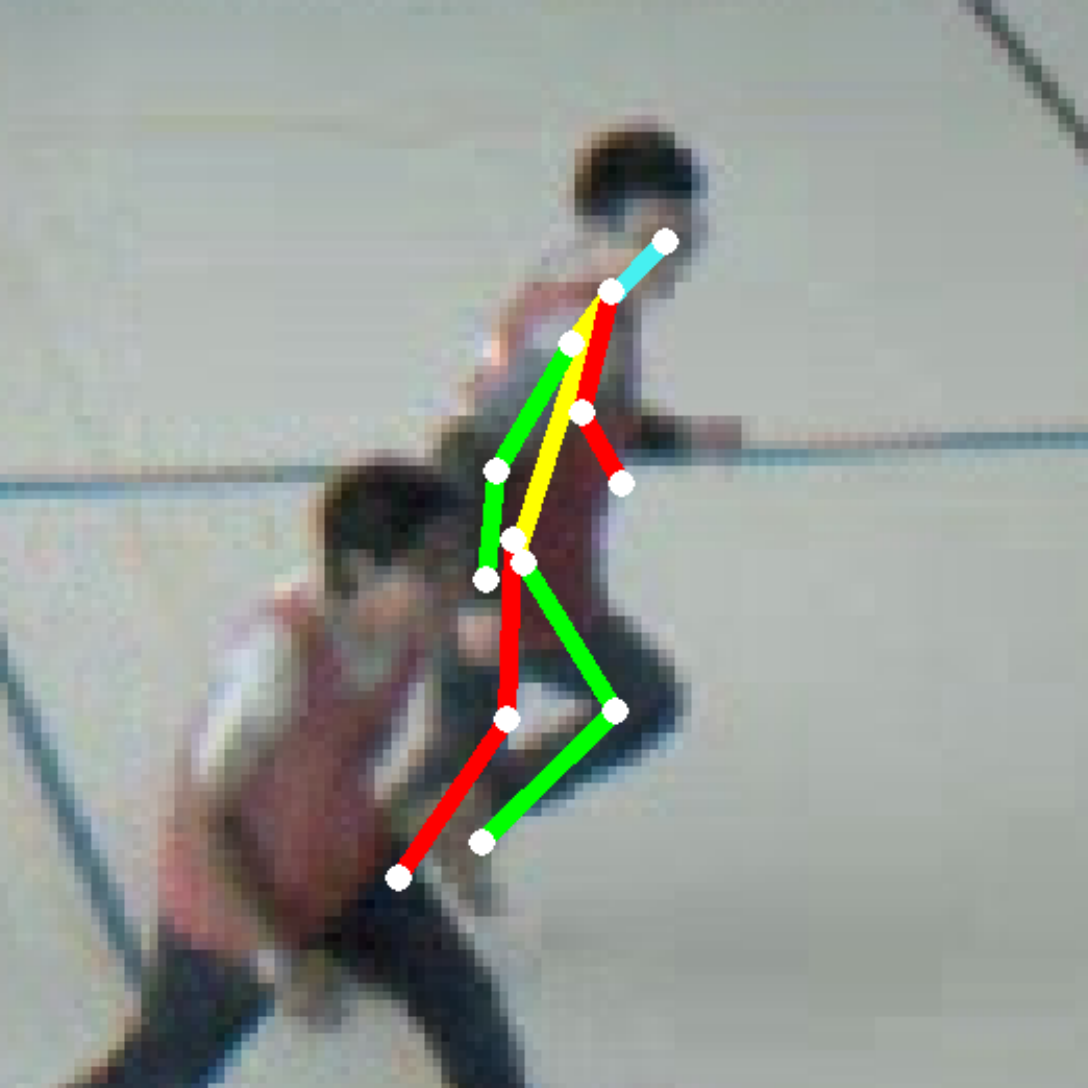}
    \end{subfigure}
    \begin{subfigure}[b]{0.16\linewidth}        
        \centering
        \includegraphics[width=\linewidth]{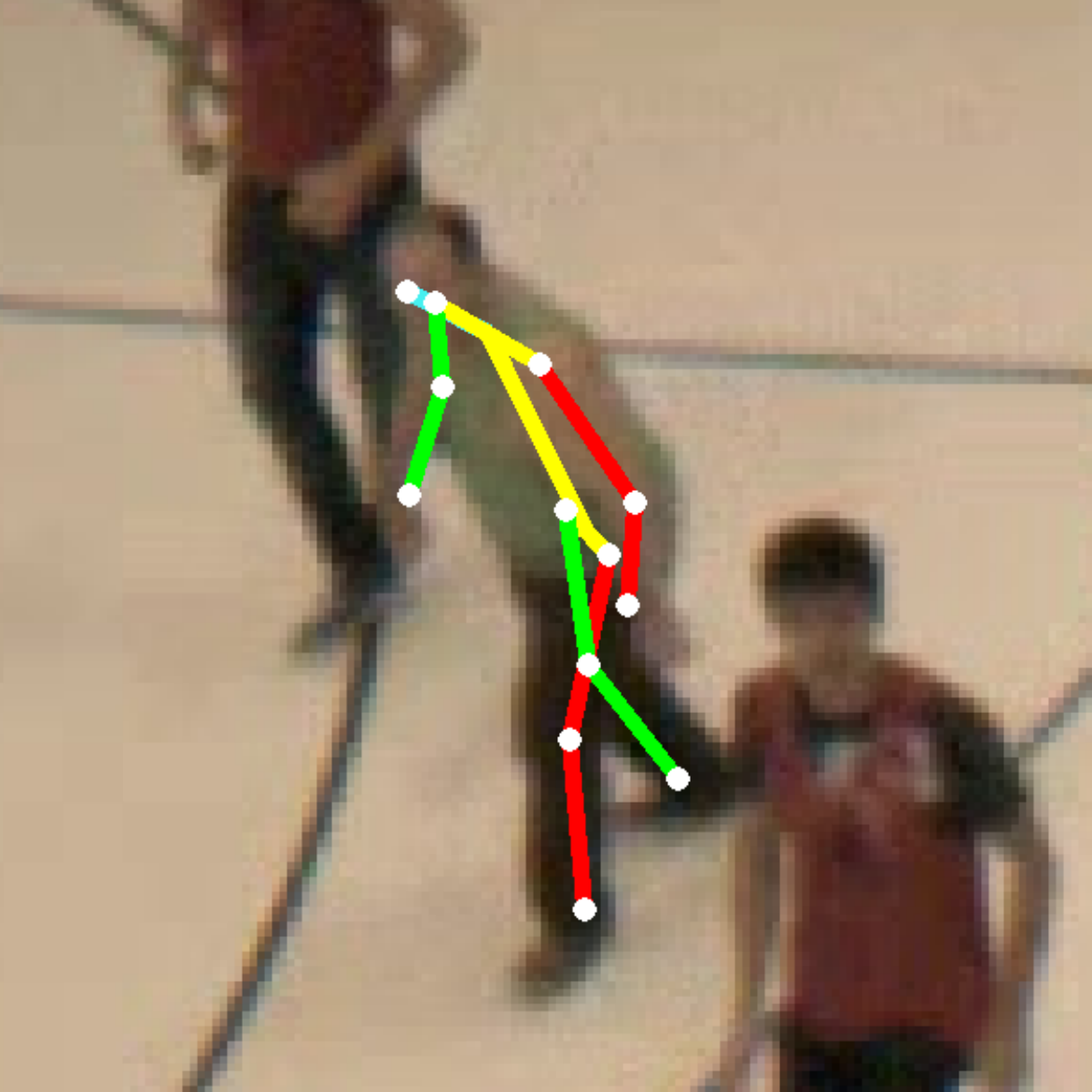}
    \end{subfigure}
    \begin{subfigure}[b]{0.16\linewidth}        
        \centering
        \includegraphics[width=\linewidth]{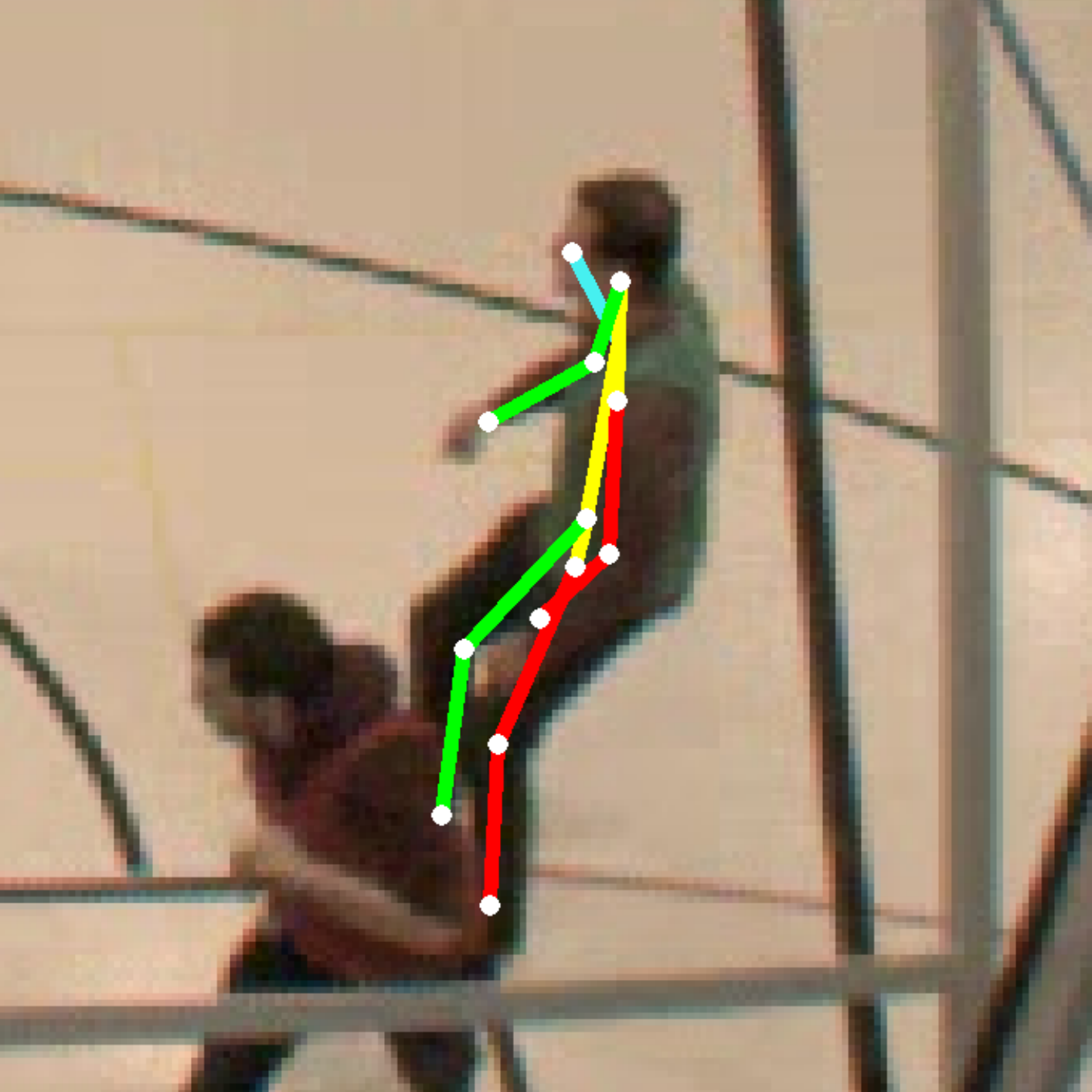}
    \end{subfigure}
    \begin{subfigure}[b]{0.16\linewidth}        
        \centering
        \includegraphics[width=\linewidth]{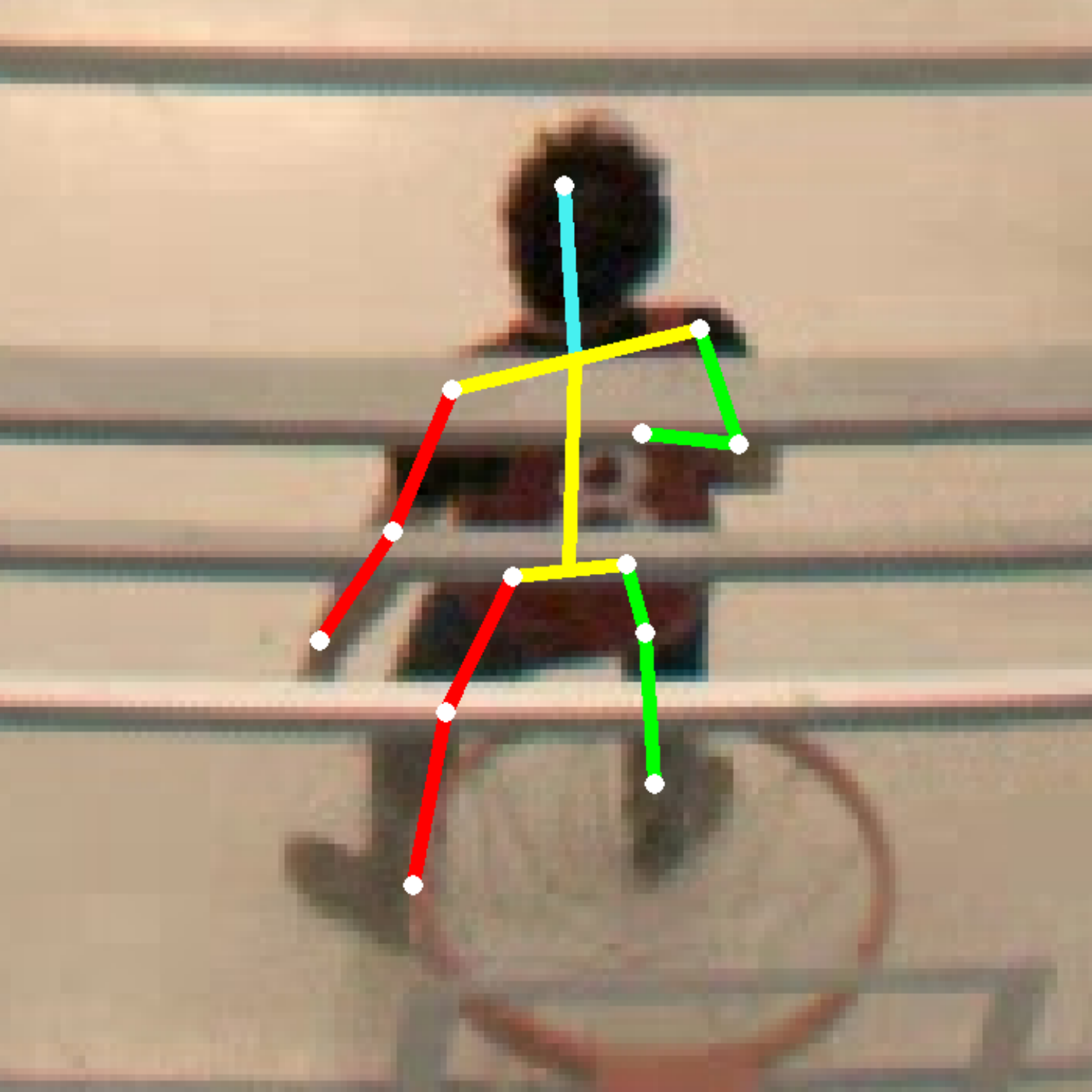}
    \end{subfigure}
    \begin{subfigure}[b]{0.16\linewidth}        
        \centering
        \includegraphics[width=\linewidth]{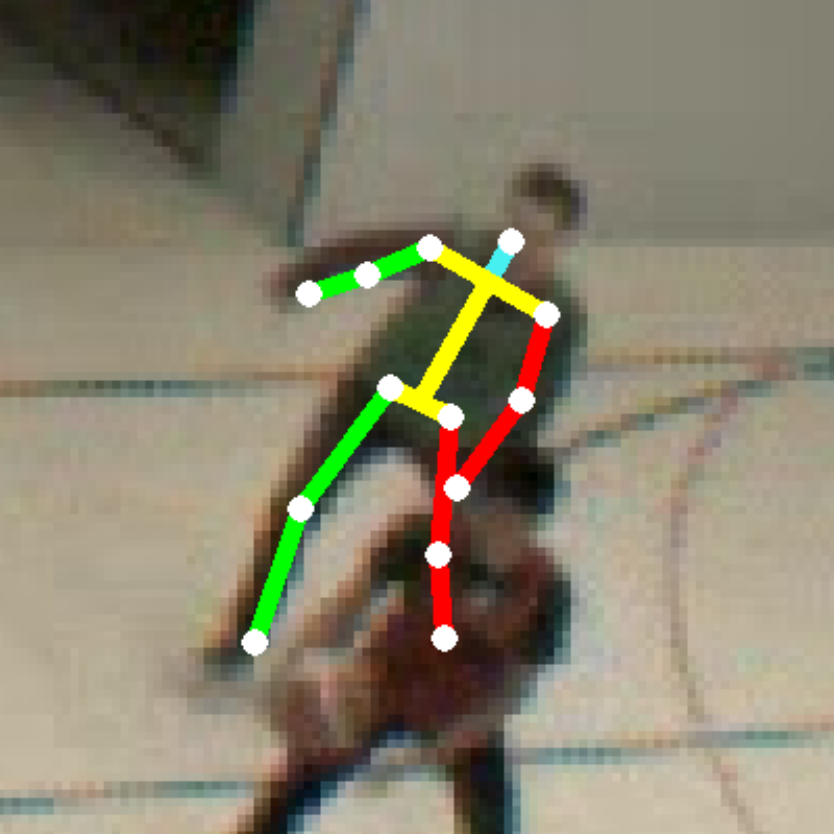}
    \end{subfigure} \\ \vspace{1mm}

    \begin{subfigure}[b]{0.16\linewidth}        
        \centering

\begin{tikzpicture}
    \draw (0, 0) node[inner sep=0] {\includegraphics[width=\linewidth]{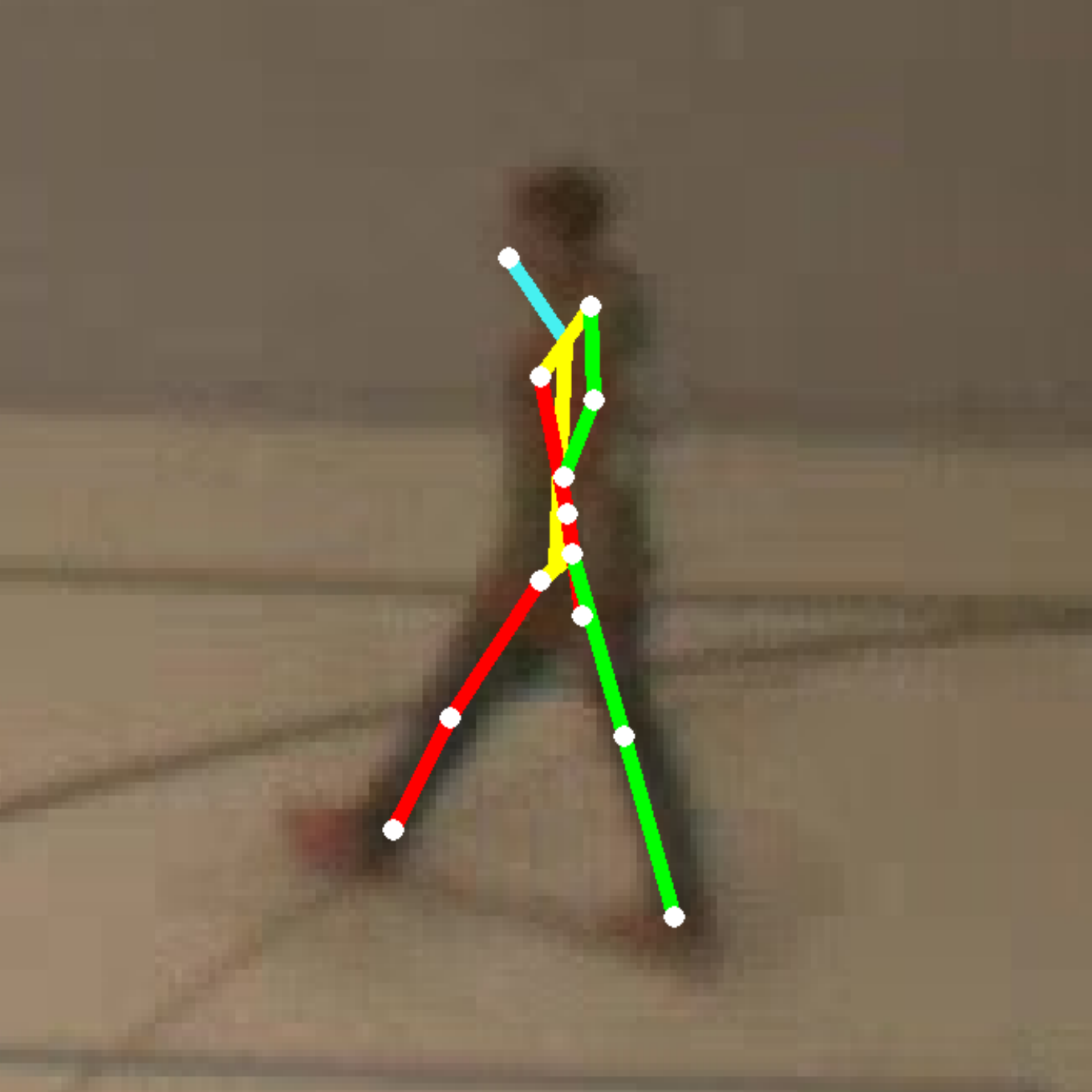}};
    \draw (-0.98, 1) node [fill=white] {b)};
\end{tikzpicture}        
        
    \end{subfigure}
    \begin{subfigure}[b]{0.16\linewidth}        
        \centering
        \includegraphics[width=\linewidth]{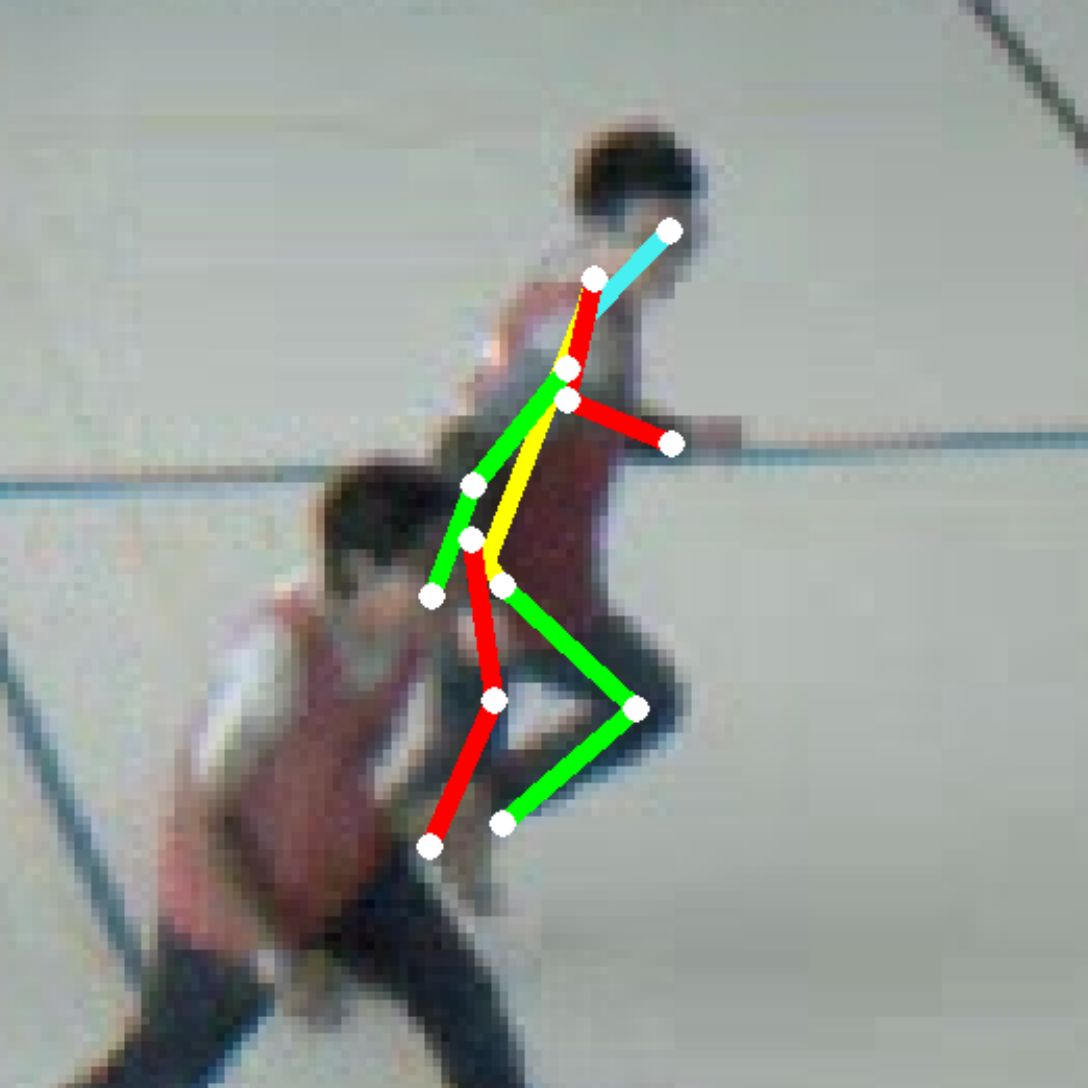}
    \end{subfigure}
    \begin{subfigure}[b]{0.16\linewidth}        
        \centering
        \includegraphics[width=\linewidth]{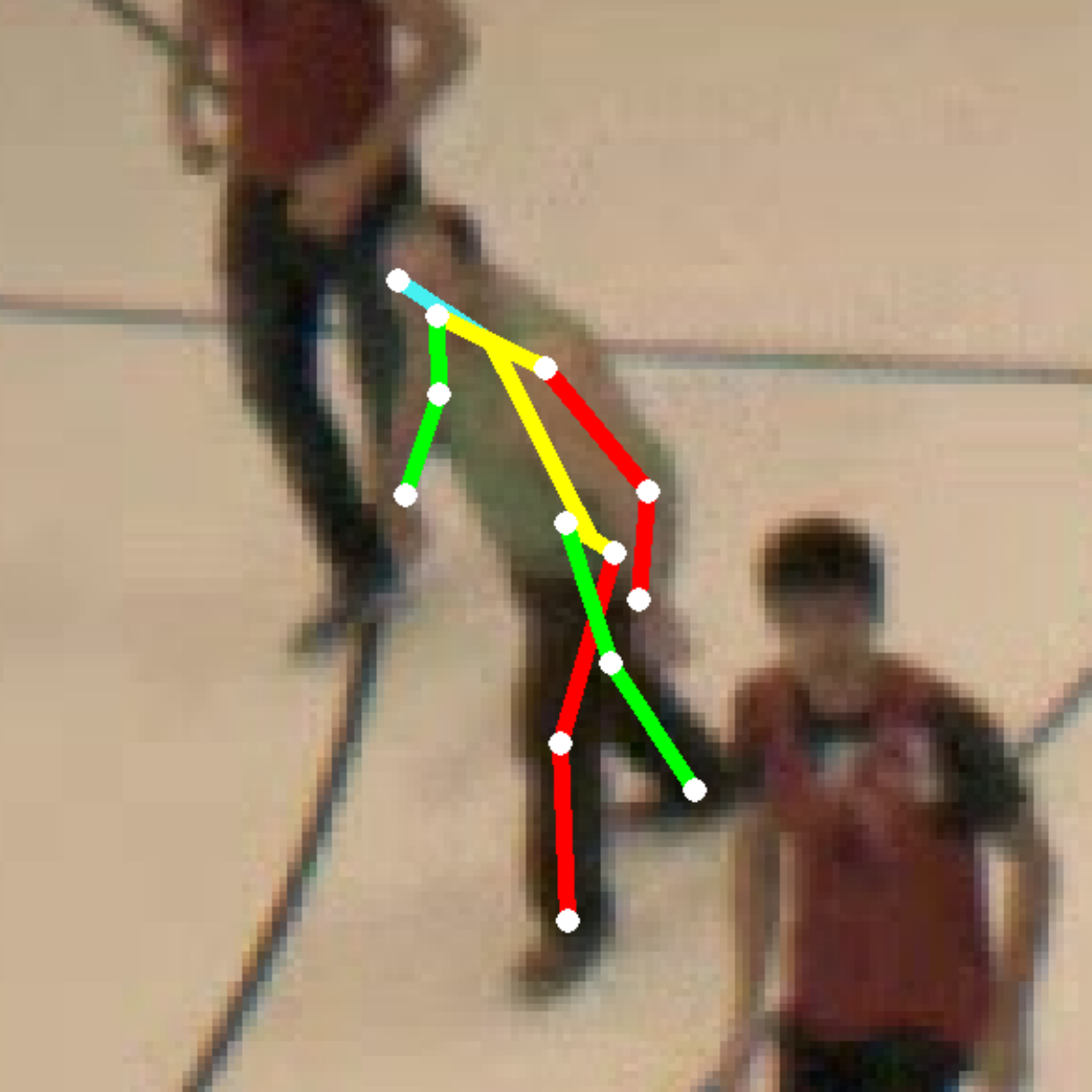}
    \end{subfigure}
    \begin{subfigure}[b]{0.16\linewidth}        
        \centering
        \includegraphics[width=\linewidth]{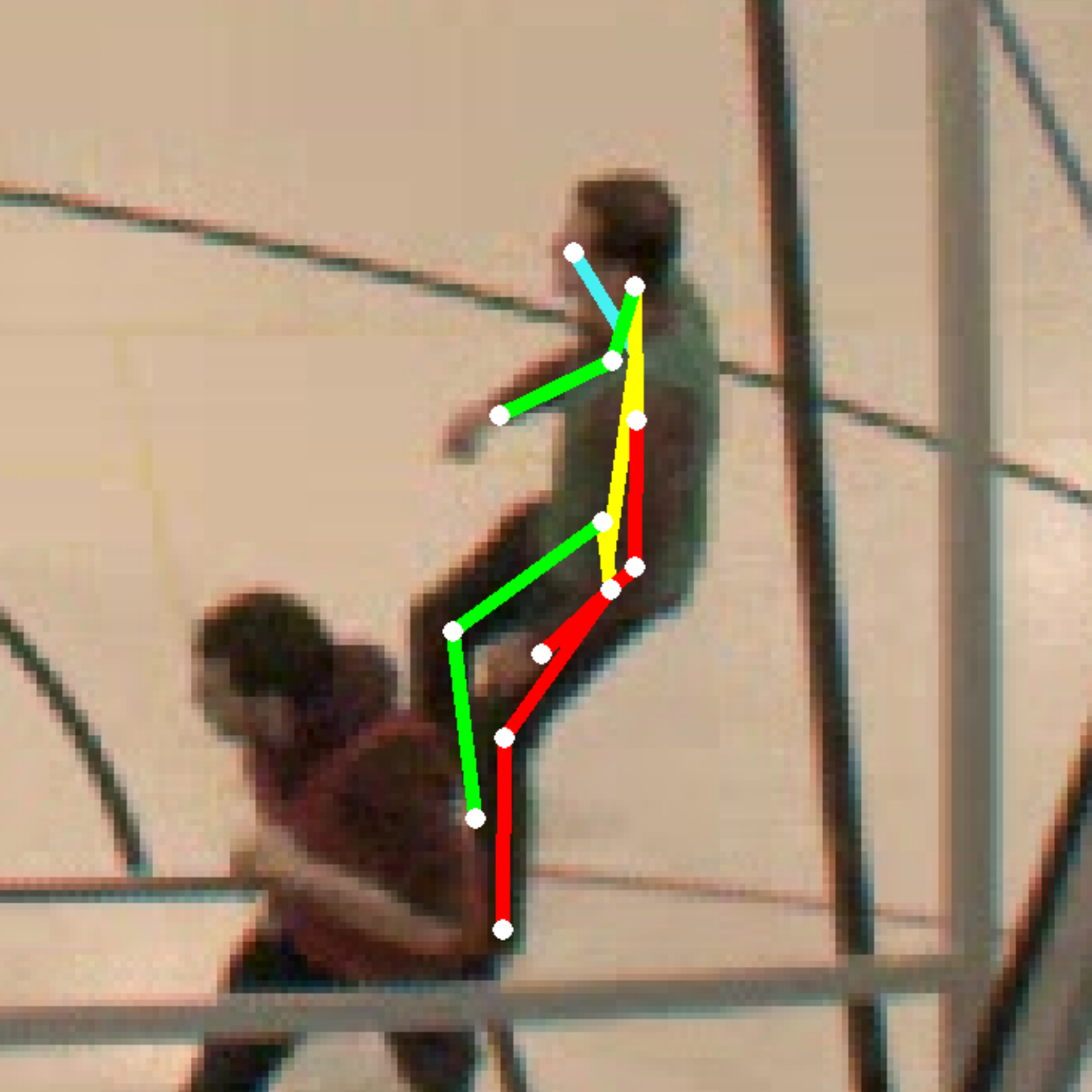}
    \end{subfigure}
    \begin{subfigure}[b]{0.16\linewidth}        
        \centering
        \includegraphics[width=\linewidth]{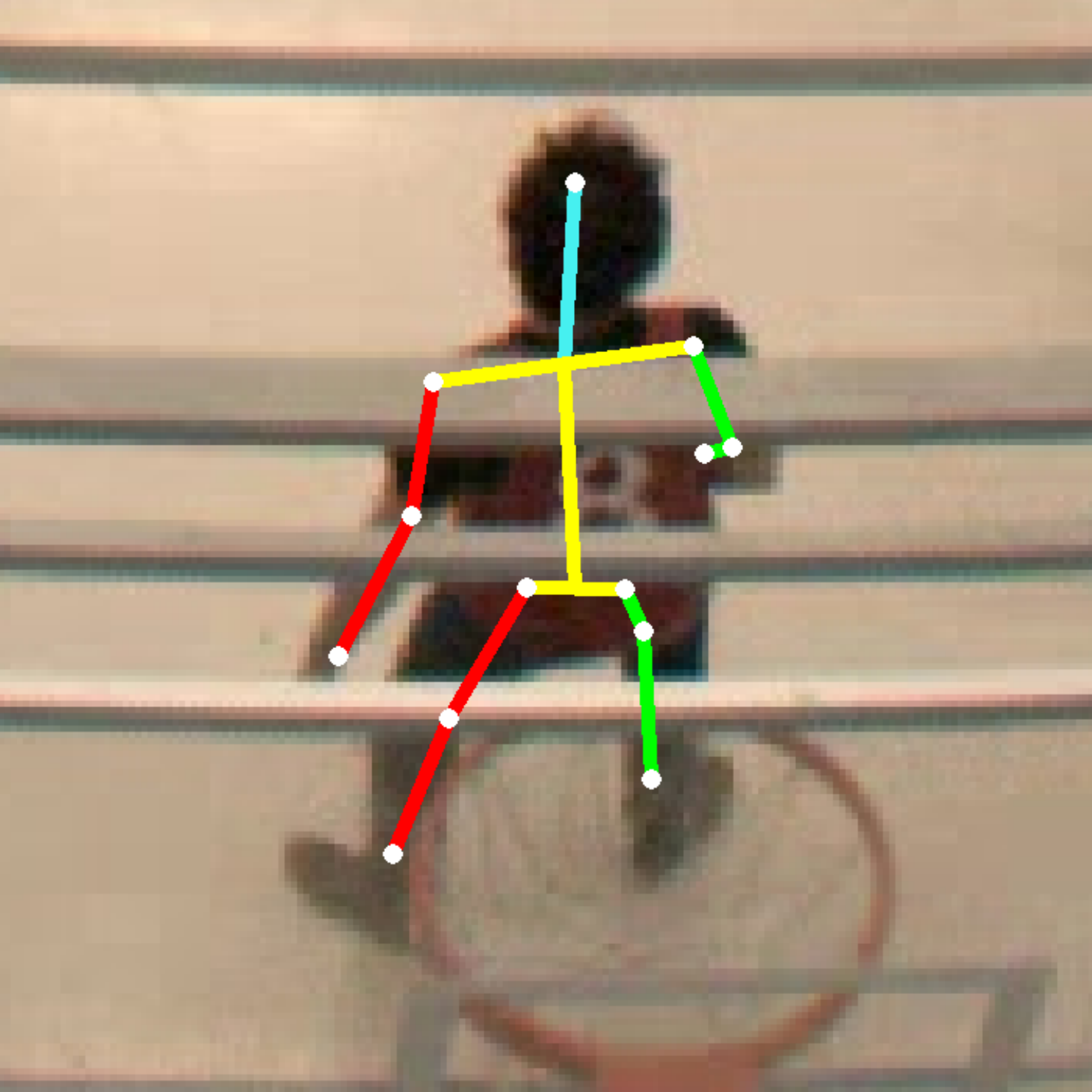}
    \end{subfigure}
    \begin{subfigure}[b]{0.16\linewidth}        
        \centering
        \includegraphics[width=\linewidth]{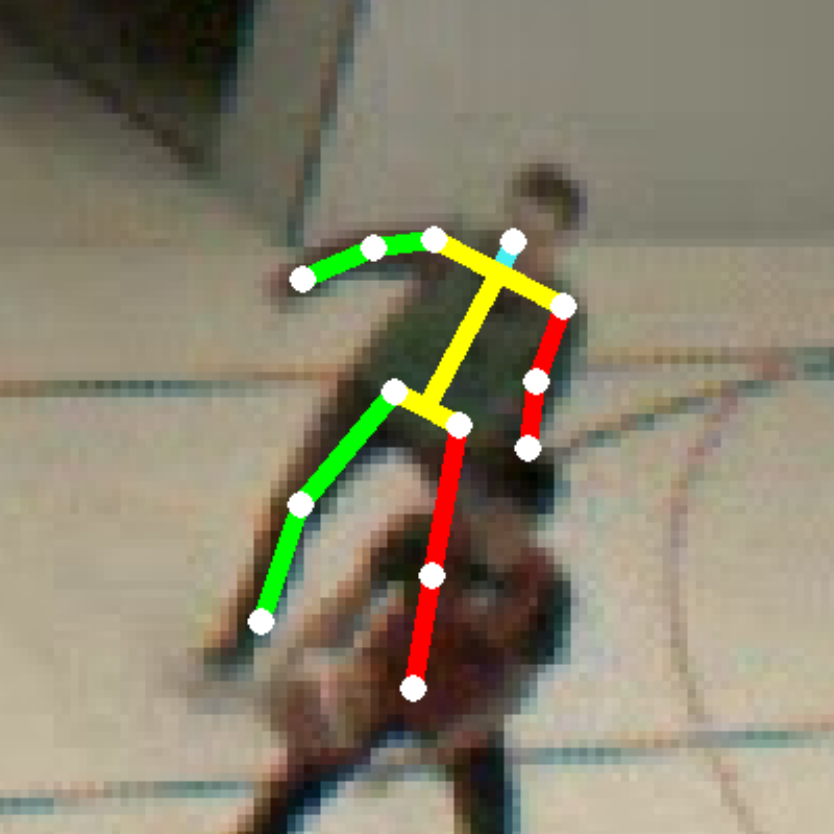}
    \end{subfigure} \\ \vspace{1mm}

    \begin{subfigure}[b]{0.16\linewidth}        
        \centering

\begin{tikzpicture}
    \draw (0, 0) node[inner sep=0] {\includegraphics[width=\linewidth]{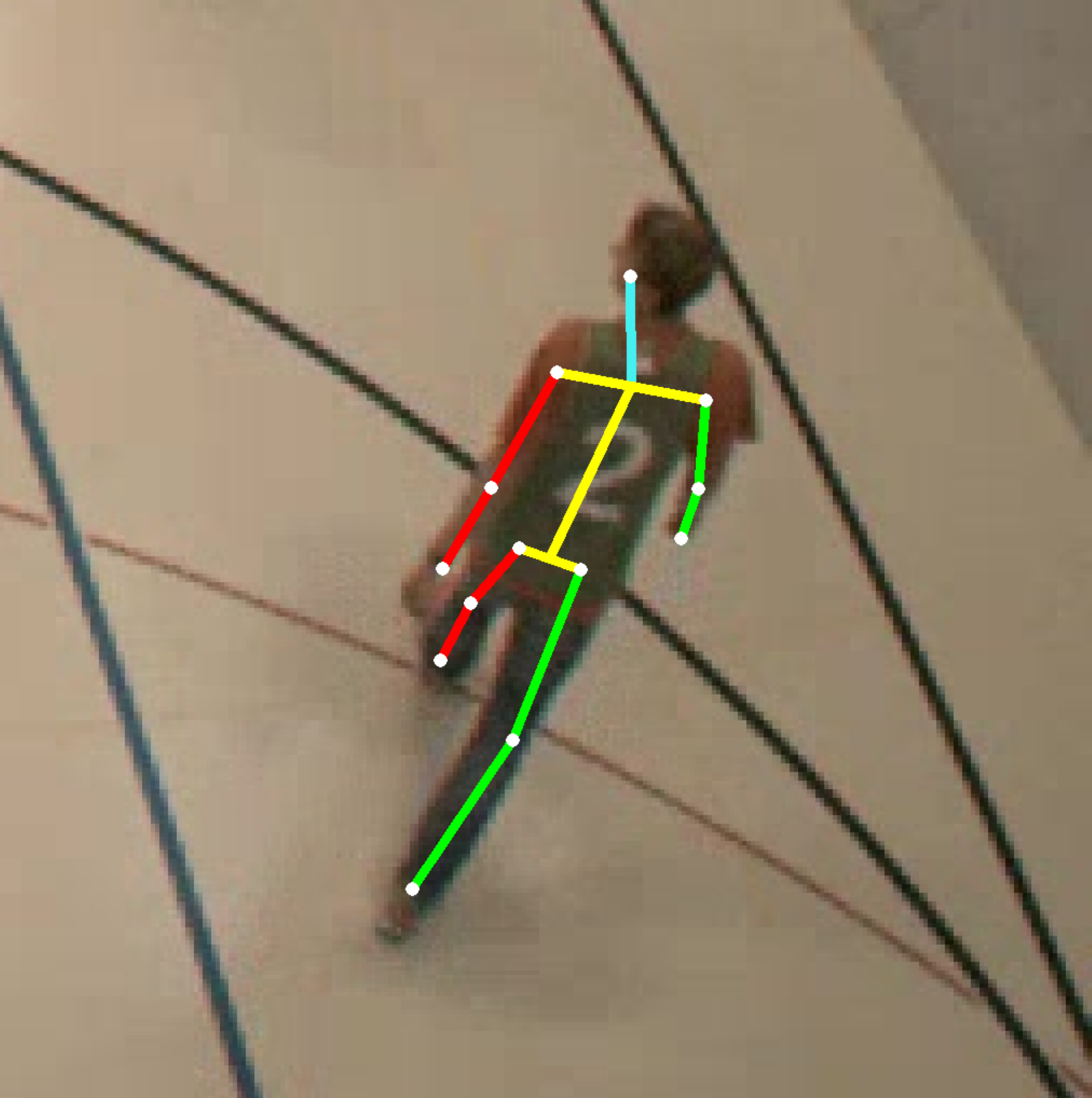}};
    \draw (-0.98, 1) node [fill=white] {c)};
\end{tikzpicture}        
        
    \end{subfigure}
    \begin{subfigure}[b]{0.16\linewidth}        
        \centering
        \includegraphics[width=\linewidth]{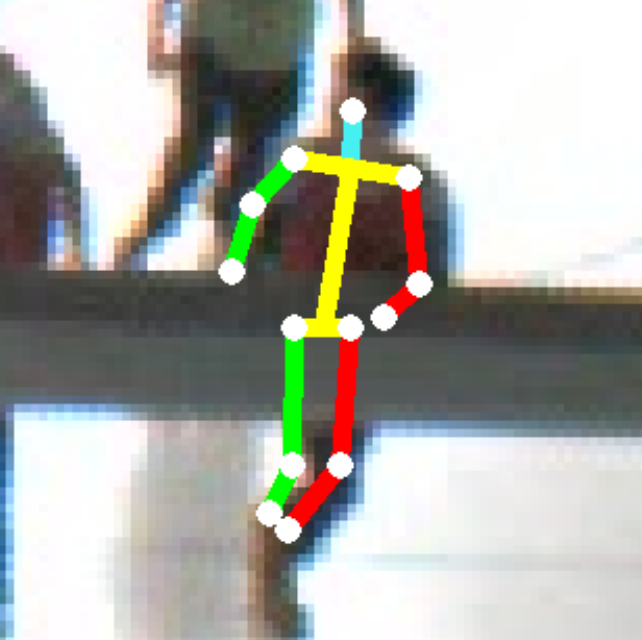}
    \end{subfigure}
    \begin{subfigure}[b]{0.16\linewidth}        
        \centering
        \includegraphics[width=\linewidth]{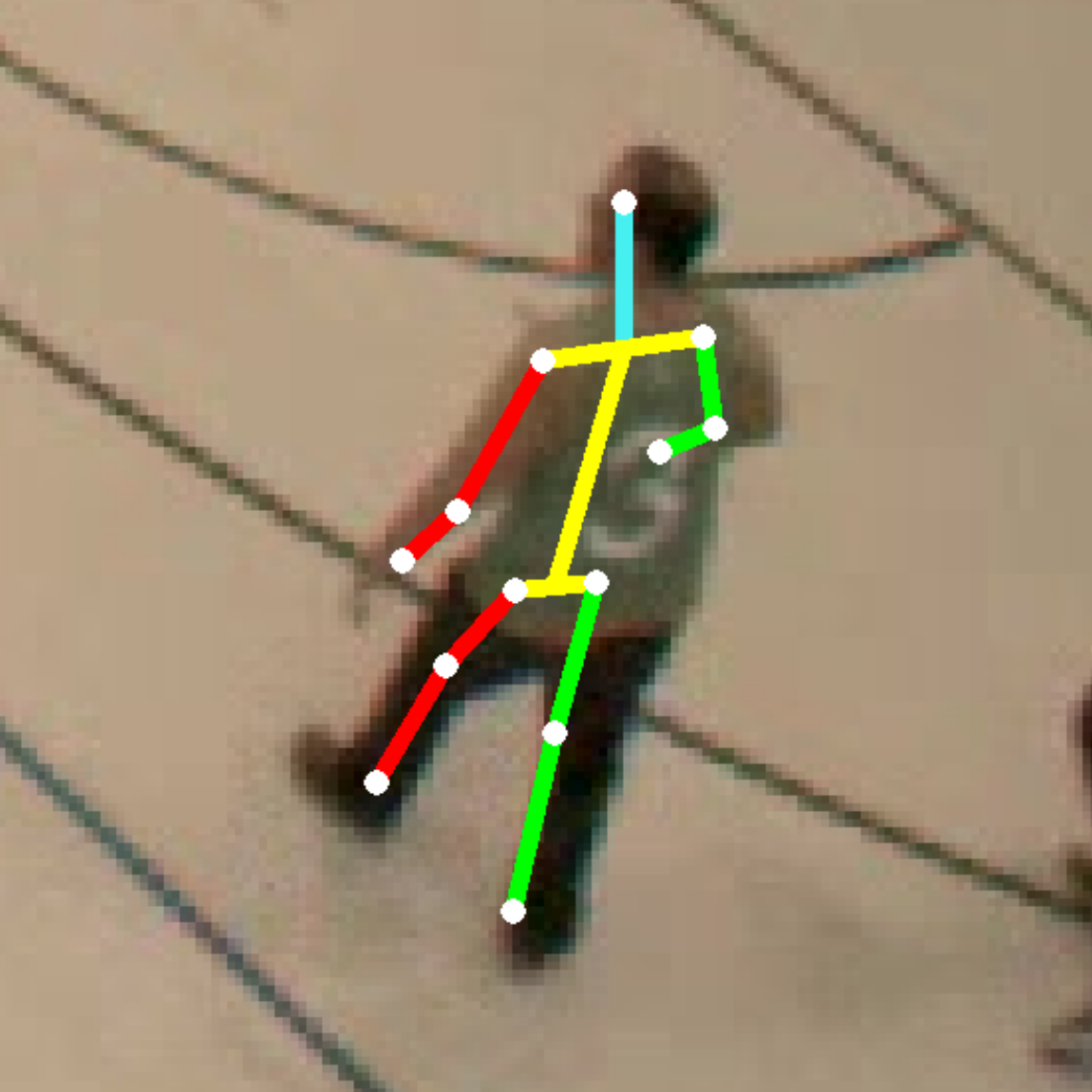}
    \end{subfigure}
    \begin{subfigure}[b]{0.16\linewidth}        
        \centering
        \includegraphics[width=\linewidth]{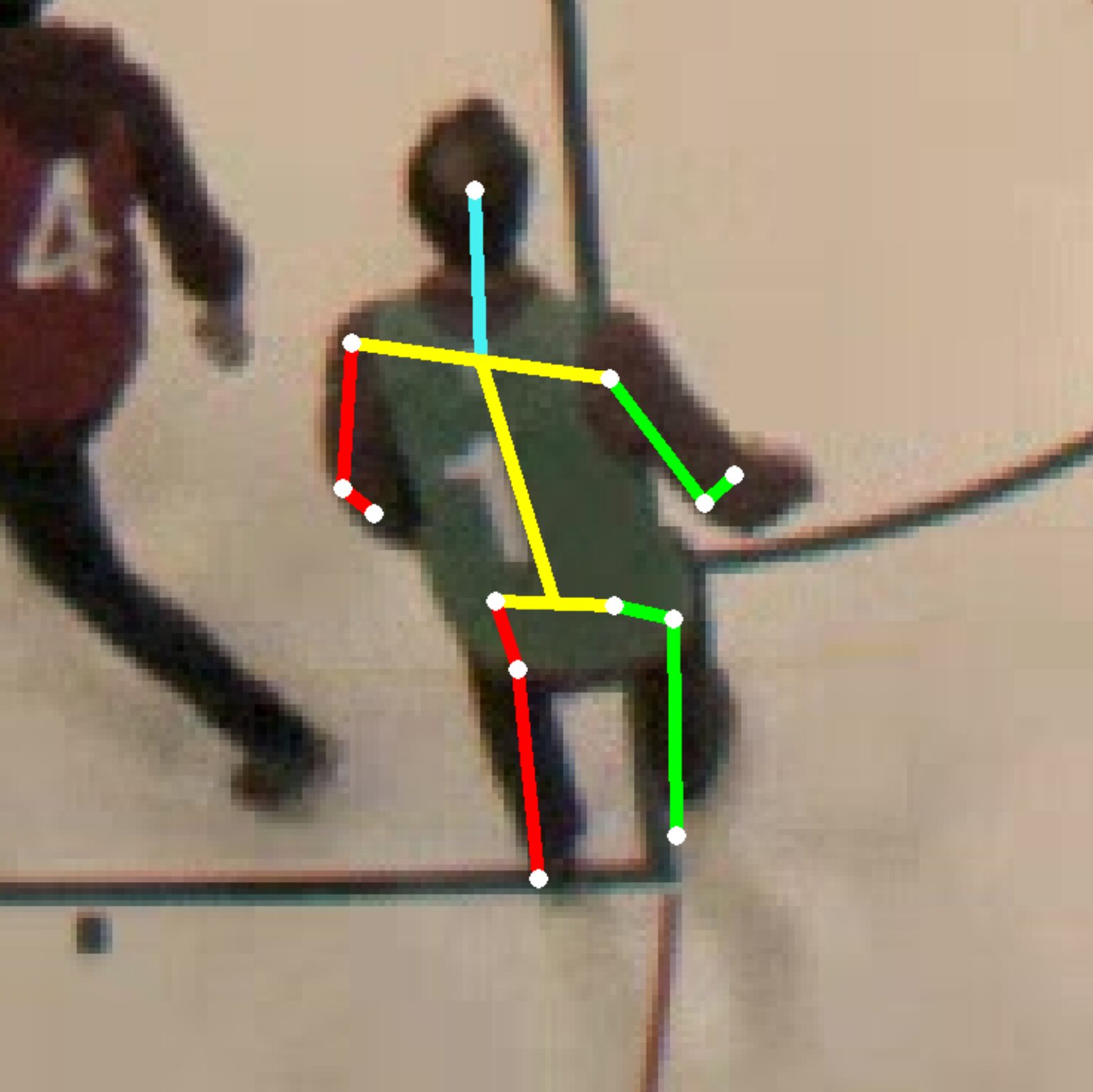}
    \end{subfigure}
    \begin{subfigure}[b]{0.16\linewidth}        
        \centering
        \includegraphics[width=\linewidth]{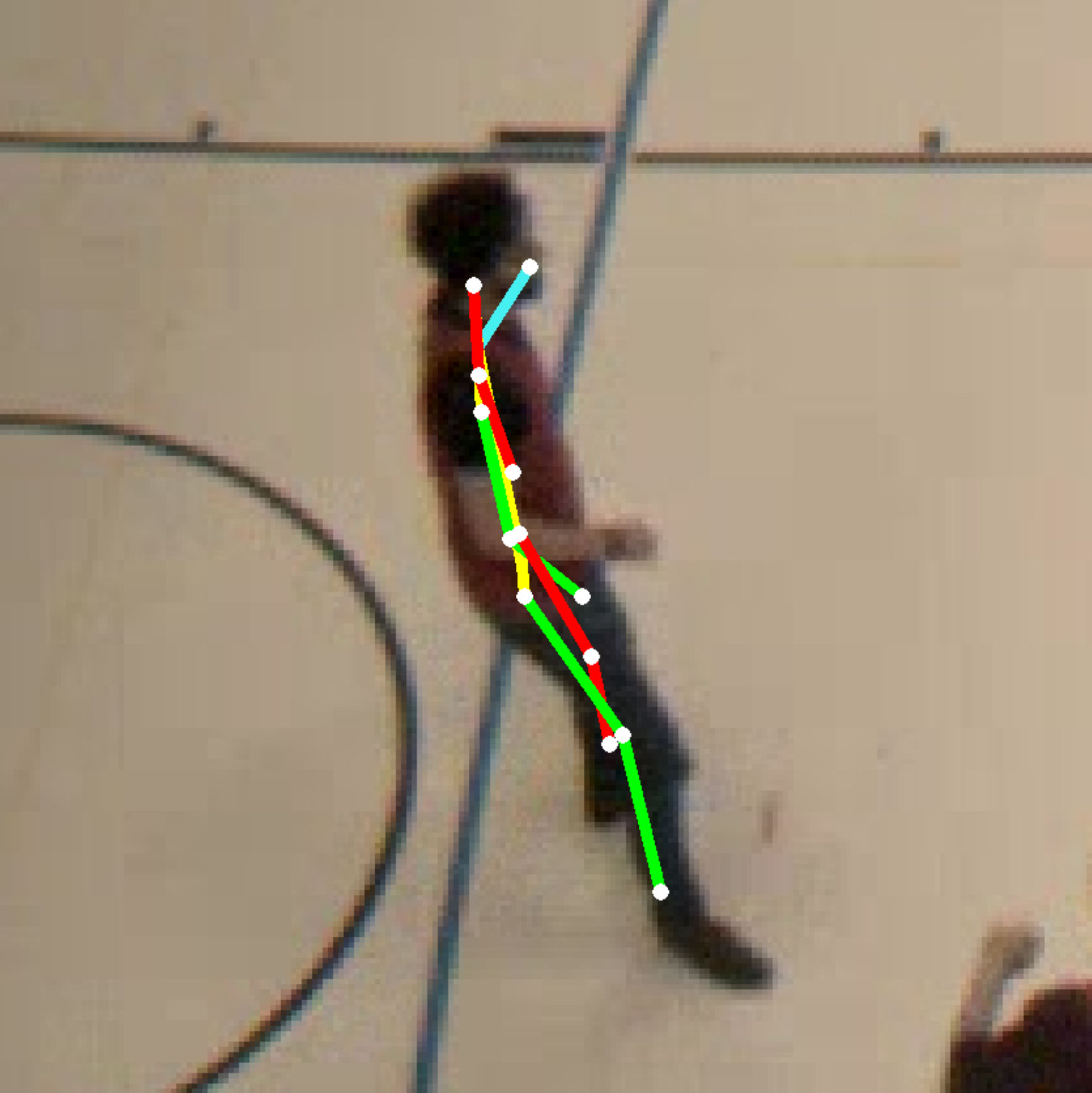}
    \end{subfigure}
    \begin{subfigure}[b]{0.16\linewidth}        
        \centering
        \includegraphics[width=\linewidth]{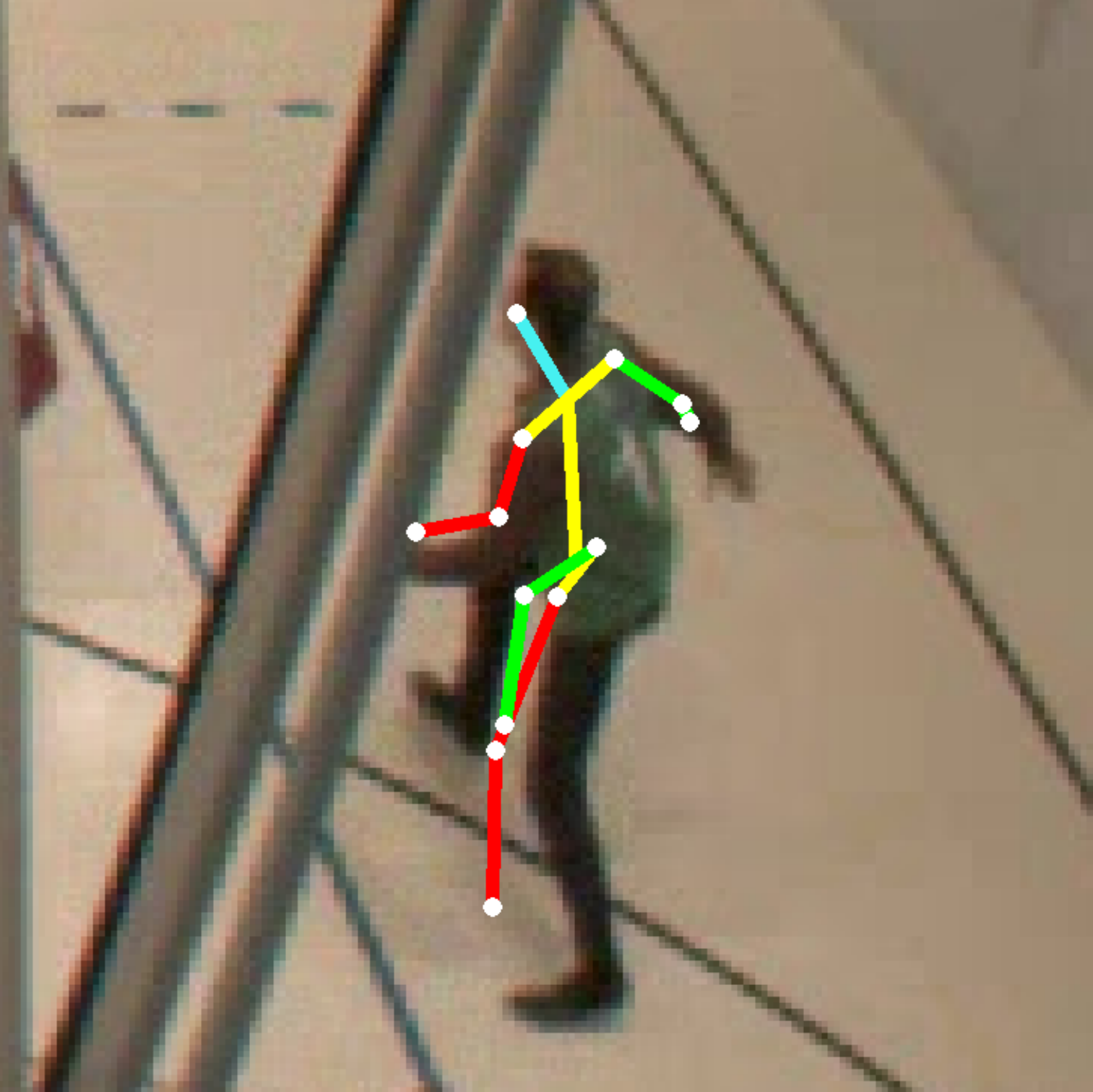}
    \end{subfigure}

    \caption{\small \textbf{Qualitative results on the unlabeled (training) set of the SportCenter dataset.} From top to bottom, superimposed 3D poses results of (a) multi-view triangulation, (b) our single-view lifting approach, and (c) same 3D pose as b) but projected into another view to visualize the depth information.}
    \label{fig:visuals_sc_sup_train}
\end{figure*}

Figure~\ref{fig:visuals_h36m2},~\ref{fig:visuals_3dhp},~\ref{fig:visuals_sc_sup} and~\ref{fig:visuals_sc_sup_train} show qualitative results for Human3.6M~\cite{Ionescu14a},  MPI-INF-3DHP~\cite{Mehta17a}, and SportCenter datasets. For the SportCenter dataset we also show the 3D pose result of triangulating 2D predictions from multiple views.

\section{Conclusion}
\label{sec:conclusion}

In this work, we have presented an effective way of using triangulation as a form of self-supervision for single view 3D pose estimation. Our approach imposes geometrical constraints on multi-view images to train models with little annotations in a semi-supervised learning setup. Making the pipeline end-to-end differentiable was key to this. We have demonstrated how the proposed robust triangulation can reliably generate pseudo labels even in crowded scenes and how to use them to supervise a single view 3D pose estimator. The experimental results, especially in the SportCenter dataset, clearly shows its advantages in learning a precise triangulated 3D pose over other triangulation methods. In future, we plan to further investigate the impact of integrating temporal or spatial constraints into our learning framework.
\bibliographystyle{splncs04}
\bibliography{string,vision,photog,learning,optim}


\end{document}